\PassOptionsToPackage{dvipsnames}{xcolor}
\documentclass[acmsmall]{acmart}
\AtBeginDocument{%
  }

\setcopyright{acmlicensed}
\copyrightyear{2018}
\acmYear{2018}
\acmDOI{XXXXXXX.XXXXXXX}

\acmJournal{TIST}
\acmVolume{37}
\acmNumber{4}
\acmArticle{111}
\acmMonth{8}

\usepackage{hyperref} 

\usepackage{graphicx}
\usepackage{tikz}

\usepackage{forest}
\usetikzlibrary{trees,positioning,shapes,shadows,arrows.meta, shadows.blur}

\usepackage{caption}
\usepackage{subcaption}

\usepackage{booktabs}
\usepackage{multirow, makecell, rotating}

\usepackage{pifont}

\usepackage{pgf}
\usepackage{pgfplots}
\usepgfplotslibrary{groupplots}
\pgfplotsset{compat=1.18}

\usepackage{ifthen}

\usepackage[most]{tcolorbox}
\tcbuselibrary{listings, breakable}
\usepackage{enumitem}

\usepackage{algorithm}
\usepackage{amsmath}
\usepackage{algpseudocode}

\usepackage{awesomebox}

\usepackage[mathlines,switch]{lineno} 

\newcommand*\colorcheck[1]{%
  \expandafter\newcommand\csname #1check\endcsname{\textcolor{#1}{\ding{52}}}%
}
\colorcheck{blue}
\colorcheck{green}
\colorcheck{yellow}
\colorcheck{orange}
\colorcheck{red}

\definecolor{mypink}{rgb}{0.0, 0.0, 0.0}
\newcommand{\news}[1]{\textcolor{mypink}{#1}} 

\newcommand*\colourxmark[1]{%
  \expandafter\newcommand\csname #1xmark\endcsname{\textcolor{#1}{\ding{56}}}%
}
\colourxmark{blue}
\colourxmark{green}
\colourxmark{red}

\newcommand*\colourtriangle[1]{%
  \expandafter\newcommand\csname #1triangle\endcsname{\textcolor{#1}{\ding{115}}}%
}
\colourtriangle{blue}
\colourtriangle{green}
\colourtriangle{red}

\newcommand{\blackcircle}[1]{%
  \tikz[baseline=(char.base)]{
    \node[shape=circle,draw,inner sep=0.5pt,fill=black,text=white] (char) {#1};
  }
}

\newcommand{\supportbar}[3][black!80]{%
  \pgfmathsetmacro{\barfraction}{min(1.0, #2/#3)}
  \pgfmathsetlengthmacro{\barwidth}{1.6cm * \barfraction} 
  \begin{tikzpicture}[baseline=0ex]
    \draw[draw=gray!50, fill=gray!10] (0,0) rectangle (1.6cm, 0.2cm);
    \fill[fill=#1] (0,0) rectangle (\barwidth, 0.2cm);
    \foreach \x in {0.4333, 0.8666, 1.3} {
      \draw[gray!70, line width=0.2pt] (\x cm, 0) -- (\x cm, 0.2cm);
    }
  \end{tikzpicture}
}

\definecolor{RYB1}{RGB}{141, 211, 199}
\definecolor{RYB2}{RGB}{255, 255, 179}
\definecolor{RYB3}{RGB}{190, 186, 218}
\definecolor{RYB4}{RGB}{251, 128, 114}

\pgfplotscreateplotcyclelist{colorbrewer-RYB}{
    {RYB1!50!black,fill=RYB1},
    {RYB2!50!black,fill=RYB2},
    {RYB3!50!black,fill=RYB3},
    {RYB4!50!black,fill=RYB4},
}

\pgfplotsset{
    flatuilight/.style={
        ybar,
        bar width=5pt,
        axis line style={black!60},
        tick style={black!60},
        tick label style={font=\scriptsize\sffamily},
        label style={font=\scriptsize\sffamily},
        grid=major,
        grid style={solid,gray!20},
        legend style={
            font=\tiny,
            draw=none,
            fill=none,
            at={(0.98,0.98)},
            yshift=-6pt,
            anchor=north east
        },
        every axis plot/.append style={
        draw=none,
        fill opacity=0.95
        },
        every node near coord/.append style={
            font=\tiny
        }
    }
}

\pgfplotsset{
    llmplot/.style={
        flatuilight,
        ybar,
        bar width=4.4pt,
        width=\linewidth,
        height=0.6\linewidth,
        title style={yshift=-145pt, font=\small},
        symbolic x coords={
            DeepSeek-R1, 
            DeepSeek-V3, 
            {Llama 3.3 70B}, 
            {Llama 3.1 405B},
            {Llama 3.1 70B}, 
            {Llama 3.1 8B}, 
            {Qwen 2.5 Coder 32B}, 
            {Qwen QwQ Preview 32B}
        },
        xtick=data,
        xtick pos=bottom,
        x tick label style={rotate=45, anchor=east, font=\tiny\sffamily, text width=9ex, xshift=0.3ex,yshift=-1.0ex, align=right},
        y tick label style={font=\scriptsize\sffamily},
        ymajorgrids=true,
        enlarge x limits=0.1,
        bar shift auto=true,
        every axis plot/.append style={line width=0.5pt},
        legend style={
            at={(1.1,1.1)}, anchor=south,
            legend columns=-1,
            font=\tiny,
            /tikz/every even column/.append style={column sep=2pt}
        },
        legend image code/.code={
            \draw[#1, line width=0.5pt] (0cm,-0.6mm) rectangle (1mm,1mm);
        },
        nodes near coords,
        every node near coord/.append style={rotate=90, anchor=west, font=\tiny\sffamily},
        cycle list name=colorbrewer-RYB, fill opacity=0.9,
    }
}

\newcommand{\High}{\textcolor{ForestGreen}{High}}
\newcommand{\Medium}{\textcolor{Goldenrod!80!Gray}{Medium}}
\newcommand{\Low}{\textcolor{BrickRed}{Low}}

\tcbuselibrary{breakable}

\newtcolorbox{findingbox}[1][]{
    colback=blue!2!gray!8,
    colframe=black!40,
    boxrule=0.5pt,
    arc=3pt,
    left=4pt, right=4pt, top=4pt, bottom=4pt,
    fonttitle=\sffamily\bfseries,
    fontupper=\sffamily\footnotesize,
    breakable,
    #1
}

\begin{document}

\title{A Survey on Inference Engines for Large Language Models: Perspectives on Optimization and Efficiency}

\author{Sihyeong Park}
\email{sihyeong@keti.re.kr}
\orcid{0000-0001-8244-4817}
\affiliation{%
  \institution{Korea Electronics Technology Institute}
  \city{Seongnam-si}
  \state{Gyeonggi-do}
  \country{South Korea}
}

\author{Sungryeol Jeon}
\affiliation{%
  \institution{Korea Electronics Technology Institute}
  \city{Seongnam-si}
  \state{Gyeonggi-do}
  \country{South Korea}}
  \orcid{0009-0003-6890-230X}
\email{wjstjdfuf98@keti.re.kr}

\author{Chaelyn Lee}
\affiliation{%
  \institution{Korea Electronics Technology Institute}
  \city{Seongnam-si}
  \state{Gyeonggi-do}
  \country{South Korea}}
\orcid{0009-0008-0417-4864}
\email{mylynchae@keti.re.kr}

\author{Seokhun Jeon}
\affiliation{%
  \institution{Korea Electronics Technology Institute}
  \city{Seongnam-si}
  \state{Gyeonggi-do}
  \country{South Korea}}
\orcid{0009-0002-4009-178X}
\email{seokhun.jeon@keti.re.kr}

\author{Byung-Soo Kim}
\affiliation{%
  \institution{Korea Electronics Technology Institute}
  \city{Seongnam-si}
  \state{Gyeonggi-do}
  \country{South Korea}}
\orcid{0009-0004-2996-2841}
\email{bskim4k@keti.re.kr}

\author{Jemin Lee}
\authornote{Corresponding Author}
\affiliation{%
  \institution{Electronics and Telecommunications Research Institute}
  \state{Daejeon}
  \country{South Korea}}
\orcid{0000-0002-9332-3508}
\email{leejaymin@etri.re.kr}

\renewcommand{\shortauthors}{Park et al.}

\begin{abstract}
Large language models (LLMs) are widely applied in chatbots, code generators, and search engines. Workload such as chain-of-throught, complex reasoning, agent services significantly increase the inference cost by invoke the model repeatedly. Optimization methods such as parallelism, compression, and caching have been adopted to reduce costs, but the diverse service requirements make it hard to select the right method. Recently, specialized LLM inference engines have emerged as a key component for integrating the optimization methods into service-oriented infrastructures. However, a systematic study on inference engines is still lacking.This paper provides a comprehensive evaluation of 25 open-source and commercial inference engines. We examine each inference engine in terms of ease-of-use, ease-of-deployment, general-purpose support, scalability, and suitability for throughput- and latency-aware computation. Furthermore, we explore the design goals of each inference engine by investigating the optimization techniques it supports. In addition, we assess the ecosystem maturity of open source inference engines and handle the performance and cost policy of commercial solutions.We outline future research directions that include support for complex LLM-based services, support of various hardware, and enhanced security, offering practical guidance to researchers and developers in selecting and designing optimized LLM inference engines. We also provide a public repository to continually track developments in this fast-evolving field: \href{https://github.com/sihyeong/Awesome-LLM-Inference-Engine}{https://github.com/sihyeong/Awesome-LLM-Inference-Engine}.
\end{abstract}



\received{20 February 2007}
\received[revised]{12 March 2009}
\received[accepted]{5 June 2009}

\maketitle


\section{Introduction} \label{sec:introduction}
Large Language Models (LLMs) are being utilized in a wide range of services, such as chatbots, code generation, and search engines, with remarkable examples including OpenAI's ChatGPT~\cite{achiam2023gpt}, GitHub Copilot~\cite{copliot}, and Google Gemini~\cite{gemini}. Building on these successes, numerous new models and services have rapidly emerged; however, this expansion introduces new challenges in deploying and serving LLMs on a scale.

Recent trends like reasoning-centric test-time scaling~\cite{snell2024scaling, kumar2025llm} and LLM-based AI agents~\cite{li2024survey, guan2024intelligent} have significantly increased both computational demands and the number of inference calls in LLM-based applications. Reasoning-centric test-time scaling replaces single-pass answer generation with multi-step reasoning or iterative self-verification to improve output quality. Also known as chain-of-thought (CoT)~\cite{wei2022chain}, self-consistency~\cite{chen2023universal}, and test-time reasoning~\cite{hao2024llm}, these methods increase accuracy by invoking the model multiple times per query, thereby raising latency and computing costs. Meanwhile, LLM-based AI agents such as AutoGPT~\cite{autogpt} and LangChain~\cite{LangChain} autonomously plan a sequence of tasks to fulfill a single user request, repeatedly calling the model within a single session. Consequently, inference efficiency has become essential for deploying both reasoning-oriented LLMs and AI agents in practice.

To manage the growing inference costs of LLMs, various optimization techniques---such as quantization~\cite{egashira2025exploiting}, lightweight architectures~\cite{xu2024foundationsurvey}, and knowledge distillation (KD)~\cite{yang2024survey}---have been adopted. In large-scale services, however, the diversity of prompt lengths, query types, and output formats often means that a single optimization method cannot cover every scenario. As a result, LLM inference engines, which offer multiple optimization strategies and handle the inference process, have become crucial infrastructure components that directly affect both service quality and cost.

Although general-purpose deep learning frameworks like PyTorch~\cite{paszke2019pytorch} and TensorFlow~\cite{abadi2016tensorflow}---originally designed to support a wide range of models, from convolutional neural networks (CNNs) to recurrent neural networks (RNNs)---are widely used for LLM inference, they prioritize broad hardware and architecture compatibility. Consequently, they do not include various specialized optimizations for LLMs or for sequential decoding. Running large-scale models on these frameworks can lead to slower performance and higher resource usage, underscoring the need for dedicated inference solutions.

Reflecting this need, a growing number of specialized LLM inference engines have emerged. They provide capabilities such as batching, streaming, and attention optimizations that are not typically found in general-purpose frameworks. However, Each engine targets different hardware such as graphics processing units (GPUs) and LLM accelerators, optimization scopes ranging from model compression to memory offloading, and intended use cases varying from real-time conversational systems to large-scale text generation. As a result, the ecosystem has become both rapidly evolving and fragmented, making it difficult to determine which optimization methods are supported by each engine and how effectively they perform under various conditions. Consequently, there is a pressing need for a comprehensive review and comparison of LLM inference engines and the optimization techniques they offer.

\begin{table}[tbp]
    \centering
    \caption{Comparison of Representative Surveys on Efficient LLM Inference}
    \label{tab:compare_survey}
    \resizebox{.95\textwidth}{!}{%
    \begin{tabular}
           { >{\centering\arraybackslash}p{.22\textwidth} 
            p{.5\textwidth} 
            >{\centering\arraybackslash}m{.18\textwidth} 
            p{.45\textwidth}  }
        \toprule
        \multicolumn{1}{c}{Survey} & 
        \multicolumn{1}{c}{Scope} & 
        \multicolumn{1}{c}{\makecell{\# of Reviewed\\ Inference Engine}} & 
        \multicolumn{1}{c}{Limitation}   \\ 
        \midrule
        Chitty-Venkata et al., JSA (2023) \cite{chitty2023survey} & \textbf{Efficient inference} - architecture design, knowledge distillation, pruning, quantization &  \redxmark &  Covers only optimization techniques for efficient inference \\  
        
        Miao et al., ArXiv (2023) \cite{miao2023towards} & \textbf{Efficient model serving }- decoding algorithms, architecture design, model compression, quantization, parallel computation, memory management, request scheduling, kernel optimizations  & 10 &  Covers only parallel computation, iteration scheduling, attention kernel support, and brief main features of the inference engine \\  
        
        Bai et al., ArXiv (2024) \cite{bai2024beyond} & \textbf{Resource-efficient model }- architecture design, pre-training, fine-tuning, inference optimization, system design & \redxmark &  Covers only model-side optimization techniques for efficient inference \\  
       
        Xu et al., ArXiv (2024) \cite{xu2024foundationsurvey} & \textbf{Resource-efficient foundation models} - foundation model, architecture design, resource-efficient algorithms, resource-efficient systems & 23 & Provides information on training and inference support in cloud and edge environments, as well as inference optimization techniques, but lacks detailed description of the inference engine. \\  
        
        Park et al., ArXiv (2024) \cite{park2024comprehensive}& \textbf{Model compression }- pruning, quantization, knowledge distillation, low-rank approximation, parameter sharing, architecture design& \redxmark & Covers only model compression techniques \\  
        
        Yuan et al., ArXiv (2024) \cite{yuan2024llm} & \textbf{Efficient inference }- model compression, fast decoding algorithm, compiler/system optimization, hardware optimization  & \redxmark &  Covers only optimization techniques for efficient inference \\  
       
        Zhu et al., TACL (2024) \cite{zhu2024survey} &\textbf{Model compression }- quantization, pruning, knowledge distillation, low-rank factorization& \redxmark & Covers only model compression techniques  \\  
       
        Wang et al., ArXiv (2024) \cite{wang2024model} & \textbf{Model compression and efficient inference} - quantization, pruning, knowledge distillation, architecture design, framework  & 6 &  Provides explanations focused on optimization features rather than the inference engine itself, and includes outdated inference engines \\  
        
        Zhou et al., ArXiv (2024) \cite{zhou2024survey} & \textbf{Efficient inference} - data-level optimization, architecture design, model compression, inference engine, serving system & 18 &  Explores optimization techniques from the perspectives of inference optimization and serving optimization, but describes only a limited set of techniques \\  
        
        Wan et al., TMLR (2024) \cite{wan2024efficient}  & \textbf{Efficient models} - model optimization schemes, data selection/engineering, framework & 18 & Describes training, fine-tuning, and inference support of the inference engine along with key features, but takes a more comprehensive view rather than focusing specifically on inference\\  
        
        Li et al., ArXiv (2024) \cite{li2024large}  & \textbf{Hardware perspective inference optimization} - hardware architecture (CPU, GPU, FPGA, ASIC, PIM/NDP), quantization, sparsity, speculative decoding, homogeneous/heterogeneous cooperation & \redxmark & Covers only hardware-aware optimization techniques for efficient inference \\  
       
        Xu et al., CSUR (2025) \cite{xu2025resource}& \textbf{Resource-efficient algorithms} - attention optimization, architecture design, pre-training, fine-tuning, inference algorithm, model compression, distributed training, serving & 18 & Covers training and inference support, but lacks sufficient description of the inference engine \\  
        
        Zheng et al., CSUR (2025) \cite{zheng2024review} & \textbf{Efficient models} - model compression, runtime optimization, on-device applications & 8 & Covers mobile and desktop support and related optimization techniques only briefly \\  

        \news{\textbf{Ours}} & \news{\textbf{Inference engine and efficient inference} - open-source and commercial inference engine, inference optimization of inference engine} & \makecell{\news{\textbf{25}}\\ \news{\textbf{(New: 13)}}} & \news{Covers only covers inference engines and their optimizations} \\  
        \bottomrule
    \end{tabular}%
    }
\end{table}

Most existing surveys on LLM optimization (Table~\ref{tab:compare_survey}) have focused on specific methods, such as model compression or hardware acceleration, and therefore have not fully explored which optimization techniques are supported by individual inference engines. In addition, many of these surveys omit recently released commercial engines. For instance, Chitty-Venkata et al.~\cite{chitty2023survey} and Yuan et al.~\cite{yuan2024llm} focus on transformer-based model compression, while Park et al.~\cite{park2024comprehensive} and Zhu et al.~\cite{zhu2024survey} examine compression methods in detail. 
Similarly, works such as Xu et al.~\cite{xu2024foundationsurvey}, Xu et al.~\cite{xu2025resource} and Wang et al.~\cite{wang2024model} discuss optimization strategies for LLM inference or serving systems in cloud or edge environments, but they lack a detailed examination of the design and implementation of each engine.
Consequently, there remains a gap in the literature for a survey that not only presents the current landscape of LLM inference engines, but also systematically links their specialized features to the optimization techniques they implement.

To fill this gap, this paper adopts a framework-centric perspective, thoroughly examining a range of LLM inference engines and categorizing the optimization techniques each one implements. In particular, it maps how these engines handle methods like quantization, caching, and parallelization, enabling readers to quickly identify engines that align with specific requirements. This paper also includes recently released commercial engines that are not covered in previous surveys, comparing their architectural goals, hardware targets, and significant features. 

\tikzset{
    basic/.style = { draw, align=center, font=\sffamily, rectangle },
    root/.style   = { basic, draw=none, rounded corners=3pt, thin, align=center,  fill=Lavender!30 }, 
    section_node/.style  = { basic, draw=none, thin, rounded corners=3pt, align=center, fill=MidnightBlue!30, text width=13em }, 
    subsection_node/.style  = { basic, draw=none, thin, rounded corners=3pt, fill=SpringGreen!30, text width=13em, align=center }, 
    subsubsection_node/.style  = { basic, draw=none, thin, rounded corners=3pt, fill=Peach!30, text width=18em, align=center }, 
    context_node/.style  = { basic, draw=none, thin, rounded corners=3pt, fill=Gray!20, text width=26em, align=center },
    edge from parent/.style = { draw=black, edge from parent fork right }
}
\begin{figure}[tbp]
    \centering
    \resizebox{.99\textwidth}{!}{
    \begin{forest}
        for tree={
            grow=0,reversed,
            calign=midpoint,
            growth parent anchor=west,
            parent anchor=east,
            child anchor=west,
            level distance=20mm,
            scale=1,
            edge path={\noexpand\path[\forestoption{edge},->, >={latex}] 
                (!u.parent anchor) -- +(10pt,0pt) |-  (.child anchor) 
                \forestoption{edge label};}
        }
        [
        {{\centering \rotatebox{90}{A Survey on Inference Engines for Large Language Models: Perspectives on Optimization and Efficiency}}}, root, s sep=3mm, anchor=center, align=center, l sep=10mm,
            [Introduction (\S\ref{sec:introduction}), section_node, edge={draw=black, ->, >={latex}}, l sep=12mm]
            [Backgrounds (\S\ref{sec:background}), section_node, edge={draw=black, ->, >={latex}}, l sep=12mm,
                [Structure and Optimizations of LLMs (\S\ref{sec:background_strucure}), subsection_node, l sep=6mm , text width=32em]
                [LLM Inference Process and Optimization (\S\ref{sec:background_inference}), subsection_node, l sep=6mm , text width=32em]
                [Inference-Aware LLM Serving Workflow (\S\ref{sec:background_serving}), subsection_node, l sep=6mm, text width=32em]
                [Emerging Trends in LLM Inference (\S\ref{sec:emerging_trends}), subsection_node, l sep=6mm, text width=32em]
            ]
            [Practical Guides to \\Inference Engines  (\S\ref{sec:practical_guide}), section_node, edge={draw=black, ->, >={latex}}, l sep=12mm,
                [Ecosystem Maturity and Sustainability Signals (\S\ref{sec:practical_guide_ecosystem}), subsection_node, l sep=6mm, text width=32em]
                [Hardware Compatibility and Platform Support (\S\ref{sec:practical_guide_hardware_platform}), subsection_node, l sep=6mm, text width=32em]
                [Design and Pricing Strategies of Commercial Inference Engines (\S\ref{sec:parctical_guide_commercial_engine}), subsection_node, l sep=6mm, text width=32em]
            ]
            [Detailed Review of \\Inference Engines (\S\ref{sec:detailed_review}), section_node, edge={draw=black, ->, >={latex}}, l sep=12mm,
                [Single-Node \& \\Heterogeneous Devices (\S\ref{sec:detailed_review}), subsection_node, l sep=6mm, 
                    [{Ollama \cite{ollama} (\S\ref{sec:detailed_review_ollama}),
                    llama.cpp \cite{llamacpp} (\S\ref{sec:detailed_review_llamacpp}),
                    MAX \cite{max} (\S\ref{sec:detailed_review_max}), \\
                    MLC LLM \cite{mlcllm} (\S\ref{sec:detailed_review_mlcllm}),  
                    PowerInfer \cite{song2024powerinfer,xue2024powerinfer} (\S\ref{sec:detailed_review_powerinfer}),
                    TGI \cite{tgi}} (\S\ref{sec:detailed_review_together}), 
                    subsubsection_node, text width=36em]
                ]
                [Single-Node \& \\Homogeneous Devices (\S\ref{sec:detailed_review}), subsection_node, l sep=6mm,
                    [{Unsloth \cite{unsloth} (\S\ref{sec:detailed_review_unsloth}),
                    llama2.c \cite{llama2c} (\S\ref{sec:detailed_review_llama2c}),
                    bitnet.cpp \cite{wang20241} (\S\ref{sec:detailed_review_bitnetcpp}), \\
                     OpenLLM \cite{openllm} (\S\ref{sec:detailed_review_openllm}),
                    LightLLM \cite{lightllm} (\S\ref{sec:detailed_review_lightllm}),   
                     NanoFlow \cite{zhu2024nanoflow} (\S\ref{sec:detailed_review_nanoflow}), \\
                     vAttention \cite{prabhu2025vattention} (\S\ref{sec:detailed_review_vattention}),
                    Sarathi-Serve \cite{agrawal2024taming} (\S\ref{sec:detailed_review_sarithiserve}),
                    Friendli Inference \cite{friendli}} (\S\ref{sec:detailed_review_friendli}), 
                    subsubsection_node , text width=36em]
                ]
                [Multi-Node \& \\Heterogeneous Devices (\S\ref{sec:detailed_review}), subsection_node, l sep=6mm, 
                    [{vLLM \cite{kwon2023efficient} (\S\ref{sec:detailed_review_vllm}), 
                    DeepSpeed-FastGen \cite{holmes2024deepspeed} (\S\ref{sec:detailed_review_deepspeed}),
                    SGLang \cite{zheng2024sglang} (\S\ref{sec:detailed_review_sglang}), \\
                    LitGPT \cite{litgpt} (\S\ref{sec:detailed_review_litpgt}),
                    LMDeploy \cite{2023lmdeploy} (\S\ref{sec:detailed_review_lmdeploy}), 
                    Fireworks AI \cite{fireworks} (\S\ref{sec:detailed_review_fireworks}), \\
                    Together Inference \cite{together}} (\S\ref{sec:detailed_review_together}), 
                    subsubsection_node, text width=36em]
                ]
                [Multi-Node \& \\Homogeneour Devices (\S\ref{sec:detailed_review}), subsection_node, l sep=6mm,
                    [{TensorRT-LLM \cite{tensorrtllm} (\S\ref{sec:detailed_review_tensorrtllm}),
                    DistServe \cite{zhong2024distserve} (\S\ref{sec:detailed_review_distserve}),
                    GroqCloud \cite{groqcloud}} (\S\ref{sec:detailed_review_groq}), 
                    subsubsection_node, text width=36em]
                ]
            ]
            [LLM Inference \\Optimization (\S\ref{sec:inference_optimization}), section_node, edge={draw=black, ->, >={latex}}, l sep=12mm,
                [Batch Optimization (\S\ref{sec:inference_optimization_batch}), subsection_node, l sep=6mm,
                    [Dynamic Batching \cite{crankshaw2017clipper, ali2020batch} (\S\ref{sec:inference_optimization_batch_dynamic}), subsubsection_node]
                    [Continuous Batching \cite{yu2022orca, he2024deferred} (\S\ref{sec:inference_optimization_batch_continuous}), subsubsection_node]
                    [Nano-batching \cite{zhu2024nanoflow} (\S\ref{sec:inference_optimization_batch_nano}), subsubsection_node]
                    [Chunked-prefills \cite{agrawal2023sarathi} (\S\ref{sec:inference_optimization_batch_chunked}), subsubsection_node]
                ]
                [Parallelism (\S\ref{sec:inference_optimization_parallelism}), subsection_node, l sep=6mm,
                    [Data Parallelism \cite{rajbhandari2020zero} (\S\ref{sec:inference_optimization_parallelism_data}), subsubsection_node]
                    [Fully Shared \\ Data Parallelism \cite{zhao2023pytorch} (\S\ref{sec:inference_optimization_parallelism_fsdp}), subsubsection_node]
                    [Expert Parallelism \cite{liu2024deepseek-v3, zhu2025megascale, balmau2025accelerating} (\S\ref{sec:inference_optimization_expert}), subsubsection_node]
                    [Tensor Parallelism \cite{stojkovic2024towards, prabhakar2024kraken} 
                    (\S\ref{sec:inference_optimization_tensor}), subsubsection_node]
                    [Pipeline Parallelism \cite{agrawal2023sarathi, hu2021pipeline, ma2024hpipe, yu2024twinpilots} (\S\ref{sec:inference_optimization_parallelism_pipeline}), subsubsection_node]
                ]
                [Compression (\S\ref{sec:inference_optimization_compression}), subsection_node, l sep=6mm,
                    [Quantization (\S\ref{sec:inference_optimization_quantization}) , subsubsection_node,
                        [{PTQ \cite{li2023fptq}, 
                        QAT \cite{chen2024efficientqat, liu2023llm}, 
                        AQLM \cite{egiazarian2024extreme}, 
                        SmoothQuant \cite{xiao2023smoothquant}, \\
                        KV Cache Quantization \cite{hooper2024kvquant, liu2024kivi}, 
                        EXL2 \cite{exl2}, 
                        EETQ \cite{eetq}, \\
                        LLM Compressor \cite{llmcompressor}, 
                        GPTQ \cite{frantar2022gptq}, 
                        Marlin \cite{frantar2025marlin}, 
                        Microscaling Format \cite{rouhani2023microscaling}}, context_node, yshift=18pt] 
                    ]
                    [Pruning (\S\ref{sec:inference_optimization_pruning}), subsubsection_node,
                        [{cuSPARSE \cite{cusparse}, 
                        Wanda \cite{sun2023simple}, 
                        Mini-GPTs \cite{valicenti2023mini}, \\
                        Token pruning \cite{fu2024lazyllm},
                        Post-Training Pruning \cite{zhao2024pruning}}, 
                        context_node]
                    ]
                    [Sparsity Optimization (\S\ref{sec:inference_optimization_sparsity}), subsubsection_node,
                        [{Structured Sparsity \cite{zheng2024learn, dong2023towards}, 
                        Dynamic Sparsity \cite{zhang2023dynamic}, \\
                        Kernel-level Sparsity \cite{xia2023flash, borvstnik2014sparse,xFormers2022,xiang2025cutespmm}, 
                        Block Sparsity \cite{gao2024seerattention}, \\
                        N:M Sparsity \cite{zhang2022learning}, 
                        MoE \cite{cai2024survey}, 
                        Sparse MoE \cite{fedus2022switch, du2022glam}, \\
                        Dynamic Token Sparsity \cite{yang2024post, fu2024lazyllm}, \\
                        Contextual Sparsity \cite{liu2023deja,akhauri2024shadowllm}}, 
                        context_node]
                    ]
                ]
                [Inference-aware Fine-Tuning (\S\ref{sec:inference_optimization_fine_tuning}), subsection_node, l sep=6mm,
                    [Full-Parameter Fine-Tuning \cite{lv2023full} (\S\ref{sec:inference_optimization_fine_tuning}), subsubsection_node]
                    [Parameter-Efficient Fine-Tuning (PEFT) (\S\ref{sec:inference_optimization_fine_tuning}), subsubsection_node,
                        [{LoRA \cite{hu2022lora, sheng2023s}, QLoRA \cite{dettmers2023qlora, zhang2023machine}}, 
                        context_node, yshift=-6pt]
                    ]
                ]
                [Caching (\S\ref{sec:inference_optimization_caching}), subsection_node, l sep=6mm,
                    [Prompt Caching \cite{zhu2024efficient} (\S\ref{sec:inference_optimization_caching_prompt}), subsubsection_node]
                    [Prefix Caching \cite{liu2024optimizing, pan2024marconi} (\S\ref{sec:inference_optimization_caching_prefix}), subsubsection_node]
                    [KV Caching \cite{pope2023efficiently}  (\S\ref{sec:inference_optimization_caching_kv}), subsubsection_node]
                ]
                [Attention Optimization (\S\ref{sec:inference_optimization_attention}), subsection_node, l sep=6mm,
                    [KV Cache Optimization (\S\ref{sec:inference_optimization_attention_paged}), subsubsection_node,
                        [{PagedAttention \cite{kwon2023efficient}, 
                        TokenAttention \cite{lightllm}, \\
                        ChunkedAttention \cite{ye2024chunkattention}}, 
                        context_node, yshift=6pt]
                    ]
                    [I/O Optimization (\S\ref{sec:inference_optimization_attention_flash}), subsubsection_node,
                        [FlashAttention \cite{dao2022flashattention,dao2023flashattention,shah2024flashattention}, context_node]
                    ]
                    [KV Cache Reuse (\S\ref{sec:inference_optimization_attention_radix}), subsubsection_node,
                        [RadixAttention \cite{zheng2024sglang}, context_node]
                    ]
                    [Attention Programming Model (\S\ref{sec:inference_optimization_attention_flex}), subsubsection_node,
                        [FlexAttention \cite{dong2024flex}, context_node]
                    ]
                    [MQA Optimization (\S\ref{sec:inference_optimization_attention_fire}), subsubsection_node,
                        [FireAttention \cite{fireworks}, context_node]
                    ]
                ]
                [Sampling Optimization (\S\ref{sec:inference_optimization_sampling}), subsection_node, l sep=6mm,
                    [Speculative Decoding (\S\ref{sec:inference_optimization_sampling}), subsubsection_node,
                        [{EAGLE \cite{li2024eagle, li2024eagle2, li2025eagle}, 
                        Medusa \cite{cai2024medusa}, 
                        ReDrafter \cite{cheng2024recurrent}}, 
                        context_node]
                    ]
                ]
                [Structured Outputs (\S\ref{sec:inference_optimization_structured_outputs}), subsection_node, l sep=6mm,
                    [Constrained Decoding (\S\ref{sec:inference_optimization_structured_outputs}), subsubsection_node,
                        [{FSM \cite{willard2023efficient}, 
                        CFG \cite{geng2023grammar, barke2024hysynth}, 
                        Outlines  \cite{outlines}, \\
                        XGrammar \cite{dong2024xgrammar}, 
                        LM Format Enforcer \cite{lm-format-enforcer}, \\
                        llguidance  \cite{llguidance}, 
                        GBNF \cite{gbnf}, 
                        OpenAI Structured Outputs \cite{openai_structure_outputs}, 
                        JSONSchemaBench \cite{geng2025generating}, \\
                        StructTest  \cite{chen2024structtest}, 
                        SoEval \cite{liu2024llms}},
                        context_node]
                    ]
                ]
            ]
            [Empirical Evaluation (\S\ref{sec:experiments}), section_node, edge={draw=black, ->, >={latex}}, l sep=12mm]
            [Future Directions and Open Challenges (\S\ref{sec:future_direction}), section_node, edge={draw=black, ->, >={latex}}, l sep=12mm]
            [Conclusion (\S\ref{sec:conclusion}), section_node, edge={draw=black, ->, >={latex}}, l sep=12mm]
        ]
    \end{forest}
    }
    \caption{Taxonomy of LLM Inference Engines and Optimizations}
    \label{fig:lit_surv}
\end{figure}

\news{
This study provides a comprehensive analysis not only of previously examined inference engines such as vLLM~\cite{kwon2023efficient}, llama.cpp~\cite{llamacpp}, MLC-LLM~\cite{mlcllm}, and Sarathi-Serve~\cite{agrawal2024taming}, but also of recently released frameworks including MAX~\cite{max}, LitGPT~\cite{litgpt}, and vAttention~\cite{prabhu2025vattention}. In contrast to previous work, which has mainly focused on presenting optimization techniques offered by each engine, we also address practical indicators such as ecosystem maturity of open-source projects and the ease of installation and deployment. Furthermore, we conduct a comparative analysis of each inference engine from the perspectives of throughput-aware, latency-aware, and scalability design, thereby presenting selection criteria suitable for real-world service environments.}

The goal is to offer practical insights for researchers and engineers who need to build or operate high-performance, cost-efficient LLM services.

As shown in Fig.~\ref{fig:lit_surv}, this paper systematically organizes the major LLM inference engines and their respective optimization methods. 
Section~\ref{sec:background} outlines the core aspects of decoder-based transformer architectures, attention mechanisms, and the standard LLM inference process. 
Section~\ref{sec:practical_guide} presents a comprehensive review of the leading LLM inference engines, including ecosystem, hardware and operating system (OS) support. In particular, commercial offerings are discussed to help readers find suitable solutions for their own service environments and deployment objectives. To this end, we analyzed various aspects of inference engines, including their ecosystem, usability, as well as their support for hardware and platforms across both edge and server environments.
Section~\ref{sec:detailed_review} offers a detailed discussion of the architectures of various LLM inference engines and the inference-specific optimization features offered by each engine. 
Section~\ref{sec:inference_optimization} classifies fundamental inference optimization techniques found in current inference engines---covering batch optimization (\S~\ref{sec:inference_optimization_batch}), parallelization (\S~\ref{sec:inference_optimization_parallelism}), model compression (\S~\ref{sec:inference_optimization_compression}), fine-tuning (\S~\ref{sec:inference_optimization_fine_tuning}), caching (\S~\ref{sec:inference_optimization_caching}), attention optimization (\S~\ref{sec:inference_optimization_attention}), sampling optimization (\S~\ref{sec:inference_optimization_sampling}) and structured outputs (\S~\ref{sec:inference_optimization_structured_outputs})---while also examining emerging trends. By synthesizing these techniques, the chapter helps readers choose the inference engine that best matches their service requirements.
Based on these discussions, Section~\ref{sec:future_direction} explores future directions and major challenges in the development of LLM inference engines. Specifically, we examine the ongoing evolution of LLMs and how inference engines accommodate these changes, with particular attention to security and compatibility across diverse hardware platforms. We present perspectives on multiple aspects, including inference engine optimization strategies, security for inference and support for diverse hardware platforms and architectures.
Finally, section~\ref{sec:conclusion} concludes the paper.

\section{Backgrounds} \label{sec:background}

To enhance the efficiency of LLM inference, it is crucial not only to select a model suited to the domain but also to choose and optimize an appropriate inference engine while taking a diverse approach to overall development. This section examines LLMs from the perspective of inference, demonstrating how tasks such as model compression and deployment strategies can be seamlessly integrated with inference engines to achieve fast and cost-effective services.

First, we review the decoder-only transformer architecture, along with various attention mechanisms and considerations related to efficient inference. Second, we explain the inference process, focusing on the prefill and decode phases, and highlight corresponding optimization techniques from the inference engine perspective. Finally, we combine these elements to provide a comprehensive overview of the entire pipeline for inference and service deployment.

\subsection{Structure and Optimizations of LLMs} \label{sec:background_strucure}

\textbf{LLM Architecture Types.}
LLMs can be broadly categorized into three types based on the Transformer architecture~\cite{vaswani2017attention}: decoder-only, encoder-decoder, and encoder-only models. The encoder-decoder model first encodes the entire input and then uses the decoder with cross-attention at each step, which leads to higher memory usage and more complex procedures during inference. Encoder-only models are suitable for tasks like classification or retrieval, but since they are optimized for one-time inference, they are not ideal for token-by-token generation.

In contrast, decoder-only models have a simpler structure and are widely adopted in recent LLMs due to their strong zero-shot performance through autoregressive training~\cite{wang2022language, wei2022emergent}. Therefore, this paper mainly focuses on the decoder-only architecture.

\begin{figure}[tbp]
    \centering
    \resizebox{0.95\textwidth}{!}{%
    \begin{minipage}[b]{0.69\textwidth}
        \centering
        \includegraphics[width=\linewidth]{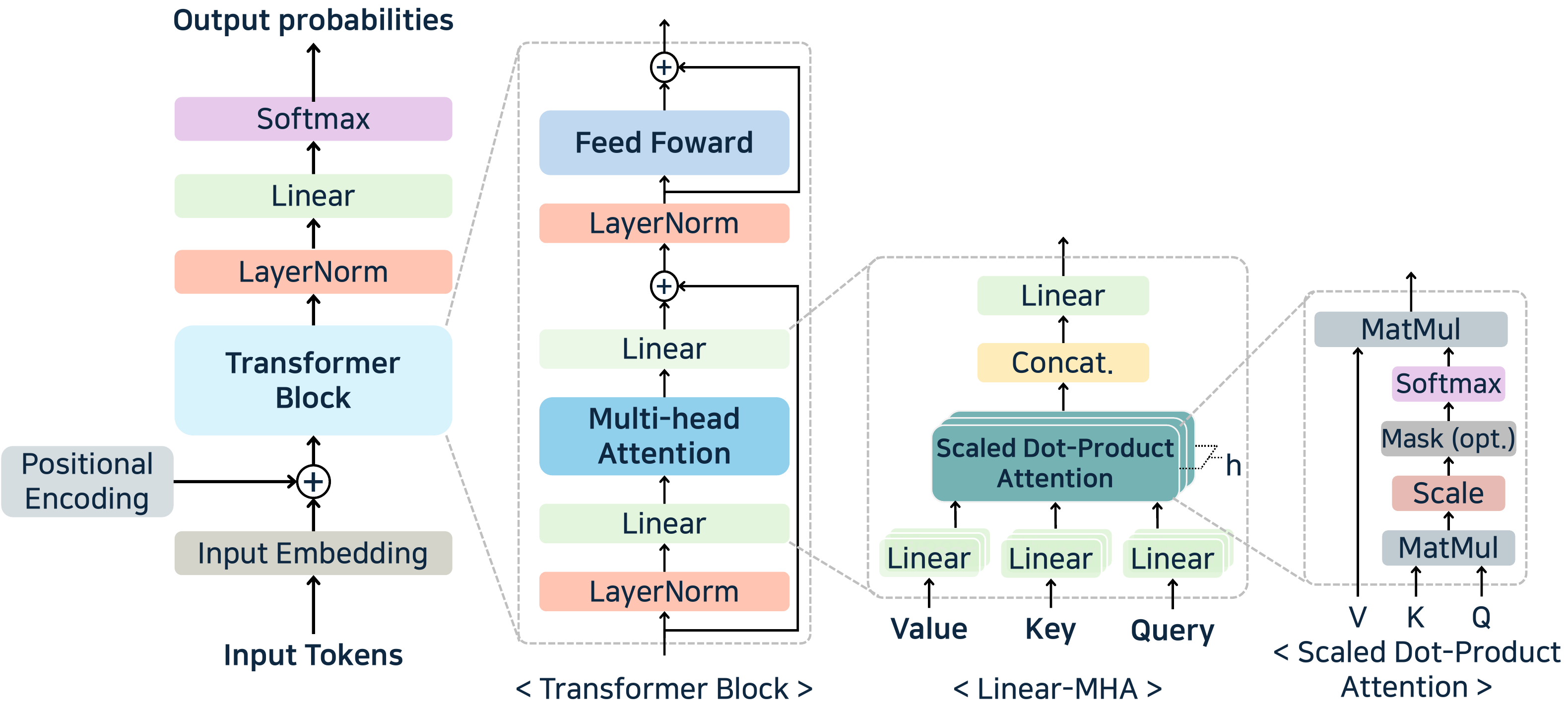}
        \captionof{figure}{Overview of Decoder-only Transformer Architecture}
        \label{fig:transformer}
    \end{minipage}
    \hspace{0.03\textwidth}
    \begin{minipage}[b]{0.31\textwidth}
        \centering
        \includegraphics[width=\linewidth]{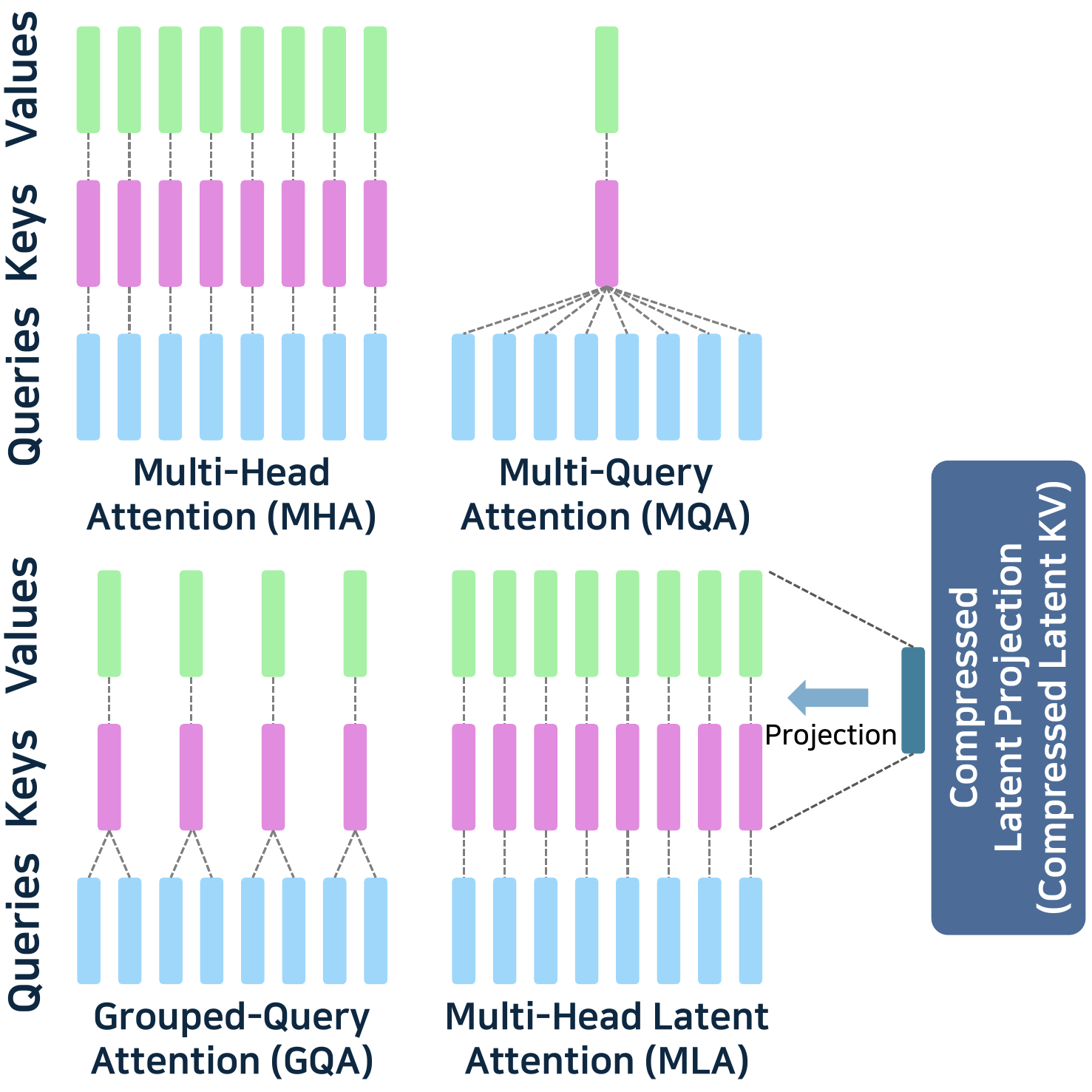}
        \captionof{figure}{Attention Mechanism}
        \label{fig:attention}
    \end{minipage}
    }
\end{figure}

\textbf{Standard Decoder-Only Architecture.}
Fig.~\ref{fig:transformer} shows the architecture of a decoder-only transformer. When a text input is received, it is first tokenized and then converted to high-dimensional vectors by an embedding layer. At this stage, positional encoding is added to incorporate token order. The resulting embeddings pass through several transformer blocks, each comprising Multi-Head Attention (MHA), a Feed-Forward Network (FFN), and residual connections. The MHA layer splits the input into Query ($\mathbf{Q}$), Key ($\mathbf{K}$), and Value ($\mathbf{V}$) vectors and performs scaled dot-product attention in parallel across multiple heads. In each head, $\mathbf{Q}-\mathbf{K}$ similarity scores are computed and applied to $\mathbf{V}$, aggregating the results. Causal masking ensures that only previously generated tokens are attended to, enabling autoregressive context learning. Next, the FFN layer refines the attention output by applying a linear transformation, expanding it to a higher-dimensional space, and using an activation function (e.g., ReLU~\cite{agarap2018deep}, GELU~\cite{hendrycks2016gaussian}, or SiLU~\cite{elfwing2018sigmoid}) before reducing it back to the original dimension. This sequence of operations increases the model's representational capacity. Both the MHA and FFN layers employ residual connections and layer normalization. Residual connections mitigate the vanishing gradients in deep networks~\cite{zhang2019improving}, and layer normalization keeps the output distributions stable, facilitating smoother training.

After the transformer block operations are finished, each input token produces a hidden state which is normalized and used to predict the next token in text generation. The hidden state is then passed through a linear layer, resulting in a logit vector over the vocabulary. Applying the softmax function converts these logits into a probability distribution, and the token with the highest probability is chosen as the next token. This procedure is repeated iteratively to generate the final text.

\textbf{Attention Structure Variants.}
In a standard transformer, MHA is employed, but recent modifications---like the one shown in Fig.~\ref{fig:attention}---have been introduced to improve inference efficiency. In MHA, each of the $N_h$ heads uses its own $\mathbf{Q, K, V}$ matrices, enabling the model to learn distinct subspace representations. However, increasing the number of heads also expands the size of the Key-Value ($\mathbf{KV}$) cache during inference, because all $\mathbf{K}$ and $\mathbf{V}$ values must be stored. To address this, Multi-Query Attention (MQA)~\cite{shazeer2019fast} was proposed. MQA retains multiple query heads while sharing a single set of $\mathbf{K}$ and $\mathbf{V}$ across all heads, thereby reducing the $\mathbf{KV}$ cache to roughly $1/N_h$ of what MHA requires. Although this approach may slightly reduce expressiveness, it significantly decreases memory usage. Grouped-Query Attention (GQA)~\cite{ainslie2023gqa} takes a middle ground by sharing K and V among head groups rather than across all heads. By tuning the number of groups ($N_g$), developers can strike a balance between memory efficiency and model performance. More recently, DeepSeek-v2~\cite{liu2024deepseek} introduced Multi-Head Latent Attention (MLA), which compresses $\mathbf{K}$ and $\mathbf{V}$ from multiple heads into a shared latent vector. This design further minimizes cache size while preserving accuracy. Because these alternative attention mechanisms alter the size and structure of the $\mathbf{KV}$ cache, inference engines must adapt accordingly. For instance, MQA and GQA require cache management that reflects shared $\mathbf{K}$ and $\mathbf{V}$, whereas MLA involves reconstructing compressed $\mathbf{K}$ and $\mathbf{V}$. As a result, the compatibility of an inference engine can vary depending on the attention structure of the model.

\textbf{Variants in Positional Embedding, Tokenization and Normalization.}
In LLMs, key architectural variants include the type of positional embedding, tokenizer choice, and the placement of normalization layers. Even well-known LLMs adopt different configurations: BLOOM~\cite{workshop2022bloom} uses Attention with Linear Biases (ALiBi)~\cite{press2021train}, while Llama~\cite{touvron2023llama, touvron2023llama2, grattafiori2024llama} and Mistral~\cite{jiang2023mistral} employ Rotary Position Embedding (RoPE)~\cite{su2024roformer}.
The selection of tokenizer also varies across models-GPT variants typically use Byte-Pair Encoding (BPE) tokenizers~\cite{zouhar2023formal}, whereas Llama~\cite{touvron2023llama, touvron2023llama2, grattafiori2024llama} and T5~\cite{raffel2020exploring} rely on SentencePiece-based unigram tokenizers~\cite{kudo2018sentencepiece}. Other architectural differences include the placement of normalization layers~\cite{rajaraman2024toward}.

\subsection{LLM Inference Process and Optimization} \label{sec:background_inference}

\begin{figure}[tbp]
    \centering
    \resizebox{0.99\textwidth}{!}{%
    \begin{minipage}[b]{0.66\textwidth}
        \centering
        \includegraphics[width=\linewidth]{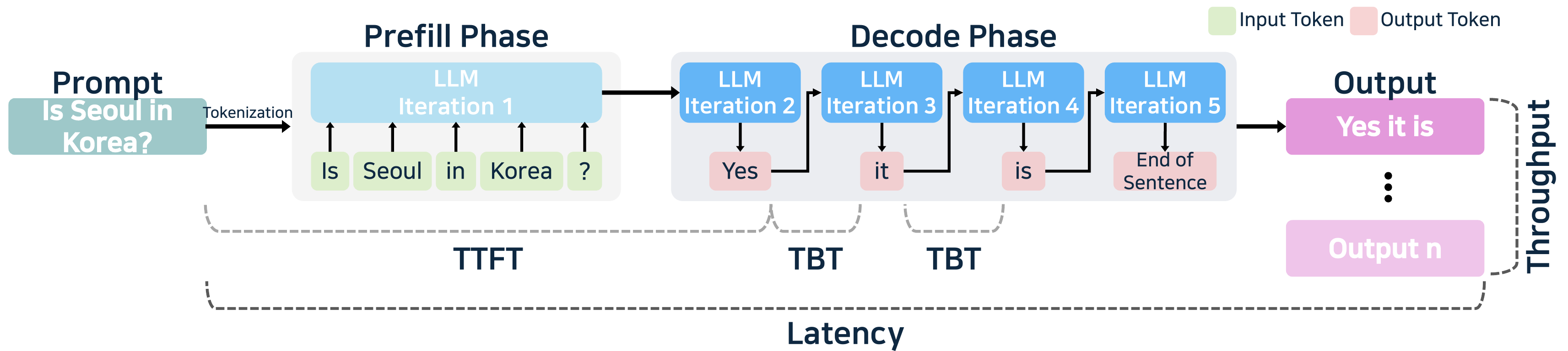}
        \captionof{figure}{LLM Inference Process}
        \label{fig:inference_preocess}
    \end{minipage}
    \hspace{0.01\textwidth}
    \begin{minipage}[b]{0.33\textwidth}
        \centering
        \includegraphics[width=\linewidth]{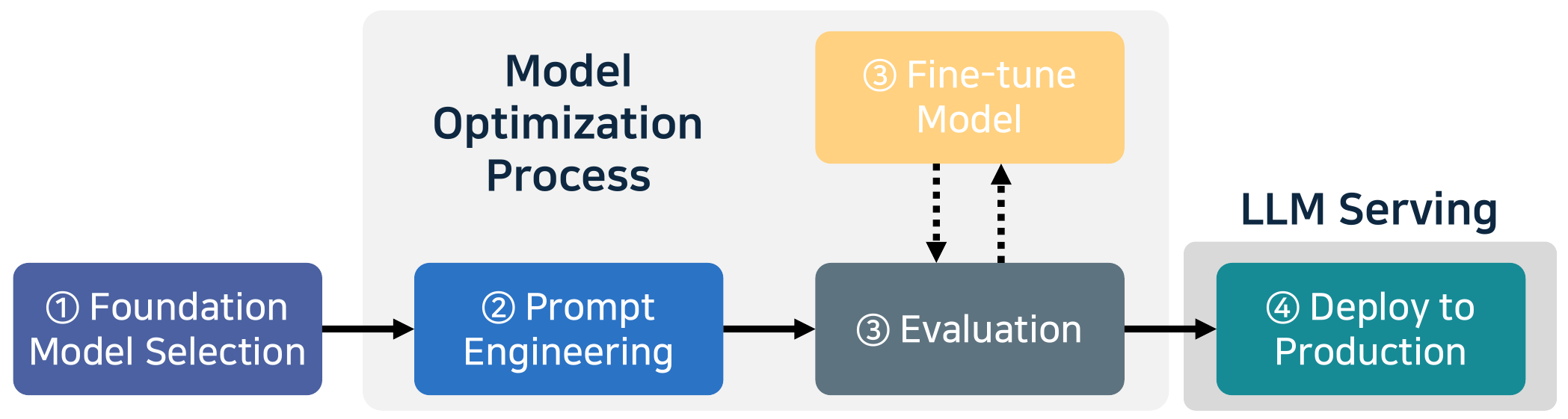}
        \captionof{figure}{Inference and Serving Process of LLM}
        \label{fig:process}
    \end{minipage}
    }
\end{figure}

LLM inference proceeds by tokenizing the user's input text and generating subsequent tokens until a stopping criterion (e.g., token limit or end-of-sequence (EOS) command) is met. As shown in Fig.~\ref{fig:inference_preocess}, this process comprises two main phases: the prefill phase, which generates the first token based on the input, and the decode phase, which sequentially produces the remaining tokens.

\textbf{Prefill phase.}
This phase processes the input text to compute the hidden state of the last token for each sample, thereby capturing the contextual and semantic meaning of the input. In decoder-only transformer models, this phase involves tokenization, embedding, and transformer block computations. Attention and FFN operations are performed on all input tokens. In this step, attention scales approximately with the square of the sequence length n ($\mathcal{O}(n^2)$), and the complexity of FFN increases with the size of the intermediate layer, resulting in large-scale array-to-array computations. $\mathbf{Q, K,}$ and $\mathbf{V}$---used to capture relationships among all input tokens---are generated immediately, loading substantial data into memory for intensive computation.

For example, as shown in Fig.~\ref{fig:inference_preocess}, if the user input is \texttt{Is Seoul in Korea?}, it is tokenized into [\texttt{Is}, \texttt{Seoul}, \texttt{in}, \texttt{Korea}, \texttt{?}], mapped to a unique token value such as [101, 4523, 1102, 2342, 63]. Position embeddings are then applied to these token IDs, converting them into high-dimensional vectors (e.g., [[0.1, 0.2, ...], [0.3, 0.5, ...], ...]). These vectors undergo attention and FFN computations, allowing the model to learn contextual relationships among tokens and refine their representations. Finally, the hidden state of the last token (\texttt{?}) is stored for use in the decode phase, where it guides the generation of subsequent tokens.

\textbf{Decode phase.}
This phase iteratively generates new tokens based on the hidden state computed during the prefill phase, following an autoregressive process in which only one token is produced at a time. In this phase, the final hidden state of the transformer block is passed through a linear transformation and a softmax function, which yields a probability distribution over the vocabulary. The token with the highest probability is selected and appended to the input sequence. Throughout this process, $\mathbf{K}$ and $\mathbf{V}$, as well as the input and output tokens, are stored on the GPU, system memory, or cache. Because $\mathbf{K}$ and $\mathbf{V}$ must be accessed and updated repeatedly, the decode phase often becomes a limitation of the memory bandwidth. Although the attention computation resembles that of the prefill phase, frequent reference to previously generated tokens increases latency, and the data to be accessed grows linearly with the sequence length.

Specifically, during the decode phase, the hidden state of the last token (\texttt{?}) saved in the prefill phase is used to predict the next token. For example, when the attention mechanism processes the newly generated token, the transformer block produces a final hidden state that is then linearly projected into a logit vector over the vocabulary. After softmax is applied, the most probable word---say, \texttt{Yes}---is chosen and appended to the existing sequence. Repeating this procedure can generate a full response, such as \texttt{Yes it is}, as shown in Fig.~\ref{fig:inference_preocess}.

\begin{table}[tbp]
    \caption{Key metrics of LLM performance}
    \label{tab:metrics}
    \centering
    \resizebox{.9\textwidth}{!}{%
    \begin{tabular}{@{}p{0.25\textwidth} p{0.3\textwidth} p{0.35\textwidth} p{0.45\textwidth}@{}}
    \toprule
    \multicolumn{1}{c}{Metric} & \multicolumn{1}{c}{Definition} & \multicolumn{1}{c}{User Perspective} & \multicolumn{1}{c}{Optimization Technique}
        \\ \midrule
        Time-to-first-token (TTFT) & 
        Time taken for the model to generate the first token  & 
        Most directly impacts the user's perception of response speed &
        Batching (\S\ref{sec:inference_optimization}), Kernel fusion, Prompt caching (\S\ref{sec:inference_optimization_caching_prompt}), Speculative decoding (\S\ref{sec:inference_optimization_sampling}) \\
        
        Time-between-tokens (TBT)  & 
        Time interval between each token  &
        Reflects the speed at which subsequent tokens are generated &
        KV caching (\S\ref{sec:inference_optimization_caching_kv}), Kernel fusion, Attention optimization (\S\ref{sec:inference_optimization_attention}), Quantization (\S\ref{sec:inference_optimization_quantization}) \\
        
        End-to-end latency  & 
        Total time from client request to complete response &
        Reflects overall response time and user experience &
        Batching (\S\ref{sec:inference_optimization}), KV caching (\S\ref{sec:inference_optimization_caching_kv}), Pruning (\S\ref{sec:inference_optimization_pruning}), Speculative decoding (\S\ref{sec:inference_optimization_sampling}), FlashAttention (\S\ref{sec:inference_optimization_attention_flash}) \\
        
        Throughput  & 
        Number of tokens processed per unit time  & 
        Represents the system's processing capacity &
        Batching (\S\ref{sec:inference_optimization}), Prefill optimization (\S\ref{sec:inference_optimization_batch_chunked}), Parallelism (TP/PP) (\S\ref{sec:inference_optimization_parallelism}), Quantization (\S\ref{sec:inference_optimization_quantization})
        \\ \bottomrule
    \end{tabular}%
    }
\end{table}

\textbf{System terms.}
The performance terms of an LLM system are illustrated in Fig.~\ref{fig:inference_preocess} and the accompanying Table~\ref{tab:metrics}. 
Time-To-First-Token (TTFT) measures the time it takes to receive a user request to generate the first token. It is especially important for how fast the system feels to the user. Time-Between-Tokens (TBT) (or Inter-token Latencies (ITL)) refers to the time it takes to generate each following token. It is often described as Time Per Output Token (TPOT), which is the average token generation speed during decoding. In addition, end-to-end latency represents the total response time for a user query and can be calculated as: 
\text{Latency} = $TTFT~+~(TBT~\times~number~of~tokens)$)
While latency gives an overall measure of responsiveness, throughput shows how many user requests the system can handle at the same time.

From a phase-wise perspective, the prefill phase affects TTFT and the decode phase impacts TBT. The latency of the prefill phase increases with input length, but can be reduced using parallel computation. On the other hand, latency in the decode phase grows with the number of generated tokens and has a more direct impact on the user experience.

\textbf{Optimization.}
Taking these performance metrics into account, LLM inference engines employ various customized optimization techniques for the prefill and decode phase. Most engines use $\mathbf{KV}$ caching to avoid redundant computation during decoding by reusing cached context and computing new operations only for the latest token. Recently, techniques such as continuous batching~\cite{yu2022orca} and hybrid batching~\cite{kamath2024pod} have been introduced to further improve decode phase efficiency, group prefill, and decode operations from multiple requests to better use GPU resources.

In addition, many inference engines reduce per-token overhead during decoding through kernel fusion~\cite{filipovivc2015optimizing, sun2024much, zhao2025hape} and hardware-specific computation kernels. Kernel fusion consolidates operations---such as LayerNorm, matrix multiplication, and activation functions---into a single GPU kernel, which decreases memory access and kernel launch overhead.

Quantization~\cite{egashira2025exploiting} is another key optimization. By representing model parameters in 8-bit or 4-bit integers instead of 16-bit or 32-bit floating-point formats, memory usage and bandwidth demands drop, especially during decoding. Quantized models can cache more tokens and handle more concurrent requests on the same hardware, often boosting the computation speed.

In general, caching, batching, kernel optimization, and quantization are fundamental to optimizing token throughput and minimizing latency in LLM inference services. Providing robust support for these techniques within an inference engine is crucial for delivering high-quality, scalable LLM solutions.

\subsection{Inference-Aware LLM Serving Workflow} \label{sec:background_serving}

LLM development typically involves gathering training data, pretraining on a large corpus, and then aligning and evaluating the resulting model. For production, inference often relies on a pretrained foundation model~\cite{xu2024foundationsurvey}. 
This complete pipeline is commonly referred to as LLM Operations (LLMOps) and, as shown in Fig.~\ref{fig:process}, consists of four.

\blackcircle{1} \textbf{Model selection.} 
Selecting a model and an inference engine that match service-level requirements, performance needs, and available hardware is crucial for a successful LLM deployment. A model might be well suited to the target domain but incompatible with a specific inference engine---so both factors must be considered together. When choosing an inference engine, it is equally important to assess the expected user concurrency and service-level objectives (SLO), then select a solution capable of meeting the necessary latency and throughput goals. Ultimately, the design principles and implementation of the inference engine dictate achievable performance, ease of integration, and general ease of use.

\blackcircle{2} \textbf{Prompt engineering.}
This step involves optimizing how the model is prompted and deployed. 
Prompt design can significantly influence model performance, as developers carefully craft system messages and user prompts to ensure consistent, high-quality outputs. This practice is known as prompt engineering~\cite{ye2023prompt}, which directs the model to produce desired responses without requiring additional computation. For example, a well-structured system prompt can adjust the model tone or decrease inappropriate responses, reducing trial-and-error during inference and contributing to more stable operation. During development, prompt templates undergo iterative testing and revision so that, in production, the model achieves the intended output with minimal further tuning.

\blackcircle{3} \textbf{Evaluation and fine-tuning.}
When prompt design is completed, the model must be evaluated to verify if it achieves the required level of performance. If not, fine-tuning can be applied to enhance accuracy or domain-specific capabilities. For example, instruction tuning~\cite{zhang2023instruction} can train the model with instruction response datasets to increase accuracy or domain-related responses. Other techniques include prompt tuning~\cite{lester2021power}, which adds task-optimized vectors to input embeddings, and prefix tuning~\cite{vos2022towards}, which modifies the model by inserting trainable parameters into hidden states at all layers. If the model size exceeds the available hardware, quantization can be used to compress activations or weights. Post-training quantization~\cite{frantar2022gptq, xiao2023smoothquant, li2023fptq} is based on a calibration dataset to calculate scaling parameters, converting weights or activations to lower precision. Alternatively, quantization-aware training~\cite{chen2024efficientqat, liu2023llm} simulates quantized conditions during training, ensuring that the model retains accuracy despite low-precision weights.

\blackcircle{4} \textbf{Deployment.}
Once an LLM achieves the desired performance level after fine-tuning, it should be prepared for production deployment. A key decision at this point involves choosing between a cloud application programming interface (API) or on-premise hosting. Cloud APIs (i.e., external LLM services) offer quick setup and the flexibility to scale with changing workloads, but they depend on external infrastructure and may raise data privacy issues. Because each query traverses a network, latency increases, making cloud APIs potentially unsuitable for latency-critical use cases. However, hosting LLMs on-premise avoids these concerns and markedly reduces latency. Eliminating network overhead accelerates response times, and keeping data inside internal systems improves privacy. Additionally, on-premise hosting allows for fine-grained control over model parameters and hardware configurations. Although it can require substantial infrastructure investment, this approach may prove more economical for large-scale services. Given these factors, it is advantageous to consider the deployment methods early in development. Aligning the model to the intended inference environment (for example, by quantization~\cite{egashira2025exploiting} or KD~\cite{yang2024survey}) and selecting an appropriate inference engine from the outset can streamline the process.

\subsection{\news{Emerging Trends in LLM Inference}} \label{sec:emerging_trends}

Traditional CNN and Deep Neural Network (DNN) workloads assume fixed input sizes and regular computation graphs. Larger batch sizes and kernel tuning therefore deliver almost linear throughput gains, while techniques such as layer partitioning, offloading, and pipelining remain reliable ways to trim latency or boost throughput. LLM inference faces different constraints because it generates token-by-token text inference.

\begin{itemize}
\item \textbf{Step-by-step execution.} 
Prefill and decode run in separate phases, and each has its own bottlenecks.

\item  \textbf{KV cache.}
Long contexts require large memory capacity and high bandwidth to store and fetch key-value pairs.

\item \textbf{Real-time requirements.} 
Long CoT prompts~\cite{wei2022chain}, structured outputs~\cite{liu2024we}, and many concurrent requests often overlap, therefore, systems must trade off latency against stability.
\end{itemize}

Recently, LLMs have moved beyond short input and output pairs to tasks such as CoT~\cite{wei2022chain}, reasoning~\cite{kumar2025llm}, long context processing~\cite{tang2024quest}, structured output~\cite{liu2024we}, high concurrency and operation under strict energy and cost limits. This shift increases the need for inference engines and system-level optimizations that target the distinct bottlenecks of the prefill and decode phases. Workloads with long input contexts often reach memory and bandwidth limits during the prefill phase, while tasks such as mathematical problem solving or code generation, which produce long outputs, hit latency limits during the decode phase.

\begin{table}[tbp]
    \caption{\news{Comparison between CNN/DNN and LLM Inference Workloads}}
    \label{tab:cnn_llm_comparison}
    \centering
    \resizebox{.9\textwidth}{!}{%
    \begin{tabular}{@{}>{\centering\arraybackslash}p{0.35\textwidth} p{0.55\textwidth} p{0.55\textwidth}@{}}
    \toprule
    \multicolumn{1}{c}{\news{Category}} & \multicolumn{1}{c}{\news{CNN/DNN Workload}} & \multicolumn{1}{c}{\news{LLM Inference Workload}} \\
    \midrule
        \news{Input/Computation Graph}
        & \news{Fixed-size inputs, regular convolution graphs; favorable to large-batch scaling; partitioning and pipelining work reliably} 
        & \news{Token streaming-based autoregressive; heterogeneity between prefill-decode phases causes interference and bottleneck shifting} \\
        
        \news{Memory Characteristics}
        & \news{Feature map reuse and locality-centric; memory hierarchy design highly predictable} 
        & \news{Dominated by KV-cache capacity/bandwidth bottlenecks} \\
        
        \news{Latency/Bottlenecks}
        & \news{Compute-bound tendency; FLOPs utilization is a key indicator} 
        & \news{Memory-bound with TBT; heterogeneous bottlenecks coexist across stages} \\
        
        \news{Batching/Scheduling}
        & \news{Large batches yield near-linear throughput scaling; partitioning and scheduling alone are effective}
        & \news{Interference between prefill-decode in mixed traffic; preventing decode hotspots is critical} \\
        
        \news{Compression/Quantization}
        & \news{INT8/FP16 maximizes kernel efficiency; serving benefits are relatively intuitive} 
        & \news{Ultra-low precision (e.g. MXFP4, 1-bit) suffers from dequantization/kernel path overhead, offsetting gains} \\
        
        \news{Offloading/Communication}
        & \news{Predictable feature-map transfers; static pipelining effective} 
        & \news{GPU-CPU-Memory transfers fluctuate dynamically; static offloading risks idle time} \\
        
        \news{Energy/Operations}
        & \news{Cluster operations driven by throughput and latency metrics} 
        & \news{Simultaneous optimization of Perf/\$, Perf/W, and SLO; Dynamic Voltage and Frequency Scaling (DVFS) effective when leveraging repetitive structures} \\
        
        \news{Edge/Distributed}
        & \news{Partitioning-scheduling co-optimization effective in collaborative inference} 
        & \news{On-device speculative inference, collaborative sharding, migration require dynamic decision-making} \\
    \bottomrule
    \end{tabular}%
    }
\end{table}

Therefore, unlike CNNs, LLM inference cannot rely on high-performance kernels and large batch sizes. It needs engine-level methods such as phase separated batching and scheduling, hardware and software co-design, quantization, and adaptive offloading. A holistic approach that integrates web services, inference engines, and system infrastructure is also essential to efficiently handle many concurrent user requests. Table~\ref{tab:cnn_llm_comparison} compares the main differences between CNN and LLM inference workloads.

Recent trends in LLM inference include the following:

\begin{itemize}
\item \textbf{Spread of CoT and inference intensive workloads.} 
CoT~\cite{wei2022chain} improves the accuracy of complex problems by explicitly generating intermediate reasoning steps, which increases the fraction of workloads that are inference-heavy at decode time. As the explanation-implementation-verification-correction loop deepens, the number of output tokens grows substantially and makes the decode stage the dominant source of latency~\cite{cai2024medusa, fu2024break}.

\item \textbf{Long-context inference.}
Many workloads now need tens of thousands or even millions of tokens, as in legal review or large codebase analysis. Prefill attention grows quadratically with sequence length, while the key value cache demands more bandwidth and memory. Both factors sharply increase TTFT~\cite{tang2024quest}.

\item \textbf{Application-specific decoding.}
Because applications differ in their priorities among accuracy, latency, and cost, fixed decoding strategies are insufficient to consistently ensure quality across domains, motivating quality of service (QoS)-guided decoding policies that adapt search and verification budgets to SLOs and cost~\cite{kakolyris2024slo}. 

\item  \textbf{Increased concurrency and mixed workload}
In real-world service environments, a single model instance typically handles multiple sessions simultaneously with mixed workload (conversation, summarization, math, code), and the autoregressive nature produces dissimilar resource profiles for prefill versus decode that interfere with one another under naive batching~\cite{hu2025shuffleinfer}. 

\item  \textbf{Collaboration across heterogeneous devices.}
The execution environment for inference has expanded beyond a single cloud GPU to encompass multiple edge nodes and heterogeneous accelerators, requiring the co-optimization of batching, partitioning, and scheduling strategies under joint communication, computation, memory, and power constraints~\cite{dai2024joint}. 
\news{Recently, a disaggregated inference~\cite{hu2025shuffleinfer} approach has been introduced to improve efficiency by separating the prefill and decode phase based on device-specific computational and memory characteristics. For example, the prefill phase, which involves intensive large-scale matrix multiplications, is allocated to high-bandwidth GPUs, while the decode phase, which requires frequent token-wise cache access, is assigned to low-latency CPUs or devices with larger memory capacity. This configuration reduces overall latency and improves resource utilization.}

\item  \news{\textbf{Expansion to the MoE Model.}
As parameter size in large-scale LLMs rise from tens to hundreds of billions, dense architectures that turn on every parameter for each token drastically increase inference floating-point operations per second (FLOPs), memory, and communication cost. Mixture of Experts (MoE) models ease this burden by activating only the top-k experts per token, keeping large capacity while trimming computation. This sparse approach shifts memory and communication overhead, increases arithmetic intensity, and reduces inference latency, cost, and energy use~\cite{dai2024deepseekmoe, llama4}.}

\item  \news{\textbf{Extension toward Multi-Agent Environments.}
With the increasing sophistication of LLM applications, the traditional paradigm in which a single model handles all requests is rapidly evolving into a multi-agent environment~\cite{li2024survey}, where multiple models (agents) collaborate to solve complex tasks. As multiple agents operate simultaneously, the overall memory requirements increase sharply, creating new challenges in efficiently sharing and coordinating input/output data and $\mathbf{KV}$ caches among agents.}
\end{itemize}

Modern LLM inference engines should expose integrated, composable optimization techniques spanning algorithms, runtime, and batch management, guided by a latency/energy model and key performance indicator (KPI) targets including Perf/\$, Perf/W, Joule/request, and SLO miss rate, allowing principled policy selection under mixed workloads and heterogeneous resources~\cite{samsi2023words, kakolyris2024slo}.

\section{Practical Guides to Inference Engines}\label{sec:practical_guide}

This section offers practical guidance on choosing an LLM inference engine by examining several key aspects. First, we look at ecosystem maturity and sustainability signals, such as how the engine is developed, licensed, and supported by its community. Next, we discuss hardware compatibility and platform support, focusing on whether the engine targets edge devices or server environments. We then explore the design and pricing strategies of commercial inference engines, including cost considerations and memory usage. Finally, we present a hardware-aware categorization of LLM inference engines, comparing engines based on their target use (edge or server), device types, and performance goals.

\begin{table}[tbp]
    \centering
    \caption{Comparison of LLM Inference Engines}
    \label{tab:frameworks}
    \resizebox{.99\textwidth}{!}{%
    \begin{tabular}{l l c c c c c c c c c c}
    \toprule
    \multirow{2.4}{*}{Frameworks} & 
    \multirow{2.5}{*}{\makecell{Organization}} & 
    \multirow{2.5}{*}{\makecell{Release\\Date}} & 
    \multirow{2.5}{*}{\makecell{Open-Source\\Support$^\dagger$}} & 
    \multicolumn{3}{c}{GitHub (Sep. 2025)} & 
    \multirow{2.5}{*}{\makecell{Supported\\Models$^\ddagger$}} & 
    \multirow{2.4}{*}{\makecell{Docs$^\ast$}} & 
    \multicolumn{3}{c}{User Forum$^\ast$$^\ast$} \\
    \cmidrule(lr){5-7} \cmidrule(lr){10-12}
    & & & & \makecell{\# Stars \footnotesize{(Rate)}} & \makecell{Star} & \makecell{Commit} & & & S & F & M \\
    \midrule
        Ollama~\cite{ollama} & 
        Community \footnotesize(Ollama)& 
        Jun. 2023 & 
        \greencheck & 
        153.0K \footnotesize{(187.2)} & 
        \scalebox{0.03}{\input{plot/ollama_star}} & 
        \scalebox{0.03}{\input{plot/ollama_commit}}&
        \supportbar{160}{185} & 
        \orangecheck  & 
        \greencheck & \redxmark & \greencheck
        \\
        llama.cpp~\cite{llamacpp} & 
        Community \footnotesize(gml.ai) & 
        Mar. 2023 & 
        \greencheck & 
        86.6K \footnotesize{(101.2)}& 
        \scalebox{0.03}{\input{plot/llamacpp_star}} & 
        \scalebox{0.03}{\input{plot/llamacpp_commit}} &
        \supportbar{70}{185} & 
        \yellowcheck & 
        \redxmark & \redxmark & \redxmark   
        \\    
        vLLM~\cite{kwon2023efficient} & 
        Academic \footnotesize(vLLM Team)& 
        Feb. 2023 & 
        \greencheck & 
        58.3K \footnotesize{(61.2)} & 
        \scalebox{0.03}{\input{plot/vllm_star}} & 
        \scalebox{0.03}{\input{plot/vllm_commit}} &
        \supportbar{122}{185} & 
        \greencheck & 
        \greencheck & \greencheck & \greencheck
        \\  
        DeepSpeed-FastGen~\cite{holmes2024deepspeed} & 
        Big Tech \footnotesize(Microsoft) & 
        Nov. 2023 & 
        \greencheck & 
        40.1K \footnotesize{(50.0)} & 
        \scalebox{0.03}{\input{plot/deepspeed_star}} & 
        \scalebox{0.03}{\input{plot/deepspeed_commit}}&
        \supportbar{11}{185} &
        \greencheck & 
        \redxmark & \redxmark & \greencheck 
        \\      
        Unsloth~\cite{unsloth} & 
        Startup \footnotesize(unsloth AI) &  
        Nov. 2023 & 
        \bluetriangle & 
        45.6K \footnotesize{(69.2)} & 
        \scalebox{0.03}{\input{plot/unsloth_star}}  & 
        \scalebox{0.03}{\input{plot/unsloth_commit}} &
        \supportbar{24}{185} & 
        \yellowcheck & 
        \greencheck & \greencheck & \redxmark 
        \\    
        MAX~\cite{max} &
        Startup \footnotesize(Modular Inc.) & 
        Apr. 2023 & 
        \bluetriangle & 
        24.8K \footnotesize{(28.4)}& 
        \scalebox{0.03}{\input{plot/max_star}} & 
        \scalebox{0.03}{\input{plot/max_commit}} &
        \supportbar{524}{185} & 
        \orangecheck & 
        \greencheck & \greencheck & \greencheck 
        \\    
        MLC LLM~\cite{mlcllm} & 
        Community \footnotesize(MLC-AI) & 
        Apr. 2023 & 
        \greencheck & 
        21.4K \footnotesize{(24.5)} & 
        \scalebox{0.03}{\input{plot/mlcllm_star}} & 
        \scalebox{0.03}{\input{plot/mlcllm_commit}} &
        \supportbar{35}{185} & 
        \orangecheck & 
        \greencheck & \redxmark & \redxmark 
        \\
        llama2.c~\cite{llama2c} & 
        Community \footnotesize(Andrej Karpathy) & 
        Jul. 2023 & 
        \greencheck & 
        18.8K \footnotesize{(23.8)} & 
        \scalebox{0.03}{\input{plot/llama2c_star}} & 
        \scalebox{0.03}{\input{plot/llama2c_commit}} &
        \supportbar{4}{185} & 
        \redxmark & 
        \greencheck & \redxmark & \redxmark 
        \\
        bitnet.cpp~\cite{wang20241} & 
        Big Tech \footnotesize(Microsoft) & 
        Oct. 2024 & 
        \greencheck & 
        22.0K \footnotesize{(53.9)} & 
        \scalebox{0.03}{\input{plot/bitnetcpp_star}} & 
        \scalebox{0.03}{\input{plot/bitnetcpp_commit}} &
        \supportbar{4}{185} & 
        \redxmark & 
        \redxmark & \redxmark & \redxmark 
        \\
        SGLang~\cite{zheng2024sglang} & 
        Academic \footnotesize(SGLang Team) & 
        Jan. 2024 & 
        \greencheck & 
        18.0K \footnotesize{(29.1)} &
        \scalebox{0.03}{\input{plot/sglang_star}} & 
        \scalebox{0.03}{\input{plot/sglang_commit}} &
        \supportbar{44}{185}  & 
        \orangecheck & 
        \greencheck & \redxmark & \greencheck 
        \\
        LitGPT~\cite{litgpt} & 
        Startup \footnotesize(Lightning AI) & 
        Jun. 2024 & 
        \greencheck & 
        12.8K \footnotesize{(14.7)} &
        \scalebox{0.03}{\input{plot/litgpt_star}} & 
        \scalebox{0.03}{\input{plot/lightllm_commit}} &
        \supportbar{38}{185}  & 
        \yellowcheck & 
        \greencheck & \redxmark & \greencheck 
        \\
        OpenLLM~\cite{openllm} & 
        Startup \footnotesize(BentoML) & 
        Apr. 2023 & 
        \bluetriangle & 
        11.8K \footnotesize{(13.3)} & 
        \scalebox{0.03}{\input{plot/openllm_star}} & 
        \scalebox{0.03}{\input{plot/openllm_commit}} &
        \supportbar{18}{185} & 
        \redxmark & 
        \greencheck & \redxmark & \redxmark 
        \\
        TensorRT-LLM~\cite{tensorrtllm} &
        Big Tech \footnotesize(NVIDIA) &
        Aug. 2023 & 
        \bluetriangle & 
        11.6K \footnotesize{(15.2)} & 
        \scalebox{0.03}{\input{plot/trt_star}} & 
        \scalebox{0.03}{\input{plot/trt_commit}}  &
        \supportbar{76}{185} & 
        \greencheck & 
        \redxmark & \greencheck & \greencheck
        \\
        TGI~\cite{tgi} & 
        Startup \footnotesize(Hugging Face) & 
        Oct. 2022 & 
        \greencheck & 
        10.5K \footnotesize{(9.8)} & 
        \scalebox{0.03}{\input{plot/tgi_star}}  & 
        \scalebox{0.03}{\input{plot/tgi_commit}} &
        \supportbar{35}{185} & 
        \orangecheck & 
        \redxmark & \greencheck & \redxmark
        \\
        PowerInfer~\cite{song2024powerinfer} &
        Academic \footnotesize(SJTU-IPADS) &
        Dec. 2023 & 
        \greencheck &
        8.3K \footnotesize{(13.0)} & 
        \scalebox{0.03}{\input{plot/powerinfer_star}}  & 
        \scalebox{0.03}{\input{plot/powerinfer_commit}} &
        \supportbar{16}{185} & 
        \redxmark & 
        \redxmark & \redxmark & \redxmark 
        \\
        LMDeploy~\cite{2023lmdeploy} &
        Startup \footnotesize(MMRazor/MMDeploy) & 
        Jun. 2023 & 
        \greencheck & 
        7.1K \footnotesize{(8.6)} & 
        \scalebox{0.03}{\input{plot/lmdeploy_star}}  & 
        \scalebox{0.03}{\input{plot/lmdeploy_commit}} &
        \supportbar{59}{185} 
        & 
        \orangecheck & 
        \greencheck & \redxmark & \redxmark
        \\
        LightLLM~\cite{lightllm} & 
        Academic \footnotesize(Lightllm Team) & 
        Jul. 2023 &
        \greencheck & 
        3.6K \footnotesize{(4.6)} &
        \scalebox{0.03}{\input{plot/lightllm_star}} & 
        \scalebox{0.03}{\input{plot/lightllm_commit}} &
        \supportbar{20}{185} & 
        \orangecheck &
        \greencheck & \redxmark & \redxmark 
        \\
        NanoFlow~\cite{zhu2024nanoflow} & 
        Academic \footnotesize(UW Efeslab) & 
        Aug. 2024 & 
        \greencheck & 
        0.8K \footnotesize{(2.3)} & 
        \scalebox{0.03}{\input{plot/nanoflow_star}} & 
        \scalebox{0.03}{\input{plot/nanoflow_commit}}  &
        \supportbar{6}{185} & 
        \redxmark & 
        \redxmark & \redxmark & \redxmark 
        \\
        DistServe~\cite{zhong2024distserve} & 
        Academic \footnotesize (PKU) & 
        Jan. 2024 & 
        \greencheck &
        0.6K \footnotesize{(1.1)} & 
        \scalebox{0.03}{\input{plot/distserve_star}}  & 
        \scalebox{0.03}{\input{plot/distserve_commit}} &
        \supportbar{3}{185} & 
        \redxmark &
        \redxmark & \redxmark & \redxmark 
        \\
        vAttention~\cite{prabhu2025vattention} & 
        Big Tech \footnotesize(Microsoft) & 
        May. 2024 & 
        \greencheck & 
        0.4K \footnotesize{(0.8)} & 
        \scalebox{0.03}{\input{plot/vattention_star}} & 
        \scalebox{0.03}{\input{plot/vattention_commit}} &
        \supportbar{3}{185} & 
        \redxmark &
        \redxmark & \redxmark & \redxmark 
        \\
        Sarathi-Serve~\cite{agrawal2023sarathi} & 
        Big Tech \footnotesize(Microsoft) &
        Nov. 2023 &
        \greencheck & 
        0.4K \footnotesize{(0.6)} &
        \scalebox{0.03}{\input{plot/sarathiserve_star}} & 
        \scalebox{0.03}{\input{plot/sarathiserve_commit}} &
        \supportbar{8}{185} & 
        \redxmark &
        \redxmark & \redxmark & \redxmark 
        \\
        \midrule
        Friendli Inference~\cite{friendli} & 
        Startup \footnotesize(FriendliAI Inc.) & 
        Nov. 2023 & 
        \redxmark & 
        -- & 
        -- & 
        -- & 
        \supportbar{185}{185} &
        \yellowcheck & 
        \redxmark & \redxmark & \greencheck 
        \\
        Fireworks AI~\cite{fireworks} &
        Startup \footnotesize(Fireworks AI, Inc.) & 
        Jul. 2023 &
        \redxmark &
        -- & 
        -- &
        -- & 
        \supportbar{185}{185} &
        \yellowcheck &
        \greencheck & \redxmark & \redxmark 
        \\
        GroqCloud~\cite{groqcloud} & 
        Startup \footnotesize(Groq Inc.) & 
        Feb. 2024 & 
        \redxmark & 
        -- & 
        -- & 
        -- & 
        \supportbar{18}{185} & 
        \redxmark & 
        \greencheck & \redxmark & \greencheck 
        \\
        Together Inference~\cite{together} & 
        Startup \footnotesize(together.ai) & 
        Nov. 2023 &
        \redxmark & 
        -- &
        -- &
        -- & 
        \supportbar{141}{185} &
        \yellowcheck & 
        \greencheck & \redxmark & \redxmark 
        \\
    \bottomrule
    \end{tabular}%
    }
    \flushleft
    \hspace{1mm} \tiny{$^\dagger$\bluetriangle indicates partial open-source support, }
    \hspace{0mm} \tiny{$^\ddagger$ Each square represents 50 models (Sep. 2025) }
    \\
    \hspace{1mm} \tiny{$^\ast$ Indicates the level of detail of the document (\yellowcheck: Simple, \orangecheck: Moderate, \greencheck: Detail), }
    \\
    \hspace{1mm} \tiny{$^\ast$$^\ast$S refers for social networking services (Discord/Slack), F refers for discussion forums (private forums/reddit), and M refers for meetups}

\end{table}

\newcommand{\spiderchart}[8]{%
\begin{subfigure}[b]{0.19\linewidth}
\centering
\vspace{1em}
\begin{tikzpicture}[scale=0.85]
  \foreach \r in {0.2,0.4,0.6,0.8,1.0} {
    \draw[gray!20] (0:\r) -- (60:\r) -- (120:\r) -- (180:\r) -- (240:\r) -- (300:\r) -- cycle;
  }
  \foreach \a in {0,60,120,180,240,300} {
    \draw[gray!40] (0,0) -- (\a:1);
  }
  \node[font=\tiny\sffamily, align=center, rotate=270] at (0:1.3)  {Ease-of-\\Use};
  \node[font=\tiny\sffamily, align=center] at (60:1.35)   {Ease-of-\\Deploy};
  \node[font=\tiny\sffamily, align=center] at (120:1.35)  {General-\\Purpose};
  \node[font=\tiny\sffamily, align=center, rotate=90] at (180:1.2)  {Scalability};
  \node[font=\tiny\sffamily, align=center] at (240:1.35)  {Throughput-\\Aware};
  \node[font=\tiny\sffamily, align=center] at (300:1.35)  {Latency-\\Aware};
  \draw[thick, #8, fill=#8!30, opacity=0.7]
    (0:#1) -- (60:#2) -- (120:#3) --
    (180:#4) -- (240:#5) -- (300:#6) -- cycle;
\end{tikzpicture}
\caption{\scriptsize #7}
\end{subfigure}
}

\begin{figure}[tbp]
    \centering
    \resizebox{0.99\textwidth}{!}{%
    \begin{minipage}{\textwidth}
        \centering
        \spiderchart{0.67}{1.00}{0.54}{0.43}{0.50}{0.50}{Ollama}{gray!70!blue}
        \spiderchart{0.27}{1.00}{0.59}{0.29}{0.67}{0.67}{llama.cpp}{gray!70!blue}
        \spiderchart{1.00}{1.00}{0.69}{1.00}{0.83}{0.67}{vLLM}{orange!90!black}
        \spiderchart{0.87}{1.00}{0.43}{1.00}{0.83}{0.50}{DeepSpeed-FastGen}{orange!90!black}
        \spiderchart{0.40}{0.50}{0.14}{0.29}{0.17}{0.33}{Unsloth}{teal!70!black}
        \spiderchart{0.73}{0.75}{0.64}{0.71}{0.67}{0.67}{MAX}{gray!70!blue}
        \spiderchart{0.60}{0.75}{0.39}{0.14}{0.83}{0.50}{MLC LLM}{gray!70!blue}
        \spiderchart{0.07}{0.25}{0.12}{0.14}{0.17}{0.17}{llama2.c}{teal!70!black}
        \spiderchart{0.00}{0.50}{0.12}{0.14}{0.17}{0.17}{bitnet.cpp}{teal!70!black}
        \spiderchart{0.67}{1.00}{0.39}{1.00}{0.67}{0.50}{SGLang}{orange!90!black}
        \spiderchart{0.47}{0.50}{0.27}{0.71}{0.50}{0.67}{LitGPT}{orange!90!black}
        \spiderchart{0.07}{0.50}{0.07}{0.43}{0.33}{0.17}{OpenLLM}{teal!70!black}
        \spiderchart{0.93}{1.00}{0.18}{1.00}{1.00}{0.67}{TensorRT-LLM}{purple!80!black}
        \spiderchart{0.60}{1.00}{0.51}{0.71}{0.50}{0.67}{TGI}{gray!70!blue}
        \spiderchart{0.00}{0.25}{0.25}{0.29}{0.50}{0.67}{PowerInfer}{gray!70!blue}
        \spiderchart{0.60}{1.00}{0.29}{1.00}{0.67}{0.33}{LMDeploy}{orange!90!black}
        \spiderchart{0.60}{0.75}{0.14}{0.57}{0.67}{0.50}{LightLLM}{teal!70!black}
        \spiderchart{0.00}{0.25}{0.12}{0.43}{0.50}{0.50}{NanoFlow}{teal!70!black}
        \spiderchart{0.00}{0.25}{0.06}{1.00}{0.83}{0.67}{DistServe}{purple!80!black}
        \spiderchart{0.00}{0.25}{0.12}{0.43}{0.67}{0.67}{vAttention}{teal!70!black}
        \spiderchart{0.00}{0.25}{0.07}{0.43}{0.50}{0.33}{Sarathi-Serve}{teal!70!black}
        \spiderchart{0.33}{1.00}{0.21}{0.57}{0.67}{0.33}{Friendli Inference}{teal!70!black}
        \spiderchart{0.33}{1.00}{0.26}{1.00}{0.33}{0.50}{Fireworks AI}{orange!90!black}
        \spiderchart{0.13}{1.00}{0.07}{0.86}{0.50}{0.17}{GroqCloud}{purple!80!black}
        \spiderchart{0.33}{1.00}{0.17}{1.00}{0.17}{0.33}{Together Inference}{orange!90!black}
    \end{minipage}
    }
    \caption{Representative characteristics comparison of LLM inference engines across six dimensions: model generality, ease of deployment and use, latency and throughput optimization, and scalability}
    \label{fig:inference_engine_spider_graph}
\end{figure}

\begin{table}[tbp]
    \centering
    \caption{Hardware Features of LLM Inference Engines}
    \label{tab:frameworks_hardware}
    \resizebox{.99\textwidth}{!}{%
    \begin{tabular}{@{}l cccc | cc ccc ccccc c c@{}}
    \toprule
    \multirow{4.5}{*}{Engines} &
      \multicolumn{4}{c|}{Supported Platform} &
      \multicolumn{2}{c}{CPU} &
      \multicolumn{3}{c}{GPU} &
      \multicolumn{5}{c}{AI Accelerators} &
      \multicolumn{1}{c}{\multirow{4.5}{*}{\begin{tabular}[c]{@{}c@{}}Mobile/\\ Edge\end{tabular}}} &
      \multicolumn{1}{c}{\multirow{4.5}{*}{ETC.}} \\ 
      \cmidrule(lr){2-5} \cmidrule(lr){6-7} \cmidrule(lr){8-10} \cmidrule(lr){11-15}
      &
      \multicolumn{1}{c}{Linux} &
      \multicolumn{1}{c}{Windows} &
      \multicolumn{1}{c}{macOS} &
      \multicolumn{1}{c|}{\begin{tabular}[c]{@{}c@{}}Web/\\ API\end{tabular}} &
      \multicolumn{1}{c}{x86-64} &
      \multicolumn{1}{c}{\begin{tabular}[c]{@{}c@{}}ARM/\\ Apple Silicon\\ \footnotesize{(Vulkan, Metal)}\end{tabular}} &
      \multicolumn{1}{c}{\begin{tabular}[c]{@{}c@{}}NVIDIA\\ \footnotesize{(CUDA)}\end{tabular}} &
      \multicolumn{1}{c}{\begin{tabular}[c]{@{}c@{}}AMD\\ \footnotesize{(ROCm, HIP)}\end{tabular}} &
      \multicolumn{1}{c}{\begin{tabular}[c]{@{}c@{}}Intel \\ \footnotesize{(SYCL)}\end{tabular}} &
      \multicolumn{1}{c}{\begin{tabular}[c]{@{}c@{}}Google \\ TPU\end{tabular}} &
      \multicolumn{1}{c}{\begin{tabular}[c]{@{}c@{}}AMD \\ Instinct\end{tabular}} &
      \multicolumn{1}{c}{\begin{tabular}[c]{@{}c@{}}Intel \\ Gaudi\end{tabular}} &
      \multicolumn{1}{c}{\begin{tabular}[c]{@{}c@{}}Huawei \\ Ascend\end{tabular}} &
      \multicolumn{1}{c}{\begin{tabular}[c]{@{}c@{}}AWS \\ Inferentia\end{tabular}} &
      \multicolumn{1}{c}{} &
      \multicolumn{1}{c}{} \\ 
      \midrule
        Ollama~\cite{ollama}
            & \greencheck & \greencheck & \greencheck & \redxmark 
            & \greencheck & \greencheck
            & \greencheck & \greencheck & \greencheck 
            & \redxmark & \greencheck & \redxmark & \redxmark & \redxmark 
            & \makecell{\greencheck\\[-4pt] { \tiny (NVIDIA Jetson)}} 
            & --   \\
        llama.cpp~\cite{llamacpp}          
            & \greencheck & \greencheck & \greencheck & \redxmark 
            & \greencheck & \greencheck 
            & \greencheck & \greencheck & \greencheck 
            & \redxmark & \greencheck & \redxmark & \greencheck & \redxmark 
            & \makecell{\greencheck\\ [-5pt]{ \tiny (Qualcomm Adreno)}} 
            & \makecell{\scriptsize{Moore Thread}\\[-4pt]  \scriptsize {MTT}} \\
        vLLM~\cite{kwon2023efficient}               
            & \greencheck & \redxmark & \redxmark & \redxmark 
            & \greencheck & \greencheck 
            & \greencheck & \greencheck & \greencheck 
            & \greencheck & \greencheck & \greencheck & \greencheck & \greencheck 
            & \makecell{\greencheck\\[-4pt] { \tiny (NVIDIA Jetson)}} 
            &  -- \\
        DeepSpeed-FastGen~\cite{holmes2024deepspeed} 
            & \greencheck & \greencheck & \redxmark & \redxmark 
            & \greencheck & \redxmark 
            & \greencheck & \redxmark & \greencheck 
            & \redxmark & \greencheck & \greencheck & \greencheck & \redxmark 
            & \redxmark 
            & \makecell{\scriptsize{Tecorigin}\\ [-4pt] \scriptsize{SDAA}} \\
        Unsloth~\cite{unsloth}            
            & \greencheck & \greencheck & \redxmark & \redxmark
            & \greencheck & \redxmark
            & \greencheck & \redxmark & \redxmark
            & \redxmark & \redxmark & \redxmark & \redxmark & \redxmark 
            & \redxmark 
            & -- \\
        MAX~\cite{max}               
            & \greencheck & \greencheck & \greencheck & \redxmark 
            & \greencheck & \greencheck 
            & \greencheck & \greencheck & \redxmark 
            & \redxmark & \redxmark & \redxmark & \redxmark & \redxmark  
            & \redxmark 
            & -- \\
        MLC LLM~\cite{mlcllm}            
            & \greencheck & \greencheck & \greencheck & \redxmark 
            & \greencheck & \greencheck 
            & \greencheck & \greencheck & \greencheck 
            & \redxmark & \redxmark & \redxmark & \redxmark & \redxmark 
            & \makecell{\greencheck\\[-5pt] \tiny {(Qualcomm Adreno,} \\[-5pt] \tiny {ARM Mali)}} 
            & -- \\
        llama2.c~\cite{llama2c}            
            & \greencheck & \greencheck & \greencheck & \redxmark 
            & \greencheck & \greencheck 
            & \redxmark & \redxmark & \redxmark 
            & \redxmark & \redxmark & \redxmark & \redxmark & \redxmark
            &  \redxmark 
            & --   \\
        bitnet.cpp~\cite{wang20241}             
            & \greencheck & \greencheck & \greencheck & \redxmark 
            & \greencheck & \greencheck 
            & \redxmark & \redxmark & \redxmark 
            & \redxmark & \redxmark  & \redxmark & \redxmark & \redxmark 
            & \redxmark
            & -- \\
        SGLang~\cite{zheng2024sglang}             
            & \greencheck & \redxmark & \redxmark & \redxmark 
            & \greencheck & \redxmark 
            & \greencheck & \redxmark & \greencheck 
            & \redxmark & \greencheck  & \greencheck & \redxmark & \redxmark 
            & \makecell{\greencheck\\[-4pt] { \tiny (NVIDIA Jetson)}} 
            & -- \\
         LitGPT~\cite{litgpt}             
           & \greencheck & \redxmark & \greencheck & \redxmark 
           & \greencheck & \redxmark 
           & \greencheck & \redxmark & \redxmark 
           & \greencheck & \greencheck  & \redxmark & \redxmark & \redxmark 
           & \redxmark 
           & -- \\
        OpenLLM~\cite{openllm}           
            & \greencheck & \redxmark & \redxmark & \redxmark
            & \redxmark & \redxmark 
            & \greencheck & \redxmark & \redxmark
            & \redxmark & \redxmark & \redxmark & \redxmark & \redxmark 
            & \redxmark 
            & -- \\
        TensorRT-LLM~\cite{tensorrtllm}       
            & \greencheck & \greencheck & \redxmark & \redxmark 
            & \redxmark & \redxmark 
            & \greencheck & \redxmark & \redxmark 
            & \redxmark & \redxmark & \redxmark & \redxmark & \redxmark  
            & \makecell{\greencheck\\ [-4pt]{ \tiny (NVIDIA Jetson)}} 
            & -- \\
        TGI~\cite{tgi}                
            & \greencheck & \redxmark & \redxmark & \redxmark 
            & \greencheck & \greencheck 
            & \greencheck & \redxmark & \greencheck 
            & \greencheck & \greencheck & \greencheck & \redxmark & \greencheck
            & \redxmark 
            & -- \\
        PowerInfer~\cite{xue2024powerinfer,song2024powerinfer}         
            & \greencheck & \greencheck & \greencheck & \redxmark 
            & \greencheck & \greencheck 
            & \greencheck & \greencheck & \redxmark 
            & \redxmark & \redxmark & \redxmark & \redxmark & \redxmark 
            & \makecell{\greencheck\\[-5pt] \tiny {(Qualcomm Snapdragon 8)}} 
            & -- \\
        LMDeploy~\cite{2023lmdeploy}           
            & \greencheck & \greencheck & \redxmark & \redxmark 
            & \greencheck & \redxmark 
            & \greencheck  & \redxmark & \redxmark 
            & \redxmark & \redxmark & \redxmark & \greencheck & \redxmark 
            & \makecell{\greencheck\\ [-4pt]{ \tiny (NVIDIA Jetson)}} 
            & -- \\
        LightLLM~\cite{lightllm}           
            & \greencheck & \redxmark & \redxmark & \redxmark 
            & \greencheck & \redxmark 
            & \greencheck & \redxmark & \redxmark 
            & \redxmark & \redxmark & \redxmark & \redxmark & \redxmark
            & \redxmark 
            & -- \\
        NanoFlow~\cite{zhu2024nanoflow}           
            & \greencheck & \redxmark & \redxmark & \redxmark
            & \greencheck & \redxmark 
            & \greencheck & \redxmark & \redxmark 
            & \redxmark & \redxmark & \redxmark & \redxmark & \redxmark 
            & \redxmark 
            & -- \\
        DistServe~\cite{zhong2024distserve}          
            & \greencheck & \redxmark & \redxmark & \redxmark 
            & \redxmark & \redxmark
            & \greencheck & \redxmark & \redxmark 
            & \redxmark & \redxmark & \redxmark & \redxmark & \redxmark 
            & \redxmark
            & --  \\
        vAttention~\cite{prabhu2025vattention}        
            & \greencheck & \redxmark & \redxmark & \redxmark 
            & \greencheck & \redxmark 
            & \greencheck & \redxmark & \redxmark
            & \redxmark & \redxmark & \redxmark & \redxmark & \redxmark
            & \redxmark 
            & -- \\
        Sarathi-Serve~\cite{agrawal2023sarathi}      
            & \greencheck & \redxmark & \redxmark & \redxmark 
            & \redxmark & \redxmark 
            & \greencheck & \redxmark & \redxmark 
            & \redxmark & \redxmark & \redxmark & \redxmark & \redxmark 
            & \redxmark 
            & -- \\
        Friendli Inference~\cite{friendli} 
            & \redxmark & \redxmark & \redxmark & \greencheck 
            & \redxmark & \redxmark 
            & \greencheck & \redxmark & \redxmark 
            & \redxmark & \redxmark & \redxmark & \redxmark & \redxmark 
            & \redxmark 
            & -- \\
        Fireworks AI~\cite{fireworks}       
            & \redxmark & \redxmark & \redxmark & \greencheck 
            & \redxmark & \redxmark 
            & \greencheck & \redxmark & \redxmark 
            & \redxmark & \greencheck & \redxmark & \redxmark & \redxmark 
            & \redxmark 
            & -- \\
        GroqCloud~\cite{groqcloud}          
            & \redxmark & \redxmark & \redxmark & \greencheck 
            & \redxmark & \redxmark 
            & \redxmark & \redxmark & \redxmark
            & \redxmark & \redxmark & \redxmark & \redxmark & \redxmark 
            & \redxmark 
            & \scriptsize{Groq LPU} \\
        Together Inference~\cite{together} 
            & \redxmark & \redxmark & \redxmark & \greencheck 
            & \redxmark & \redxmark 
            & \greencheck & \redxmark & \redxmark 
            & \redxmark & \redxmark & \redxmark & \redxmark & \redxmark
            & \redxmark 
            & -- \\ 
        \bottomrule
    \end{tabular}%
    }
\end{table}

Table~\ref{tab:frameworks} provides a summary of the LLM inference engines examined in this paper, and Fig.~\ref{fig:inference_engine_spider_graph} offers a visual representation of the characteristics of each engine. 
\textsf{\small{General-Purpose}} is a composite metric derived from the number of supported models in Table~\ref{tab:frameworks} and the range of hardware platforms in Table~\ref{tab:frameworks_hardware}. A higher score indicates broader compatibility with diverse models and hardware.

\textsf{\small{Ease-of-Deploy}} measures how easily an engine can be installed via the Python package installer (pip), Debian Advanced Package Tool (APT), Homebrew~\cite{homebrew}, customized through source builds, Docker~\cite{docker} or Conda~\cite{conda} environments or prebuilt binaries. A higher rating suggests simpler, faster installation and deployment. 

\textsf{\small{Ease-of-Use}} evaluates both documentation quality and user community activity level (as shown in Table~\ref{tab:frameworks}).

\textsf{{Latency-Aware}} and \textsf{\small{Throughput-Aware}} represent the each engine's support for latency- and throughput-specific optimization techniques, respectively, based on the metrics in Table~\ref{tab:metrics} (\S\ref{sec:background}) and the optimization features in Table~\ref{tab:frameworks_optimization} (\S\ref{sec:inference_optimization}). Higher values imply more robust capabilities to optimize in those areas.

Lastly, \textsf{\small{Scalability}} indicates how effectively an engine accommodates edge, server, and multi-node environments. Higher scores indicate suitability for large-scale LLM workloads.

For commercial inference engines, some metric scores may be lower because they rely on publicly available information.

By referring to Fig.~\ref{fig:inference_engine_spider_graph}, users can determine which LLM inference engine best matches their service needs and deployment settings.

\subsection{Ecosystem Maturity and Sustainability Signals} \label{sec:practical_guide_ecosystem}

This section discusses non-technical indicators related to the current status of LLM inference engines.
As shown in Table~\ref{tab:frameworks}, LLM inference engines can be categorized into open-source and closed-source commercial tools. For open-source tools, we analyze sustainability based on the types of development and maintenance organization, open software licenses, and the maturity of user support.

\textbf{Development and Maintenance Organizations.}
Open-source inference engines are mainly developed and maintained by big tech companies, startups, communities or academic institutions. Among the \textbf{21} inference engines surveyed, the number of inference engines by organization type is: \texttt{\small {Academic}} (\textbf{6}), \texttt{\small {Startup}} (\textbf{6}), \texttt{\small {Big Tech}} (\textbf{5}), and \texttt{\small {Community}} (\textbf{4}). While the difference is not large, this shows that LLM inference engines are being developed and maintained by a variety of organizations. 
Regardless of the organization, most open-source projects use permissive licenses such as MIT or Apache 2.0, making them easy to adopt and use.

Projects maintained by community groups may face challenges in long-term maintenance, which could limit the integration of new technologies. Some projects like Unsloth~\cite{unsloth}, MAX~\cite{max}, OpenLLM~\cite{openllm}, and TensorRT-LLM~\cite{tensorrtllm}, which are led by big tech or startups, only release parts of their source code.

While open-source engines are developed by diverse groups such as big tech, startups, communities, and academia, most commercial LLM inference engines are developed and run by startups. This is because startups can move quickly and develop specialized technologies to enter the market faster with differentiated, high-performance services.

\textbf{User Preference.}
We measured user preference for open-source LLM inference engines using GitHub statistics such as total stars, daily average growth rate, and star growth trends over time. In this study, we considered a project to be highly popular if it gained more than 25 stars per day on average. Projects like Ollama~\cite{ollama} (\textbf{187.2}), llama.cpp~\cite{llamacpp} (\textbf{101.2}), Unsloth~\cite{unsloth} (\textbf{69.2}), vLLM~\cite{kwon2023efficient} (\textbf{61.2}), bitnet.cpp~\cite{wang20241} (\textbf{53.9}) and DeepSpeed-FastGen~\cite{holmes2024deepspeed} (\textbf{50.0}) meet this criterion and have tens of thousands of total stars, indicating high interest and rapid adoption by the community.

On the other hand, some projects have a large number of total stars but show slower recent growth. For example, TGI~\cite{tgi}, TensorRT-LLM~\cite{tensorrtllm}, and OpenLLM~\cite{openllm} each have more than 10K stars, but their daily growth is below 25, and their growth curves are flat after an initial spike. This may suggest that they received attention early, but are now facing difficulties in maintaining community interest. Possible reasons include limited usability or closed ecosystems.

This kind of analysis helps estimate the future growth potential of projects, providing a long-term perspective when choosing engines for practical or research use.

\textbf{Ease of Use.}
We evaluate the user-friendliness of LLM inference engines based on the quality of documentation and the availability of user forums. Our analysis shows that top projects like vLLM~\cite{kwon2023efficient}, and DeepSpeed-FastGen~\cite{holmes2024deepspeed}, TensorRT-LLM~\cite{tensorrtllm} provide well-written documentation, and vLLM~\cite{kwon2023efficient} and MAX~\cite{max}, LitGPT~\cite{litgpt} have active community channels (e.g., Discord, forums), making onboarding and troubleshooting easier. This is closely related to their high star counts and rapid user adoption.

In contrast, projects such as bitnet.cpp~\cite{wang20241}, OpenLLM~\cite{openllm}, PowerInfer~\cite{xue2024powerinfer}, NanoFlow~\cite{zhu2024nanoflow}, DistServe~\cite{zhong2024distserve}, etc. have limited documentation or lack community channels. This is also reflected in their slow star growth, indicating a higher entry barrier for users. Projects with poor documentation and no forums tend to have lower popularity and slower growth. 

These results suggest that beyond technical performance, user support systems are important factors in engine selection and community growth.

\textbf{Development Activity.}
We evaluated the development activity of LLM inference engines based on GitHub commit trends and the number of supported models. By considering both indicators together, we achieved more reliable results than simply counting commits alone.
Projects such as llama.cpp~\cite{llamacpp}, vLLM~\cite{kwon2023efficient}, and DeepSpeed-FastGen~\cite{holmes2024deepspeed} show consistent and frequent updates in their commit histories, while also supporting a wide range of LLM models.
On the other hand, engines like TGI~\cite{tgi} and TensorRT-LLM~\cite{tensorrtllm}, which gained many stars early on, show relatively stagnant commit activity and limited model support. This may indicate lower flexibility for future feature extensions. 
In particular, projects such as OpenLLM~\cite{openllm} and PowerInfer~\cite{xue2024powerinfer}, which have a narrow range of supported models or only short-term commit activity, show signs of limited technical adaptability, which can be a constraint for real-world applications.

Overall, the number of GitHub stars and commit activity show similar patterns, suggesting that user interest and active development often go hand-in-hand. Inference engines that are frequently updated and support diverse models are more likely to be well-maintained over the long term.

\subsection{Hardware Compatibility and Platform Support} \label{sec:practical_guide_hardware_platform}

\textbf{Hardware and OS Support.}
As shown in Table~\ref{tab:frameworks_hardware}, each inference engine is designed with different goals and target systems. Some inference engines support various hardware types, while others are optimized for a single platform. These hardware compatibility differences affect performance-related features such as quantization data formats, kernel fusion, and support for multi-node or multi-GPU configurations. Therefore, for optimal service performance, inference engines should be selected based on their compatibility with the intended hardware setup.
In addition, Table~\ref{tab:frameworks_hardware} summarizes the OS and hardware support status for each inference engine.

Most inference engines operate in Linux environments, with some additionally supporting Windows or macOS. Commercial engines often provide web-based inference services, but they also enable on-premise deployments. These platform differences can affect both development complexity and inference performance, depending on the range of software capabilities.

\textbf{CPU-Based Inference.}
While many engines include CPU-based inference, non-edge-focused solutions typically employ the CPU for specific tasks---such as offloading operations or handling model weights---rather than as the primary compute resource.

\textbf{Edge and Server Environments.}
On edge devices (e.g., mobiles and Internet of Things (IoTs) systems), limited compute and memory resources require inference engines to focus on lightweight design. These engines reduce model size and apply techniques like quantization to minimize memory usage and enable execution on low-power hardware. Mobile and edge-oriented engines may need to run entirely on CPUs or leverage AI accelerators embedded in system-on-chip (SoC) platforms, such as Neural Processing Units (NPUs) or Digital Signal Processors (DSPs). For example, Apple Core ML~\cite{coreml} and Google AI Edge SDK~\cite{edgesdk} allow deployment of transformer operations to dedicated hardware on consumer devices. Edge inference engines include Ollama~\cite{ollama}, llama.cpp~\cite{llamacpp}, and MLC LLM~\cite{mlcllm}, and in particular, MLC LLM provides compiler technology for various edge hardware.

Conversely, server-side inference engines are optimized for multi-GPU environments to handle high volumes of requests. They rely on distributed computing techniques such as model and pipeline parallelism to spread large models across devices, and they use large batch sizes and dynamic scheduling to maximize hardware utilization.As AI accelerators such as Intel Max~\cite{intelmax}, Google TPU~\cite{jouppi2023tpu}, AMD Instinct~\cite{smith2024amd}, and Intel Gaudi~\cite{kaplan2024intel} are adopted as replacements for NVIDIA GPUs in inference servers, more and more engines are offering heterogeneous hardware backends. Server inference engines include TensorRT-LLM~\cite{tensorrtllm}, vLLM~\cite{kwon2023efficient}, DeepSpeed-FastGen~\cite{holmes2024deepspeed}, etc., and provide optimization techniques for throughput or latency.

\begin{figure}[tbp]
    \centering
    \resizebox{.6\textwidth}{!}{%
        \begin{tikzpicture}[font=\sffamily]
    
        \fill[gray!10!blue!10, rounded corners=5pt]    (0,0) rectangle (5,1.6);
        \fill[gray!10!teal!10, rounded corners=5pt]    (5,0) rectangle (10,1.6);
        \fill[gray!5!orange!10, rounded corners=5pt]   (0,-1.6) rectangle (5,0);
        \fill[gray!5!purple!10, rounded corners=5pt]   (5,-1.6) rectangle (10,0);
    
        \node at (2.5,0.8) {
            \parbox[c][\height][c]{5cm}{
                \centering\tiny
                llama.cpp~\cite{llamacpp},
                MAX~\cite{max},
                MLC LLM~\cite{mlcllm}, \\
                Ollama~\cite{ollama},
                PowerInfer~\cite{xue2024powerinfer},
                TGI~\cite{tgi}
            }
        };
        \node at (7.5,0.8) {
            \parbox[c][\height][c]{5cm}{
                \centering\tiny
                bitnet.cpp~\cite{wang20241},
                LightLLM~\cite{lightllm},
                llama2.c~\cite{llama2c},
                NanoFlow~\cite{zhu2024nanoflow},
                OpenLLM~\cite{lightllm},
                Sarathi-Serve~\cite{agrawal2024taming},
                Unsloth~\cite{unsloth},
                vAttention~\cite{prabhu2025vattention},
                Friendli Inference~\cite{friendli}
            }
        };
        \node at (2.5,-0.8) {
            \parbox[c][\height][c]{5cm}{
                \centering\tiny
                DeepSpeed-FastGen~\cite{holmes2024deepspeed},
                LitGPT~\cite{litgpt},
                LMDeploy~\cite{2023lmdeploy},
                SGLang~\cite{zheng2024sglang},
                vLLM~\cite{kwon2023efficient},
                Fireworks AI~\cite{fireworks}, \\
                Together Inference~\cite{together}
            }
        };
        \node at (7.5,-0.8) {
            \parbox[c][\height][c]{5cm}{
                \centering\tiny
                DistServe~\cite{zhong2024distserve},
                TensorRT-LLM~\cite{tensorrtllm},
                GroqCloud~\cite{groqcloud}
            }
        };
    
        \node at (5,2.25) {\scriptsize{\textbf{Device Support}}};
        \node[rotate=90] at (-0.8,0) {\scriptsize{\textbf{Scale}}};
    
        \node[rotate=90] at (-0.3,0.8) {\scriptsize{Single-Node}};
        \node[rotate=90] at (-0.3,-0.8) {\scriptsize{Multi-Node}};
        \node at (2.5,1.9) {\scriptsize{Heterogeneous Devices}};
        \node at (7.5,1.9) {\scriptsize{Homogeneous Devices}};
    
        \end{tikzpicture}
    }
    \caption{A taxonomy of LLM inference engines categorized by scalability and hardware support}
    \label{fig:scale-device-matrix}
\end{figure}

\textbf{Scalability and Device Types.}
Fig.~\ref{fig:scale-device-matrix} groups the inference engines from Table~\ref{tab:frameworks} according to their hardware characteristics. The \textit{X}-axis distinguishes between support for a single device type versus multiple types, while the \textit{Y}-axis shows whether each engine supports single-node or multi-node configurations. A single node generally includes one to eight GPUs, whereas multi-node systems connect multiple such nodes.

Single-node inference engines emphasize intra-node optimization for CPUs, consumer-level GPUs, or edge/IoT devices. Ollama~\cite{ollama} and llama.cpp~\cite{llamacpp} focus on consumer-level hardware (e.g., laptops and PCs), and MLC LLM~\cite{mlcllm} targets efficient inference on various edge platforms. By contrast, multi-node inference engines handle both inter-node and intra-node computations, optimizing scalability and performance for multi-user workloads. Representative inference engines belonging to this category include vLLM~\cite{kwon2023efficient}, TensorRT-LLM~\cite{tensorrtllm}, and SGLang~\cite{zheng2024sglang}.

Inference engines supporting heterogeneous devices can operate with multiple hardware types beyond GPUs, allowing developers to choose hardware based on application requirements. In contrast, engines that support only homogeneous devices---such as those specialized for NVIDIA GPUs or Groq LPU~\cite{abts2020think}---can deliver high performance through custom kernels and low-level optimizations, though their narrower hardware support may limit portability.

\subsection{Design and Pricing Strategies of Commercial Inference Engines} \label{sec:parctical_guide_commercial_engine}

\textbf{Cloud Services and Model Coverage.}
Commercial inference engines offer cloud-based services that simplify the setup of LLM applications and underlying hardware, compared to many open-source solutions. In particular, Friendli Inference~\cite{friendli}, Fireworks AI~\cite{fireworks}, and Together Inference~\cite{together} support a broader model range than most open-source inference engines, covering not only LLMs but also image, audio, and multimodal models, and they facilitate rapid adoption of newly released models.

A key advantage of commercial inference engines is that they can provide various model and hardware support customized to the scale of the service, reducing the cost and complexity of server deployment and maintenance. Unlike some open-source engines---whose maintenance may be inconsistent or whose licensing might shift to paid models if resources become constrained---commercial services generally guarantee updates and enhancements over a specified duration, ensuring reliable long-term operation.

\textbf{Hardware Variety and Specialization.}
Among these services, Friendli Inference~\cite{friendli} and Together Inference~\cite{together} focus on optimizing inference for NVIDIA GPUs, whereas GroqCloud~\cite{groqcloud} leverages the proprietary Groq LPU AI accelerator~\cite{abts2020think}. Fireworks AI supports a broader range of hardware, including AMD Instinct MI300X~\cite{smith2024amd}, and meets privacy and reliability standards through relevant certifications.

\begin{figure}[tbp]
    \centering
    \resizebox{.75\textwidth}{!}{%
        \begin{subfigure}[b]{0.49\textwidth}
            \centering
            \begin{tikzpicture}
            \begin{axis}[
                llmplot,
                ylabel={Output tokens/s},
                ylabel style={at={(axis description cs:-0.15,0.08)}, anchor=west},
                ymin=0, ymax=650,
                ytick={0,100,...,600},
                title={\shortstack{(a) Inference Throughput (Output Tokens/sec)}},
            ]
            \addplot+[bar shift=-6.75pt] coordinates {
              (DeepSeek-R1,64.9) 
              (DeepSeek-V3,0) 
              ({Llama 3.3 70B},176.1)
              ({Llama 3.1 405B},0) 
              ({Llama 3.1 70B},196.5) 
              ({Llama 3.1 8B},486.9)
              ({Qwen 2.5 Coder 32B},0) 
              ({Qwen QwQ Preview 32B},0)
            };
            \addplot+[bar shift=-2.25pt] coordinates {
              (DeepSeek-R1,93.9) 
              (DeepSeek-V3,62.3) 
              ({Llama 3.3 70B},123.5)
              ({Llama 3.1 405B},70.5) 
              ({Llama 3.1 70B},134.1) 
              ({Llama 3.1 8B},225.6)
              ({Qwen 2.5 Coder 32B},61.7) 
              ({Qwen QwQ Preview 32B},64.6)
            };
            \addplot+[bar shift=2.25pt] coordinates {
              (DeepSeek-R1,275.1) 
              (DeepSeek-V3,0) 
              ({Llama 3.3 70B},275.3)
              ({Llama 3.1 405B},0) 
              ({Llama 3.1 70B},0) 
              ({Llama 3.1 8B},452.6)
              ({Qwen 2.5 Coder 32B},323.3) 
              ({Qwen QwQ Preview 32B},314.1)
            };
            \addplot+[bar shift=6.75pt] coordinates {
              (DeepSeek-R1,107.2) 
              (DeepSeek-V3,35.4)
              ({Llama 3.3 70B},154.7)
              ({Llama 3.1 405B},108.8) 
              ({Llama 3.1 70B},228) 
              ({Llama 3.1 8B},319)
              ({Qwen 2.5 Coder 32B},85) 
              ({Qwen QwQ Preview 32B},65.1)
            };
            \legend{Friendli Inference, Fireworks AI, GroqCloud, Together Inference}
            \end{axis}
            \end{tikzpicture}
        \end{subfigure}
        \hspace{-8mm}
        \begin{subfigure}[b]{0.49\textwidth}
            \centering
            \begin{tikzpicture}
            \begin{axis}[
                llmplot,
                ylabel={Latency (s)},
                ylabel style={at={(axis description cs:-0.15,0.2)}, anchor=west},
                ymin=0, ymax=1.1,
                ytick={0,0.1,...,1.0},
                title={\shortstack{(b) Initial Response Latency (TTFT (s))}},
            ]
            \addplot+[bar shift=-6.75pt] coordinates {
              (DeepSeek-R1,0.43)
              (DeepSeek-V3,0) 
              ({Llama 3.3 70B},0.32)
              ({Llama 3.1 405B},0) 
              ({Llama 3.1 70B},0.33) 
              ({Llama 3.1 8B},0.28)
              ({Qwen 2.5 Coder 32B},0) 
              ({Qwen QwQ Preview 32B},0)
            };
            \addplot+[bar shift=-2.25pt] coordinates {
              (DeepSeek-R1,0.72) 
              (DeepSeek-V3,0.84)
              ({Llama 3.3 70B},0.55)
              ({Llama 3.1 405B},0.72)
              ({Llama 3.1 70B},0.49) 
              ({Llama 3.1 8B},0.34)
              ({Qwen 2.5 Coder 32B},0.35) 
              ({Qwen QwQ Preview 32B},0.43)
            };
            \addplot+[bar shift=2.25pt] coordinates {
              (DeepSeek-R1,0.53)
              (DeepSeek-V3,0) 
              ({Llama 3.3 70B},0.65)
              ({Llama 3.1 405B},0)
              ({Llama 3.1 70B},0)
              ({Llama 3.1 8B},0.39)
              ({Qwen 2.5 Coder 32B},0.56)
              ({Qwen QwQ Preview 32B},0.34)
            };
            \addplot+[bar shift=6.75pt] coordinates {
              (DeepSeek-R1,0.41)
              (DeepSeek-V3,0.73) 
              ({Llama 3.3 70B},0.77)
              ({Llama 3.1 405B},0.55)
              ({Llama 3.1 70B},0.31) 
              ({Llama 3.1 8B},0.2)
              ({Qwen 2.5 Coder 32B},0.49)
              ({Qwen QwQ Preview 32B},0.6)
            };
            \end{axis}
            \end{tikzpicture}
        \end{subfigure}
    }

    \caption{Comparison of Inference Performance across commercial LLM inference engines}
    \label{fig:throughput-ttft}
\end{figure}

\begin{table}[tbp]
    \centering
    \begin{minipage}[t]{0.45\linewidth}
        \centering
        \resizebox{\linewidth}{!}{%
        \begin{tabular}{@{}l cc cc cc cc@{}}
        \toprule
        \multirow{2.5}{*}{Models} & \multicolumn{2}{c}{Friendli AI$^\dagger$} & \multicolumn{2}{c}{Fireworks AI} & \multicolumn{2}{c}{GroqCloud} & \multicolumn{2}{c}{Together AI$^\ddagger$} \\
        \cmidrule(lr){2-3} \cmidrule(lr){4-5} \cmidrule(lr){6-7} \cmidrule(lr){8-9}
            & Input & Output & Input & Output & Input & Output & Input & Output \\
        \midrule
        DeepSeek-R1              & 3.00  & 7.00   & 3.00  & 8.00   & 0.75$^\ast$ & 0.99$^\ast$ & 3.00  & 7.00 \\
        DeepSeek-V3              & --     & --     & 0.90   & 0.90   & --   & --   & 1.25  & 1.25 \\
        Llama 3.3 70B            & 0.60   & 0.60   & --     & --     & 0.59 & 0.79 & 0.88  & 0.88 \\
        Llama 3.1 405B           & --     & --     & 3.00   & 3.00   & --   & --   & 3.50  & 3.50 \\
        Llama 3.1 70B            & 0.60   & 0.60   & --     & --     & --   & --   & 0.88  & 0.88 \\
        Llama 3.1 8B             & 0.10   & 0.10   & --     & --     & 0.05 & 0.08 & 0.18  & 0.18 \\
        Llama 4 Maveric          & --     & --     & 0.22    & 0.88     & 0.20 & 0.60 & 0.27  & 0.85 \\
        Qwen 2.5 Coder 32B       & --     & --     & --     & --     & 0.79 & 0.79 & 0.80  & 0.80 \\
        Qwen QwQ Preview 32B     & --     & --     & --     & --     & 0.29 & 0.39 & 1.20  & 1.20 \\
        OpenAI gpt OSS 120b      & --     & --     & 0.15   & 0.60     & 0.15 & 0.75 & 0.15  & 0.60 \\
        \bottomrule
        \end{tabular}%
        }

        \vspace{-0.1mm}
        \flushleft
        \hspace{0mm} \tiny{$^\dagger$Llama is Instruct model, $^\ddagger$Turbo mode price} \\
        \hspace{0mm} \tiny{$^\ast$DeepSeek-R1 Distill Llama 70B}
        \caption{Pricing by Model in Commercial LLM Engines (\$/1M tokens)}
        \label{tab:framework_price}
    \end{minipage}
    \hspace{2mm}
    \begin{minipage}[t]{0.43\linewidth}
        \centering
        \resizebox{\linewidth}{!}{%
        \begin{tabular}{@{}lcccc@{}}
        \toprule
        \multicolumn{1}{c}{Hardwares} & Friendli AI & Fireworks AI & GroqCloud$^\dagger$ & Together AI \\
        \midrule
        NVIDIA A100 80GB     & 2.9 & 2.9  & --   & 1.30 \\
        NVIDIA H100 80GB     & 3.9 & 5.8  & --   & 2.29 \\
        NVIDIA H200 141GB    & 4.5  & 6.99 & --   & 3.79 \\
        NVIDIA B200 180GB    & 8.9  & 11.99 & --   & 5.50 \\
        AMD MI300X           & --  & 4.99 & --   & --   \\
        Groq LPU             & --  & --   & --   & --   \\
        \bottomrule
        \end{tabular}%
        }
        \vspace{-0.1mm}
        \flushleft
        \hspace{0mm} \tiny{$^\dagger$Charging prices based on tokens and requests per model rather than per device} \\
        \caption{Pricing by Hardware Type in Commercial LLM Engines (\$/hour for 1 device)}
        \label{tab:hardware_price}
    \end{minipage}
\end{table}

\textbf{Performance and Cost Trade-offs.}
When selecting a commercial engine, it is also necessary to consider hardware support and cost. Commercial inference engines typically aim for low latency and high throughput by implementing batch optimization, request pipelining, and other techniques that offer faster and more streamlined deployment compared to open-source alternatives. Fig.~\ref{fig:throughput-ttft} and Table~\ref{tab:framework_price} show the inference performance and costs for various models (e.g., reasoning (DeepSeek-R1~\cite{guo2025deepseek}), MoE (DeepSeek-V1~\cite{bi2024deepseek}), large-scale (Llama 3~\cite{grattafiori2024llama}), code generation (Qwen 2.5 Coder~\cite{hui2024qwen2}), multimodal (Qwen QWQ~\cite{team2024qwq})) using different commercial engines~\cite{artificalanalysis}. Additionally, Table~\ref{tab:hardware_price} summarizes the hardware costs provided by each commercial engine. Even when using the same hardware, the cost may vary depending on the degree of kernel and compute optimization in each engine. 

\section{Detailed Review of Inference Engines} \label{sec:detailed_review}

This section provides a detailed literature review of the 25 inference engines listed in~Table \ref{tab:frameworks}. For each engine, we describe its architecture, key features, and distinctive traits. We also explain, engine by engine, the representative characteristics shown in Fig.~\ref{fig:inference_engine_spider_graph}'s six-axis radar plots.

\subsection{Ollama} \label{sec:detailed_review_ollama}

Ollama~\cite{ollama} is a Go programming language~\cite{go}-based inference engine designed to run LLMs in local environments, enabling users without technical background to easily test and deploy models. Consequently, it primarily targets single-GPU setups rather than multi-GPU systems, relying on llama.cpp as its core backend.

Ollama is composed of two primary components: a \texttt{\small {client}} and a \texttt{\small {server}}. The \texttt{\small {client}} sends requests to the \texttt{\small {server}} via a command-line interface (CLI), while the \texttt{\small {server}} includes an HTTP server and a llama.cpp~\cite{llamacpp} backend. The HTTP server manages \texttt{\small {client-server}} communication, and the llama.cpp backend loads the model and processes inference requests.

The inference engine supports a variety of models---such as Llama~\cite{grattafiori2024llama}, Falcon~\cite{almazrouei2023falcon}, Mistral~\cite{jiang2023mistral}, and DeepSeek-R1~\cite{guo2025deepseek}---and is important to quickly adapt to newly released models. It uses both GGUF~\cite{gguf} and Safetensors~\cite{safetensor} formats for model inference and provides model customization via a \textit{Modelfile}. In addition, Ollama offers a REST API that allows users to manage and execute models through HTTP requests, making it suitable for chat, text generation, and other applications. Integration options include Open WebUI~\cite{openwebui}, SwiftChat~\cite{swiftchat}, Google Cloud, and oterm~\cite{oterm}, extending its deployment capabilities in mobile, cloud, and local environments.

However, Ollama prioritizes user accessibility over advanced inference optimizations, meaning it lacks features such as memory optimization, multi-GPU functionality, and multi-node support. In return, it delivers broad compatibility by supporting not only NVIDIA GPUs but also AMD GPUs and ARM platforms.

\begin{findingbox}[]
   {\small \textbf{Representative Characteristics Summary}}
    \begin{itemize}[leftmargin=1.2em, label=--, itemsep=0em]
        \item \textbf{General-Purpose [\Medium]}: Supports popular community models and both NVIDIA and AMD GPUs, but lacks multi-GPU or edge specialization.
        \item \textbf{Ease-of-Deploy [\High]}: One-line installation via Homebrew, pip, or Docker makes setup extremely simple.
        \item \textbf{Ease-of-Use [\Medium]}: A concise CLI and REST API, plus GUI integrations such as Open WebUI, lower the entry barrier for non-experts.
        \item \textbf{Latency-Aware [\Medium]}: The engine provides no Flash- or KV-cache optimizations, so single-token latency remains higher.
        \item \textbf{Throughput-Aware [\Medium]}: Single-GPU operation without batching strategies limits sustained throughput.
        \item \textbf{Scalability [\Medium]}: Designed for local single-GPU use and cannot extend to multi-node deployments.
    \end{itemize}
\end{findingbox}

\subsection{llama.cpp} \label{sec:detailed_review_llamacpp}

llama.cpp~\cite{llamacpp} is a C++ library for LLM inference that runs models on CPUs without a GPU. Consequently, it depends on minimal external software and operates efficiently on diverse hardware architectures. It supports quantization for multiple data types (e.g., 1.5-bit, 4-bit, 8-bit), reducing memory usage, and boosting efficiency.

llama.cpp also introduces the Georgi Gerganov Unified Format (GGUF)~\cite{gguf} for streamlined LLM storage and deployment. GGUF consolidates model parameters, structure, and metadata into a single file, improving the Georgi Gerganov Machine Learning (GGML)~\cite{ggml} format by providing better flexibility and compatibility. This approach standardizes model storage and simplifies deployment.

llama.cpp supports a range of hardware platforms---including x86, ARM, and NVIDIA GPUs---and uses the GGML context to configure these backends. It provides hardware-specific kernel and graph optimizations that facilitate efficient inference. Additionally, llama.cpp extends usability with subprojects such as \texttt{\small {llama-cli}} for command-line execution, llama-server for OpenAI API compatible HTTP-serving, and lightweight runners like \texttt{\small {llama-run}} and \texttt{\small {llama-simple}}.

\begin{findingbox}[]
    {\small \textbf{Representative Characteristics Summary}}
    \begin{itemize}[leftmargin=1.2em, label=--, itemsep=0em]
        \item \textbf{General-Purpose [\Medium]}: Runs on x86, ARM CPUs and NVIDIA GPUs with several quantization formats for broad hardware reach.
        \item \textbf{Ease-of-Deploy [\High]}: A single static binary or minimal CMake build keeps external dependencies near zero.
        \item \textbf{Ease-of-Use [\Low]}: CLI helpers and an OpenAI-style server exist, but documentation is concise and community-driven.
        \item \textbf{Latency-Aware [\Medium]}: Optional FlashAttention kernels and GPU offload reduce token delay on capable devices.
        \item \textbf{Throughput-Aware [\Medium]}: Multithreading and continuous batching boost CPU throughput, although distributed support is minimal.
        \item \textbf{Scalability [\Low]}: Optimized for single-node execution and lacks native cluster features.
    \end{itemize}
\end{findingbox}

\subsection{vLLM} \label{sec:detailed_review_vllm}

vLLM~\cite{kwon2023efficient} is a high-performance LLMs serving library, focusing on fast token generation and low latency. Its \texttt{\small {PagedAttention}} mechanism enhances memory efficiency by storing  $\mathbf{KV}$ cache in non-contiguous memory blocks, preventing the fragmentation issues associated with contiguous storage.

vLLM is built around \texttt{\small {AsyncLLM}} for asynchronous request handling, an OpenAI-compatible API server, and an \texttt{\small {EngineCore}} that conducts inference. A ZeroMQ~\cite{zeromq}-based multiprocessing API server overlaps operations between \texttt{\small {AsyncLLM}} and the API layer. \texttt{\small {EngineCore}} features modules for scheduling and model execution, enabling concurrent handling of CPU-heavy tasks (e.g., tokenization, multimodal input management, and token detokenization) alongside the main execution loop for improved throughput. Its symmetric architecture reduces inter-process overhead and supports optimized tensor parallelism.

Additionally, vLLM supports FlashAttention-3~\cite{shah2024flashattention} to further reduce inference latency. It employs a distributed system architecture for multi-GPU workload distribution, leveraging Megatron-LM's tensor parallelism~\cite{shoeybi2019megatron}. Beyond CPU and GPU support, vLLM is compatible with AWS Inferentia \cite{inferentia} and Google TPU~\cite{jouppi2023tpu}, extending its capabilities to multimodal inference.

\news{vLLM provides a batch decoding technique, called cascade inference~\cite{cascade-inference}, which utilizes shared prefixes to efficiently manage memory bandwidth. During LLM inference, when multiple requests contain identical prefixes, recomputing the prefix segment for each request incurs substantial memory and time overhead. This issue becomes more severe as the prefix length increases and the number of concurrent requests grows.}

\news{Cascade inference separates the common prefix from each request's individual suffix and stores the $\mathbf{KV}$ cache for the prefix in the GPU's shared memory, allowing multiple requests to reference it simultaneously. As a result, redundant prefix computations are eliminated, significantly reducing both latency and memory consumption. In vLLM, cascade inference can be toggled on or off through a dedicated flag, but is typically designed to activate automatically based on detected input patterns.}

\begin{findingbox}[]
    {\small \textbf{Representative Characteristics Summary}}
    \begin{itemize}[leftmargin=1.2em, label=--, itemsep=0em]
        \item \textbf{General-Purpose [\High]}: Serves a wide range of LLMs across GPUs, TPUs, and AWS Inferentia accelerators.
        \item \textbf{Ease-of-Deploy [\High]}: Docker images and a pip package simplify setup, but distributed configuration still requires manual steps.
        \item \textbf{Ease-of-Use [\High]}: OpenAI-compatible endpoints and an active community streamline application integration.
        \item \textbf{Latency-Aware [\Medium]}: FlashAttention-3 and PagedAttention aggressively cut attention-time latency.
        \item \textbf{Throughput-Aware [\High]}: AsyncLLM scheduling and ZeroMQ multiprocessing maintain high token-per-second rates.
        \item \textbf{Scalability [\High]}: Built-in tensor parallelism enables multi-GPU and multi-node clustering.
    \end{itemize}
\end{findingbox}

\subsection{DeepSpeed-FastGen} \label{sec:detailed_review_deepspeed}

DeepSpeed-FastGen~\cite{holmes2024deepspeed} is an LLM inference engine integrating Microsoft DeepSpeed Inference \cite{aminabadi2022deepspeed} and DeepSpeed Model Implementations for Inference (MII) \cite{deepspeedmii}. It optimizes memory usage to enable efficient model inference.

DeepSpeed-FastGen deploys DeepSpeed MII for its frontend and backend, handling requests through features like dedicated query/response APIs, continuous batching, and a model pipeline. Internally, it leverages DeepSpeed Inference to support hardware-optimized kernels (e.g., NVIDIA CUDA), as well as Blocked $\mathbf{KV}$-Cache and tensor parallelism.

A major feature is the \texttt{\small {Dynamic SplitFuse}} technique, which splits long prompts into smaller segments and processes them in multiple forward passes, improving throughput and reducing latency. By maintaining a consistent forward pass size, system processing efficiency increases. DeepSpeed-FastGen also offers replica-level load balancing, distributing inference workloads across multiple nodes. Compared to single-node inference, multi-node deployments can deliver significant speedups in query processing.

\begin{findingbox}[]
    {\small \textbf{Representative Characteristics Summary}}
    \begin{itemize}[leftmargin=1.2em, label=--, itemsep=0em]
        \item \textbf{General-Purpose [\Medium]}: DeepSpeed-MII front end supports numerous HuggingFace checkpoints and custom models.
        \item \textbf{Ease-of-Deploy [\High]}: A containerized launcher is available, though model conversion and registry are still required.
        \item \textbf{Ease-of-Use [\High]}: MII-style APIs are clear, but some DeepSpeed configuration know-how is assumed.
        \item \textbf{Latency-Aware [\Medium]}: Dynamic SplitFuse splits long prompts to cap worst-case latency.
        \item \textbf{Throughput-Aware [\High]}: Continuous batching, blocked KV-cache, and tensor parallelism keep GPUs saturated.
        \item \textbf{Scalability [\High]}: Replica-level load balancing supports efficient multi-node service.
    \end{itemize}
\end{findingbox}

\subsection{Unsloth} \label{sec:detailed_review_unsloth}

Unsloth~\cite{unsloth} is a engine focused on efficient fine-tuning and inference for LLMs. It achieves rapid fine-tuning and reduced memory usage through techniques such as Low-rank adaptation (LoRA)~\cite{hu2022lora} and Quantized-LoRA (QLoRA)~\cite{dettmers2023qlora} while preserving model accuracy. All kernels are implemented in OpenAI Triton~\cite{tillet2019triton}, further enhancing the execution speed of LLM. Although Unsloth integrates modules such as xFormers~\cite{xFormers2022} to accelerate transformer operations. This approach allows for flexible customization of attention blocks and other modules, providing greater adaptability for diverse use cases.

For compatibility, Unsloth supports both GGUF~\cite{gguf} and vLLM~\cite{kwon2023efficient} formats and offers a straightforward API for creating inference services. However, it currently runs only on NVIDIA GPUs, and advanced optimization features such as multi-GPU and multi-node support are exclusive to the paid version. The open-source release is restricted to single-GPU setups and supports a limited number of models.

\begin{findingbox}[]
    {\small \textbf{Representative Characteristics Summary}}
    \begin{itemize}[leftmargin=1.2em, label=--, itemsep=0em]
        \item \textbf{General-Purpose [\Low]}: Provides GGUF and vLLM model formats but currently restricts execution to NVIDIA GPUs.
        \item \textbf{Ease-of-Deploy [\Medium]}: A single pip install delivers both fine-tuning and inference capabilities.
        \item \textbf{Ease-of-Use [\Medium]}: High-level Python APIs are simple, though advanced documentation is still limited.
        \item \textbf{Latency-Aware [\Medium]}: Triton-fused kernels shorten attention steps, trimming token latency moderately.
        \item \textbf{Throughput-Aware [\Low]}: xFormers integration helps single-GPU throughput; distributed execution is paywalled.
        \item \textbf{Scalability [\Low]}: The open-source edition runs on a single GPU and omits multi-node features.
    \end{itemize}
\end{findingbox}

\subsection{MAX} \label{sec:detailed_review_max}

Modular Accelerated Xecution (MAX)~\cite{max} is an integrated platform aimed at simplifying the creation and deployment of high-performance AI endpoints, while maintaining flexibility across diverse hardware setups. It offers a graph compiler and runtime capable of accelerating generative AI models through hardware-agnostic libraries. By compiling models into optimized computation graphs, MAX enhances execution efficiency and reduces latency for better performance.

MAX is built on the Mojo programming language~\cite{mojo}. Mojo extends Python with system programming features from C, C++, and CUDA via Multi-Level Intermediate Representation (MLIR)~\cite{lattner2021mlir}, enabling high performance on CPUs, GPUs, and specialized AI accelerators.

MAX comprises two main components: the \texttt{\small {MAX Engine}} (an inference library and runtime) and \texttt{\small {MAX Serve}}, a serving utility for model deployment. \texttt{\small {MAX Serve}} hosts LLMs and provides OpenAI API compatible REST endpoints in both local and cloud environments. It applies continuous heterogeneous batching and multi-step scheduling to maximize GPU utilization and ensure stable performance, particularly for large-scale workloads. Internally, \texttt{\small {MAX Serve}} integrates the \texttt{\small {MAX Engine}}, which utilizes its graph compiler and runtime to accelerate models on CPUs and GPUs.

Currently, MAX supports inference workloads across both local and cloud environments and operates on CPUs and NVIDIA GPUs.

\begin{findingbox}[]
    {\small \textbf{Representative Characteristics Summary}}
    \begin{itemize}[leftmargin=1.2em, label=--, itemsep=0em]
        \item \textbf{General-Purpose [\Medium]}: Mojo's MLIR compiler targets CPUs, GPUs, and future accelerators from one model graph.
        \item \textbf{Ease-of-Deploy [\High]}: Docker images and a CLI exist, but users still package models into MAX Serve.
        \item \textbf{Ease-of-Use [\High]}: REST endpoints are easy to consume, yet Mojo tooling is early-stage for newcomers.
        \item \textbf{Latency-Aware [\Medium]}: Ahead-of-time graph compilation fuses kernels and shortens critical paths.
        \item \textbf{Throughput-Aware [\Medium]}: Continuous heterogeneous batching and multi-step scheduling keep devices busy.
        \item \textbf{Scalability [\High]}: Operates on local and cloud machines with experimental multi-GPU support.
    \end{itemize}
\end{findingbox}

\subsection{MLC LLM} \label{sec:detailed_review_mlcllm}

MLC LLM~\cite{mlcllm} is a compiler and high-performance deployment engine for LLMs, designed to enable model development, optimization, and deployment across multiple platforms. It supports inference not only on NVIDIA and AMD GPUs but also on mobile and edge devices such as iOS and Android, unifying server and edge environments into a single LLM engine. The provided engine, \texttt{\small {MLCEngine}}, delivers high throughput and low latency in server environments and also supports lightweight local deployment.

Achieving platform-wide LLM acceleration requires extensive GPU programming and runtime compatibility. To address this, MLC LLM builds on Apache TVM~\cite{chen2018tvm}, generating GPU libraries automatically for each hardware and platform. It integrates LLM-specific optimizations such as continuous batching~\cite{yu2022orca} and speculative decoding~\cite{leviathan2023fast, xia2024unlocking}, and employs FlashInfer~\cite{ye2025flashinfer} to accelerate NVIDIA GPUs. MLC LLM either converts and quantizes foundation model weights or loads pre-converted weights, using the \texttt{\small {model-weights-mlc}} module for operator fusion, memory allocation, and hardware-specific optimizations; the \texttt{\small {model-lib}} component then constructs platform-native runtimes for each device. MLC LLM offers a range of deployment modes---Python APIs, OpenAI-compatible APIs, REST servers, and WebLLM~\cite{ruan2024webllm}---ensuring broad portability across cloud and local platforms.

\begin{findingbox}[]
    {\small \textbf{Representative Characteristics Summary}}
    \begin{itemize}[leftmargin=1.2em, label=--, itemsep=0em]
        \item \textbf{General-Purpose [\Medium]}: A single engine serves desktop, mobile, and WebLLM runtimes across NVIDIA and AMD GPUs.
        \item \textbf{Ease-of-Deploy [\High]}: The installer script compiles TVM kernels for each target automatically.
        \item \textbf{Ease-of-Use [\Medium]}: Python and REST APIs plus a web demo provide moderate integration effort.
        \item \textbf{Latency-Aware [\Medium]}: FlashInfer kernels and continuous batching enable low-latency generation.
        \item \textbf{Throughput-Aware [\High]}: Speculative decoding and operator fusion lift tokens-per-second on GPUs.
        \item \textbf{Scalability [\Medium]}: Generates native runtimes for edge devices through to cloud servers.
    \end{itemize}
\end{findingbox}

\subsection{llama2.c} \label{sec:detailed_review_llama2c}

llama2.c~\cite{llama2c} is an inference engine designed to run small Llama2~\cite{touvron2023llama2}-based models in a single C file. It comprises approximately 700 lines of C code and can load models trained with PyTorch~\cite{paszke2019pytorch} for inference. 

The inference engine focuses on small-scale domains and is intended for educational use and features a simple structure. Rather than implementing advanced optimization techniques, it only includes the essential code needed for LLM inference. Parallel processing is limited to OpenMP-based multithreading and runs exclusively on CPUs, without support for GPU execution or distributed environments.

\begin{findingbox}[]
    {\small \textbf{Representative Characteristics Summary}}
    \begin{itemize}[leftmargin=1.2em, label=--, itemsep=0em]
        \item \textbf{General-Purpose [\Low]}: Runs only small Llama-2 checkpoints on CPUs for education and demonstration.
        \item \textbf{Ease-of-Deploy [\Low]}: Compiles in seconds with no external libraries for rapid experimentation.
        \item \textbf{Ease-of-Use [\Low]}: Approximately 700 lines of readable C code make learning and modification easy.
        \item \textbf{Latency-Aware [\Low]}: Only basic OpenMP threading is present, leaving high per-token latency.
        \item \textbf{Throughput-Aware [\Low]}: No batching, GPU support, or cache management reduces sustained throughput.
        \item \textbf{Scalability [\Low]}: Designed for a single CPU host with no distributed or GPU pathway.
    \end{itemize}
\end{findingbox}

\subsection{bitnet.cpp} \label{sec:detailed_review_bitnetcpp}

bitnet.cpp~\cite{wang20241} is a CPU-only inference engine developed in the context of one-bit LLM research. Built based on llama.cpp~\cite{llamacpp}, it focuses on fast, lossless inference of ternary models (BitNet b1.58~\cite{ma2024era}) while minimizing power consumption. The project offers three kernel types---\texttt{\small {I2\_S}}, \texttt{\small {TL1}}, and \texttt{\small {TL2}}---optimized for both x86 and ARM processors.

The \texttt{\small {I2\_S}} kernel converts full-precision weights to a two-bit format offline, then restores the original values during inference to accelerate general matrix-vector multiply (GEMV) operations. This approach reduces memory and bandwidth and also improves performance in multithreading systems. The \texttt{\small {TL1}} kernel compresses every two weights into a four-bit index and employs a lookup table (LUT) with nine precomputed activation values based on the T-MAC~\cite{wei2025t} method, allowing large models to run efficiently even with limited thread environments. \texttt{\small {TL2}} compresses every three weights into a five-bit index, shrinking the model size to one-sixth of the \texttt{\small {TL1}} footprint and making it suitable for environments with tight memory or bandwidth constraints.

bitnet.cpp supports only local CPU execution and relies on multithreading rather than distributed parallelism for acceleration. In addition to BitNet b1.58~\cite{ma2024era}, it can run the Llama 3 8B~\cite{grattafiori2024llama} and Falcon 3~\cite{falcon3} family models, but it does not yet support broader hardware platforms or large-scale distributed deployments.

\begin{findingbox}[]
    {\small \textbf{Representative Characteristics Summary}}
    \begin{itemize}[leftmargin=1.2em, label=--, itemsep=0em]
        \item \textbf{General-Purpose [\Low]}: Runs only on local CPUs and supports a narrow model set (BitNet b1.58 plus a few Llama 3 and Falcon variants), so overall hardware and model diversity is limited.
        \item \textbf{Ease-of-Deploy [\Medium]}: Ships as a self-contained C++ binary that builds with minimal dependencies and requires no GPU drivers, enabling rapid installation on almost any x86 or ARM host. 
        \item \textbf{Ease-of-Use [\Low]}: While the CLI closely mirrors llama.cpp, documentation and community examples are still sparse, which raises the learning curve for first-time users. 
        \item \textbf{Latency-Aware [\Low]}: The engine focuses on memory-bandwidth reduction rather than dedicated latency techniques; single-token delay remains governed by CPU core speed.
        \item \textbf{Throughput-Aware [\Low]}: Multithreaded I2\_S, TL1, and TL2 kernels use 2- to 5-bit weight compression to boost GEMV throughput compared with full-precision CPU baselines.
        \item \textbf{Scalability [\Low]}: All acceleration is confined to one multicore server; there is no support for multi-socket or distributed execution across nodes.
    \end{itemize}
\end{findingbox}

\subsection{SGLang} \label{sec:detailed_review_sglang}

Structured Generation Language for LLMs~(SGLang)~\cite{zheng2024sglang} is a system designed to execute LLMs efficiently by overcoming limitations found in existing inference engines, including multimodal input handling, parallel processing, and $\mathbf{KV}$ cache reuse. To achieve this, SGLang uses multi-call structures and introduces Language Model Programs (\texttt{\small {LM Programs}}), which support various model types (vision, embedding, reward models) as well as multi-node operation.

The inference engine comprises a frontend and a backend (runtime) and provides an OpenAI-compatible API. SGLang's frontend, written in Python, enables flexible authoring of \texttt{\small {LM Programs}} using conventional control flow and libraries, enhancing developer ease. Meanwhile, the backend applies execution optimizations that include RadixAttention-based $\mathbf{KV}$ cache management and structured decoding with compressed finite state machines, enabling rapid inference. These methods allow SGLang to outperform existing inference engines in throughput and excel in tasks such as agent control and logical reasoning.

SGLang provides both an interpreter and a compiler. The interpreter manages prompt states as streams and asynchronously handles fundamental operations to improve synchronization and parallelism. It also tracks program execution paths, enabling further compiler optimizations. After compiling these programs into computation graphs, the SGLang graph executor rewrites the graph or establishes static execution plans.

For further optimization, SGLang employs a \texttt{\small {Zero-Overhead Batch Scheduler}}, similar to NanoFlow's Nano-batching strategy~\cite{zhu2024nanoflow}, to increase parallelism in model inference. It also features a cache-aware load balancer that improves prefix cache hit rates, thus boosting overall throughput.

\begin{findingbox}[]
    {\small \textbf{Representative Characteristics Summary}}
    \begin{itemize}[leftmargin=1.2em, label=--, itemsep=0em]
        \item \textbf{General-Purpose [\Medium]}: Language-Model Programs manage multimodal models and support multi-node execution.
        \item \textbf{Ease-of-Deploy [\High]}: Requires source compilation and CUDA toolchain configuration before use.
        \item \textbf{Ease-of-Use [\Medium]}: Python DSL is flexible but introduces a learning curve with stream-based semantics.
        \item \textbf{Latency-Aware [\Medium]}: RadixAttention and compressed finite-state decoding reduce tail latency.
        \item \textbf{Throughput-Aware [\Medium]}: The Zero-Overhead Batch Scheduler maximizes overlap, achieving extreme throughput.
        \item \textbf{Scalability [\High]}: Cache-aware load balancing enables cluster execution, though tooling is still maturing.
    \end{itemize}
\end{findingbox}

\subsection{LitGPT} \label{sec:detailed_review_litpgt}

LitGPT~\cite{litgpt} is an end-to-end framework that covers fine-tuning, inference, testing and deployment. Built on nanoGPT~\cite{nanogpt}, Lit-LLaMA~\cite{litllama}, and Lightning Fabric~\cite{lightning_fabric}, it supports pretrained models for rapid prototyping. 

LitGPT scales from a single GPU to multi-GPU and multi-node environments, offering distributed parallelism through Fully Sharded Data Parallelism (FSDP)~\cite{zhao2023pytorch} and faster computation with FlashAttention-2~\cite{dao2023flashattention}. This framework also includes memory and speed optimizations via quantization~\cite{egashira2025exploiting} and LoRA~\cite{hu2022lora}, and it can run LLMs on Google TPUs through the PyTorch/XLA compiler~\cite{pytorch_xla}.

\begin{findingbox}[]
    {\small \textbf{Representative Characteristics Summary}}
    \begin{itemize}[leftmargin=1.2em, label=--, itemsep=0em]
        \item \textbf{General-Purpose [\Low]}: Supports NVIDIA GPUs, AMD Instinct, and Google TPU, but is primarily optimized for NVIDIA GPUs.
        \item \textbf{Ease-of-Deploy [\Medium]}: Offers easy installation via pip and provides prebuilt packages.
        \item \textbf{Ease-of-Use [\Medium]}: Provides brief manuals and maintains community through forums and meet-ups.
        \item \textbf{Latency-Aware [\Medium]}: Reduces response time with FlashAttention-2, speculative decoding, and KV caching.
        \item \textbf{Throughput-Aware [\Medium]}: Increases overall throughput with FSDP and batching optimizations.
        \item \textbf{Scalability [\High]}: Extends from a single-GPU setup to multi-GPU and multi-node deployments.
    \end{itemize}
\end{findingbox}

\subsection{OpenLLM} \label{sec:detailed_review_openllm}

OpenLLM~\cite{openllm} is a platform for the straightforward execution and deployment of open-source LLMs and custom models through simple commands. Designed as a cloud-based solution that overcomes the scalability and high-load issues of existing platforms like Ollama~\cite{ollama}, OpenLLM targets multi-user support, high throughput, and low latency. This makes it well suited for deploying LLMs on cloud or on-premise servers and for building LLM-based applications. A key advantage is data security, achieved via a Bring Your Own Cloud (BYOC) model.

OpenLLM provides an OpenAI-compatible API server that simplifies LLM execution and employs vLLM~\cite{kwon2023efficient} and BentoML~\cite{bentoml} as backends to maintain high throughput in large-scale environments. It uses \texttt{\small {Bento}}, a custom file format developed by BentoML, which packages source code, models, data files, and dependencies into a single entity. These \texttt{\small {Bento}} objects can be transformed into container images for convenient deployment.

\begin{findingbox}[]
    {\small \textbf{Representative Characteristics Summary}}
    \begin{itemize}[leftmargin=1.2em, label=--, itemsep=0em]
        \item \textbf{General-Purpose [\Low]}: Combines vLLM and BentoML back ends to run varied open-source models in the cloud.
        \item \textbf{Ease-of-Deploy [\Medium]}: One command converts a model into a Bento image deployable in any BYOC environment.
        \item \textbf{Ease-of-Use [\Low]}: CLI, web UI, and OpenAI-style endpoints cut application integration time sharply.
        \item \textbf{Latency-Aware [\Low]}: FlashAttention from vLLM lowers core latency; additional cloud overhead may remain.
        \item \textbf{Throughput-Aware [\Medium]}: Bento containers batch requests continuously and scale horizontally.
        \item \textbf{Scalability [\Medium]}: Multi-tenant support is built-in, while multi-node GPU pods require custom orchestration.
    \end{itemize}
\end{findingbox}

\subsection{TensorRT-LLM} \label{sec:detailed_review_tensorrtllm}

TensorRT-LLM~\cite{tensorrtllm} is a inference engine to optimize inference on NVIDIA GPUs and is part of NVIDIA's NeMO~\cite{kuchaiev2019nemo} end-to-end generative AI development ecosystem. It includes compilation and optimization libraries to boost model inference performance. During compilation, the TensorRT~\cite{tensorrt} compiler analyzes the computation graph to select optimal kernels, fusing them to minimize memory overhead. This allows maximal exploitation of CUDA kernels and Tensor Cores, and supports various low-precision operations for faster inference.

Models for inference can be trained using NVIDIA NeMo or PyTorch~\cite{paszke2019pytorch}, or sourced from pretrained weights on platforms like Hugging Face, and must be converted to a TensorRT-compatible format using the Model Definition API. Although TensorRT-LLM primarily uses TensorRT as its backend, it also includes Python and C++ backends for NVIDIA Triton Inference Server~\cite{tritoninferenceserver}, providing an end-to-end solution for online LLM deployment. A PyTorch backend is available experimentally. With support from NVIDIA Collective Communication Library (NCCL)~\cite{nccl}, TensorRT-LLM offers distributed inference via tensor parallelism and pipeline parallelism in multi-GPU environments. For optimized serving, in-flight batching groups incoming requests dynamically.

To overcome performance constraints of ring-based All-Reduce topologies in multi-node environments, TensorRT-LLM introduces a multishot approach that harnesses NVSwitch's multicast capabilities, reducing latency by up to 3$\times$. However, TensorRT-LLM is limited to NVIDIA GPUs, restricting hardware scalability.

\begin{findingbox}[]
    {\small \textbf{Representative Characteristics Summary}}
    \begin{itemize}[leftmargin=1.2em, label=--, itemsep=0em]
        \item \textbf{General-Purpose [\Low]}: Targets NVIDIA GPUs exclusively, limiting hardware diversity.
        \item \textbf{Ease-of-Deploy [\High]}: Model conversion and Triton back-end registration add setup steps despite helper scripts.
        \item \textbf{Ease-of-Use [\High]}: Sample Python and C++ code exist, but NeMo and Triton familiarity helps.
        \item \textbf{Latency-Aware [\Medium]}: Kernel fusion on Tensor Cores delivers very low single-token latency.
        \item \textbf{Throughput-Aware [\High]}: In-flight batching and pipeline parallelism maintain high throughput on large models.
        \item \textbf{Scalability [\High]}: NVSwitch multicast and NCCL enable efficient multi-GPU and multi-node deployment.
    \end{itemize}
\end{findingbox}

\subsection{Hugging Face TGI} \label{sec:detailed_review_tgi}

Hugging Face Text Generation Inference (TGI)~\cite{tgi}  is a toolkit for deploying and serving LLMs, supporting diverse inference workloads and integrating with backends like vLLM~\cite{kwon2023efficient} and TensorRT-LLM~\cite{tensorrtllm}. It accommodates various hardware platforms, including NVIDIA GPUs, AWS Inferentia~\cite{inferentia}, and Intel Gaudi~\cite{kaplan2024intel} letting users choose suitable backends for their hardware. Built in Rust, TGI's backend supports streaming and concurrency, efficiently handling high LLM traffic.

TGI comprises three key components: a \texttt{\small {router}}, \texttt{\small {launcher}} and \texttt{\small {model server}}. The \texttt{\small {router}} is an HTTP server that manages client requests (supporting Hugging Face's custom APIs and the OpenAI Message API), batching incoming requests with a queue, scheduler and a memory block allocator. The \texttt{\small {launcher}} spins up one or more \texttt{\small {model server}} and shards models based on parameters from the \texttt{\small {router}}. The \texttt{\small {model server}}---implemented in Python---receives Google Remote Procedure Call (gRPC)~\cite{grpc}-based requests for model loading and inference.

To optimize inference, TGI employs quantization, RoPE scaling~\cite{liu2023scaling}, Safetensors~\cite{safetensor}, and \texttt{\small {Zero Config}} for automatic configuration depending on hardware and model. It also leverages Flashinfer~\cite{ye2025flashinfer} and Flashdecoding~\cite{hong2024flashdecoding++} to deliver fast performance on long prompts. For observability, it connects with tools like Prometheus~\cite{turnbull2018monitoring} and Grafana~\cite{chakraborty2021grafana}. When running models on multiple devices, TGI synchronizes using NVIDIA NCCL~\cite{nccl} Although it supports tensor parallelism for multi-device inference, only certain LLM models are currently compatible.

\begin{findingbox}[]
    {\small \textbf{Representative Characteristics Summary}}
    \begin{itemize}[leftmargin=1.2em, label=--, itemsep=0em]
        \item \textbf{General-Purpose [\Medium]}: Swappable vLLM or TensorRT-LLM back ends cover NVIDIA, Inferentia, and Gaudi hardware.
        \item \textbf{Ease-of-Deploy [\High]}: A single launcher auto-configures hardware and downloads model weights.
        \item \textbf{Ease-of-Use [\Medium]}: Supports custom HF APIs and OpenAI messages with built-in monitoring hooks.
        \item \textbf{Latency-Aware [\Medium]}: FlashInfer and Flashdecoding accelerate long-sequence generation.
        \item \textbf{Throughput-Aware [\Medium]}: Router and scheduler batch inputs continuously for high request volume.
        \item \textbf{Scalability [\High]}: Model sharding and NCCL permit multi-GPU serving across nodes.
    \end{itemize}
\end{findingbox}

\subsection{PowerInfer} \label{sec:detailed_review_powerinfer}

PowerInfer~\cite{song2024powerinfer} is an LLM inference system built by extending llama.cpp~\cite{llamacpp}, designed to run LLMs on a single consumer-grade GPU. Running LLMs without model compression techniques often leads to accuracy loss and memory limitations. CPU-GPU offloading methods suffer from high PCIe latency, which slows down the inference. Additionally, speculative decoding becomes inefficient with small batch sizes and can degrade model performance.

To address these limitations, PowerInfer leverages the observation that neuron activations in LLMs follow a power-law distribution. It separates frequently activated neurons (\texttt{\small {hot neurons}}) from less active ones (\texttt{\small {cold neurons}}). \texttt{\small {Hot neurons}} are loaded onto the GPU for fast computation, while \texttt{\small {cold neurons}} are handled on the CPU. This design reduces GPU memory usage and minimizes CPU-GPU data transfer. PowerInfer uses an offline profiling step to identify hot and \texttt{\small {cold neurons}} based on their activation frequency, and an online predictor to determine which neurons are active for each input.

PowerInfer uses a hybrid approach for inference, comprising offline and online components. In the offline phase, it analyzes neuron activation patterns (\texttt{\small {Insight-1}}) and classifies neurons into \texttt{\small {hot}} and \texttt{\small {cold}} categories using the activation data. It then performs neuron assignment optimization through Integer Linear Programming (ILP) to maximize memory utilization. In the online component, neurons are assigned to the GPU or CPU based on predefined policies, and distributed computations are performed via GPU and CPU executors. 

PowerInfer also introduces neuron-aware sparse operators to overcome the limitations of existing sparse computation libraries. These operators can directly handle irregular tensors at the neuron level without format conversion, and are optimized for both GPU and CPU execution.

As a result, PowerInfer enables efficient LLM inference without fully loading the model into GPU memory, making it a practical solution for memory-constrained local environments.

Recently, PowerInfer-2~\cite{xue2024powerinfer} has been proposed to further extend this approach to mobile devices such as smartphones. PowerInfer-2 extends PowerInfer's capabilities to scenarios involving memory-constrained mobile devices. Relying on the same \texttt{\small {hot-cold neuron}} algorithm, it partitions matrix operations by neuron clusters and allocates them efficiently between the CPU and NPU, implementing I/O pipeline optimizations for faster inference. During the offline phase, PowerInfer-2 generates an execution plan adapted to neuron activation patterns, hardware constraints, and batch sizes. In the online inference phase, it uses neuron caching along with an NPU-based prefill stage and a CPU-NPU hybrid decoding phase, thus boosting overall performance.

\begin{findingbox}[]
    {\small \textbf{Representative Characteristics Summary}}
    \begin{itemize}[leftmargin=1.2em, label=--, itemsep=0em]
        \item \textbf{General-Purpose [\Low]}: Extends llama.cpp for single consumer GPUs and desktop scenarios.
        \item \textbf{Ease-of-Deploy [\Low]}: Pre-built Docker images simplify setup on one GPU.
        \item \textbf{Ease-of-Use [\Low]}: Basic scripts are provided, though neuron-level tuning remains manual.
        \item \textbf{Latency-Aware [\Medium]}: Hot-cold neuron separation removes some transfers but PCIe overhead persists.
        \item \textbf{Throughput-Aware [\Medium]}: Neuron-aware sparse operators moderately raise tokens per second.
        \item \textbf{Scalability [\Low]}: Designed for a single GPU with CPU assist and no cluster capability.
    \end{itemize}
\end{findingbox}

\subsection{LMDeploy} \label{sec:detailed_review_lmdeploy}

LMDeploy~\cite{2023lmdeploy} is an inference and serving engine that incorporates several optimization techniques, including continuous batching~\cite{yu2022orca}, dynamic split and fuse, and high-performance CUDA kernels. In addition to facilitating efficient inference, it provides features such as quantization, fine-tuning, and multi-model services across multiple machines and cards, enabling straightforward and effective service deployment in various contexts.

To support high throughput in interactive LLM inference, LMDeploy offers an engine called \texttt{\small {TurboMind}}, which is built on NVIDIA FasterTransformer~\cite{fastformer}. \texttt{\small {TurboMind}} includes efficient LLM implementations, a Persistent Batch module, and a $\mathbf{KV}$ Cache Manager, all accessible through a simple API. The Persistent Batch module manages continuous batching with a fixed number of batch slots. When a request arrives, it occupies one of these slots, and upon completion, the slot is freed. Meanwhile, the $\mathbf{KV}$ Cache Manager functions as a memory pool, applying a Least Recently Used (LRU) policy to decide which sequence cache to evict when additional memory is required.

In addition to \texttt{\small {TurboMind}}, LMDeploy provides a developer-friendly engine named \texttt{\small {lmdeploy.pytorch}}, which offers a PyTorch-like environment while sharing the same service interface as \texttt{\small {TurboMind}}. It performs model loading, adapter integration, cache management, and parallel processing through an Engine object composed of three components. \texttt{\small {ModelAgent}} encapsulates the model, \texttt{\small {Scheduler}} handles resource allocation and sequence tracking, and \texttt{\small {RequestManager}} manages input and output for requests. In particular, the Scheduler uses a mechanism similar to vLLM's PagedAttention~\cite{kwon2023efficient} to allocate and release blocks based on the sequence length and supports S-LoRA~\cite{sheng2023s}, enabling multiple LoRA adapters to operate within limited memory.

Although LMDeploy features both \texttt{\small {TurboMind}} for high-performance inference and \texttt{\small {lmdeploy.pytorch}} for easier development, it currently supports only NVIDIA GPU environments.

\begin{findingbox}[]
    {\small \textbf{Representative Characteristics Summary}}
    \begin{itemize}[leftmargin=1.2em, label=--, itemsep=0em]
        \item \textbf{General-Purpose [\Low]}: Includes TurboMind and PyTorch engines but remains NVIDIA-only.
        \item \textbf{Ease-of-Deploy [\High]}: Docker images and a serve script ease installation, though driver matching is needed.
        \item \textbf{Ease-of-Use [\Medium]}: A unified API toggles between high-performance and development modes.
        \item \textbf{Latency-Aware [\Medium]}: KV-cache LRU and dynamic split-and-fuse significantly reduce prompt latency.
        \item \textbf{Throughput-Aware [\Medium]}: Persistent batching and continuous scheduling keep GPUs fully occupied.
        \item \textbf{Scalability [\High]}: Supports multiple GPUs per node; multi-node orchestration is still experimental.
    \end{itemize}
\end{findingbox}

\subsection{LightLLM} \label{sec:detailed_review_lightllm}

LightLLM~\cite{lightllm} is a Python-based, lightweight, and highly scalable LLM inference engine that addresses performance, scheduling, and memory inefficiencies in existing solutions. Using a three-process asynchronous collaboration approach, it separates tokenization, model inference, and detokenization to boost GPU utilization.

LightLLM replaces PagedAttention~\cite{kwon2023efficient} with \texttt{\small {TokenAttention}} and introduces \texttt{\small {Efficient Router Scheduling}}.  LightLLM uses an \texttt{\small {Efficient Router}} to manage GPU memory at a fine-grained, token-level granularity depending on whether it is in the prefill or decode phase. This router employs a custom algorithm to batch tokens appropriately. Additionally, the scheduling and model inference stages are merged, removing the communication overhead between the scheduler and the model-RPC. LightLLM also integrates OpenAI Triton~\cite{tillet2019triton} to optimize service scheduling kernels.

The inference engine consists of multiple modules, each running as a separate process (e.g., \texttt{\small {Metric Server}}, \texttt{\small {Health Server}}, \texttt{\small {HTTP Server}}, \texttt{\small {Router}}). These modules communicate via ZeroMQ~\cite{zeromq} or RPC. The \texttt{\small {Cache Manager}} stores multimodal inference results, while the \texttt{\small {Visual Server}} handles multimodal requests.

LightLLM also features a \texttt{\small {CacheTensorManager}} class to handle the allocation and deallocation of Torch tensors. By maximizing inter-layer tensor sharing during runtime and permitting memory sharing across distinct CUDA graphs, it reduces overall memory usage. A \texttt{\small {ModelBackend}} defines the mechanism and operations needed for prefill or decode requests from the router. Each backend maintains its own model object, supporting parallel existence of multiple backends. The model class performs computations on the device and includes tensor parallelism support.

\begin{findingbox}[]
    {\small \textbf{Representative Characteristics Summary}}
    \begin{itemize}[leftmargin=1.2em, label=--, itemsep=0em]
        \item \textbf{General-Purpose [\Low]}: TokenAttention backend offers a lightweight footprint for NVIDIA GPUs.
        \item \textbf{Ease-of-Deploy [\High]}: Manual source builds and custom dependencies increase setup complexity.
        \item \textbf{Ease-of-Use [\Medium]}: Multi-process ZeroMQ architecture and minimal docs raise the learning barrier.
        \item \textbf{Latency-Aware [\Medium]}: Triton-optimized kernels and router fusion shorten critical-path latency.
        \item \textbf{Throughput-Aware [\Medium]}: Efficient router scheduling and memory sharing maintain high TPS.
        \item \textbf{Scalability [\Medium]}: Multiple back ends can run concurrently; cluster scaling is manual.
    \end{itemize}
\end{findingbox}

\subsection{NanoFlow} \label{sec:detailed_review_nanoflow}

NanoFlow~\cite{zhu2024nanoflow} is a high-performance inference engine that improves LLM throughput by introducing \texttt{\small {Nano-batching}} and supporting co-scheduling of operations for intra-device parallelism. Traditional systems process pipelines sequentially, often underutilizing hardware resources.

By dividing batches into smaller  \texttt{\small {nano-batches}}, NanoFlow boosts optimization flexibility. It can also estimate GPU memory usage to check whether additional requests fit. If necessary, it offloads $\mathbf{KV}$ cache data to lower memory tiers---like system memory or disk---maximizing overall resource usage.

To implement \texttt{\small {Nano-batching}}, NanoFlow classifies LLM service operations into three types: \texttt{\small {memory-bound operations}} like self-attention computations, \texttt{\small {compute-bound operations}} such as General Matrix Multiplication (GEMM), and \texttt{\small {network-bound operations}} such as AllReduce. Then analyzes the resource requirements of each operation and the corresponding iterations or latencies to pinpoint performance characteristics and bottlenecks. Based on these findings, NanoFlow maximizes hardware parallelism to achieve higher throughput.

NanoFlow consists of three primary components. The \texttt{\small {global batch scheduler}} collects all incoming requests, creates dense batches in high-performance sizes (determined by offline profiling), and uses continuous batching~\cite{yu2022orca} technique to fill these batches dynamically. It also applies chunked prefill~\cite{agrawal2023sarathi} operations and a discrete batching approach, selecting only the batch sizes that were identified as optimal rather than arbitrary ones. By prioritizing throughput rather than focusing solely on latency, this method exploits available memory to process more requests in parallel.

Next, the intra-device parallelism engine enables fine-grained parallel operations for \texttt{\small {Nano-batching}}, along with execution unit scheduling to reduce interference among tasks. Lastly, the \texttt{\small {KV cache manager}} oversees the decoding status of every request, estimates future memory usage (assuming an average decode length), and manages GPU memory to prevent out-of-memory issues. If predicted usage does not exceed the GPU limits, the request is accepted; otherwise, it is deferred.

However, NanoFlow's \texttt{\small {Nano-batching}} mechanism requires additional setup---such as per-model schedule optimization---and may need pipeline adjustments or kernel re-implementation for new models. It also introduces overhead, potentially lowers efficiency for individual operations due to smaller batch sizes, and remains dependent on NVIDIA GPUs.

\begin{findingbox}[]
    {\small \textbf{Representative Characteristics Summary}}
    \begin{itemize}[leftmargin=1.2em, label=--, itemsep=0em]
        \item \textbf{General-Purpose [\Low]}: Operates solely on NVIDIA GPUs and demands per-model nano-schedule tuning.
        \item \textbf{Ease-of-Deploy [\Low]}: Research-grade code requires custom schedule files and environment tweaks.
        \item \textbf{Ease-of-Use [\Low]}: Sparse documentation and pipeline modifications limit accessibility.
        \item \textbf{Latency-Aware [\Medium]}: Memory forecasting and KV offload avoid OOM stalls, indirectly cutting latency.
        \item \textbf{Throughput-Aware [\Medium]}: Nano-batching plus intra-device parallelism greatly boost throughput.
        \item \textbf{Scalability [\Medium]}: Confined to a single node without distributed scheduling.
    \end{itemize}
\end{findingbox}

\subsection{DistServe} \label{sec:detailed_review_distserve}

DistServe~\cite{zhong2024distserve} is a serving system designed to efficiently run LLM inference across multiple GPU clusters while keeping latency low. It breaks down LLM inference requests at a granular level to enable parallel execution, thereby boosting throughput and resource utilization. Traditional inference engines handle prefill and decode on a single device, causing resource interference and pipeline inefficiencies. By decoupling them and applying both intra-operation and inter-operation parallelization via SwiftTransformer~\cite{swifttransformer}, DistServe reduces overhead.

DistServe also addresses large model sizes, such as a 175B-parameter model that can require 350GB of memory. It uses a low node-affinity placement algorithm for batch allocation, relying on NVLink when computations for a given stage remain on the same node. Online scheduling further manages workloads in real time to meet latency SLO requirements.

DistServe consists of a batching algorithm module, a \texttt{\small {RESTful API  frontend}}, an \texttt{\small {orchestration layer}}, and a \texttt{\small {parallel execution engine}}. The batching module provides a simulator and algorithms to optimally distribute requests based on particular models and cluster setups. The \texttt{\small {RESTful API  frontend}} supports an OpenAI-compatible interface and accepts user inputs such as maximum output length and temperature. The \texttt{\small {orchestration layer}} manages prefill and decode instances, handles request dispatching, and coordinates $\mathbf{KV}$ cache transfers. For inter-node GPU communication, DistServe uses NCCL~\cite{nccl}, while intra-node transfers rely on asynchronous memory copy. Individual instances run as GPU workers through Ray~\cite{moritz2018ray}, driven by a \texttt{\small {parallel execution engine}}.

Because DistServe is intended for large GPU clusters, its parallel strategies and resource allocations can be difficult to adapt to smaller-scale or resource-constrained settings (e.g., single or few-GPU systems), potentially limiting performance in those scenarios.

\begin{findingbox}[]
    {\small \textbf{Representative Characteristics Summary}}
    \begin{itemize}[leftmargin=1.2em, label=--, itemsep=0em]
        \item \textbf{General-Purpose [\Low]}: Aims at very large models across multi-GPU clusters.
        \item \textbf{Ease-of-Deploy [\Low]}: Requires Ray cluster setup and NVLink topology awareness.
        \item \textbf{Ease-of-Use [\Low]}: Placement-algorithm tuning and orchestration add complexity for operators.
        \item \textbf{Latency-Aware [\Medium]}: Decoupled prefill and decode phases reduce tail latency under load.
        \item \textbf{Throughput-Aware [\High]}: Intra- and inter-operation parallelization plus low node-affinity batching maximize throughput.
        \item \textbf{Scalability [\High]}: Designed for multi-node clusters, scaling to hundreds of GPUs.
    \end{itemize}
\end{findingbox}

\subsection{vAttention} \label{sec:detailed_review_vattention}

vAttention~\cite{prabhu2025vattention} is a inference engine for dynamically managing $\mathbf{KV}$ cache memory during LLM inference. Built on Sarathi-Serve~\cite{agrawal2023sarathi}, it includes components such as \texttt{\small {sarathi-lean}}, a \texttt{\small {vattention memory allocator}}, and a \texttt{\small {custom Unified Virtual Memory (UVM) driver}}. These elements support both PagedAttention~\cite{kwon2023efficient} and vAttention-style memory management.

vAttention addresses the complexity and performance limitations linked to virtual contiguity in PagedAttention---commonly used in transformer-based LLMs. It enhances performance (especially in prefill-bound workloads) while staying compatible with existing kernels. To achieve this, vAttention modifies PyTorch~\cite{paszke2019pytorch} caching allocator to introduce virtual tensors, reserving virtual memory buffers without allocating physical memory from the start.

Unlike PagedAttention, where LLM serving systems must manually handle mappings between $\mathbf{KV}$ cache and dynamic memory blocks, vAttention integrates memory allocation and computation and enables predictive page allocation. It separates virtual and physical memory usage via low-level CUDA APIs (rather than \texttt{\small {cudaMalloc}}), and supports optimizations that target NVIDIA's Hopper architecture through FlashAttention-3~\cite{shah2024flashattention}, restricting it to NVIDIA GPUs.

vAttention is implemented as a Python library that wraps CUDA/C++ extension libraries that interfacing with the CUDA driver. During model serving, each worker sets up vAttention based on model parameters and page group sizes, allocating virtual tensors as needed. It checks whether the $\mathbf{KV}$ cache is mapped to physical memory before launching kernels, tracking page allocations during both prefill and decode. Only when all current pages are used does it allocate new pages and it frees or reclaims pages once a request ends.

\begin{findingbox}[]
    {\small \textbf{Representative Characteristics Summary}}
    \begin{itemize}[leftmargin=1.2em, label=--, itemsep=0em]
        \item \textbf{General-Purpose [\Low]}: Tailored for NVIDIA Hopper GPUs, limiting portability.
        \item \textbf{Ease-of-Deploy [\Low]}: CUDA driver patches and custom UVM setup complicate installation.
        \item \textbf{Ease-of-Use [\Low]}: Experimental wrapper and minimal docs hamper quick adoption.
        \item \textbf{Latency-Aware [\Medium]}: Predictive page allocation hides memory-map costs and speeds prefill.
        \item \textbf{Throughput-Aware [\Medium]}: Integrated KV memory and compute paths provide moderate gains.
        \item \textbf{Scalability [\Medium]}: Currently supports only single-node, single-GPU execution.
    \end{itemize}
\end{findingbox}

\subsection{Sarathi-Serve} \label{sec:detailed_review_sarithiserve}

Sarathi-Serve~\cite{agrawal2024taming} is a high-performance inference scheduler built on vLLM~\cite{kwon2023efficient} to address the trade-off between throughput and latency in LLM inference. It relies on FlashAttention-2~\cite{dao2023flashattention} and FlashInfer~\cite{ye2025flashinfer} as backends to enhance decode-stage throughput in multi-GPU and multi-node environments.

Previous systems, such as Orca~\cite{yu2022orca} and vLLM~\cite{kwon2023efficient}, faced generation stalls---where decode requests wait because of prolonged prefill---and pipeline inefficiencies---where insufficient parallelism at the request level left GPU resources underused. Sarathi-Serve tackles these problems via chunked prefill and stall-free scheduling, cutting down TBT while offering high throughput and minimal TBT latency.

Sarathi-Serve decides the maximum number of tokens (\texttt{\small {token budget}}) in each batch based on TBT SLOs and chunked prefill overhead. Under strict latency requirements, it sets a smaller token budget and splits prompts into smaller chunks, lowering tail latency at the cost of some overall system efficiency. Under looser latency constraints, it raises the token budget to improve prefill efficiency. With token budgets like 2,048 or 512, Sarathi-Serve provides efficient inference for varying SLO conditions.

\begin{findingbox}[]
    {\small \textbf{Representative Characteristics Summary}}
    \begin{itemize}[leftmargin=1.2em, label=--, itemsep=0em]
        \item \textbf{General-Purpose [\Low]}: Extends vLLM scheduling to multiple model categories.
        \item \textbf{Ease-of-Deploy [\Low]}: A simple CLI launches servers, yet CUDA and NCCL versions must align.
        \item \textbf{Ease-of-Use [\Low]}: Interactive SLO slider lets users trade latency for throughput with ease.
        \item \textbf{Latency-Aware [\Medium]}: Chunked prefill and stall-free scheduling keep TBT consistently low.
        \item \textbf{Throughput-Aware [\Medium]}: Token-budget batching adapts to workload for maximum throughput.
        \item \textbf{Scalability [\Medium]}: Multi-GPU and multi-node deployments are supported via FlashAttention-2.
    \end{itemize}
\end{findingbox}

\subsection{Friendli Inference} \label{sec:detailed_review_friendli}

Friendli Inference~\cite{friendli} is a commercial LLM inference engine built on top of Orca~\cite{yu2022orca} and designed to enhance inference through features such as iteration batching. It supports both web- and API-based serving via \texttt{\small {Friendli Container}} and \texttt{\small {Friendli Serverless/Dedicated Endpoints}}, with the latter focusing on stable service by managing traffic and adhering to service-level agreements (SLAs). Users can integrate Friendli AI solutions with Amazon SageMaker~\cite{sagemakerai} and gRPC~\cite{grpc} inference servers, while monitoring is facilitated through tools like Grafana~\cite{chakraborty2021grafana}.

For optimization, Friendli Inference allows serving multiple LoRA~\cite{hu2022lora} models on a single GPU, maximizing utilization of different user-defined models. It introduces \texttt{\small {TCache}}, a GPU load reduction technique that caches frequently accessed results to maintain a high TTFT compared to conventional frameworks. Quantization techniques are also used to further improve the inference performance. However, Friendli Inference primarily targets NVIDIA GPUs, limiting its support on other hardware platforms.

\begin{findingbox}[]
    {\small \textbf{Representative Characteristics Summary}}
    \begin{itemize}[leftmargin=1.2em, label=--, itemsep=0em]
        \item \textbf{General-Purpose [\Low]}: Focuses on hosting multiple LoRA models on NVIDIA GPUs.
        \item \textbf{Ease-of-Deploy [\High]}: Container and serverless options plus SageMaker integration simplify rollout.
        \item \textbf{Ease-of-Use [\Medium]}: Web console with Grafana metrics streamlines monitoring and management.
        \item \textbf{Latency-Aware [\Medium]}: TCache lowers time-to-first-token, though deeper latency tooling is not described.
        \item \textbf{Throughput-Aware [\Medium]}: Iteration batching and quantization deliver strong tokens-per-second performance.
        \item \textbf{Scalability [\Medium]}: Dedicated and serverless GPU instances scale horizontally, but non-GPU back ends are absent.
    \end{itemize}
\end{findingbox}

\subsection{Fireworks AI} \label{sec:detailed_review_fireworks}

Fireworks AI~\cite{fireworks} is a inference platform for rapid and efficient serving of both LLMs and image models, supporting inference, fine-tuning, and deployment. It offers a simple interface and APIs---compatible with services like LangChain~\cite{LangChain} and the OpenAI API---and can operate in serverless, on-demand, or enterprise environments. Beyond NVIDIA GPUs, Fireworks AI also supports LLM inference on AMD Instinct MI300X~\cite{smith2024amd}, broadening its hardware compatibility.

To meet diverse throughput and latency demands, Fireworks AI uses multiple parallelization and optimization techniques, including multi/group query attention optimization, sharding, quantization, and continuous batching. It provides deployment configurations specifically tailored for low latency, high throughput, or long input/output sequences. In particular, Fireworks AI employs its own MQA~\cite{shazeer2019fast} model and a custom CUDA kernel (\texttt{\small {FireAttention}}) to further accelerate inference.

To ensure service reliability, Fireworks AI is SOC 2 Type II~\cite{aicpa2022soc2} and HIPAA~\cite{marron2024implementing} certified, ensuring privacy, security, availability, processing integrity and confidentiality.

\begin{findingbox}[]
    {\small \textbf{Representative Characteristics Summary}}
    \begin{itemize}[leftmargin=1.2em, label=--, itemsep=0em]
        \item \textbf{General-Purpose [\Low]}: Serves LLMs and vision models on NVIDIA GPUs and AMD MI300X accelerators.
        \item \textbf{Ease-of-Deploy [\High]}: Cloud console offers API keys and quick templates for inference or fine-tuning.
        \item \textbf{Ease-of-Use [\Medium]}: OpenAI-compatible endpoints and LangChain adapters reduce integration effort.
        \item \textbf{Latency-Aware [\Medium]}: FireAttention kernels and low-latency deployment profiles minimize response time.
        \item \textbf{Throughput-Aware [\Medium]}: Multi-query attention, sharding, and continuous batching provide high TPS.
        \item \textbf{Scalability [\High]}: Serverless and enterprise clusters scale elastically with SOC 2 and HIPAA compliance.
    \end{itemize}
\end{findingbox}

\subsection{GroqCloud} \label{sec:detailed_review_groq}

GroqCloud~\cite{groqcloud} is an AI infrastructure platform focused on delivering high-performance, low-latency inference for LLMs through a specialized hardware architecture and software stack. It aims to resolve bottlenecks and non-deterministic behaviors commonly found in conventional GPU systems by offering inference services powered by the Groq Language Processing Unit (LPU)~\cite{abts2022software, abts2020think}. Built specifically for AI inference, the Groq LPU achieves lower latency and higher throughput than traditional GPUs. Its Tensor Streaming Processor (TSP) architecture~\cite{abts2020think} statically schedules and locks the model execution path at compile time, eliminating runtime variability and enabling predictable response times.

A key advantage of Groq LPU is its ability to maintain optimal performance even with a batch size of one, making it well suited for latency-sensitive applications like financial trading and autonomous driving. Through its \texttt{\small {TruePoint}} technology, the Groq LPU delivers near-FP32 precision when using FP16 or INT8 computations. For high-throughput workloads, GroqCloud provides an asynchronous batching API, Flex Processing for scalable throughput, and deterministic QoS scheduling to meet a range of SLAs. Additionally, its kernel-less compiler approach removes the need for manual kernel optimization, simplifying development, and lowering maintenance overhead.

However, because GroqCloud relies on static compilation, it may have limited flexibility in dynamically adjusting batch sizes or handling complex runtime branching.

\begin{findingbox}[]
    {\small \textbf{Representative Characteristics Summary}}
    \begin{itemize}[leftmargin=1.2em, label=--, itemsep=0em]
        \item \textbf{General-Purpose [\Low]}: Runs exclusively on Groq LPUs, limiting hardware flexibility.
        \item \textbf{Ease-of-Deploy [\High]}: Fully managed cloud API hides all compilation and runtime details.
        \item \textbf{Ease-of-Use [\Low]}: Deterministic QoS and simple REST calls facilitate rapid integration.
        \item \textbf{Latency-Aware [Low]}: TSP architecture yields sub-millisecond latency even at batch size 1.
        \item \textbf{Throughput-Aware [\Medium]}: Flex Processing sustains high throughput without hurting single-request latency.
        \item \textbf{Scalability [\High]}: LPU pods scale horizontally under the same deterministic schedule.
    \end{itemize}
\end{findingbox}

\subsection{Together Inference} \label{sec:detailed_review_together}

Together Inference~\cite{together} is part of the Together AI platform, offering high-performance LLM inference with an emphasis on speed, cost, and accuracy. To enhance LLM serving, it implements transformer-optimized CUDA kernels, quantization~\cite{egashira2025exploiting}, and speculative decoding~\cite{leviathan2023fast}. Together Inference provides different model configurations to meet diverse needs, from maximum performance and full-precision accuracy to lower cost and higher throughput. It supports dedicated instances, serverless deployments, and multi-GPU environments; however, it is optimized exclusively for NVIDIA GPU-based services.

\begin{findingbox}[]
   {\small \textbf{Representative Characteristics Summary}}
    \begin{itemize}[leftmargin=1.2em, label=--, itemsep=0em]
        \item \textbf{General-Purpose [\Low]}: Offers multiple accuracy-versus-cost tiers, but only on NVIDIA GPUs.
        \item \textbf{Ease-of-Deploy [\High]}: Provides serverless endpoints and dedicated instances through a web UI.
        \item \textbf{Ease-of-Use [\Medium]}: OpenAI-style API eases migration for existing clients.
        \item \textbf{Latency-Aware [\Medium]}: Speculative decoding and custom CUDA kernels reduce median and tail latencies.
        \item \textbf{Throughput-Aware [\Low]}: Quantization, optimized attention, and continuous batching significantly raise TPS.
        \item \textbf{Scalability [\High]}: Multi-GPU instances scale vertically; adding endpoints enables horizontal growth.
    \end{itemize}
\end{findingbox}

\section{\news{LLM Inference Optimization}} \label{sec:inference_optimization}

LLM inference performance depends not only on the model size and hardware environment but also on various inference optimization techniques. The specialized LLM inference engines introduced earlier seek low latency and high throughput by employing parallelism, efficient memory management, kernel optimization, and quantization.

This section explains key inference optimization strategies, including parallel processing, memory optimization, latency, and throughput optimization. In addition to the methods provided by existing inference engines, we also examine recent research findings on inference optimization. Table~\ref{tab:frameworks_optimization} summarizes the optimization techniques supported by each LLM inference engine.

\begin{table}[tbp]
    \centering
    \caption{Optimizations of LLM Inference Engines}
    \label{tab:frameworks_optimization}
    
    \resizebox{.99\textwidth}{!}{%
    \begin{tabular}{l cccc ccccc ccc cccc cc c c}
    \toprule
    \multirow{9.5}{*}{Engines} 
        & \multicolumn{4}{c}{\makecell{Batch\\Optimization}} 
        & \multicolumn{5}{c}{\makecell{Parallelism}} 
        & \multicolumn{3}{c}{\makecell{Compression}} 
        & \makecell{Fine-\\tuning}
        & \multicolumn{3}{c}{\makecell{Caching}}
        & \multicolumn{2}{c}{\makecell{Attention\\Optimization}}
        & \multicolumn{1}{c}{\makecell{Sampling\\Optimization}} 
        & \multicolumn{1}{c}{\makecell{Output\\Optimization}} \\
        \cmidrule(lr){2-5} \cmidrule(lr){6-10} \cmidrule(lr){11-13}  \cmidrule(lr){14-14} \cmidrule(lr){15-17} 
        \cmidrule(lr){18-19} \cmidrule(lr){20-20} \cmidrule(lr){21-21}     
        & \rotatebox{90}{Dynamic Batching}
        & \rotatebox{90}{Continuous Batching}
        & \rotatebox{90}{Nano Batching}
        & \rotatebox{90}{Chunked-prefills}
        & \rotatebox{90}{Data Parallelism}
        & \rotatebox{90}{\makecell{Fully Sharded \\ Data Parallel}}
        & \rotatebox{90}{\news{Expert Parallelism}}
        & \rotatebox{90}{Tensor Parallelism}
        & \rotatebox{90}{Pipeline Parallelism}
        & \rotatebox{90}{\makecell{Quantization$^\dagger$/\\Support \\Quantized Model}}
        & \rotatebox{90}{Pruning}
        & \rotatebox{90}{Sparsity$^\ddagger$}
        & \rotatebox{90}{LoRA}
        & \rotatebox{90}{Prompt Caching}
        & \rotatebox{90}{Prefix Caching}
        & \rotatebox{90}{KV Caching}
        & \rotatebox{90}{PagedAttention}
        & \rotatebox{90}{FlashAttention}
        & \rotatebox{90}{Speculative Decoding}
        & \rotatebox{90}{Structured Outputs$^\ast$} \\
    \midrule
    Ollama~\cite{ollama}  
        & \redxmark & \redxmark & \redxmark & \redxmark 
        & \redxmark & \redxmark & \redxmark & \greencheck & \greencheck 
        & \greencheck & \greencheck & \makecell{\greencheck \\[-7pt] \tiny (MoE, SL)}
        & \greencheck
        & \greencheck & \redxmark & \greencheck
        & \redxmark   & \greencheck 
        & \greencheck & \makecell{\greencheck \\[-7pt] \tiny (GG)} \\

    llama.cpp~\cite{llamacpp}  
        & \redxmark & \greencheck & \redxmark & \redxmark 
        & \redxmark & \redxmark & \redxmark & \greencheck & \greencheck 
        & \greencheck & \redxmark & \makecell{\greencheck \\[-7pt] \tiny (MoE, SL)}
        & \greencheck
        & \greencheck & \redxmark & \greencheck
        & \redxmark  & \greencheck 
        & \greencheck & \makecell{\greencheck \\[-7pt] \tiny (GG)} \\
        
    vLLM~\cite{kwon2023efficient}  
        & \redxmark & \greencheck & \redxmark & \greencheck
        & \greencheck & \greencheck &  \greencheck & \greencheck & \greencheck
        & \makecell{\greencheck \\[-7pt] \tiny (A, A$^*$, B, D, G, L, M)} & \greencheck & \makecell{\greencheck \\[-7pt] \tiny (BS, MoE, N:M)} 
        & \greencheck
        & \redxmark & \greencheck & \greencheck
        & \greencheck  & \makecell{\greencheck \\[-7pt] \tiny (v3)} 
        & \greencheck & \makecell{\greencheck \\[-7pt] \tiny (LM, OA, OL, XG)} \\
        
    DeepSpeed-FastGen~\cite{holmes2024deepspeed}  
        & \redxmark & \greencheck & \redxmark & \greencheck
        & \greencheck & \greencheck &  \greencheck & \greencheck & \greencheck
        & \greencheck & \makecell{\greencheck \\[-7pt] \tiny (S, U)} & \makecell{\greencheck \\[-7pt] \tiny (N:M, NxM, MoE, SA)}  
        & \greencheck 
        & \redxmark & \redxmark & \greencheck
        & \greencheck & \makecell{\greencheck \\[-7pt] \tiny (v2)} 
        & \redxmark & \redxmark \\

    Unsloth~\cite{unsloth}  
        & \redxmark & \redxmark & \redxmark & \redxmark 
        & \redxmark & \redxmark & \redxmark & \redxmark & \redxmark
        & \makecell{\greencheck \\[-7pt] \tiny (B)} & \redxmark & \redxmark
        & \greencheck
        & \redxmark & \redxmark & \greencheck
        & \redxmark & \greencheck 
        & \redxmark & \redxmark \\

    MAX~\cite{max}  
        & \redxmark & \greencheck & \redxmark & \greencheck 
        & \redxmark & \redxmark & \greencheck & \greencheck & \redxmark
        & \greencheck & \redxmark & \makecell{\greencheck \\[-7pt] \tiny (MoE)}
        & \greencheck
        & \redxmark & \greencheck & \greencheck
        & \greencheck & \makecell{\greencheck \\ [-7pt]\tiny (v3)} 
        & \greencheck & \makecell{\greencheck \\[-7pt] \tiny (XG)} \\

    MLC-LLM~\cite{mlcllm}  
        & \redxmark & \greencheck & \redxmark & \greencheck 
        & \redxmark & \redxmark &  \redxmark  & \greencheck & \greencheck
        & \makecell{\greencheck \\ [-7pt]\tiny (A)} & \redxmark & \makecell{\greencheck \\[-7pt] \tiny (MoE)}
        & \redxmark
        & \redxmark & \greencheck & \greencheck
        & \greencheck  & \redxmark
        & \greencheck & \makecell{\greencheck \\[-7pt] \tiny (XG)} \\

    llama2.c~\cite{llama2c}  
        & \redxmark & \redxmark & \redxmark & \redxmark 
        & \redxmark & \redxmark &  \redxmark  & \redxmark & \redxmark
        & \greencheck & \redxmark & \redxmark
        & \redxmark
        & \redxmark & \redxmark & \greencheck
        & \redxmark   & \redxmark 
        & \redxmark & \redxmark \\

    bitnet.cpp~\cite{wang20241}  
        & \redxmark & \redxmark & \redxmark & \redxmark 
        & \redxmark & \redxmark &  \redxmark  & \redxmark & \redxmark
        & \makecell{\greencheck \\ [-7pt]\tiny (A, G)} & \redxmark & \makecell{\greencheck \\[-7pt] \tiny (MoE)}
        & \redxmark
        & \redxmark & \redxmark & \greencheck
        & \redxmark   & \redxmark 
        & \redxmark & \redxmark \\

    SGLang~\cite{zheng2024sglang}  
        & \redxmark & \greencheck & \redxmark & \greencheck
        & \greencheck & \greencheck & \greencheck & \greencheck & \redxmark
        & \makecell{\greencheck \\ [-7pt]\tiny (A, B, G, L, M)} & \greencheck & \makecell{\greencheck \\[-7pt] \tiny (DSA, N:M, MoE)}
        & \greencheck
        & \redxmark & \greencheck & \greencheck
        & \greencheck  & \redxmark 
        & \greencheck & \makecell{\greencheck \\[-7pt] \tiny (LL, OL, XG)} \\

    LitGPT~\cite{litgpt}  
        & \redxmark & \greencheck & \redxmark & \redxmark
        & \greencheck & \greencheck & \redxmark & \greencheck & \redxmark
        & \makecell{\greencheck \\ [-7pt]\tiny (B)} & \redxmark & \makecell{\greencheck \\[-7pt] \tiny (MoE)}
        & \greencheck
        & \redxmark & \redxmark & \greencheck
        & \redxmark   & \makecell{\greencheck \\ [-7pt] \tiny (v2)} 
        & \greencheck & \redxmark \\

    OpenLLM~\cite{openllm}  
        & \redxmark & \greencheck & \redxmark & \redxmark
        & \greencheck & \redxmark & \redxmark & \redxmark & \redxmark
        & \makecell{\greencheck \\[-7pt] \tiny (B, G)} & \redxmark & \redxmark
        & \redxmark
        & \redxmark & \redxmark & \redxmark
        & \redxmark  & \redxmark
        & \redxmark & \redxmark \\

    TensorRT-LLM~\cite{tensorrtllm}  
        & \greencheck & \greencheck & \redxmark & \greencheck
        & \greencheck & \redxmark & -- & \greencheck & \greencheck
        & \makecell{\greencheck \\[-7pt]  \tiny (A, G, S, W)} & \greencheck & \makecell{\greencheck \\[-7pt] \tiny (BSA, MoE)}
        & \greencheck
        & \greencheck & \redxmark & \greencheck
        & \greencheck  & \redxmark
        & \greencheck & \makecell{\greencheck \\[-7pt] \tiny (XG)} \\
        
    TGI~\cite{tgi}  
        & \redxmark & \greencheck & \redxmark & \redxmark 
        & \redxmark & \redxmark & \greencheck & \greencheck & \redxmark
        & \makecell{\greencheck \\[-7pt] \tiny (A, E, E$^*$, G, M)} & \greencheck & \makecell{\greencheck \\[-7pt] \tiny (N:M, MoE)}
        & \greencheck
        & \redxmark & \greencheck & \greencheck
        & \greencheck & \makecell{\greencheck \\ [-7pt] \tiny (v2)} 
        & \greencheck & \makecell{\greencheck \\[-7pt] \tiny (OL)} \\

    PowerInfer~\cite{song2024powerinfer, xue2024powerinfer}  
        & \redxmark & \greencheck & \redxmark & \redxmark 
        & \greencheck & \redxmark & \redxmark & \redxmark & \greencheck
        & \greencheck & \redxmark & \greencheck
        & \greencheck
        & \greencheck & \redxmark & \greencheck
        & \redxmark & \greencheck 
        & \greencheck & \makecell{\greencheck \\[-7pt] \tiny (GG)} \\

    LMDeploy~\cite{2023lmdeploy}  
        & \redxmark & \greencheck & \redxmark & \greencheck 
        & \redxmark & \redxmark & \greencheck & \greencheck & \redxmark
        & \makecell{\greencheck \\ [-7pt] \tiny (A, G, S)} & \greencheck & \makecell{\greencheck \\[-7pt] \tiny (MoE)}
        & \greencheck
        & \redxmark & \greencheck & \greencheck
        & \greencheck  & \redxmark
        & \redxmark & \makecell{\greencheck \\[-7pt] \tiny (PT)} \\

    LightLLM~\cite{lightllm}  
        & \greencheck & \redxmark & \redxmark & \greencheck 
        & \redxmark & \redxmark &\greencheck & \greencheck & \redxmark
        & \greencheck & \redxmark & \makecell{\greencheck \\[-7pt] \tiny (MoE)}
        & \redxmark
        & \greencheck & \redxmark & \greencheck
        & \redxmark   & \makecell{\greencheck \\ [-7pt] \tiny (v1)}
        & \redxmark & \makecell{\greencheck \\[-7pt] \tiny (OL, XG)} \\

    NanoFlow~\cite{zhu2024nanoflow}  
        & \redxmark & \greencheck & \greencheck & \greencheck
        & \greencheck & \redxmark & \redxmark  & \redxmark & \redxmark
        & \redxmark & \redxmark & \redxmark
        & \redxmark
        & \redxmark & \redxmark & \greencheck
        & \redxmark & \redxmark 
        & \redxmark & \redxmark \\

    DistServe~\cite{zhong2024distserve}  
        & \greencheck & \greencheck & \redxmark & \greencheck 
        & \redxmark & \redxmark & \redxmark & \greencheck & \greencheck
        & \redxmark & \redxmark & \redxmark
        & \redxmark
        & \redxmark & \redxmark & \greencheck
        & \greencheck & \makecell{\greencheck \\ [-7pt] \tiny (v1)}
        & \redxmark & \redxmark \\

    vAttention~\cite{prabhu2025vattention}  
        & \greencheck & \redxmark & \redxmark & \redxmark
        & \greencheck & \redxmark & \redxmark & \greencheck & \greencheck
        & \greencheck & \greencheck & \makecell{\greencheck \\[-7pt] \tiny (N:M)}
        & \greencheck
        & \redxmark & \redxmark & \greencheck
        & \greencheck  & \makecell{\greencheck \\[-7pt] \tiny (v2)} 
        & \redxmark & \redxmark \\

    Sarathi-Serve~\cite{agrawal2024taming}  
        & \redxmark & \redxmark & \redxmark & \greencheck 
        & \redxmark & \redxmark & \redxmark & \greencheck & \greencheck
        & \redxmark & \redxmark & \makecell{\greencheck \\[-7pt] \tiny (MoE)}
        & \redxmark
        & \redxmark & \redxmark & \greencheck
        & \redxmark & \makecell{\greencheck \\[-7pt] \tiny (v2)} 
        & \redxmark & \redxmark \\

    Friendli Inference~\cite{friendli}  
        & -- & \greencheck & -- & --
        & -- & -- & -- & \greencheck & \greencheck
        & \greencheck & -- & \makecell{\greencheck \\[-7pt] \tiny (MoE)}
        & \greencheck
        & -- & -- & --
        & --& -- 
        & \greencheck & \greencheck \\

    Fireworks AI~\cite{fireworks}  
        & -- & \greencheck & -- & --
        & -- & -- & -- & -- & --
        & \greencheck & \greencheck & \makecell{\greencheck \\[-7pt] \tiny (MoE)}
        & \greencheck
        & \greencheck & \greencheck & \greencheck
        & --  & -- 
        & \greencheck & \makecell{\greencheck \\[-7pt] \tiny (OA)} \\

    Groq Cloud~\cite{groqcloud}  
        & \redxmark & -- & -- & --
        & \greencheck & -- & -- & \greencheck & \greencheck
        & \greencheck & \greencheck & \makecell{\greencheck \\[-7pt] \tiny (MoE)}
        & --
        & -- & -- & --
        & -- & -- 
        & \greencheck & \makecell{\greencheck \\[-7pt] \tiny (OA)} \\

    Together Inference~\cite{together}  
        & -- & -- & -- & --
        & -- & \greencheck  & -- & -- & --
        & \greencheck & -- & \makecell{\greencheck \\[-7pt] \tiny (MoE)}
        & \greencheck
        & \greencheck & -- & --
        & -- & \makecell{\greencheck \\ [-7pt] \tiny (v3)} 
        & \greencheck & \greencheck \\
    \bottomrule
    \end{tabular}%
    }
    \vspace{-0.5mm}
    \flushleft
    \hspace{1mm} \tiny{$^\dagger$A: AWQ, A$^*$: AQLM, B: bitsandbytes, D: DeepSpeedFP, E: EXL2, E$^*$: EETQ, G: GPTQ, L: LLM Compressor, M: Marlin, S: SmoothQuant, W: Weight-only}
    \\
    \hspace{1mm} \tiny{$^\ddagger$BS: Block Sparse, BSA: Block Sparse Attention, DSA: Double Sparse Attention, SL: SparseLLM}
    \\
    \hspace{1mm} \tiny{$^\ast$GG: GBNF, LL: llguidance, LM: lm-formet-enforcer, OA: OpenAI API, OL: outlines, PT: PyTorch, XG: XGrammar}
\end{table}

\begin{figure}[tbp]
    \centering
    \includegraphics[width=.95\linewidth]{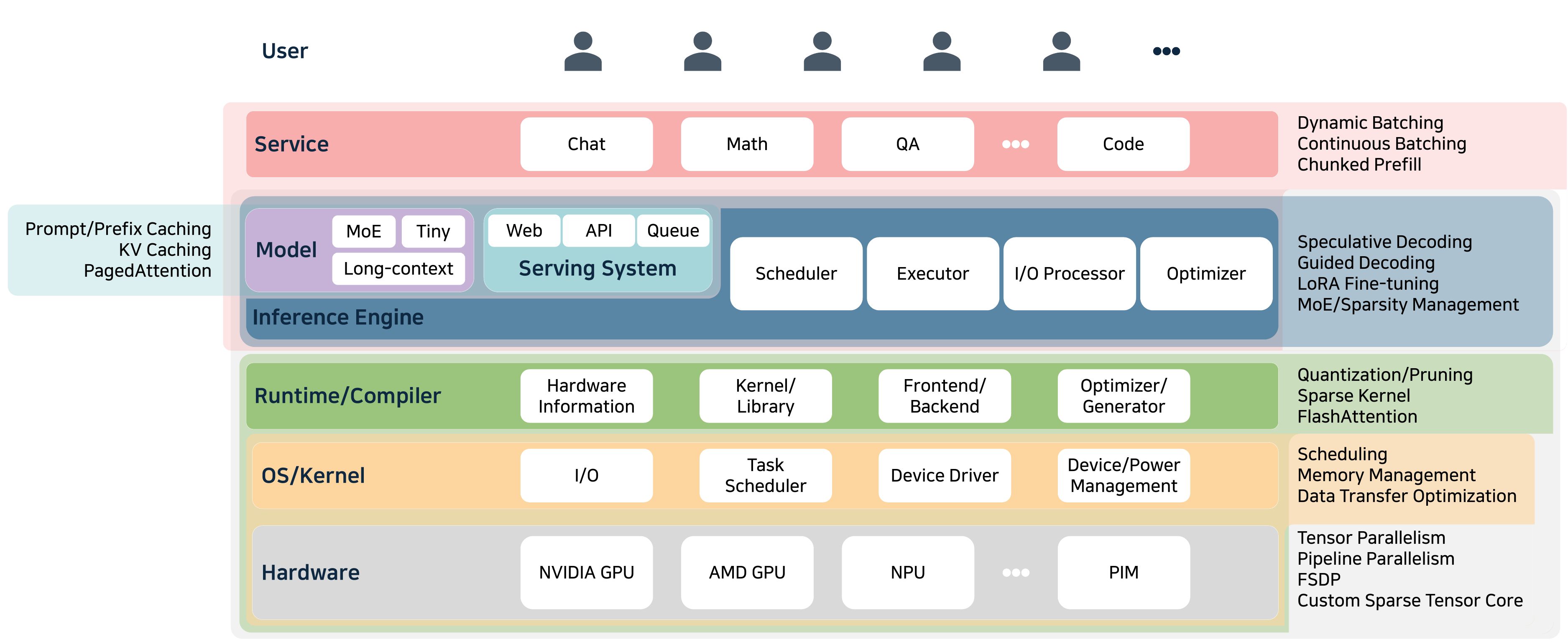}
    \caption{\news{Optimization Techniques across the LLM Inference}}
    \label{fig:optimization_overview}
\end{figure}

\news{Figure~\ref{fig:optimization_overview} presents full system-level layering of the LLM inference optimization techniques summarized in Table~\ref{tab:frameworks_optimization}. Effective LLM inference optimization is not attainable by improving a single hardware or software component, it requires an end-to-end strategy that spans from the service layer through the hardware layer, embracing co-design and cross-layer coordination.}

At the \texttt{Service} layer, techniques such as dynamic batching~\cite{crankshaw2017clipper, ali2020batch} and continuous batching~\cite{yu2022orca, he2024deferred} are used to efficiently handle multiple user requests. These batching optimizations balance latency and throughput and serve as critical factors to ensure quality in large-scale service environments.

The \texttt{Model} and \texttt{Serving System} layer, optimizations are applied to align with model architecture and serving methods. Prompt Caching~\cite{zhu2024efficient}, Prefix Caching~\cite{liu2024optimizing, pan2024marconi}, and $\mathbf{KV}$ Caching~\cite{pope2023efficiently} reduce memory usage and computation for long-context processing, while MoE~\cite{cai2024survey,fedus2022switch,du2022glam} and sparsity-based methods~\cite{yang2024post, fu2024lazyllm} decrease internal computation to allow efficient inference.

The \texttt{Inference Engine} layer directly improves token-level generation. Speculative Decoding and Guided Decoding~\cite{li2024eagle, cai2024medusa, cheng2024recurrent} accelerate token prediction, while fine-tuning methods such as LoRA~\cite{hu2022lora,dettmers2023qlora} reduce computational resource demands.

At the \texttt{Runtime} and \texttt{Compiler} layer, optimizations are performed at the graph and kernel level, including quantization~\cite{li2023fptq, chen2024efficientqat, egiazarian2024extreme, xiao2023smoothquant}, pruning~\cite{cusparse,sun2023simple,fu2024lazyllm}, sparse kernels~\cite{zhang2023dynamic,xia2023flash,gao2024seerattention,zhang2022learning}, and kernel fusion~\cite{filipovivc2015optimizing, sun2024much, zhao2025hape}. These techniques maximize FLOPs utilization and memory bandwidth efficiency, ensuring stable high performance on actual hardware.

The \texttt{OS} and \texttt{Kernel} layer manages resources and data transfer. Task scheduling and memory management enhance the utilization of GPU, I/O, and network resources, while ensuring that memory-intensive structures, such as the $\mathbf{KV}$ cache, operate efficiently.

The \texttt{Hardware} layer provides the foundation for large-scale parallel execution. Parallelism~\cite{rajbhandari2020zero,zhao2023pytorch,stojkovic2024towards,agrawal2023sarathi} and device-to-device interconnection distribute model parameters and computation across multiple accelerators, while specialized features such as sparse tensor cores~\cite{fan2025spinfer} provide direct support for sparsity optimization.

\begin{table}[tbp]
    \caption{\news{Optimization Techniques in LLM Inference: Points and Suitable Workloads}}
    \label{tab:llm_inference_optimizations}
    \centering
    \resizebox{.85\textwidth}{!}{%
    \begin{tabular}{@{}>{\centering\arraybackslash}p{0.3\textwidth} p{0.48\textwidth} p{0.48\textwidth}@{}}
    \toprule
    \multicolumn{1}{c}{\news{Optimization Technique}} & \multicolumn{1}{c}{\news{Inference Optimization Point}} & \multicolumn{1}{c}{\news{Suitable Workload}} \\
    \midrule
        \news{Batch Optimizations (\S\ref{sec:inference_optimization_batch})}
        & \news{Increase GPU utilization by grouping requests; balance latency vs. throughput}
        & \news{Multi-user chatbot services, API serving, large-scale concurrent requests} \\
        
        \news{Parallelism (\S\ref{sec:inference_optimization_parallelism})}
        & \news{Distribute model/compute across GPUs and nodes; enable handling of ultra-large models}
        & \news{70B+ model serving, distributed clusters, enterprise/research-scale LLMs} \\
        
        \news{Compression (\S\ref{sec:inference_optimization_compression})}
        & \news{Reduce memory footprint and FLOPs; enable execution on resource-limited hardware}
        & \news{Edge devices, mobile deployment, personal or lightweight inference servers} \\
        
        \news{Inference-aware Fine-tuning (\S\ref{sec:inference_optimization_fine_tuning})}
        & \news{Train only small adapter layers for domain-specific adaptation}
        & \news{Enterprise-specific chatbots, domain QA, customized LLMs with low training cost} \\
        
        \news{Caching (\S\ref{sec:inference_optimization_caching})}
        & \news{Reuse prefix and past tokens to avoid redundant computation}
        & \news{RAG-based QA, conversational assistants with repeated context, long-context summarization} \\
        
        \news{Attention Optimization (\S\ref{sec:inference_optimization_attention})}
        & \news{Reduce attention complexity from $O(n^2)$ to $O(n)$ or $O(n\log n)$; efficient long-sequence processing}
        & \news{Long-context document QA, summarization, code assistants, multi-turn dialogue} \\
        
        \news{Decoding Algorithm (\S\ref{sec:inference_optimization_sampling}, \S\ref{sec:inference_optimization_structured_outputs})}
        & \news{Accelerate token generation via prediction and guidance strategies}
        & \news{Real-time chatbots, math, voice assistants, interactive applications requiring low latency} \\
    \bottomrule
    \end{tabular}%
    }
\end{table}

Each optimization technique must be considered not only by its placement, but also in terms of the bottlenecks it alleviates and the workloads it best supports. Table~\ref{tab:llm_inference_optimizations} categorizes the commonly used optimization techniques and maps them to their performance targets and suitable workloads. For instance, batching optimizations increase GPU utilization and throughput, making them effective for multi-user chatbots and API-based services. In contrast, compression techniques focus on reducing computation and memory, which is particularly valuable for resource-constrained edge devices.

\begin{table}[tbp]
    \caption{\news{Summary of Main Optimization Focus by LLM Inference Engine}}
    \label{tab:llm_platform_summary}
    \centering
    \resizebox{.9\textwidth}{!}{%
    \begin{tabular}{@{}>{\centering\arraybackslash}p{0.6\textwidth} p{0.75\textwidth}@{}}
    \toprule
    \multicolumn{1}{c}{\news{Engine}} & \multicolumn{1}{c}{\news{Main Optimization Focus}} \\
    \midrule
        \news{vLLM~\cite{kwon2023efficient}, TGI~\cite{tgi}, TensorRT-LLM~\cite{tensorrtllm}, DeepSpeed-FastGen~\cite{holmes2024deepspeed}, LMDeploy~\cite{2023lmdeploy}, DistServe~\cite{zhong2024distserve}}
        & \news{Batch scheduling, KV caching, attention optimization, decoding acceleration; optimized for large-scale serving} \\
        
        \news{Ollama~\cite{ollama}, llama.cpp~\cite{llamacpp}, llama2.c~\cite{llama2c}, bitnet.cpp~\cite{wang20241}, PowerInfer~\cite{song2024powerinfer}}
        & \news{Compression (quantization, pruning), lightweight KV caching; optimized for edge and personal inference} \\
        
        \news{SGLang~\cite{zheng2024sglang}, MLC-LLM~\cite{mlcllm}, MAX~\cite{max}, LitGPT~\cite{litgpt}, OpenLLM~\cite{openllm}, unsloth~\cite{unsloth}, NanoFlow~\cite{zhu2024nanoflow}, vAttention~\cite{prabhu2025vattention}}
        & \news{Experimental attention variants and decoding strategies; suitable for research and flexible customization} \\
        
        \news{LightLLM~\cite{lightllm}, Sarathi-Serve~\cite{agrawal2024taming}, Friendli Inference~\cite{friendli}}
        & \news{Long-context attention optimization and speculative decoding; optimized for real-time long-sequence serving} \\
        
        \news{GroqCloud~\cite{groqcloud}, Fireworks AI~\cite{fireworks}, Together Inference~\cite{together}}
        & \news{Ultra-low latency decoding; optimized for real-time interactive services} \\
    \bottomrule
    \end{tabular}%
    }
\end{table}

Based on Table~\ref{tab:llm_platform_summary}, the LLM inference engines discussed in this paper can be classified as follows: serving-oriented engines such as vLLM~\cite{kwon2023efficient} TGI~\cite{tgi}, and TensorRT-LLM~\cite{tensorrtllm} provide batch scheduling, caching, attention optimization, and decoding acceleration, making them suitable for large-scale services. Lightweight execution engines such as llama.cpp~\cite{llamacpp}, Ollama~\cite{ollama}, and PowerInfer~\cite{song2024powerinfer} leverage quantization and pruning to provide efficient inference on personal devices or edge environments. Research- and experiment-focused engines such as SGLang~\cite{zheng2024sglang} and MLC-LLM~\cite{mlcllm} support diverse attention variants and decoding strategies, while LightLLM~\cite{lightllm} and Sarathi-Serve~\cite{agrawal2024taming} are optimized for long-context processing and real-time interaction. Commercial engines such as GroqCloud~\cite{groqcloud}, Fireworks AI~\cite{fireworks}, and Together Inference~\cite{together} employ low-latency decoding optimizations to enable industrial-scale real-time services.

Ultimately, LLM inference optimization is determined by the interaction of technologies across all layers, and this paper systematically presents which layers of the system stack are supported by each inference engine.


\subsection{Batch Optimization} \label{sec:inference_optimization_batch}

\begin{figure}[tbp]
    \centering
    \resizebox{0.95\textwidth}{!}{%
    \begin{minipage}{1.0\textwidth}
        \centering
        \begin{subfigure}[b]{0.31\textwidth}
            \includegraphics[width=\linewidth]{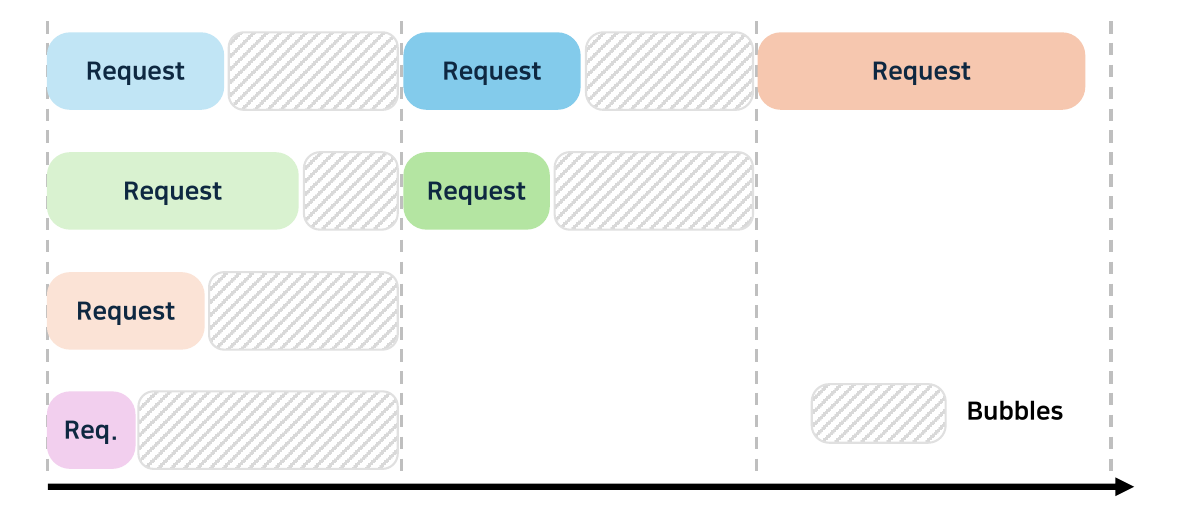}
            \caption{Static Batching}
            \label{fig:batching_static}
        \end{subfigure}
        \hspace{0.005\textwidth}
        \begin{subfigure}[b]{0.31\textwidth}
            \includegraphics[width=\linewidth]{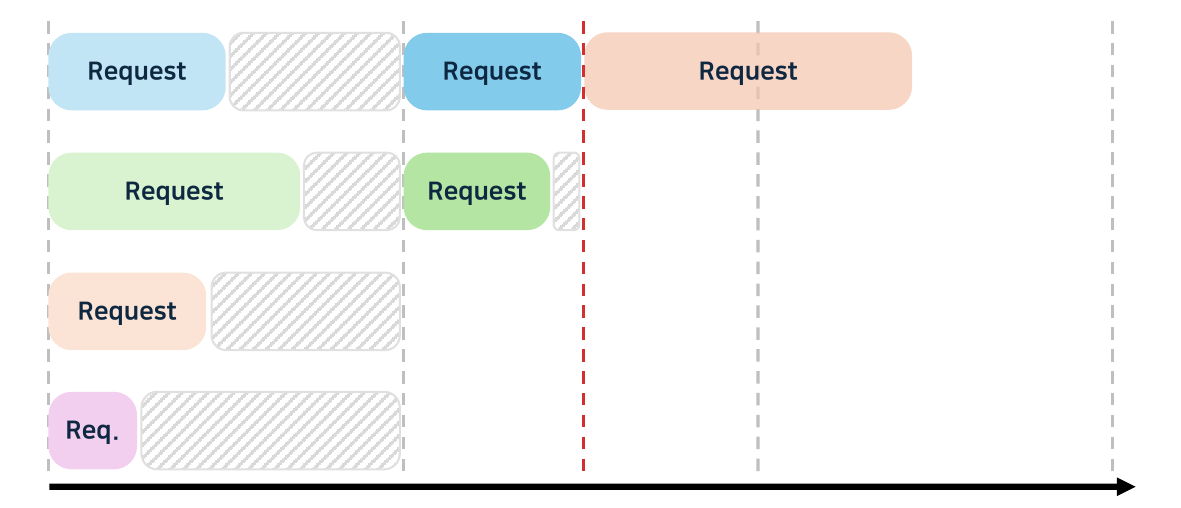}
            \caption{Dynamic Batching}
            \label{fig:batching_dynamic}
        \end{subfigure}
        \hspace{0.005\textwidth}
        \begin{subfigure}[b]{0.31\textwidth}
            \includegraphics[width=\linewidth]{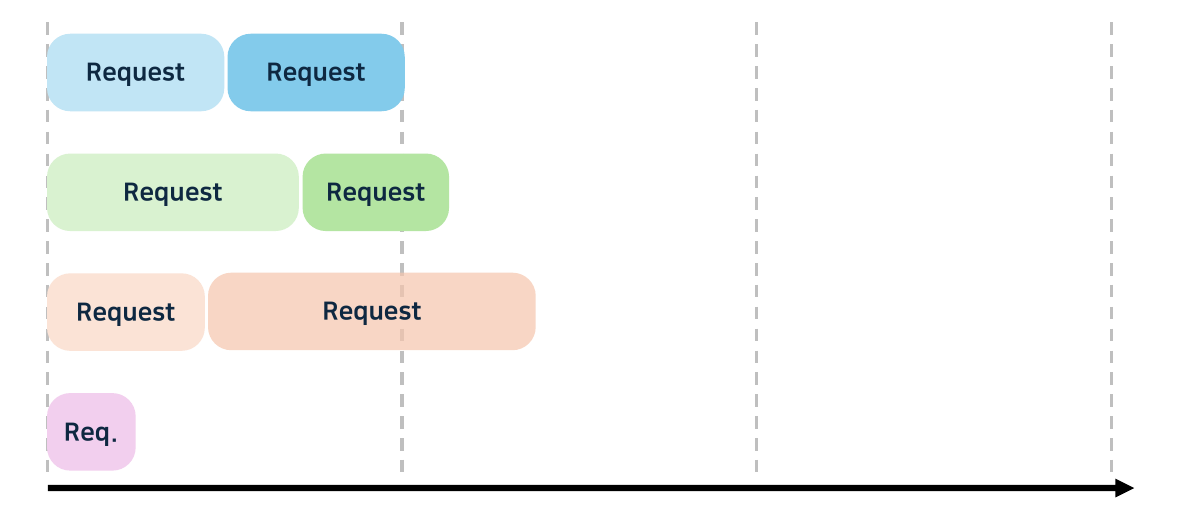}
            \caption{Continuous Batching}
            \label{fig:batching_continuous}
        \end{subfigure}
    \end{minipage}
    }
    \caption{Comparison of Batching Strategies}
    \label{fig:batching}
\end{figure}

In LLM inference, batching groups multiple input requests for simultaneous processing, boosting hardware utilization and throughput. Efficient batch processing is essential to maximize computational parallelism and latency.

Hence, finding the optimal batch size is essential. Smaller batches reduce response time but may underuse hardware resources, while larger batches yield higher throughput but risk longer response times. Various methods have been proposed to select optimal batch sizes~\cite{patel2024splitwise,guldogan2024multi,sheng2024fairness}, and inference engines typically provide mechanisms to explore the best batch size based on workload and SLOs.

Beyond batch size, the scheduling method also significantly influences inference performance. 
As shown in Fig.~\ref{fig:batching}~(\subref{fig:batching_static}), static batching processes a fixed number of requests, potentially increasing latency as new requests must wait until a batch completes. Dynamic batching~\cite{crankshaw2017clipper, ali2020batch} and continuous batching~\cite{yu2022orca, he2024deferred}, in contrast, adapt the batch in real time, often reducing latency and increasing overall efficiency.

\subsubsection{Dynamic Batching} \label{sec:inference_optimization_batch_dynamic}

Dynamic batching~\cite{crankshaw2017clipper, ali2020batch} alleviates the latency and hardware underutilization issues of static batching. As shown in Fig.~\ref{fig:batching}~(\subref{fig:batching_dynamic}) new requests are immediately added to an ongoing batch, enabling more flexible and efficient inference.

Unlike static batching, dynamic batching reconstructs batches based on incoming requests and available hardware resources, adaptively determining batch sizes. When a new request arrives, it can be merged with an existing batch or appended to an ongoing process to optimize resource usage.

Several parameters must be tuned to implement dynamic batching effectively, including the maximum batch wait time, minimum batch size, and batch size limits. Although dynamic batching can minimize latency by reducing batch size in real time, new requests can only be added after the current batch finishes. It may also introduce overhead for dynamically resizing batches and degrade performance if requests have widely varying prompt or output token lengths.

\subsubsection{Continuous Batching} \label{sec:inference_optimization_batch_continuous}

Continuous batching~\cite{yu2022orca, he2024deferred} is similar to dynamic batching~\cite{crankshaw2017clipper, ali2020batch} but allows new requests to join an ongoing batch without interruption, minimizing latency. Fig.~\ref{fig:batching}~(\subref{fig:batching_continuous}) shows how requests are continuously inserted to maximize GPU and memory efficiency.

Orca~\cite{yu2022orca} implements continuous batching via Iteration-Level Scheduling and Selective Batching. Iteration-Level Scheduling forms batches each iteration while accommodating new requests on the fly. Selective Batching focuses only on batchable transformer operations, enabling immediate results for completed requests and lowering both the average response time and the waiting time.

However, continuous batching requires sophisticated scheduling that new requests can be integrated without disrupting active processing. Efficient $\mathbf{KV}$ cache management is crucial, often involving methods such as PagedAttention~\cite{kwon2023efficient}. Inference engines such as llama.cpp~\cite{llamacpp}, DeepSpeed-FastGen~\cite{holmes2024deepspeed}, and vLLM~\cite{kwon2023efficient} use continuous batching techniques derived from Orca.

\subsubsection{Nano-batching} \label{sec:inference_optimization_batch_nano}

\begin{figure}[tbp]
  \centering
  \resizebox{0.9\textwidth}{!}{%
      \begin{minipage}[b]{0.48\linewidth}
        \centering
        \includegraphics[width=\linewidth]{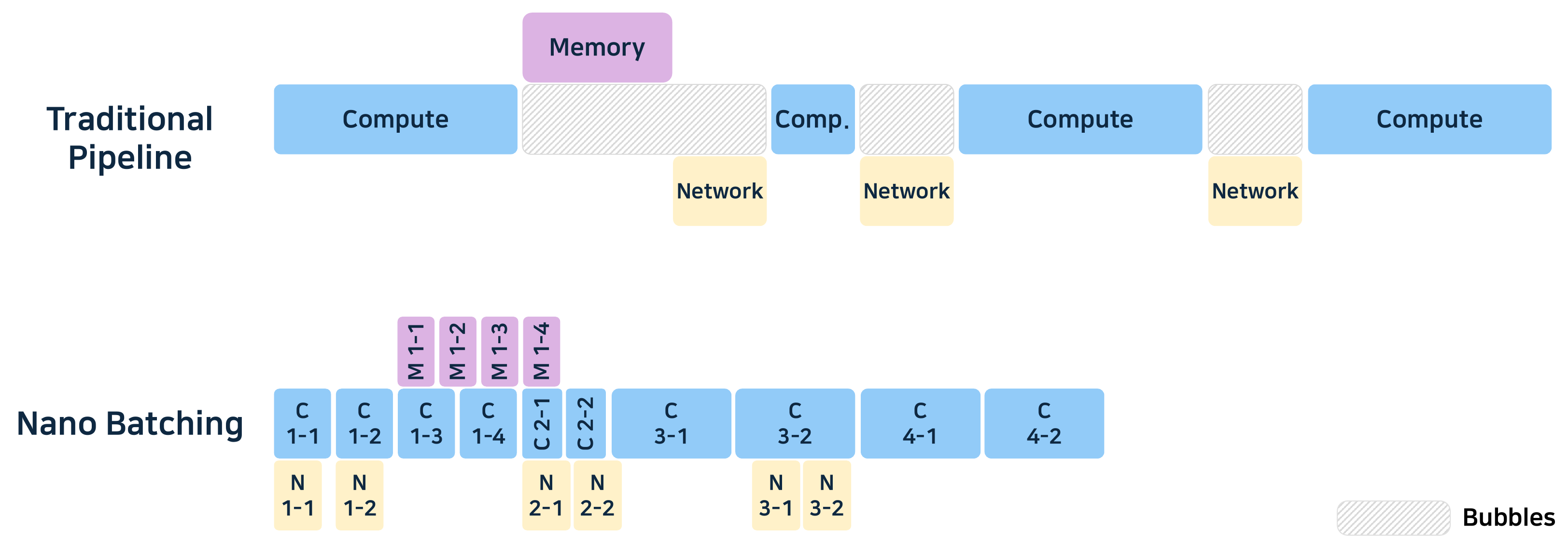}
        \caption{Nano-batching}
        \label{fig:nano_batching}
      \end{minipage}
      \hfill
      \begin{minipage}[b]{0.48\linewidth}
        \centering
        \includegraphics[width=\linewidth]{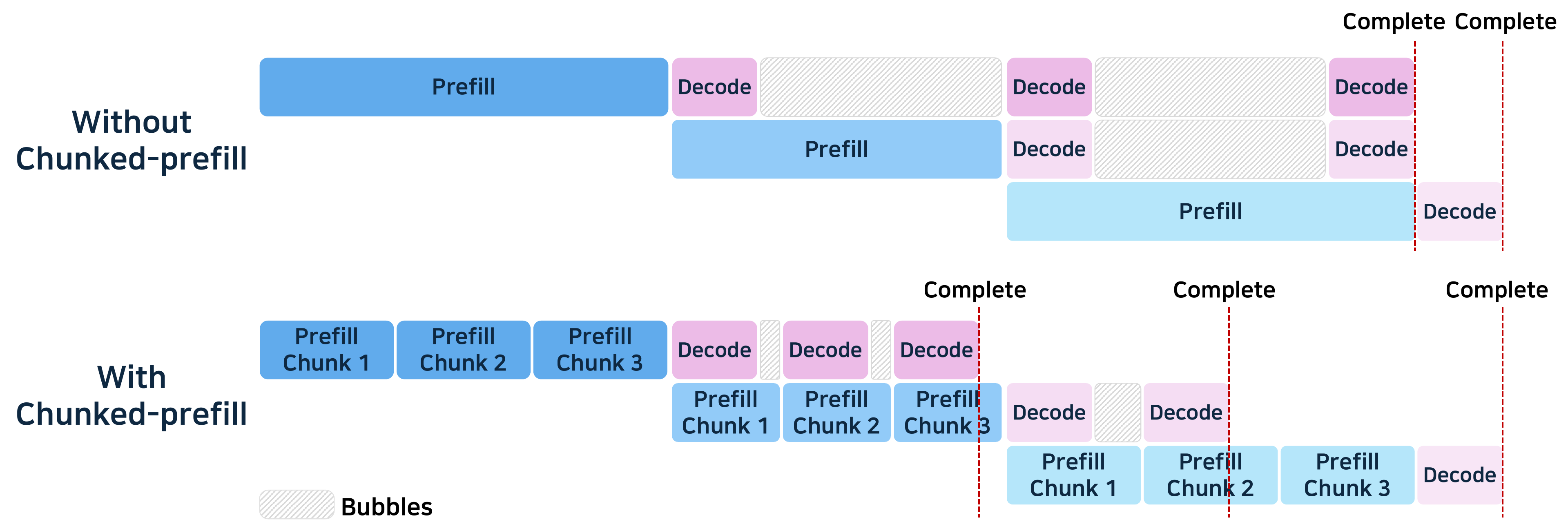}
        \caption{Chunked-prefills}
        \label{fig:chunked_prefills}
      \end{minipage}
  }
\end{figure}

Nano-batching, introduced by NanoFlow~\cite{zhu2024nanoflow}, maximizes resource utilization and throughput by running \texttt{\small {computation, memory}}, \texttt{\small {and network-bound operations}} in parallel on a single device. Traditional inference engines batch tasks at the request level, but NanoFlow divides them at the operation level, as illustrated in Fig.~\ref{fig:nano_batching}.

Operation units include attention and $\mathbf{KV}$ generation, GEMM, and collective communication for multi-GPU synchronization. NanoFlow dynamically adjusts nano-batch sizes to optimize each resource type, employing a scheduling approach that merges topological sorting and greedy search based on hardware resources and kernel optimizations.

By breaking operations into smaller nano batches, tasks can overlap and run concurrently with user-server network operations, boosting resource utilization and increasing throughput. However, nano-batching demands complex scheduling and can incur additional communication overhead when operations are spread across multiple GPUs.

\subsubsection{Chunked-prefills} \label{sec:inference_optimization_batch_chunked}

Chunked prefills~\cite{agrawal2023sarathi} addresses pipeline inefficiencies that arise in dynamic or continuous batching, especially in multi-GPU environments where the memory-bound decode phase might sit idle if a batch is empty. Processing long prompts in one step can also increase latency and hinder shorter requests.

Fig.~\ref{fig:chunked_prefills} illustrates chunked prefills, which splits long prompts into multiple segments and processes them incrementally. The decoding for the first segment can begin immediately while subsequent segments undergo prefill, allowing these phases to run concurrently and improving resource usage.

However, chunked prefills adds scheduling complexity by requiring more granular batch management. $\mathbf{KV}$ cache usage can also surge due to simultaneous prefill and decode execution. For example, DeepSpeed-FastGen~\cite{holmes2024deepspeed}'s Dynamic SplitFuse splits prompts to generate tokens earlier, while Sarathi-Serve~\cite{agrawal2024taming} stall-free scheduling immediately admits new requests, eliminating wait time and boosting efficiency.

\subsection{Parallelism} \label{sec:inference_optimization_parallelism}

\begin{figure}[tbp]
    \centering
    \resizebox{.85\textwidth}{!}{%
    \begin{minipage}{1.0\textwidth}
        \centering
        \begin{subfigure}[b]{0.15\textwidth}
            \includegraphics[width=\linewidth]{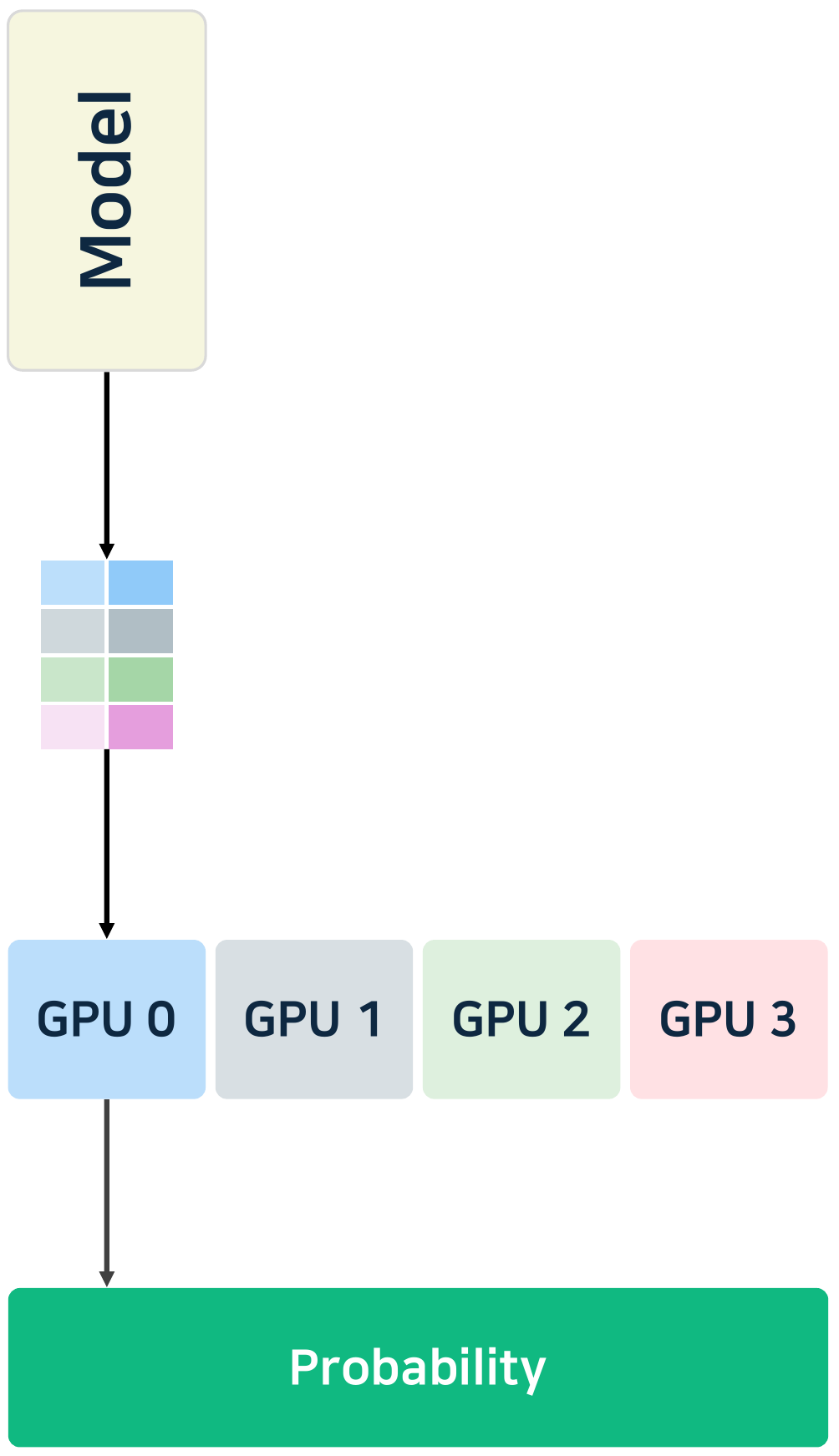}
            \caption{No Parallelism}
            \label{fig:parallelism_no}
        \end{subfigure}
        \hspace{.05\textwidth}
        \begin{subfigure}[b]{0.15\textwidth}
            \includegraphics[width=\linewidth]{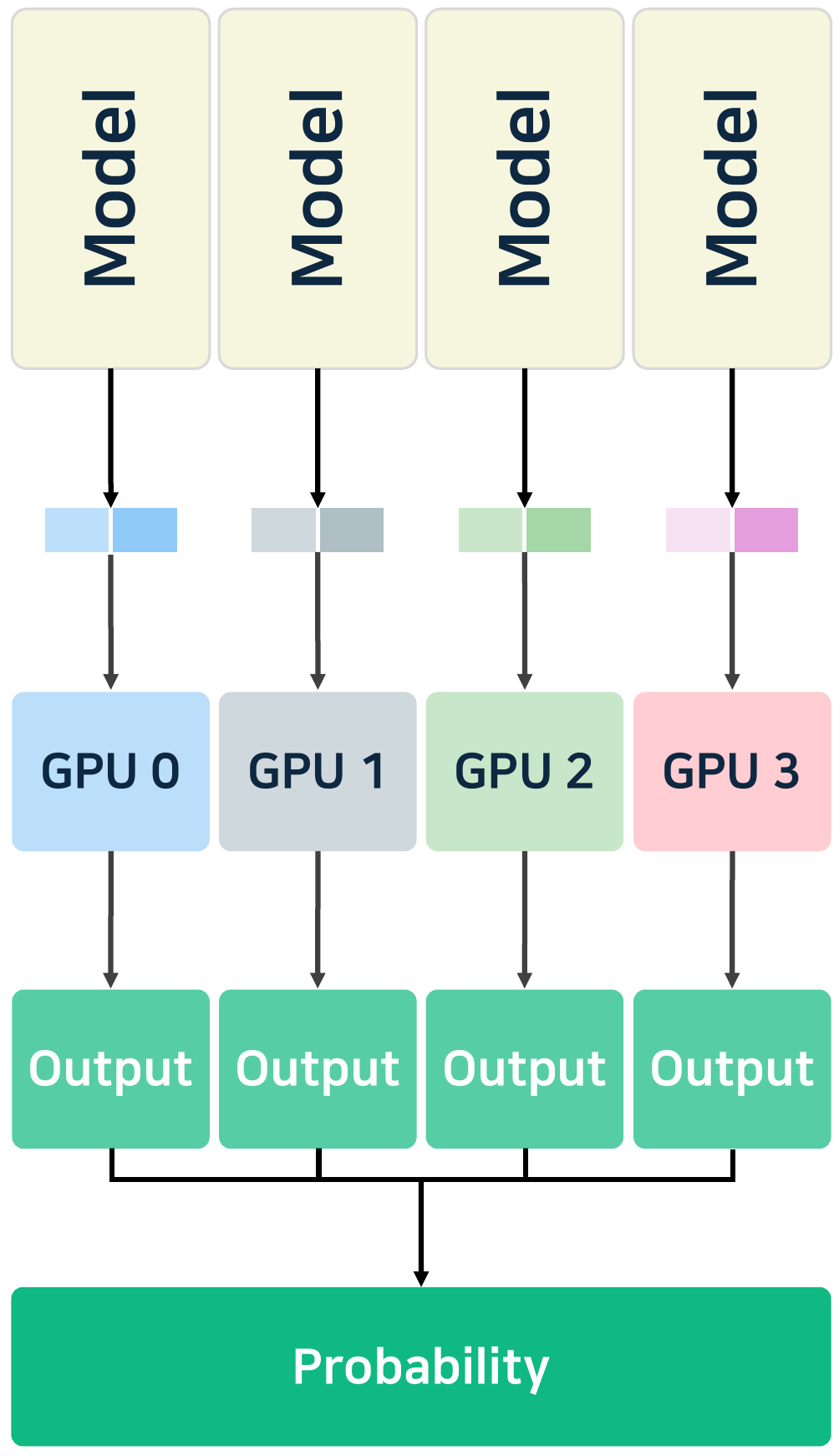}
            \caption{Data Parallelism (DP)}
            \label{fig:parallelism_data}
        \end{subfigure}
        \hspace{.05\textwidth}
        \begin{subfigure}[b]{0.19\textwidth}
            \includegraphics[width=\linewidth]{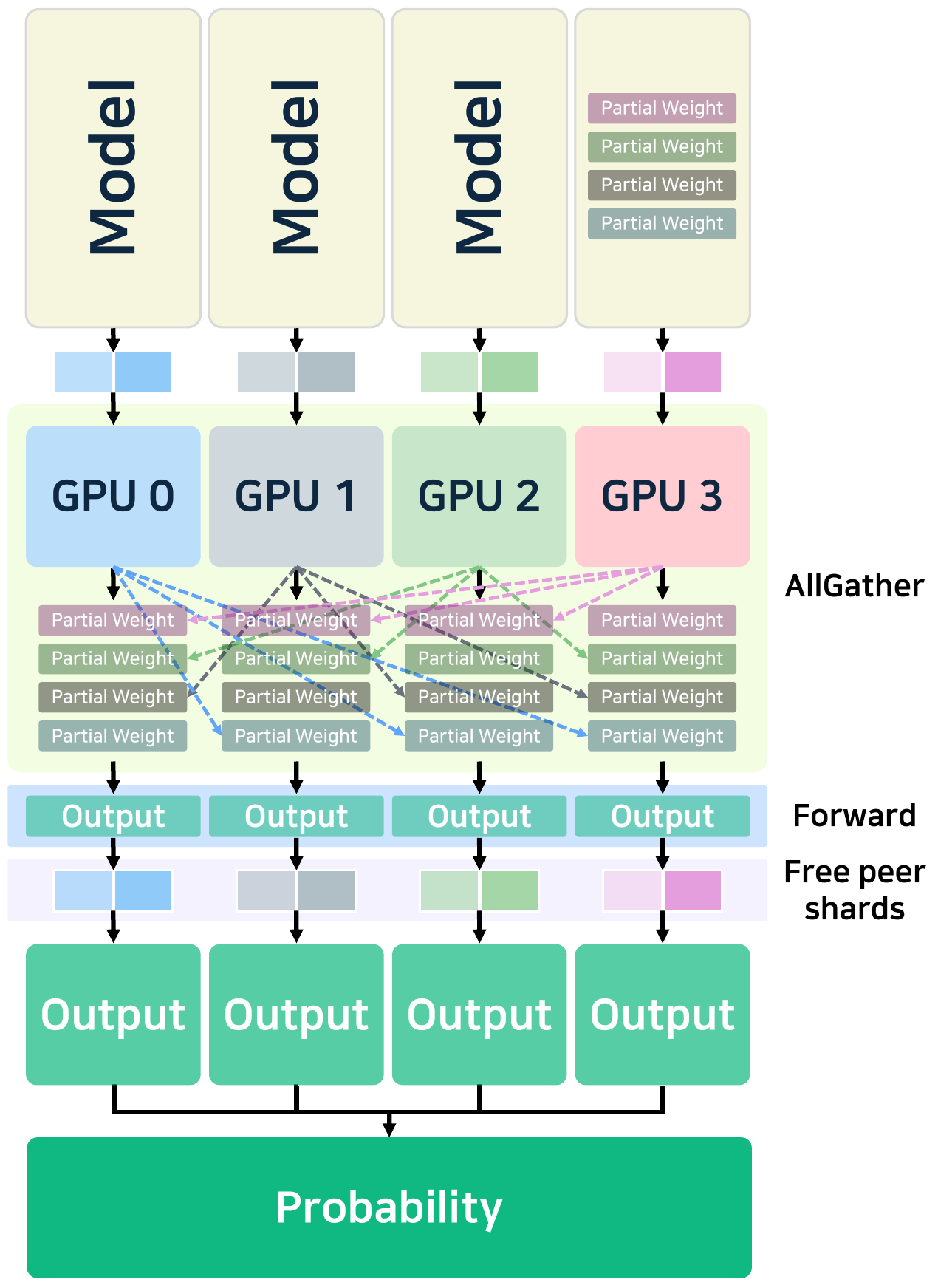}
            \caption{Fully Sharded Data Parallel~(FSDP)}
            \label{fig:parallelism_fsdp}
        \end{subfigure}
        \hspace{.05\textwidth}
        \begin{subfigure}[b]{0.15\textwidth}
            \includegraphics[width=\linewidth]{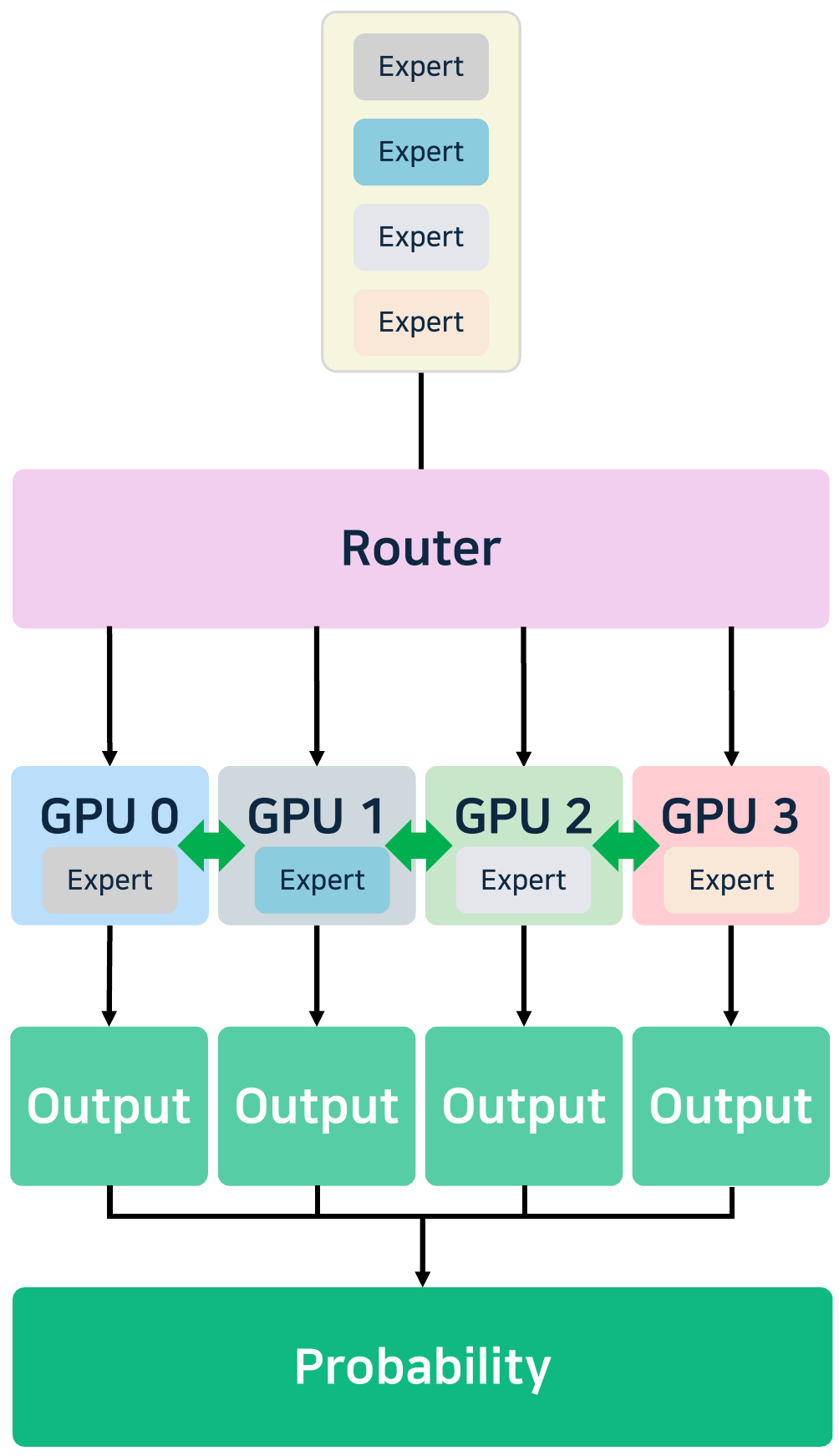}
            \caption{\news{Expert Parallelism~(EP)}}
            \label{fig:parallelism_ep}
        \end{subfigure}

        \vskip\baselineskip 

        \begin{subfigure}[b]{0.39\textwidth}
            \includegraphics[width=\linewidth]{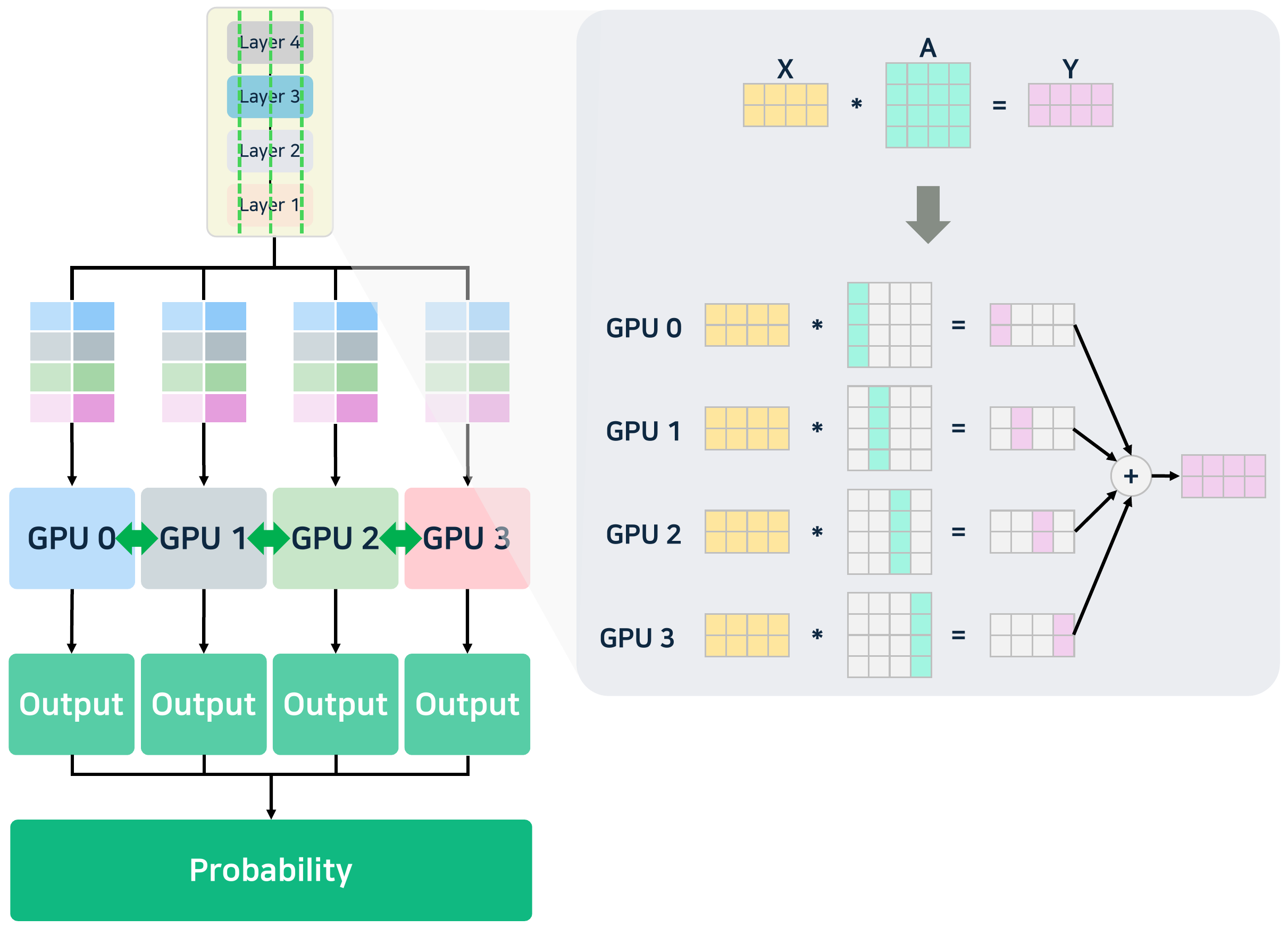}
            \caption{Tensor Parallelism (TP)}
            \label{fig:parallelism_tensor}
        \end{subfigure}
        \hfill
        \begin{subfigure}[b]{0.55\textwidth}
            \includegraphics[width=\linewidth]{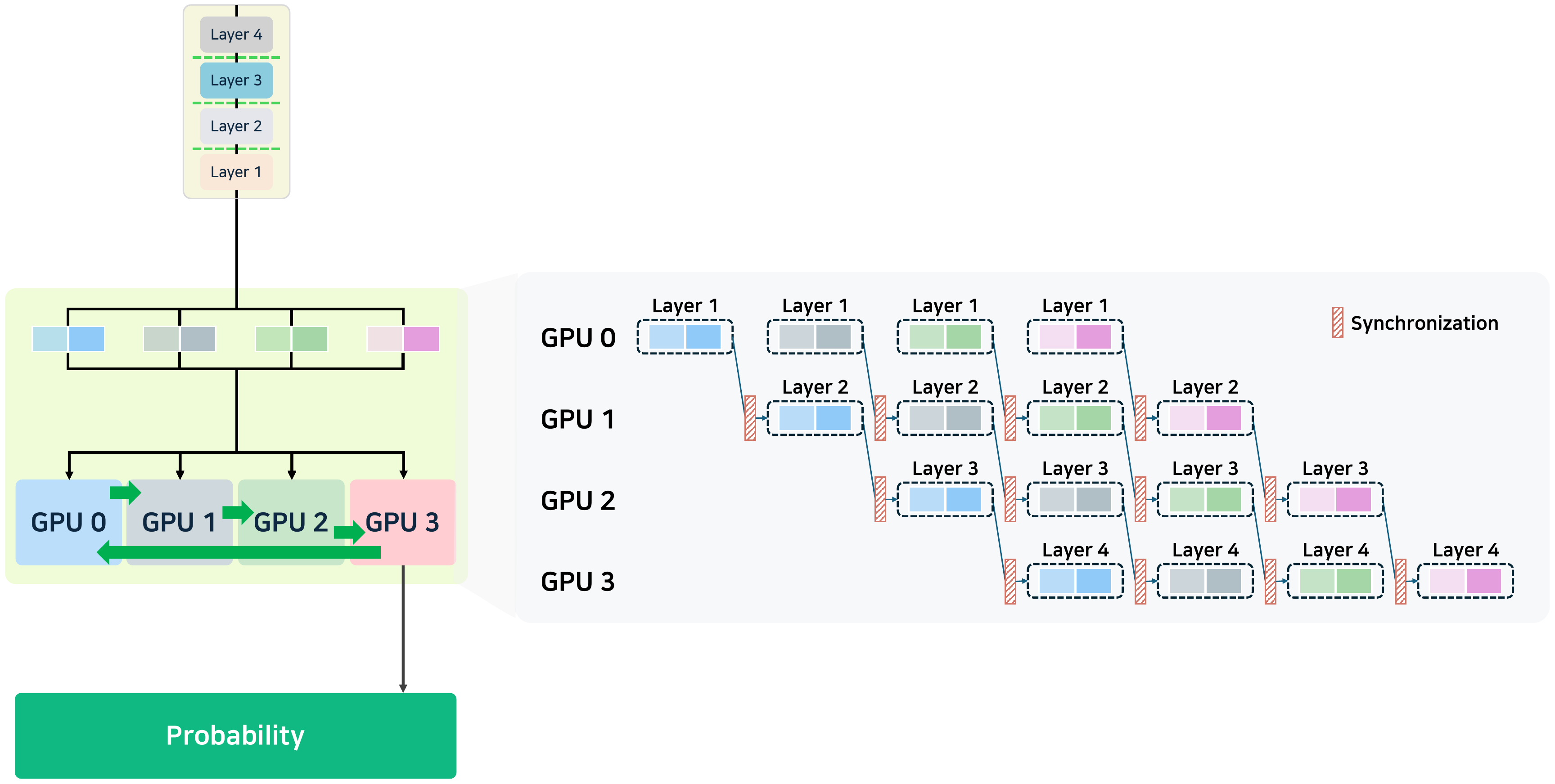}
            \caption{Pipeline Paralleism (PP)}
            \label{fig:parallelism_pipeline}
        \end{subfigure}
    \end{minipage}
    }
    \caption{Comparison of Parallelism Strategies}
    \label{fig:parallelism}
\end{figure}

Because LLMs may contain billions or even trillions of parameters, relying on a single GPU or similar hardware for inference has become increasingly challenging. As a result, distributed and parallel processing across multiple devices or nodes is vital for reducing latency and maximizing hardware utilization.

Parallelism strategies in LLM inference differ in their implementation and performance based on server architecture and hardware configuration. Factors such as the number of available GPUs or accelerators, interconnect bandwidth, memory hierarchy, and computational capacity influence how effectively tensor parallelism (TP)~\cite{stojkovic2024towards, prabhakar2024kraken}, data parallelism (DP)~\cite{rajbhandari2020zero}, FSDP~\cite{zhao2023pytorch}, and pipeline parallelism (PP)~\cite{agrawal2023sarathi, hu2021pipeline} can be employed. 
\news{With the widespread adoption of MoE models, the expert parallelism (EP)~\cite{liu2024deepseek-v3, zhu2025megascale, balmau2025accelerating} has also been introduced, in which the expert modules selected by each token are distributed across multiple devices and executed in parallel.}
Furthermore, hybrid approaches can combine multiple strategies to further improve performance~\cite{xuanlei2024hetegen, chen2023ee}. 

For example, in multi-node clusters, inter-node communication latency may become a bottleneck, necessitating techniques like communication compression or asynchronous scheduling to maintain high performance. Conversely, in single-node, multi-GPU setups, shared memory, and high-speed interconnects (e.g., NVLink, NVSwitch) allow for more efficient synchronization and workload distribution. Ultimately, the success of any parallelization strategy depends on balancing computation and communication overhead, emphasizing the need to tailor the approach to each unique hardware environment and model architecture. To address these challenges, researchers are actively developing methods to automatically explore and identify optimal parallelism strategies for each system~\cite{li2024automatically, miao2022galvatron}. Various parallelism mechanisms are shown in Fig.~\ref{fig:parallelism}.

In addition to inter-device parallelization, internal device parallelization strategies also significantly impact LLM inference performance. A representative example is the distribution of GEMM operations across multiple thread blocks, where the conventional approach partitions the matrix into fixed-size tiles and maps each block to a single tile. However, when the number of tiles does not divide evenly by the number of thread blocks, idle threads emerge, leading to wasted computational resources. To mitigate this inefficiency, the Stream-K~\cite{osama2023stream} method has been proposed. Instead of partitioning by tiles, Stream-K decomposes the multiply-accumulate (MAC) loop into finer-grained units and distributes them evenly, thereby maximizing hardware utilization. Experiments implemented in CUTLASS report up to a 14.7$\times$ performance improvement for FP16 operations on NVIDIA GPUs. Currently, several inference engines, including vAttention~\cite{prabhu2025vattention}, llama.cpp~\cite{llamacpp}, Ollama~\cite{ollama}, TensorRT-LLM~\cite{tensorrtllm}, and SGLang~\cite{zheng2024sglang}, support Stream-K-based GEMM parallelization.


\subsubsection{Data Parallelism} \label{sec:inference_optimization_parallelism_data}

As shown in Fig.~\ref{fig:parallelism}~(\subref{fig:parallelism_data}), DP~\cite{rajbhandari2020zero} replicates the same model across multiple GPUs or nodes. A mini-batch is split among available hardware devices, each performing inference independently on its share of the data. Once the computations are completed, the outputs (or weights) are collected into a single device to produce the final results.

Although this method is straightforward to implement and features relatively low communication overhead---since synchronization happens only after inference---DP can become impractical if the entire model must reside on each device, especially for massive LLMs. Furthermore, if hardware devices differ significantly in performance, the overall system may experience bottlenecks.

\subsubsection{Fully Sharded Data Parallelism} \label{sec:inference_optimization_parallelism_fsdp}

FSDP~\cite{zhao2023pytorch} is a parallelism technique designed to reduce memory usage and improve training efficiency when working with LLMs. Unlike traditional data parallelism, where each device holds a full copy of the model parameters and optimizer states, FSDP shards the model's parameters, gradients, and optimizer states across multiple devices. This removes duplicated memory usage and allows larger models to be trained on the same hardware resources.

As shown in Fig.~\ref{fig:parallelism}~(\subref{fig:parallelism_fsdp}), FSDP works by gathering all the parameters of a layer on each GPU right before that layer is executed. This allows full computation to happen on each GPU. The full parameters are temporarily loaded into memory only during this operation and are removed right after the layer finishes. This approach does not split the operation itself, making it simple to implement and compatible with most models.

However, since parameters must be all-gathered at every layer, there is a communication overhead. This can lead to performance issues during inference, especially for workloads with small batch sizes or where low latency is important. Also, if a layer requires more memory than what a single GPU can handle during the all-gather phase, it cannot be run.

During training, FSDP brings large memory savings by sharding activations and parameters. But during inference, there is no gradient or activation recomputation, thus memory savings are smaller. Therefore, the use of FSDP during inference should be decided based on model size.

FSDP is natively supported in PyTorch and works well with its autograd engine, checkpointing, and mixed precision training. It can also be combined flexibly with other parallel strategies such as hybrid parallelism. Among the LLM inference engines we studied, vLLM\cite{kwon2023efficient}, DeepSpeed-FastGen\cite{holmes2024deepspeed}, and SGLang~\cite{zheng2024sglang}, LitGPT~\cite{litgpt} support FSDP.

\subsubsection{\news{Expert Parallelism}} \label{sec:inference_optimization_expert}

\news{As MoE models~\cite{cai2024survey} have become more widespread, the expert parallelism~(EP)~\cite{liu2024deepseek-v3, zhu2025megascale, balmau2025accelerating} has been proposed to enable efficient inference for this architecture. In MoE models, only a subset of experts with the highest scores are activated for each token; therefore, replicating all experts across all devices, as done in traditional parallelization methods, leads to unnecessary memory usage and computational overhead.}

\news{As illustrated in Fig.~\ref{fig:parallelism}~(\subref{fig:parallelism_ep}), EP distributes the set of experts across multiple devices, where each device retains only the weights of its assigned experts and processes only the tokens corresponding to those experts. Specifically, the router (or gate) network computes scores for all experts, selects the top-$k$ experts, and transfers the input tokens to the devices hosting those experts via \texttt{All-to-All} communication. This approach significantly reduces memory and computation overhead compared to full model replication.}

\news{However, because the frequency of experts calls varies across tasks, load imbalance can occur among devices. DeepSeek-v3~\cite{liu2024deepseek-v3} addresses this issue by introducing the redundant experts strategy, which replicates frequently used experts, and by employing Hierarchical Load Balancing and Global Load Balancing mechanisms to maintain balanced workloads even in multi-node environments.}

\news{Both vLLM~\cite{kwon2023efficient} and SGLang~\cite{zheng2024sglang}, TensorRT-LLM~\cite{tensorrtllm} experimentally support EP to enhance the efficiency of MoE model inference.}


\subsubsection{Tensor Parallelism} \label{sec:inference_optimization_tensor}

TP~\cite{stojkovic2024towards, prabhakar2024kraken}, also known as model parallelism or sharding, divides specific LLM operations (e.g., matrix multiplication, attention, fully connected (FC) layers) across multiple hardware devices. Each device processes a slice of the operation, and intermediate results are merged afterward.

For example, Fig.~\ref{fig:parallelism}~(\subref{fig:parallelism_tensor}) shows a situation in which four GPUs handle the matrix operation \(\textit{X} \times \textit{A} = \textit{Y}\). The matrix \(\textit{A}\) is partitioned among the GPUs, either row-wise or column-wise, and the computations are reconciled via collective communication (e.g., All-Reduce or All-Gather).

By distributing large computations, tensor parallelism speeds up inference and reduces the memory footprint of each device, since individual GPUs do not need to store all weights. However, frequent inter-device communication can increase overhead, and suboptimal partitioning may reduce efficiency. Inference engines such as vLLM~\cite{kwon2023efficient}, DeepSpeed-FastGen~\cite{holmes2024deepspeed}, and TensorRT-LLM~\cite{tensorrtllm} often integrate techniques to address these challenges~\cite{shoeybi2019megatron, oh2024exegpt, xuanlei2024hetegen}.

To mitigate communication bottlenecks in tensor parallelism---particularly the performance degradation observed in TTFT---recent research has proposed communication compression techniques that reduce overhead and enhance inference speed~\cite{hansen2024communication}.

\subsubsection{Pipeline Parallelism} \label{sec:inference_optimization_parallelism_pipeline}

PP~\cite{agrawal2023sarathi, hu2021pipeline} assigns different parts (layers) of an LLM to different GPUs. The input data are split into micro-batches which traverse this \texttt{\small {pipeline}} of layers sequentially. As illustrated in Fig.~\ref{fig:parallelism}~(\subref{fig:parallelism_pipeline}), if a transformer model has four layers and there are four GPUs available, each GPU is responsible for one layer.

This arrangement can reduce memory usage by distributing layers across devices and can also accelerate inference by overlapping operations. However, communication overhead occurs when intermediate results move between devices, and the initial pipeline stages remain underutilized until the pipeline is \texttt{\small {warmed up}}. Various pipeline optimization techniques have been proposed to mitigate these concerns~\cite{ma2024hpipe, yu2024twinpilots}.

Various inference frameworks support PP, including Ollama~\cite{ollama}, llama.cpp~\cite{llamacpp}, vLLM~\cite{kwon2023efficient}, and Friendli Inference~\cite{friendli}.

\subsection{Compression} \label{sec:inference_optimization_compression}

As LLMs grow larger, conducting inference on a single GPU or server node becomes increasingly difficult. To mitigate this issue, model compression techniques---such as quantization~\cite{egashira2025exploiting}, KD~\cite{yang2024survey}, and pruning~\cite{zhu2024survey, kim2024efficient}, sparsity optimization~\cite{fan2025spinfer, shin2024sparseinfer}---have emerged. Among these, quantization is particularly important for saving memory and increasing inference speed, thereby reducing power consumption and cost. Pruning and sparsity optimization can enhance computational efficiency and inference speed, and several inference engines offer support for these techniques. Although they are closely tied to training or fine-tuning, inference engines must still ensure proper kernel selection and execution when running quantized models.

\subsubsection{Quantization} \label{sec:inference_optimization_quantization}

\begin{figure}[tbp]
    \centering
    \resizebox{0.7\textwidth}{!}{%
    \begin{subfigure}[t]{0.35\textwidth}
        \includegraphics[width=\linewidth]{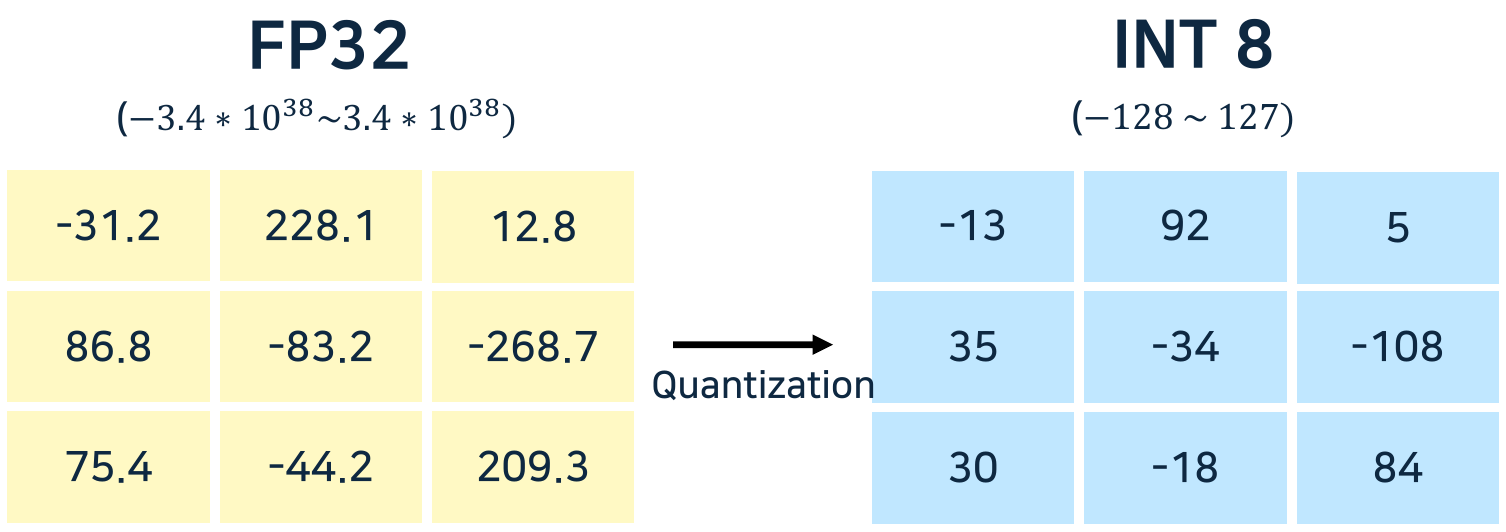}
        \caption{Quantization Scheme}
        \label{fig:quantization_scheme}
    \end{subfigure}
    \hspace{1mm} 
    \begin{subfigure}[t]{0.23\textwidth}
        \includegraphics[width=\linewidth]{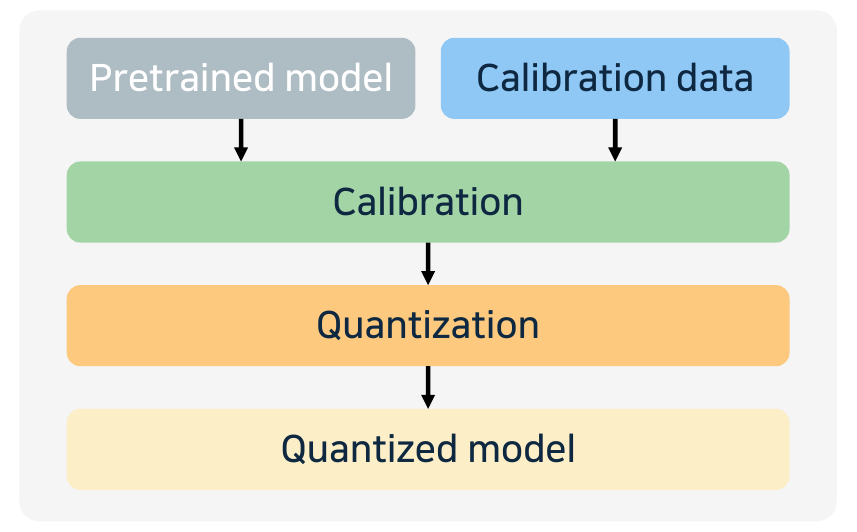}
        \caption{Post Training Quantization (PTQ)}
        \label{fig:quantization_ptq}
    \end{subfigure}
    \hspace{1mm}
    \begin{subfigure}[t]{0.23\textwidth}
        \includegraphics[width=\linewidth]{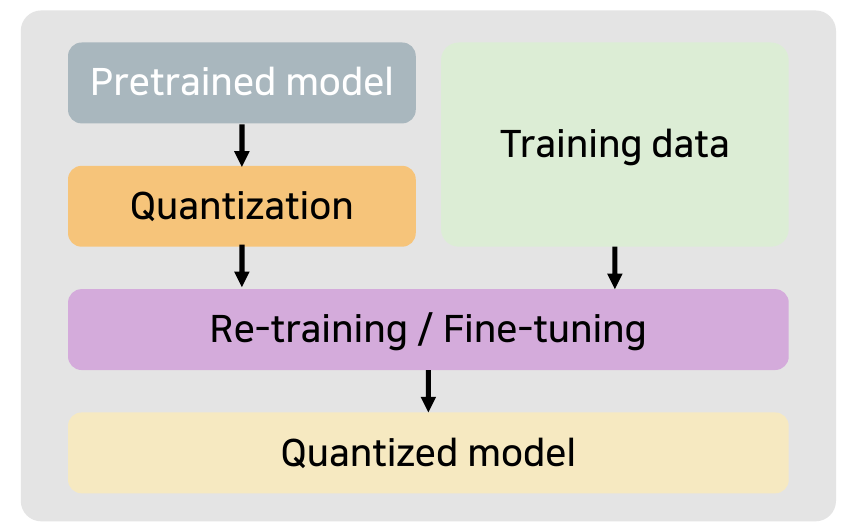}
        \caption{Quantization-Aware Training (QAT)}
        \label{fig:quantization_qat}
    \end{subfigure}
    }
    \caption{Quantization Methods}
    \label{fig:quantization}
\end{figure}

\textbf{Quantization Algorithm.}
Quantization converts pretrained FP32 or FP16 models into lower-precision floating-point formats (e.g., FP4, FP8) or integer formats (e.g., INT4, INT8), as illustrated in Fig.~\ref{fig:quantization}~(\subref{fig:quantization_scheme}). By representing fewer distinct numerical values, quantization can substantially accelerate matrix multiplications and reduce memory requirements with only minor performance trade-offs.

To convert high-precision models into lower-bit representations, methods like Absolute Max Quantization~\cite{dettmers2022gpt3} calculate a scale factor from the tensor's absolute maximum value. During dequantization, the model's values are approximated by applying the scale factor to the quantized data. Although this method is straightforward and hardware-efficient, it can be sensitive to outliers. Approaches like Affine Quantization~\cite{ma2024affinequant} address this issue by adjusting the distribution more flexibly.

Depending on the workflow, several methods can be applied, such as post-training quantization (PTQ)~\cite{xiao2023smoothquant, li2023fptq}, which applies quantization after model training (Fig.~\ref{fig:quantization}~(\subref{fig:quantization_ptq}), or quantization-aware training (QAT)~\cite{chen2024efficientqat, liu2023llm}, which integrates quantization into the training process (Fig.~\ref{fig:quantization}~(\subref{fig:quantization_qat}) PTQ can be used with pretrained models, making it straightforward and quick to implement. When applying PTQ, a small calibration dataset of representative sample data is used to refine the quantization parameters in the pretrained model. This dataset helps determine the activation distributions within each layer, which are then used to define the quantization parameters, such as clipping ranges and scale factors. However, it may result in accuracy degradation due to quantization effects. In contrast, QAT incorporates quantization operations into the training process, allowing gradient information to be considered and thereby preserving accuracy more effectively. However, QAT involves additional steps---such as fine-tuning quantization parameters, retraining, and adjusting training strategies---which increase overall training costs~\cite{chen2024efficientqat}. Consequently, even in services where high accuracy is required, PTQ is used more frequently in practice, given these realistic constraints.

A range of quantization methods can be used for LLMs and are supported by various inference engines. Generalized Post-Training Quantization (GPTQ)~\cite{frantar2022gptq} offers weight-only quantization, optimizing per-layer scales with error compensation to minimize accuracy degradation. Activation-Aware Weight Quantization (AWQ)~\cite{lin2024awq} improves weight quantization by grouping weights according to activation distributions, increasing accuracy.

Additive Quantization of Language Models~(AQLM)~\cite{egiazarian2024extreme} quantizes both weights and activations using PTQ only, avoiding QAT and achieving high performance with lower overhead than GPTQ. SmoothQuant~\cite{xiao2023smoothquant} normalizes activation and weight distributions to reduce clipping during quantization, leading to stable PTQ-based activation quantization while reducing latency and memory usage.

Additionally, $\mathbf{KV}$ cache quantization~\cite{hooper2024kvquant, liu2024kivi} has been proposed to minimize memory usage in long-context scenarios, allowing a balance between memory efficiency and generation speed while minimizing the impact on model quality. KVQuant~\cite{hooper2024kvquant} applies ultra-low precision quantization with minimal accuracy drop using Pre-Channel and Pre-RoPE key quantization, non-uniform $\mathbf{KV}$ cache quantization, and Per-Vector Dense-and-Sparse Quantization. KIVI~\cite{liu2024kivi} quantizes key caches per channel and value caches per token, achieving 2-bit quantization.

\textbf{Kernel Code and Hardware Support.}
Many inference engines integrate external quantization tools. bitsandbytes~\cite{bitsandbytes} is a CUDA-based Python library that supports 8-bit and 4-bit quantization; it supported in engines such as vLLM~\cite{kwon2023efficient} and SGLang~\cite{zheng2024sglang}, LitGPT~\cite{litgpt}. DeepSpeed FP is DeepSpeed's library for 6-bit and 8-bit weight-only quantization and is partially supported in vLLM~\cite{kwon2023efficient} on NVIDIA GPUs.

ExLlamaV2~(EXL2)~\cite{exl2} is an inference library for consumer GPUs, offering flexible 2-bit to 8-bit quantization similar to GPTQ, along with the option to mix quantization at different bit-levels per layer. EETQ~\cite{eetq} provides a straightforward and efficient approach to INT8 weight-only PTQ for transformer models. Both EXL2 and EETQ are supported in TGI~\cite{tgi}.

LLM Compressor~\cite{llmcompressor} is a quantization library designed for the vLLM environment, allowing both weight-only and activation quantization. It supports mixed-precision modes (e.g., W4A16, W8A16) and integrates techniques such as simple PTQ, GPTQ~\cite{frantar2022gptq}, and SmoothQuant~\cite{xiao2023smoothquant}. Inference engines such as vLLM~\cite{kwon2023efficient} and SGLang~\cite{zheng2024sglang} can employ LLM Compressor for quantization.

Mixed Auto-Regressive Linear Kernel~(Marlin)~\cite{frantar2025marlin} is a highly optimized kernel for FP16$\times$INT4 matrix multiplication. Designed to maximize inference speed, it can theoretically deliver up to four times the performance of FP16 by fully utilizing GPU global memory, cache, shared memory, and tensor cores. Implemented at the NVIDIA Parallel Thread Execution~(PTX)~\cite{ptx} assembly level, Marlin depends on NVIDIA GPUs and is supported by vLLM~\cite{kwon2023efficient}, SGLang~\cite{zheng2024sglang}, and TGI~\cite{tgi}.

\begin{table}[tbp]
    \centering
    \caption{Data Type Support in LLM Inference Engines}
    \label{tab:frameworks_data_type}
    
    \resizebox{.75\textwidth}{!}{%
    \begin{tabular}{lcccccccc cccc}
    \toprule
    \multirow{2.5}{*}{Engines} & \multicolumn{12}{c}{Data Type} \\ 
    \cmidrule(lr){2-5} \cmidrule(lr){6-6} \cmidrule(lr){7-7} \cmidrule(lr){8-9} \cmidrule(lr){10-13}   
     & FP32 & FP16 & FP8 & FP4  & NF4 & BF16 & INT8 & INT4 & MXFP8 & MXFP6 & MXFP4 & MXINT8\\ 
    \midrule
    Ollama~\cite{ollama}           
        & \greencheck & \greencheck & \greencheck & \redxmark 
        & \redxmark 
        & \greencheck 
        & \greencheck & \redxmark 
        & \redxmark & \redxmark & \redxmark & \redxmark \\
    llama.cpp~\cite{llamacpp}         
        & \greencheck & \greencheck & \redxmark & \redxmark 
        & \redxmark
        & \redxmark 
        & \greencheck & \greencheck
        & \redxmark & \redxmark & \redxmark & \redxmark \\
    vLLM~\cite{kwon2023efficient} 
        & \greencheck & \greencheck & \greencheck & \greencheck 
        & \greencheck
        & \greencheck 
        & \greencheck & \greencheck 
        & \redxmark & \redxmark & \redxmark & \redxmark \\
    DeepSpeed-FastGen~\cite{holmes2024deepspeed} 
        & \greencheck & \greencheck & \redxmark & \redxmark 
        & \redxmark
        & \redxmark 
        & \greencheck & \greencheck 
        & \redxmark & \redxmark & \redxmark & \redxmark \\
    Unsloth~\cite{unsloth}          
        & \greencheck & \greencheck & \greencheck & \redxmark 
        & \greencheck
        & \greencheck 
        & \greencheck & \greencheck 
        & \redxmark & \redxmark & \redxmark & \redxmark \\
    MAX~\cite{max}          
        & \greencheck & \greencheck & \greencheck & \redxmark 
        & \redxmark
        & \greencheck 
        & \greencheck & \greencheck 
        & \redxmark & \redxmark & \redxmark & \redxmark \\
    MLC LLM~\cite{mlcllm}           
        & \greencheck & \greencheck & \greencheck & \redxmark 
        & \redxmark
        & \redxmark
        & \greencheck & \greencheck 
        & \redxmark & \redxmark & \redxmark & \redxmark \\
    llama2.c~\cite{llama2c}            
        & \greencheck & \redxmark & \redxmark & \redxmark 
        & \redxmark
        & \redxmark 
        & \greencheck & \redxmark 
        & \redxmark & \redxmark & \redxmark & \redxmark \\
    bitnet.cpp~\cite{wang20241}            
        & \greencheck & \greencheck & \redxmark & \redxmark 
        & \redxmark
        & \greencheck 
        & \greencheck & \redxmark 
        & \redxmark & \redxmark & \redxmark & \redxmark \\
    SGLang~\cite{zheng2024sglang}            
        & \greencheck & \greencheck & \greencheck & \redxmark 
        & \greencheck
        & \greencheck 
        & \greencheck & \greencheck 
        & \redxmark & \redxmark & \redxmark & \redxmark \\
    LitGPT~\cite{litgpt}            
        & \greencheck & \greencheck & \redxmark & \greencheck 
        & \greencheck
        & \redxmark 
        & \greencheck & \redxmark 
        & \redxmark & \redxmark & \redxmark & \redxmark \\
    OpenLLM~\cite{openllm}          
        & \greencheck & \greencheck & \redxmark & \redxmark 
        & \redxmark
        & \redxmark 
        & \greencheck & \redxmark 
        & \redxmark & \redxmark & \redxmark & \redxmark \\
    TensorRT-LLM~\cite{tensorrtllm}     
        & \greencheck & \greencheck & \greencheck & \redxmark 
        & \redxmark
        & \greencheck 
        & \greencheck & \greencheck 
        & \greencheck & \redxmark & \greencheck & \redxmark \\
    TGI~\cite{tgi}               
        & \greencheck & \greencheck & \greencheck & \greencheck 
        & \greencheck
        & \greencheck 
        & \redxmark & \redxmark 
        & \redxmark & \redxmark & \redxmark & \redxmark \\
    PowerInfer~\cite{song2024powerinfer}        
        & \greencheck & \greencheck & \redxmark & \redxmark 
        & \redxmark
        & \greencheck 
        & \greencheck & \greencheck 
        & \redxmark & \redxmark & \redxmark & \redxmark \\
    LMDeploy~\cite{2023lmdeploy}        
        & \greencheck & \greencheck & \greencheck & \redxmark 
        & \redxmark
        & \greencheck 
        & \greencheck & \greencheck
        & \redxmark & \redxmark & \redxmark & \redxmark \\
    LightLLM~\cite{lightllm}          
        & \greencheck & \greencheck & \redxmark & \redxmark 
        & \redxmark
        & \greencheck 
        & \greencheck & \redxmark 
        & \redxmark & \redxmark & \redxmark & \redxmark \\
    NanoFlow~\cite{zhu2024nanoflow}          
        & \redxmark & \greencheck & \redxmark & \redxmark
        & \redxmark
        & \greencheck 
        & \redxmark & \redxmark 
        & \redxmark & \redxmark & \redxmark & \redxmark \\
    DistServe~\cite{zhong2024distserve}        
        & \greencheck & \greencheck & \redxmark & \redxmark 
        & \redxmark
        & \redxmark 
        & \redxmark & \redxmark
        & \redxmark & \redxmark & \redxmark & \redxmark \\
    vAttention~\cite{prabhu2025vattention}    
        & \greencheck & \greencheck & \greencheck & \redxmark
        & \redxmark
        & \greencheck 
        & \greencheck & \greencheck
        & \redxmark & \redxmark & \redxmark & \redxmark \\
    Sarathi-Serve~\cite{agrawal2024taming}    
        & \greencheck & \greencheck & \redxmark & \redxmark 
        & \redxmark
        & \greencheck 
        & \redxmark & \redxmark
        & \redxmark & \redxmark & \redxmark & \redxmark \\
    Friendli Inference~\cite{friendli} 
        & \greencheck & \greencheck & \greencheck & \redxmark
        & \redxmark
        & \greencheck 
        & \greencheck & \greencheck 
        & \redxmark & \redxmark & \redxmark & \redxmark \\
    Fireworks AI~\cite{fireworks}      
        & \redxmark & \greencheck & \greencheck & \redxmark 
        & \redxmark
        & \redxmark 
        & \redxmark & \redxmark 
        & \redxmark & \redxmark & \redxmark & \redxmark \\
    GroqCloud~\cite{groqcloud}      
        & \greencheck & \greencheck & \redxmark & \redxmark 
        & \redxmark
        & \redxmark 
        & \greencheck & \redxmark 
        & \redxmark & \redxmark & \redxmark & \redxmark \\
    Together Inference~\cite{together} 
        & \redxmark & \greencheck & \greencheck & \redxmark 
        & \redxmark
        & \redxmark 
        & \redxmark & \greencheck 
        & \redxmark & \redxmark & \redxmark & \redxmark \\ 
    \bottomrule
    \end{tabular}%
    }
\end{table}

These quantization techniques are closely connected to the hardware supported by each inference engine. As indicated in Table~\ref{tab:frameworks_data_type}, every engine accommodates specific data types, which in turn govern how quantized models are executed.

In particular, data types based on block-level scaling in the 4-8-bit range---such as the microscaling (MX) format~\cite{rouhani2023microscaling}---have been proposed to balance training and inference performance, accuracy, and framework compatibility. Examples include MXFP8 and MXINT8. Each MX block consists of a single scale value X and a set of compressed values ($P_1, P_2,\dots,P_k$). The scale format (e.g., E8M0) and element format (e.g., FP4, FP6, FP8, INT8) can be independently configured.

Unlike traditional FP8 or INT8 formats, which require a single tensor-level scaling factor to match the dynamic range of the entire tensor, MX formats split the tensor into smaller sub-blocks and assign separate scale values to each, thereby circumventing the limitations of sub-8-bit formats. Hardware platforms such as the Qualcomm Cloud AI 100~\cite{chatha2021qualcomm} and NVIDIA GPUs based on the Hopper architecture support MX formats, and some inference engines (e.g., TensorRT-LLM~\cite{tensorrtllm}) offer support for them.

\subsubsection{Pruning} \label{sec:inference_optimization_pruning}

\begin{figure}[tbp]
    \centering
    \begin{minipage}[t]{0.42\linewidth}
        \centering
        \includegraphics[width=0.95\linewidth]{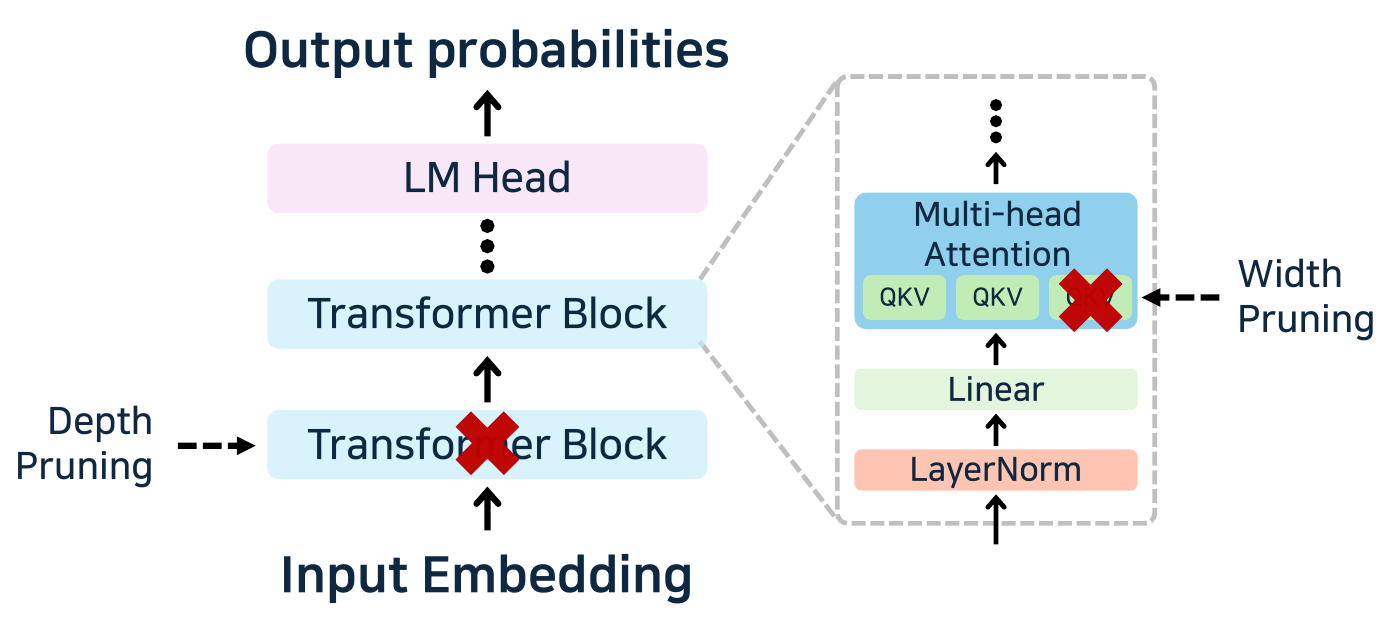}
        \captionof{figure}{Pruning in Transformer-based Model}
        \label{fig:pruning_llm}
    \end{minipage}
    \hspace{2mm}
    \begin{minipage}[t]{0.43\linewidth}
        \centering
        \includegraphics[width=\linewidth]{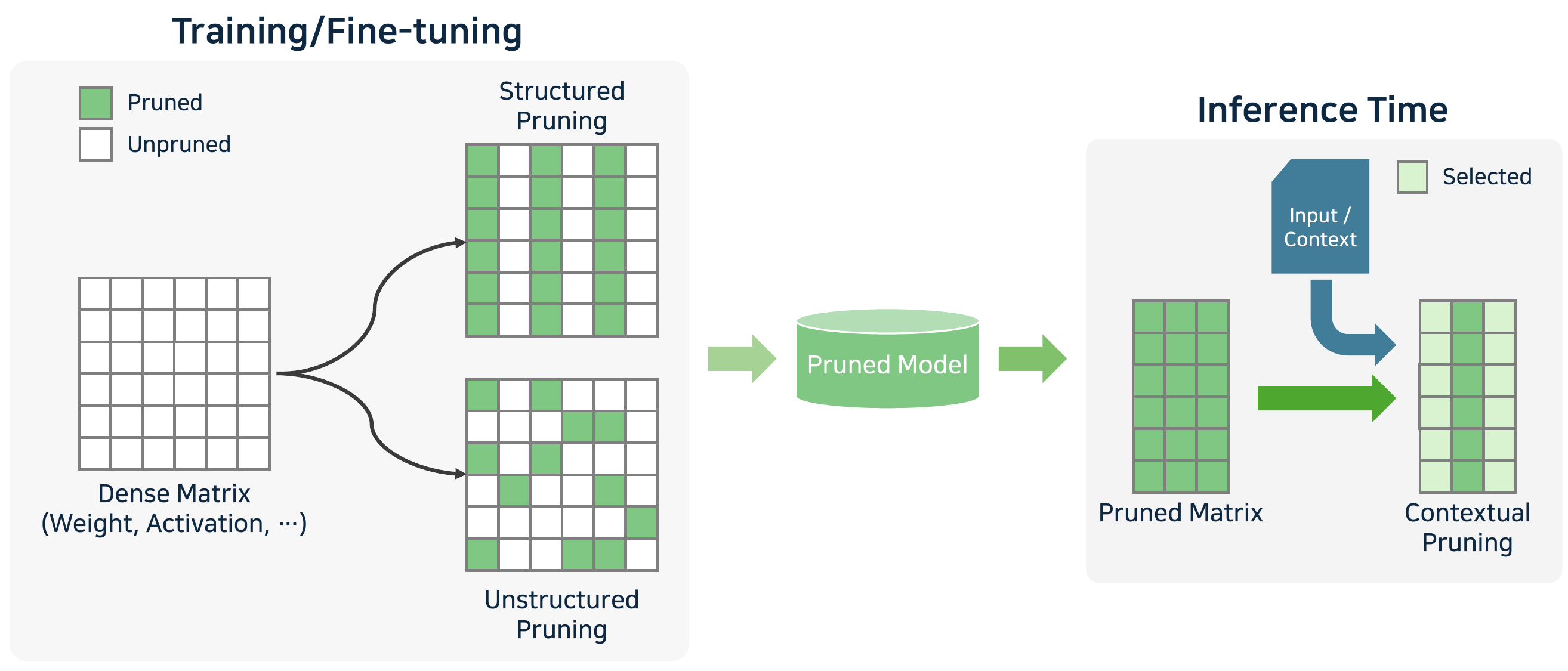}
        \captionof{figure}{Three Types of Pruning: Structured, Unstructured, and Contextual}
        \label{fig:pruning_mechanisms}
    \end{minipage}
\end{figure}

\textbf{Overview and Inference Benefits of Pruning.}
Pruning~\cite{zhu2024survey, kim2024efficient} is a model compression technique that removes less important parameters to reduce model size, often targeting weights or attention heads, as shown in Fig.~\ref{fig:pruning_llm} in Transformer-based LLMs. It can be applied after training by zeroing out and removing certain weights, applied dynamically during training/fine-tuning~\cite{liu2025pat}, or performed through one-shot pruning~\cite{shao2024one, frantar2023sparsegpt} to rapidly shrink large models.

From an inference perspective, pruning reduces the number of parameters, thereby improving memory utilization, bandwidth efficiency, and cache usage. As more weights become zero, sparse computation can reduce actual computation costs. However, to fully exploit sparse computation, the inference engine or compute library must support kernels capable of skipping or efficiently handling zero weights.

\textbf{Three Types of Pruning.}
Pruning methods in LLMs generally fall into three categories: structured pruning, unstructured pruning, and contextual pruning~\cite{bai2024beyond}, and each pruning method is illustrated in Fig.~\ref{fig:pruning_mechanisms}. Structured pruning eliminates groups of parameters with fixed structures, such as convolutional filters or neuron channels. Since the matrix dimensions are physically reduced, the inference speed can be improved even without specialized sparse computation kernels. However, inference graphs or kernels must be adjusted to match the pruned architecture. One of the structured pruning methods is LLM-Pruner~\cite{ma2023llm}, which prunes low-importance structures based on gradient information.

Unstructured pruning removes individual weights according to their importance scores. Although this reduces model size and FLOPS, the use of random sparsity patterns can limit performance gains on dense matrix multiplication kernels, requiring optimized sparse kernels for effective acceleration. For example, NVIDIA CUDA provides sparse operations via cuSPARSE~\cite{cusparse}. Among the unstructured pruning methods, Wanda~\cite{sun2023simple} prunes weights based on the product of weight magnitude and input activation.

Contextual pruning dynamically assesses the importance of weights depending on the input context or domain, selectively removing or retaining weights. This approach adapts the model to specific inputs, skips unnecessary computation paths, and improving inference efficiency. Although sparse computation may yield smaller performance gains compared to other methods, it can enhance domain-specific accuracy. Implementing contextual pruning requires inference engines to support conditional branching or precompiled kernels with logic for bypassing certain layers. For example, Mini-GPTs~\cite{valicenti2023mini} applied contextual pruning to Phi-1.5~\cite{li2023textbooks} and Opt-1.3~\cite{zhang2022opt} using legal and medical QA datasets to prune linear, activation, and embedding layers.

\textbf{Recent Advances: Post-Training and Token Pruning.}
Like quantization, post-training pruning has been a research focus for LLMs, given the high cost of training or fine-tuning. Recent efforts involve unstructured and semi-structured post-training pruning algorithms to address the Multiple Removal Problem (MRP) by pruning large quantities of weights at the LLM layer level~\cite{zhao2024pruning}.

In LLM inference, as the input token length increases, the TTFT generally increases as well. One proposed solution is token pruning~\cite{fu2024lazyllm}, which selectively computes $\mathbf{KV}$ representations solely for tokens judged important for next-token prediction---without requiring extra training or fine-tuning. The remaining tokens are deferred and only computed if needed, reducing initial computation costs and improving TTFT.

\textbf{Engine-Level Support for Pruned Models.}
Among the inference engines discussed in this paper, fewer than half directly support pruning. Most rely on NVIDIA pruning libraries, with DeepSpeed-FastGen~\cite{holmes2024deepspeed} explicitly supporting row, head, sparse, and structured/unstructured pruning through the DeepSpeed backend. Other engines generally only support running pre-pruned models.

\subsubsection{Sparsity Optimization} \label{sec:inference_optimization_sparsity}

\begin{figure}[tbp]
    \centering
    \hspace*{\fill}
    \begin{minipage}[b]{0.56\textwidth}
        \centering
        \begin{subfigure}[t]{0.33\textwidth}
            \includegraphics[width=\linewidth]{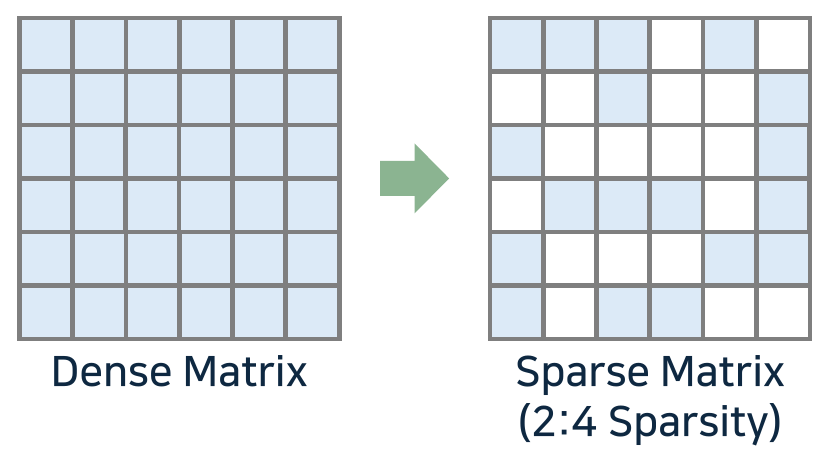}
            \caption{Structured Sparsity - N:M sparsity}
            \label{fig:structured_sparsity}
        \end{subfigure}
        \hspace{2mm}
        \begin{subfigure}[t]{0.61\textwidth}
            \includegraphics[width=\linewidth]{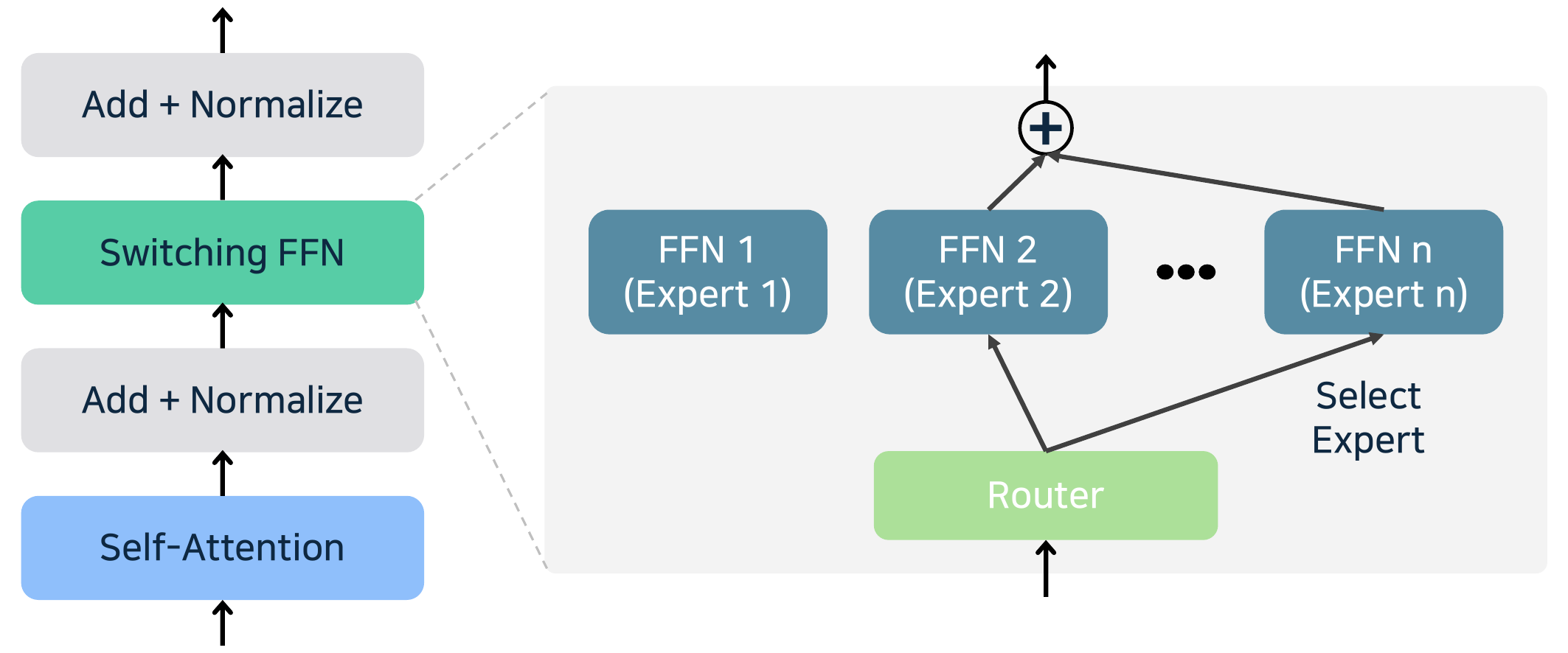}
            \caption{Dynamic Sparsity - MoE}
            \label{fig:dynamic_sparsity_moe}
        \end{subfigure}
        \captionof{figure}{Sparsity Optimizations}
        \label{fig:sparsity}
    \end{minipage}
    \begin{minipage}[b]{0.4\textwidth}
        \centering
        \includegraphics[width=0.9\linewidth]{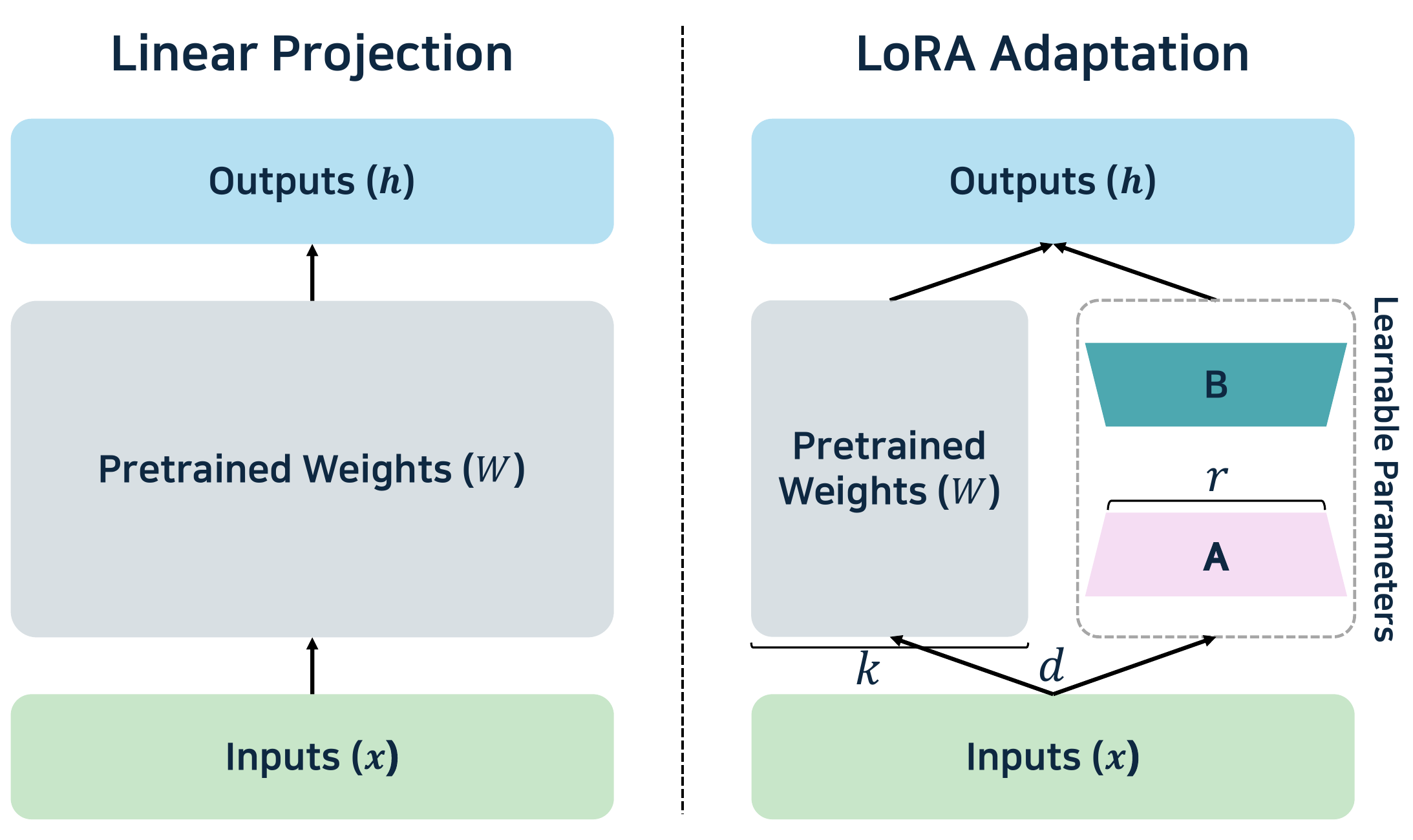}
        \captionof{figure}{LoRA}
        \label{fig:lora}
    \end{minipage}
    \hspace*{\fill}
\end{figure}

\textbf{Sparsity Optimization Overview.}
Sparsity optimization~\cite{fan2025spinfer, shin2024sparseinfer} is a technique that reduces computational costs and speeds up inference by increasing the number of zero values in model weights or activations. This approach can reduce memory usage and improve compute performance when supported by hardware or inference engines that allow sparse operations. Although it shares the same goal as pruning, sparsity optimization focuses on designs sparse model structures or applies predefined sparse patterns to achieve computational efficiency. Sparsity can be applied to the attention mechanism (e.g., sparse attention patterns) or to the weights of individual heads. Pruning can induce sparsity and models that are already sparse can be further refined through pruning. Sparsity optimization techniques include structured sparsity~\cite{zheng2024learn, dong2023towards}, dynamic sparsity~\cite{zhang2023dynamic}, and kernel-level sparsity~\cite{xia2023flash, borvstnik2014sparse}.

\textbf{Structured Sparsity.}
Structured sparsity~\cite{zheng2024learn, dong2023towards} imposes sparsity on weights or tensor values in fixed patterns, simplifying hardware-level optimizations. Typical examples are N:M sparsity~\cite{zhang2022learning} as shown in Fig.~\ref{fig:sparsity}~(\subref{fig:structured_sparsity}), where n values within an m-sized block remain active; and block sparsity~\cite{gao2024seerattention} which divides weight matrices into blocks and removes values to create sparsity. On NVIDIA GPUs and similar hardware, these static patterns can be optimized at the time of model compilation. However, rigid patterns may negatively affect model performance.

\textbf{Dynamic Sparsity and Mixture-of-Experts.}
Dynamic sparsity~\cite{zhang2023dynamic} activates only the computations required at run time based on input tokens, skipping unnecessary operations to improve efficiency. A prominent example is MoE~\cite{cai2024survey}, which replaces the MLP with multiple FFNs (experts) but activates only a subset of them, depending on the input tokens, as shown in Fig.~\ref{fig:sparsity}~(\subref{fig:dynamic_sparsity_moe}). This reduces the number of computations per token and enables large models to run more efficiently. To support MoE, the inference engine must offer gating mechanisms and flexible architectures for dynamic routing. Example MoE models include Mixtral 8x7B~\cite{jiang2024mixtral}, DeepSpeed-MoE~\cite{rajbhandari2022deepspeed}, and DeepSeek-R1~\cite{guo2025deepseek}, which allow more tokens to be processed or trained with limited time or resources. Most inference engines provide MoE support and related optimizations.

\textbf{Enhancements to Sparse MoE Training.}
In dense expert models, all experts are activated for every input, leading to a more complicated computation. Sparse MoE~\cite{fedus2022switch, du2022glam} complements this by using only some of them, such as selecting only top-k experts. To address reduced specialization or unstable training due to sparse-gating, SMoE-Dropout~\cite{chen2023sparse} employs randomized router networks that gradually increase the number of active experts throughout training to refine the model.

\textbf{Token and Contextual Sparsity.}
Another example of dynamic sparsity is dynamic token sparsity~\cite{yang2024post, fu2024lazyllm} which avoids computing attention for all tokens by focusing on a subset. In addition to contextual pruning~\cite{valicenti2023mini}, contextual sparsity~\cite{liu2023deja,akhauri2024shadowllm} has also been proposed, which selectively activates only a subset of attention heads or MLP parameters depending on the input. In contextual sparsity research such as Deja Vu~\cite{liu2023deja}, a lightweight sparsity method was employed to dynamically skip computations based on the input context, addressing the high cost of verifying whether contextual sparsity truly exists for each input.

\textbf{Kernel-Level Sparsity.}
Kernel-level sparsity checks for zero values in the computation kernels and skips them. For example, sparse matrix-dense matrix multiplication (SpMM)~\cite{borvstnik2014sparse} kernels or rely on libraries like cuTeSpMM~\cite{xiang2025cutespmm} to utilize NVIDIA GPU Tensor Cores. The xFormers~\cite{xFormers2022} library also includes CUDA kernels for memory-efficient attention, sparse attention, and block-sparse attention.

\textbf{Support in Inference Engines.}
Several engines already integrate such methods. vLLM~\cite{kwon2023efficient}, SGLang~\cite{zheng2024sglang} and TGI~\cite{tgi} support N:M sparsity, with vLLM also offering block sparsity. SGLang applies Double Sparse Attention~\cite{yang2024post}, a post-training sparse attention approach that prioritizes essential tokens (token sparsity) in self-attention while determining important feature channels offline (channel sparsity). DeepSpeed-FastGen~\cite{holmes2024deepspeed} adopts the Sparse Attention technique to introduce block-level sparsity in self-attention, reducing compute and memory usage through a variety of patterns and custom modifications. TensorRT-LLM~\cite{tensorrtllm} employs Block Sparse Attention to accelerate SpMM through block-structured sparsity, relying on the Sparse Tensor Cores featured in NVIDIA GPUs from the Ampere architecture onward.



\subsection{\news{Inference-Aware Fine-tuning}} \label{sec:inference_optimization_fine_tuning}

LLMs typically rely on foundation models pretrained on large-scale datasets to perform diverse inference tasks. However, for domain-specific or task-specific optimization, fine-tuning can significantly boost model performance.

\textbf{Parameter-Efficient Fine-Tuning.} 
Fine-tuning is a technique originally applied in CNN that modifies the parameters of a pre-trained model. It can be divided into full-parameter fine-tuning~\cite{lv2023full}, where every parameter is updated, and parameter-efficient fine-tuning (PEFT)~\cite{ding2023parameter}, where only a subset of parameters is adjusted. Because full fine-tuning requires substantial hardware resources, LLMs generally favor PEFT methods that update only part of the model. Although fine-tuning mainly targets model training, it directly affects LLM inference performance as well. Fine-tuning can be implemented by inserting adapter networks (additional layers) to recalibrate parameters~\cite{hu2023llm} or by supplying domain-specific data through prompt engineering~\cite{ye2023prompt}.

\textbf{Low-rank adaptation.} 
LoRA~\cite{hu2022lora, sheng2023s} is a representative PEFT approach. Instead of updating the entire model, LoRA keeps the original weights frozen and trains additional low-rank matrices to adjust the model parameters. Research indicates that despite the large dimensions of the FC layers in LLMs, the effective dimensionality required for adaptation is relatively low. LoRA exploits this by approximating weight updates with low-rank matrices, greatly reducing the training cost. As shown in Fig.~\ref{fig:lora}, LoRA retains the pre-trained weight matrix and trains only two small matrices, $A$ and $B$. Rather than directly computing updates for the full weight matrix ($d\times k$, $d:$ input dimension, $k:$ output dimension), LoRA approximates $A\times B$ by low rank ($r$) matrix multiplication ($d\times r$ and $r\times k$).

A major advantage of LoRA is the ability to swap in different LoRA modules for varying tasks, facilitating quick adaptation without retraining the entire model. Moreover, since LoRA only updates the small low-rank matrices, training is faster and demands less memory. Merging these trained modules with the original weights generally does not degrade inference speed, although merging may limit the capacity for multi-task processing in real time.

\textbf{Quantized Low-rank adaptation.}
As LLM sizes increase, additional methods like Quantized LoRA (QLoRA)~\cite{dettmers2023qlora, zhang2023machine} have been introduced, combining 4-bit quantization with LoRA fine-tuning. QLoRA backpropagates through a 4-bit quantized model, preserving LoRA's benefits while reducing memory usage. This enables running large models on a single device, offering efficient options for both training and inference.

\textbf{Support in Inference Engines.}
Ollama~\cite{ollama}, llama.cpp~\cite{llamacpp}, Friendli Inference~\cite{friendli}, and Fireworks AI~\cite{fireworks} support LoRA, morevoer DeepSpeed-FastGen~\cite{holmes2024deepspeed}, SGLang~\cite{zheng2024sglang}, TensorRT-LLM~\cite{tensorrtllm}, Unsloth~\cite{unsloth}, and vLLM~\cite{kwon2023efficient} support QLoRA. Especially, vLLM~\cite{kwon2023efficient}, TensorRT-LLM~\cite{tensorrtllm}, TGI~\cite{tgi}, LMDeploy~\cite{2023lmdeploy}, Friendli Inference~\cite{friendli}, and Together Inference~\cite{together} also provide Multi-LoRA functionality, enabling simultaneous serving of multiple user-customized models.

\subsection{Caching} \label{sec:inference_optimization_caching}

In LLMs containing billions to trillions of parameters, repeatedly generating hundreds to millions of tokens requires substantial computation and memory resources. To address this, most inference engines employ various caching strategies that reduce redundant computations and lower latency. Caching can be applied to several LLM components, and different caching optimizations can be combined and used together.

\subsubsection{Prompt Caching} \label{sec:inference_optimization_caching_prompt}

\begin{figure}[tbp]
    \centering
    \resizebox{.99\textwidth}{!}{%
    \begin{minipage}[t]{0.46\textwidth}
        \centering
        \includegraphics[width=\linewidth]{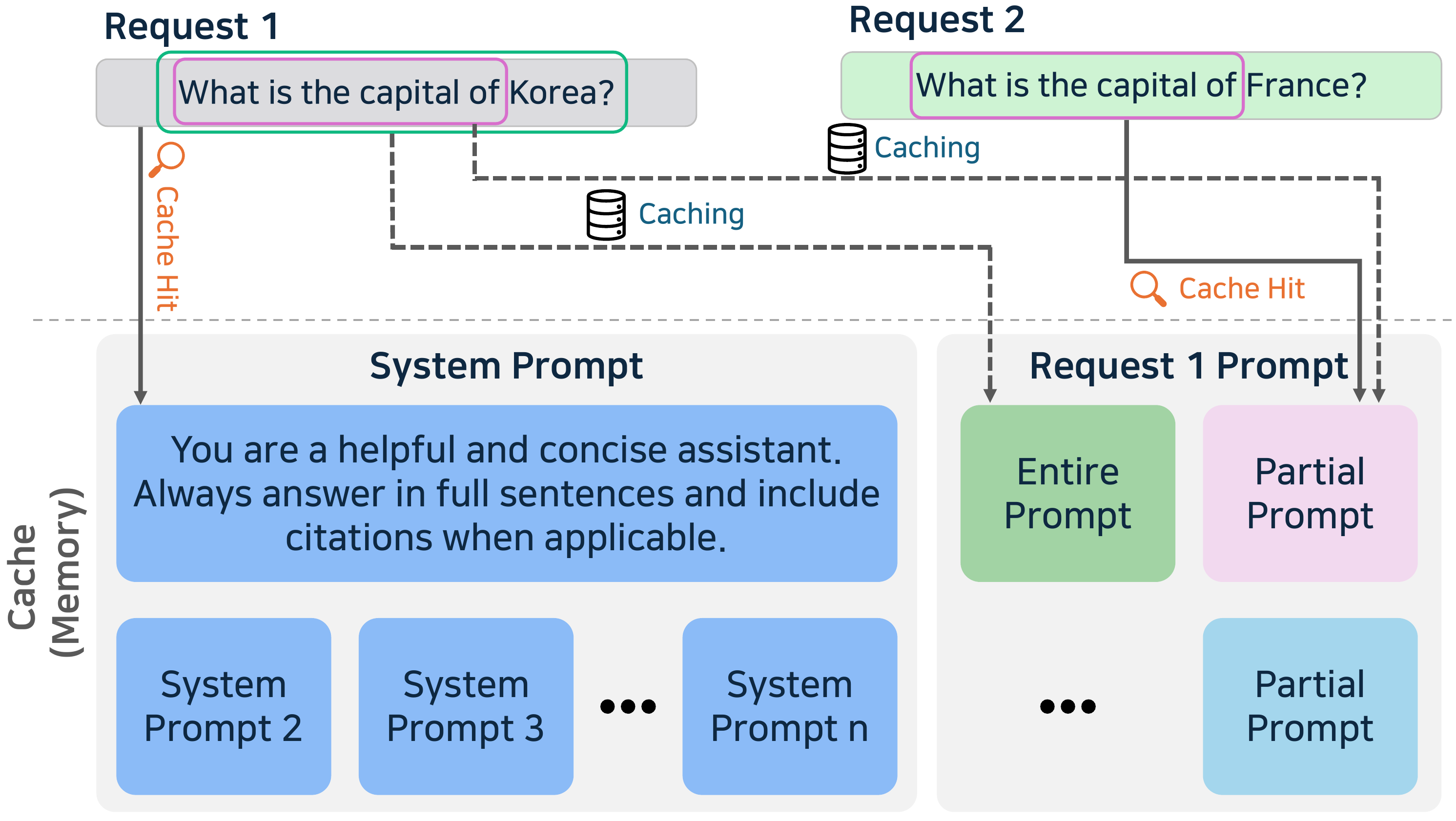}
        \captionof{figure}{Prompt Caching}
        \label{fig:prompt_caching}
    \end{minipage}
    \hspace{0.04\textwidth}
    \begin{minipage}[t]{0.48\textwidth}
        \centering
        \includegraphics[width=\linewidth]{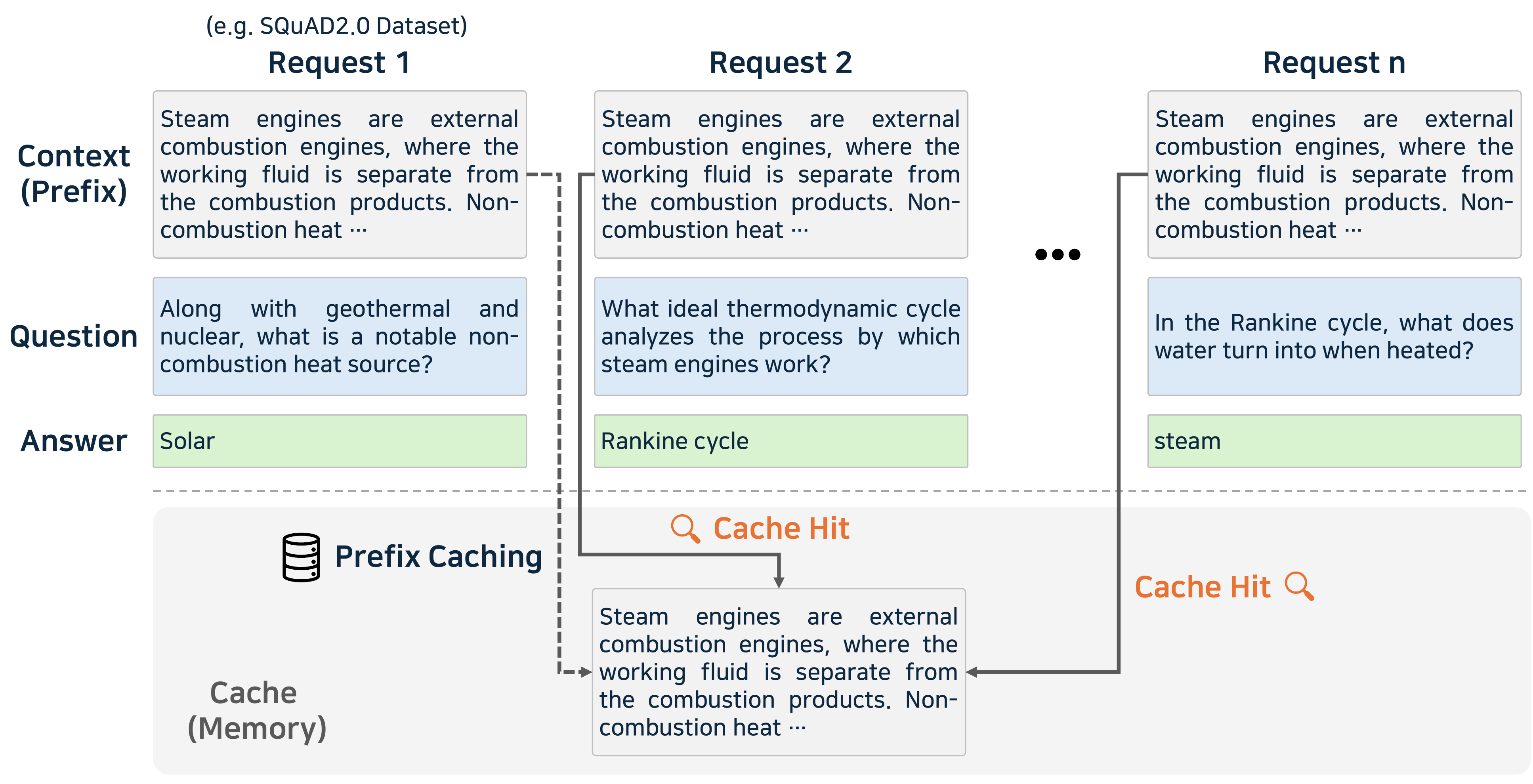}
        \captionof{figure}{Prefix Caching}
        \label{fig:prefix_caching}
    \end{minipage}
    }
\end{figure}

A large portion of LLM prompts may include frequently reused text. Identical content, such as system messages or common instructions, often appears multiple times---particularly in conversational agents, coding assistants, or extensive document processing. To optimize this, a technique known as Prompt Cache~\cite{gim2024prompt,zhu2024efficient} has been introduced. The Prompt Cache stores attention states for frequently used text segments in advance, and when the same segment reappears in a prompt, it reuses those stored attention results and only computes the new segments, thereby expediting inference. However, because the transformer architecture applies positional encoding, the attention states depend on the position of each segment, meaning that reuse is only possible if the segment appears in the same position. To overcome this limitation, Prompt Markup Language (PML) was proposed. It explicitly defines a prompt's structure, identifies reusable segments (prompt modules), and assigns unique position IDs to each module. PML functions as a schema for module positions and hierarchy, offering an interface for generating and reusing attention states at the module level. An example of prompt caching is shown in Fig.~\ref{fig:prompt_caching}.

\textbf{Prompt Caching in Commercial AI Services.}
Commercial services such as ChatGPT~\cite{chatgpt} and Claude~\cite{claude} also employ prompt caching. ChatGPT routes API requests with the same prompt to the same server that previously handled them, allowing reuse of cached results instead of recomputing from the first token. This approach has reduced latency by as much as 80\% for long prompts. To increase cache hit rates, ChatGPT places static components (e.g., instructions, examples) at the beginning of the prompt, while dynamic user content goes at the end. ChatGPT applies caching to prompts exceeding 1,024 tokens and triggers cache hits in 128-token segments. It retains cached prompts for five to sixty minutes, depending on system load, automatically evicting any prompts that remain unused. Claude employs a similar format, providing up to four cache breakpoints that split prompts into multiple cacheable segments.

\textbf{Cache Hit Prediction.}
A method has also been proposed to improve the accuracy of prompt caching~\cite{zhu2024efficient}. This study recommends predicting the effectiveness of caching based on embedding similarity. In single-turn question-answer scenarios, the study uses embeddings refined through knowledge distillation to determine whether cached responses can be reused. Cosine similarity is computed between prompt embeddings, and a model is trained to decide if the same response can be reapplied. This research also presents finite sample guarantees for loss functions like Binary Cross Entropy (BCE) and Squared Log Difference (SLD).

\textbf{Support in Inference Engines.}
Several inference engines, including Ollama~\cite{ollama}, llama.cpp~\cite{llamacpp} and TensorRT-LLM~\cite{tensorrtllm}, support prompt caching. TensorRT-LLM~\cite{tensorrtllm} provides system prompt caching, while Ollama~\cite{ollama} offers optimized prompt caching even in multi-user environments.

\subsubsection{Prefix Caching} \label{sec:inference_optimization_caching_prefix}

Prefix Caching~\cite{liu2024optimizing, pan2024marconi} is conceptually similar to prompt caching~\cite{gim2024prompt, zhu2024efficient} but focuses on caching only the common prefix segments that reappear across multiple requests, rather than caching the entire prompt, as shown in Fig.~\ref{fig:prefix_caching}. During batched inference, when multiple prompts share the same prefix, the computation for that shared segment can be reused to improve efficiency. For instance, in question-answering tasks where the same system prompt or few-shot examples are repeatedly used, caching these portions of the prompt during the prefill phase can reduce overall inference time.

However, prefix caching typically accelerates only the prefill phase and does not affect the decode phase. Consequently, if the primary bottleneck stems from extended decoding for very long responses, the performance gains from prefix caching may be limited. In addition, if a new request does not share a prefix with any existing request, the caching advantage diminishes.

\textbf{Support in Inference Engines.}
vLLM~\cite{kwon2023efficient} provides Automatic Prefix Caching (APC), which stores the $\mathbf{KV}$ cache from previous requests and reuses it whenever a new request shares a prefix with an existing one, thus skipping attention computations for the shared segments. TGI~\cite{tgi} employs high-performance data structures instead of basic string matching to speed up prefix lookups and applies chunking code to optimize memory usage; it also integrates prefix caching with Flashdecoding~\cite{hong2024flashdecoding++} kernels to support rapid inference for lengthy sequences. MAX~\cite{max} employs a PagedAttention~\cite{kwon2023efficient}-based mechanism to apply prefix caching and improve inference efficiency, and other engines, such as LMDeploy~\cite{2023lmdeploy}, also include prefix caching features.

\subsubsection{KV Caching} \label{sec:inference_optimization_caching_kv}

\begin{figure}[tbp]
    \centering
    \includegraphics[width=.65\linewidth]{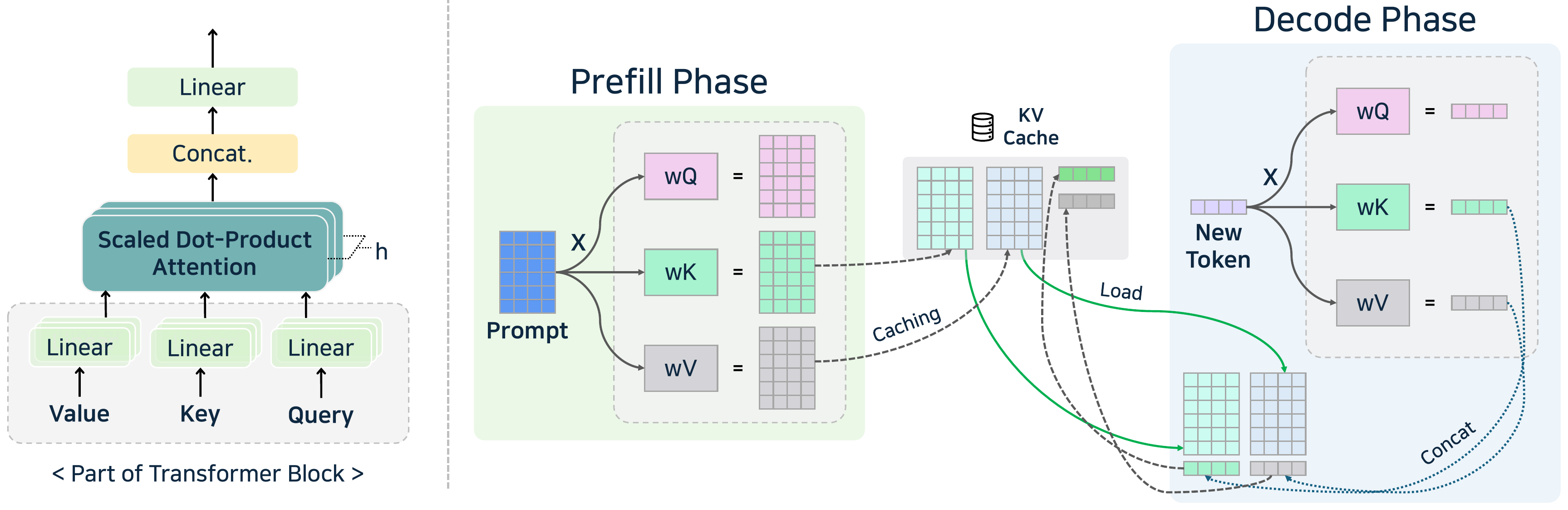}
    \caption{KV Caching}
    \label{fig:kv_caching}
\end{figure}

In self-attention of the Transformer, each token attends to all preceding tokens, resulting in a time complexity of $\mathcal{O}(n^2)$. To reduce this overhead, $\mathbf{KV}$ Caching~\cite{pope2023efficiently, ge2023model, wang2024modelkv} was proposed. By storing the K and V matrices produced in each token step and reusing them for subsequent tokens, inference can run in $\mathcal{O}(n)$ time. The $\mathbf{KV}$ Cache operation in the prefill and decode phases is as shown in Fig.~\ref{fig:kv_caching}. This provides significant efficiency gains when large batch size processing or multi-turn conversation scenarios. $\mathbf{KV}$ caching also works well alongside other optimizations such as Prefix Caching~\cite{liu2024optimizing, pan2024marconi} and Speculative Decoding~\cite{leviathan2023fast, liu2024online, spector2023accelerating}.

\textbf{Increasing KV Cache Size.}
However, the memory needed for $\mathbf{KV}$ caching increases significantly with longer contexts because both $\mathbf{K}$ and $\mathbf{V}$ must be stored in memory. The memory needed for LLM inference can be expressed as $S_{KV}/\text{token} = 2 \times n_{\text{layers}} \times (n_{\text{heads}} \times d_{\text{head}}) \times precision$ and the total size of KV Cache can be expressed as $\Sigma~S_{KV} = s_{\text{batch}} \times l_{\text{seq}} \times 2 \times n_{\text{layers}} \times s_{\text{hidden}} \times precision$.

In the equation, $n_{layers}$ represents the number of layers, $n_{heads}$ represents the number of attention heads, and $d_{head}$ represents the dimension of the heads. $s_{batch}$ is the batch size, $l_{seq}$ is the sequence length, $s_{hidden}$ is the hidden size, $precision$ is precision bytes such as FP16, and each KV cache size ($S_{KV}$, $\Sigma~S_{KV}$ is in bytes. Multiplication by 2 in the equation accounts for both the $\mathbf{K}$ and $\mathbf{V}$ components.

Reducing batch size to decrease $\mathbf{KV}$ cache memory lowers memory usage but also reduces hardware utilization and throughput. Limiting the sequence length forces the recomputation of some $\mathbf{KV}$ caches, causing inefficiency. Reducing the depth of the model may compromise performance; hence, current research focuses on attention mechanisms, cache compression, and quantization techniques.

\textbf{KV Cache Optimizations.}
As shown in Fig.~\ref{fig:attention}, attention mechanisms like GQA~\cite{ainslie2023gqa} and MQA~\cite{shazeer2019fast} reduce the number of $\mathbf{Q}$ heads or reuse $\mathbf{KV}$ heads, naturally shrinking $\mathbf{KV}$ cache size. Research also focuses on compressing $\mathbf{KV}$ caches. MiniCache~\cite{liu2024minicache} uses depth-wise compression, observing that middle-to-deep layer $\mathbf{KV}$ caches are highly similar. H2O~\cite{zhang2023h2o} leverages sparsity of the attention matrix to reuse only essential parts while retaining the same output tokens. FlexGen~\cite{sheng2023flexgen} profiles the structural properties of the attention heads and proposes adaptive $\mathbf{KV}$ cache compression. As explained in Section~\ref{sec:inference_optimization_quantization} above, quantization can also be applied to reduce the $\mathbf{KV}$ cache size~\cite{hooper2024kvquant,liu2024kivi}.

As context windows and model sizes grow, GPU memory alone can become insufficient to store $\mathbf{KV}$ caches, leading to offloading solutions. InfiniGen~\cite{lee2024infinigen} prefetches necessary $\mathbf{KV}$ segments into CPU memory speculatively, while CacheGen~\cite{liu2024cachegen} encodes and streams $\mathbf{KV}$ caches as compressed bitstreams in distributed systems, reducing network transfer latency.

\textbf{Support in Inference Engines.}
Many inference engines implement $\mathbf{KV}$ caching. LMDeploy~\cite{2023lmdeploy} and Ollama~\cite{ollama} use INT8 or FP16 quantization to reduce the cache size. DeepSpeed-FastGen~\cite{holmes2024deepspeed} applies ZeRO-Inference~\cite{aminabadi2022deepspeed}-based offloading and 4-bit quantization to increase throughput in resource-constrained settings. TensorRT-LLM~\cite{tensorrtllm} offers various optimizations, including a paged $\mathbf{KV}$ cache, $\mathbf{KV}$ cache quantization, a circular $\mathbf{KV}$ cache, and $\mathbf{KV}$ cache reuse. It saves memory using an LRU-based eviction strategy and $\mathbf{KV}$-aware routing or scheduling that forwards requests to instances with the required $\mathbf{KV}$ cache already in place. vLLM~\cite{kwon2023efficient} implements $\mathbf{KV}$ Cache Preemption, allowing some requests to be preempted to free up space. Although preempted requests required recomputation and may increase end-to-end latency, this approach can enhance overall system stability.

\subsection{Attention Optimization} \label{sec:inference_optimization_attention}

Attention lies at the core of transformer-based LLMs, but its high memory and computational costs make optimization essential for efficient service deployment. As the sequence length grows, the time required for attention operations rises quadratically, and calculating and storing $\mathbf{Q}$, $\mathbf{K}$ and $\mathbf{V}$ matrices consumes substantial memory. Increasing memory efficiency and batch size is therefore, critical for managing multiple requests and boosting overall throughput.

To improve inference performance, most LLM inference engines utilize a $\mathbf{KV}$ Cache mechanism~\cite{liu2024cachegen,strati2024dejavu} that stores the $\mathbf{K}$ and $\mathbf{V}$ vectors from previously generated tokens. This approach prevents redundant computations when producing subsequent tokens, thereby reducing both inference latency and computational overhead.

LLM inference engines offer a range of optimization techniques for efficiently storing and retrieving the $\mathbf{KV}$ cache, as well as methods to improve the use of $\mathbf{Q}$, $\mathbf{K}$ and $\mathbf{V}$ vectors are utilized during inference.

\subsubsection{KV Cache Optimization: PagedAttention} \label{sec:inference_optimization_attention_paged}

\begin{figure}[tbp]
    \centering
    \resizebox{0.99\textwidth}{!}{%
        \begin{minipage}[b]{0.39\textwidth}
            \centering
            \includegraphics[width=\linewidth]{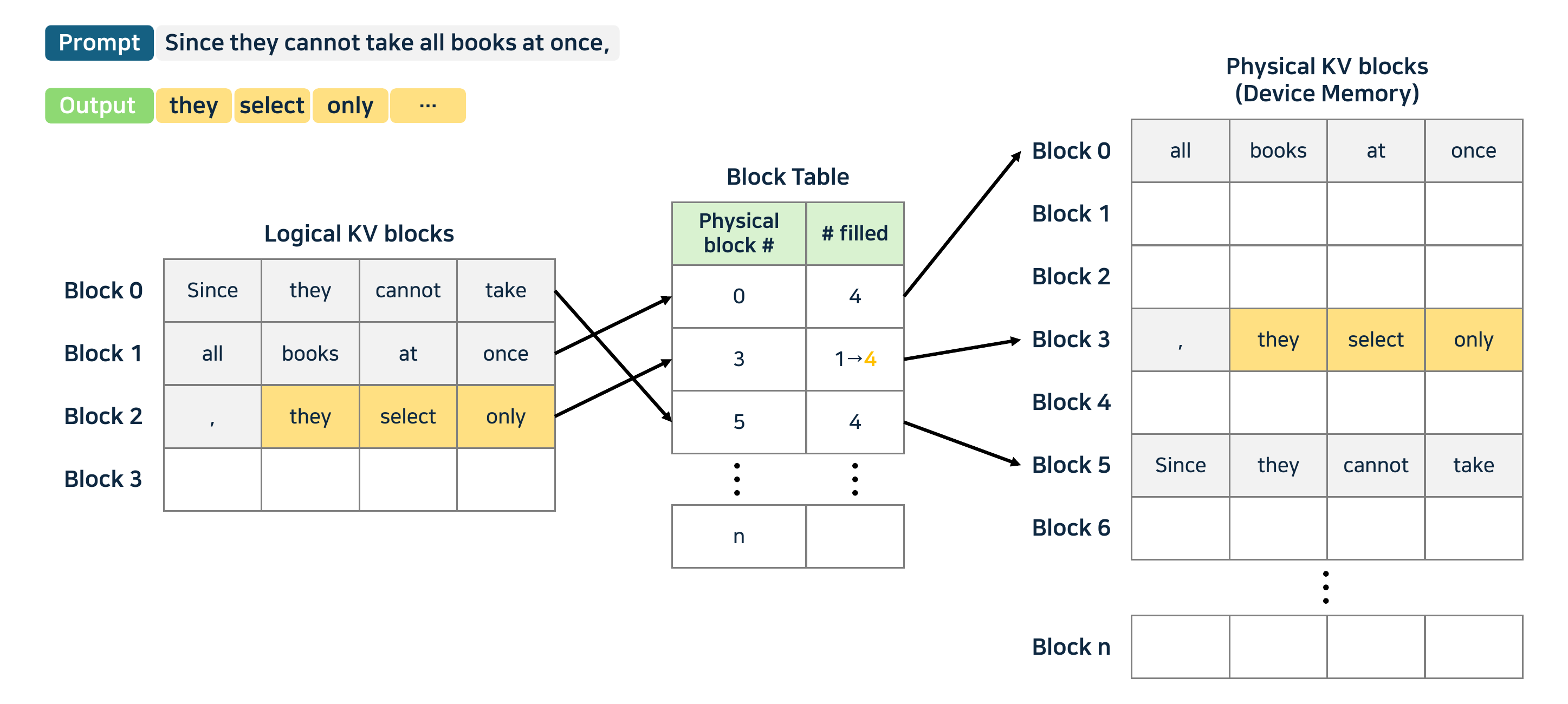}
            \caption{PagedAttention}
            \label{fig:pagedattentnion}
        \end{minipage}
        \hspace{0.01\textwidth}
        \begin{minipage}[b]{0.57\textwidth}
            \centering
            \includegraphics[width=\linewidth]{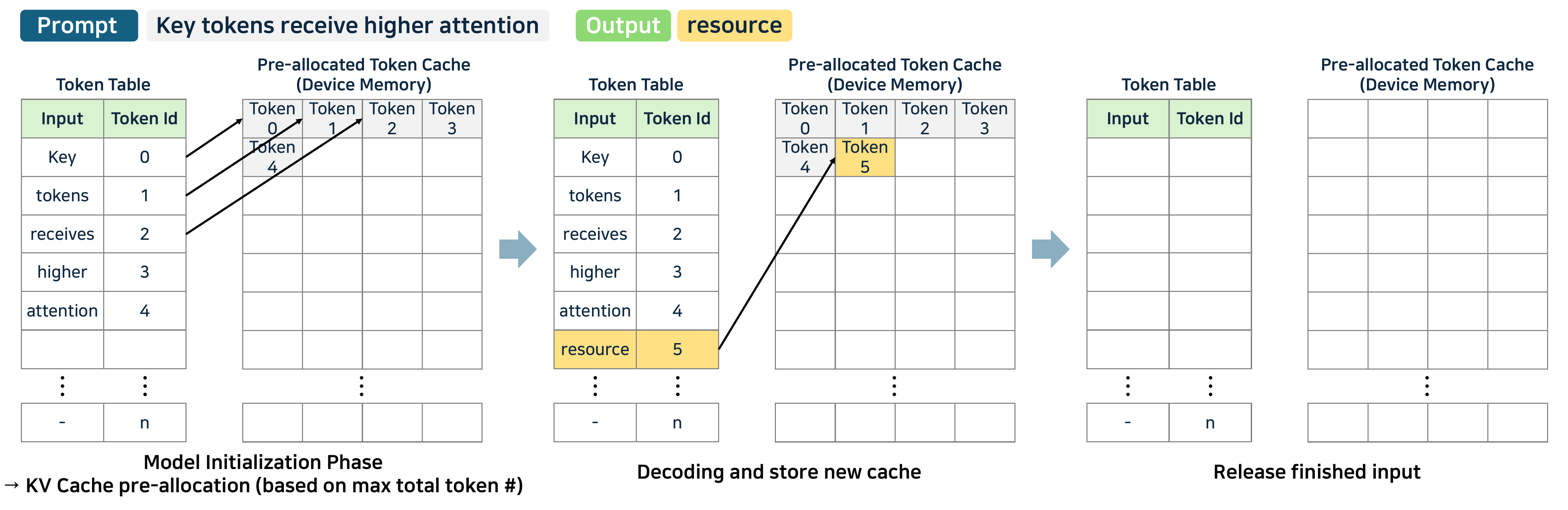}
            \caption{TokenAttention}
            \label{fig:tokenattentnion}
        \end{minipage}
    }
\end{figure}

Effective $\mathbf{KV}$ cache management is crucial for enhancing LLM inference performance. Conventional inference engines often allocate a contiguous $\mathbf{KV}$ cache based on the maximum sequence length, which can lead to internal and external fragmentation as sequence lengths vary. These fragmentation issues reduce memory efficiency and limit parallelism.

PagedAttention~\cite{kwon2023efficient} addresses this problem by adopting a Linux-style paging mechanism, as shown in Fig.~\ref{fig:pagedattentnion}. It partitions the $\mathbf{KV}$ cache into smaller pages and uses a page table to map logical blocks to physical blocks. Newly needed blocks are allocated on demand, and memory from completed requests is quickly reclaimed for new requests. Requests with identical prompts share the same $\mathbf{KV}$ cache block, further saving memory. Various inference engines, such as DeepSpeed-FastGen~\cite{holmes2024deepspeed} and vLLM~\cite{kwon2023efficient}, MAX~\cite{max}, SGLang~\cite{zheng2024sglang} integrate PagedAttention to improve inference efficiency.

LightLLM~\cite{lightllm} introduces TokenAttention, which manages $\mathbf{KV}$ cache at the token level (Fig.~\ref{fig:tokenattentnion}). Instead of a page table, TokenAttention uses a token table to track each token's actual storage location. It preallocates $\mathbf{KV}$ cache based on a user-defined maximum token limit and assigns continuous memory regions to new requests. This strategy minimizes fragmentation and improves resource utilization through fine-grained memory management.

Other approaches, such as ChunkedAttention~\cite{ye2024chunkattention}, reduce $\mathbf{KV}$ cache duplication by recognizing that system prompts often repeat. ChunkedAttention splits the \textit{KV} cache into smaller chunks and organizes them with a Prefix-Aware $\mathbf{KV}$ Cache (PAKV) structure, enabling requests with the same prefix to share cache blocks. This further enhances memory efficiency.

\subsubsection{I/O Optimization: FlashAttention} \label{sec:inference_optimization_attention_flash}

\begin{figure}[tbp]
    \centering
    \resizebox{.85\linewidth}{!}{%
    \begin{minipage}[t]{0.43\linewidth}
        \centering
        \includegraphics[width=0.99\linewidth]{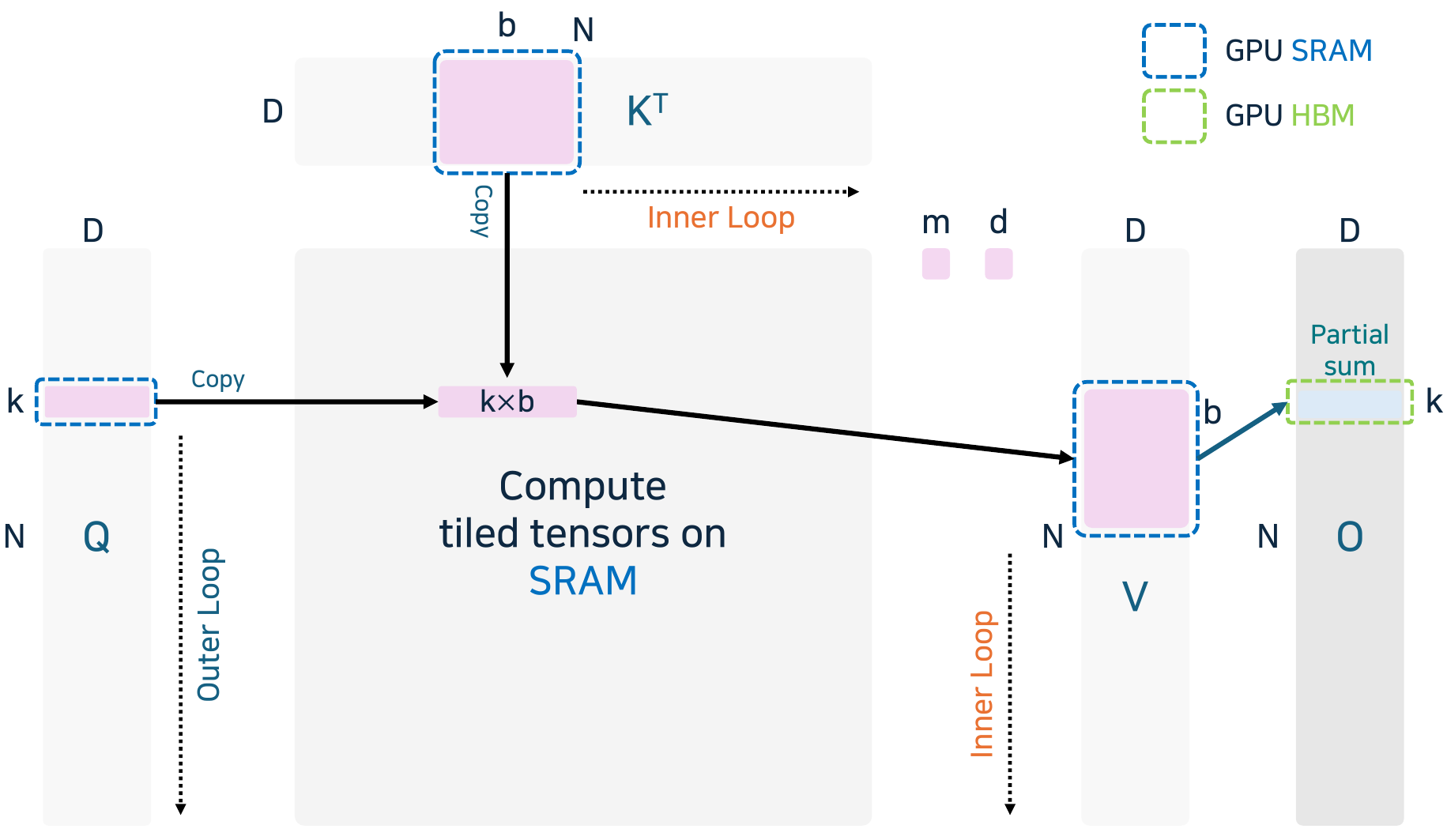}
        \captionof{figure}{FlashAttention}
        \label{fig:flashattention}
    \end{minipage}
    \hspace{-6mm}
    \begin{minipage}[t]{0.60\linewidth}
        \vspace{-105pt}
        \hspace{30pt} 
        \centering
        \resizebox{.83\linewidth}{!}{%
        \begin{minipage}{\linewidth}
        \begin{algorithm}[H]
            \caption{Simple FlashAttention}
            \begin{algorithmic}
                \For{$k \gets 1$ to $N$}
                    \For{$i \gets 1$ to $\#\text{tiles}$}
                        \State $x_i \gets Q[k, :] K^\top[:,\, (i - 1)b : ib]$  \Comment{\footnotesize{Copy to SRAM}}
                        \State $m_i^{(\text{partial})} \gets \max_{j = 1}^{b} \big(x_i[j]\big)$ \Comment{\footnotesize {Compute on SRAM}}
                        \State $m_i \gets \max\big(m_{i - 1},\, m_i^{(\text{partial})}\big)$ \Comment{\footnotesize {Compute on SRAM}}
                        \State $d_i' \gets d_{i - 1}'  e^{m_{i - 1} - m_i}
                            + \sum_{j = 1}^{b} e^{x_i[j] - m_i}$ \Comment{\footnotesize {Compute on SRAM}}
                        \State $o_i' \gets o_{i - 1}' \frac{d_{i - 1}'}{d_i'} e^{m_{i - 1} - m_i}
                            + \sum_{j = 1}^{b} \frac{e^{x_i[j] - m_i}}{d_i'} V[j + (i - 1)b,\,:]$
                        \State \hfill\parbox{0.8\linewidth}{\raggedleft\Comment{\footnotesize Compute on SRAM}}
                    \EndFor
                    \State $O[k, :] \gets o_i'$ \Comment{\footnotesize {Copy to HBM}}
                \EndFor
            \end{algorithmic}
            \label{algo:flash}
        \end{algorithm}
        \end{minipage}
        }
    \end{minipage}
    }
\end{figure}

During LLM inference, attention requires $\mathcal{O}(n^2)$ computations for a sequence of length $n$. This involves forming a score matrix through the dot product of $\mathbf{Q}$, $\mathbf{K}$, and $\mathbf{V}$, which is highly memory-intensive due to frequent data transfers between memory hierarchies on GPUs.

FlashAttention~\cite{dao2022flashattention} reduces unnecessary data transfers by splitting $\mathbf{Q}$, $\mathbf{K}$, and $\mathbf{V}$ into smaller blocks. Unlike approaches that compute the entire attention matrix before applying softmax, FlashAttention applies an online softmax per tile, avoiding redundant writes to memory. It also fuses matrix multiplication and softmax into a single pass, thereby decreasing kernel invocation overhead. 

In order to illustrate the core concept of attention operation fusion in FlashAttention, we present the essential idea in a simplified manner, as shown in Algorithm~\ref{algo:flash} and  Fig.~\ref{fig:flashattention}. In fused attention, only a small subset of intermediate results is maintained on-chip at each step, which enables memory-efficient self-attention. This design scales linearly with the sequence length $N$ while respecting GPU shared memory limits. 

We define $b$ as the block size (also referred to as the tile width). The total number of tiles along the sequence dimension is given by $\#\text{tiles} = \Bigl\lceil \frac{N}{b} \Bigr\rceil$. The term $x_i \in \mathbb{R}^b$ denotes the pre-softmax logits for tile $i$. We let $m_i$ be the global maximum value over all tiles from $1$ to $i$, and $m_i^{\text{(partial)}}$ represent the maximum value within the partial tile $i$. The variable $d_i$ is the cumulative denominator for the softmax computation up to tile $i$. 
Finally, $o_i$ is the partial output vector (corresponding to $O[k, :]$) that accumulates results up to tile $i$. Each iteration computes the local logits $x_i$ for the current tile and updates the softmax scaling factor $m_i$, cumulative denominator $d_i$, and partial output $o_i$.

Because $x_i$, $m_i$, $d_i$, and $o_i$ have small and fixed sizes ($\mathcal{O}(b)$ or $\mathcal{O}(D)$), they can reside in shared memory during kernel execution. This formulation ensures that the computation is both numerically stable and compatible with parallel tiling, making it ideal for long-sequence Transformer inference or training.

FlashAttention-2~\cite{dao2023flashattention} further optimizes GEMM and related non-matrix operations. It merges certain scaling steps and allows parallelization along the sequence length dimension, which is beneficial when the batch size and number of attention heads are small. This strategy maximizes GPU utilization for long sequences, but initially suffered from low GPU usage in GEMM on NVIDIA H100 GPUs.

To address that, FlashAttention-3~\cite{shah2024flashattention} introduces asynchronous computation and low-precision arithmetic. By splitting data transfer and computation into separate GPU warps and using a producer-consumer model, it overlaps softmax operations with Warp Group Matrix Multiply-Accumulate (WGMMA), reducing latency.

Many LLM inference engines now include support for FlashAttention variants, with vLLM~\cite{kwon2023efficient} compatible up to FlashAttention-3~\cite{shah2024flashattention}. Because these optimizations are closely related to NVIDIA GPU architectures, researchers are exploring ways to adapt similar principles for other hardware~\cite{brandon2023striped,hong2024flashdecoding++}, aiming to generalize attention acceleration across diverse platforms.

\subsubsection{KV Cache Reuse: RadixAttention} \label{sec:inference_optimization_attention_radix}

RadixAttention is an optimization technique proposed by SGLang~\cite{zheng2024sglang} that enables the automatic reuse of the KV cache across multiple operations. Traditional inference engines flush all related KV caches once a request finishes, which prevents reuse between requests and slows performance. To address this limitation, SGLang manages KV caches with a radix tree-based LRU mechanism that allows fast matching, insertion, and deletion, and it applies cache-aware scheduling to handle diverse reuse patterns efficiently. The approach is compatible with continuous batching~\cite{yu2022orca}, PagedAttention~\cite{kwon2023efficient}, and tensor parallelism, and it adds only minimal time and memory overhead, even for cache misses.

RadixAttention maps token sequences to their corresponding KV cache tensors through a radix tree, while the cache itself is stored in a non-contiguous, page-based memory layout. The tree resides in CPU memory and incurs little maintenance cost. When continuous batching is enabled, the nodes referenced by active batches cannot be deleted. Each node maintains a reference counter that tracks how many active requests use it, and nodes are removed only when this counter reaches zero.

RadixAttention also applies in multi-GPU environments. Under tensor parallelism, each GPU keeps its own shared KV cache and sub-tree without inter-GPU synchronization. The SGLang Router builds and manages a meta-tree by combining all GPU sub-trees. When a new batch arrives, the Router performs prefix matching on the meta-tree and selects dispatch policies based on request affinity. After processing, the router and the workers update their local trees independently, and the Router updates the meta-tree during periods of low system load to maintain consistency.

\subsubsection{Attention Programming Model: FlexAttention} \label{sec:inference_optimization_attention_flex}

Numerous attention optimizations have been proposed to accelerate attention operations, but most rely on manually writing hardware-specific kernels, complicating implementation and testing. To improve applicability and flexibility, FlexAttention~\cite{dong2024flex} was introduced.

FlexAttention is a general-purpose, flexible programming model that allows developers to implement diverse attention optimizations with only minimal additional code in PyTorch~\cite{paszke2019pytorch}. Observing that most attention variants can be expressed by modifying the intermediate score matrix before the softmax stage, FlexAttention accepts two callable functions---\texttt{\small {score\_mod}} and \texttt{\small {mask\_mod}}---together with the tensor input. During compilation, PyTorch automatically converts these functions into template-based handwritten attention kernels, and both forward and backward graphs are generated through the PyTorch autograd machinery. Operator fusion occurs automatically in this process, producing optimized kernel code without any low-level kernel development.

FlexAttention also supports sparsity optimization through BlockMask, which records block-level sparsity in the mask matrix to reduce computational load and memory usage while enabling flexible composition of multiple attention variants. The framework allows for independent or combined implementation of techniques such as Relative Positional Embedding~\cite{shaw2018self}, ALiBi~\cite{press2021train}, and FlashAttention~\cite{dao2022flashattention}.

FlexAttention is supported in PyTorch 2.5 and later, and Unsloth~\cite{unsloth} provides application of various kernels based on FlexAttention.

\subsubsection{MQA Optimization: FireAttention} \label{sec:inference_optimization_attention_fire}

FireAttention is an FP16- and FP8-based optimization technique developed by Fireworks AI~\cite{fireworks} to improve the performance of MoE models. Implemented as a custom CUDA kernel for MQA, it efficiently leverages memory bandwidth across a wide range of batch sizes and sequence lengths on NVIDIA H100 GPUs. Designed for multi-GPU environments, FireAttention achieves higher requests per second (RPS) and lower token-generation latency than traditional LLM inference engines.

FireAttention V2 adds FP16 and FP8 prefill-kernel support for the NVIDIA Hopper architecture and introduces multi-host deployment optimizations. In long-context, online inference workloads, it delivers up to an 8$\times$ increase in throughput and up to a 12$\times$ reduction in latency relative to existing engines.

FireAttention V3 further refines key matrix-multiplication and attention operations by providing dedicated kernels that extend support beyond NVIDIA GPUs to AMD MI300 hardware.

\subsection{Sampling Optimization} \label{sec:inference_optimization_sampling}

Because LLMs generate text autoregressively, longer input sequences increase the amount of computation and prolong user wait times. Moreover, memory I/O latency rather than raw computation often becomes the main bottleneck, affecting performance in interactive AI systems or real-time translation scenarios, where quick responses are critical.

Speculative decoding~\cite{leviathan2023fast, xia2024unlocking} accelerates token generation by leveraging an optimization concept inspired by speculative execution in computer processors. In speculative execution, operations are performed in parallel before it is confirmed whether they are needed---much like branch prediction.

\begin{figure}[tbp]
    \centering
    \includegraphics[width=.8\linewidth]{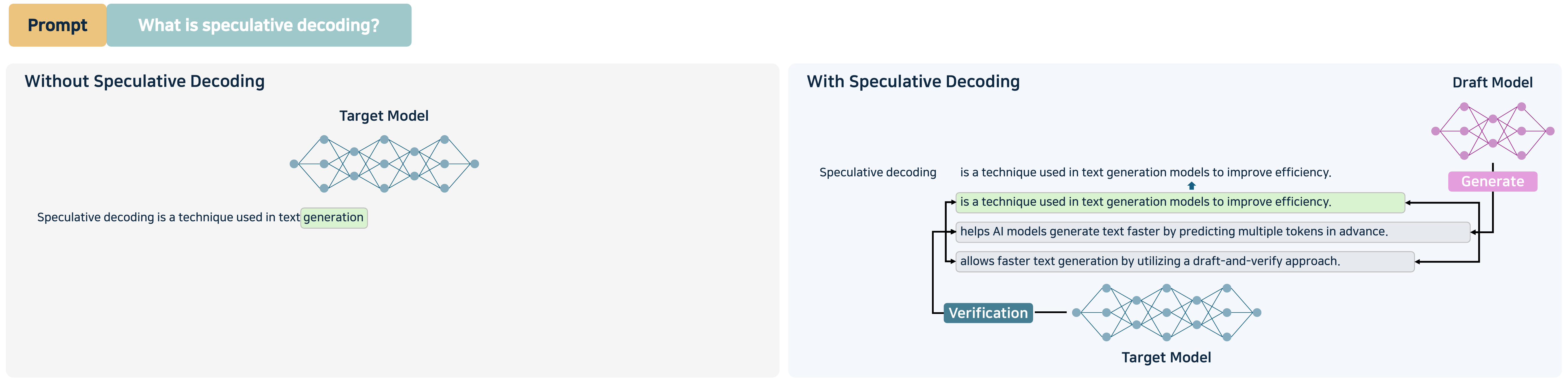}
    \caption{Speculative Decoding}
    \label{fig:speculative_decoding}
\end{figure}

Fig.~\ref{fig:speculative_decoding} illustrates this concept in LLMs through two models: a high-accuracy target model (the original LLM) and a lighter and faster draft model. The draft model generates candidate tokens that the target model later validates. This mechanism, known as speculative sampling, allows the target model to check up to \(K\) tokens at once and either accept or reject them. If certain tokens are rejected, additional tokens are sampled from an adjusted probability distribution. Parallel validation of these candidate tokens constitutes speculative decoding, speeding up generation without altering the architecture of the original model.

\textbf{Speculative Decoding Model.}
Several inference engines can employ multiple draft models for speculative decoding. Although lightweight LLMs are commonly used as drafts, some systems adopt models built on the Extrapolation Algorithm for Greater Language-model Efficiency (EAGLE)~\cite{li2024eagle}. EAGLE is a speculative sampling framework that reduces feature-level uncertainty by using a one-step-ahead token sequence and performs accurate feature prediction with minimal overhead. It achieves speculative sampling through a tree-structured draft that employs tree attention, and it is lightweight enough for real-world deployment by adding only one transformer-decoder layer as a plug-in module.

Because the acceptance rate of draft tokens depends on both position and context, EAGLE-2~\cite{li2024eagle2} estimates this rate with a confidence score from the draft model and dynamically adjusts the draft-tree structure to increase the number of accepted tokens.

Following recent test-time scaling trends, EAGLE-3~\cite{li2025eagle} was introduced to overcome the feature-prediction constraints in the original EAGLE~\cite{li2024eagle}, which limited token-prediction flexibility and reduced the benefits of data augmentation. EAGLE-3 removes these constraints and directly predicts tokens. By simulating multi-stage generation during training, it maximizes input flexibility for draft models and achieves higher speed-up ratios as training data scales.

\textbf{Optimization for Speculative Decoding.}
Research continues to refine speculative execution. Tree-based Speculative Inference~\cite{miao2024specinfer} simultaneously generates multiple candidate sequences, increasing the likelihood that the target model will approve tokens. To address difficulties in adapting draft models to changing input distributions, online speculative decoding~\cite{liu2024online} gradually aligns a draft model with the target model using knowledge distillation. This allows training and deployment of different draft models for various input patterns, and routes queries to the most suitable draft model, increasing token acceptance rates. One study proposes accelerating draft model inference by applying MXFP4 weight-only quantization~\cite{georganas2025ml}. This research introduces Multi-Level Speculative Decoding with Quantized Draft Models (ML-SpecQD), a method that combines quantization with staged speculative decoding by delegating token generation in the draft model to an even smaller, quantized draft model. Google also applied speculative decoding to its AI search function, which improved the inference speed more than two times. 

\news{Most existing speculative decoding methods have been trained and evaluated primarily on short contexts, which can lead to a significant performance drop when applied to inputs consisting of thousands to tens of thousands of tokens. To address this limitation, LongSpec~\cite{yanglongspec} has been proposed. LongSpec is designed to mitigate several bottlenecks encountered by conventional speculative decoding in long-context scenarios. First, it introduces a lightweight draft model that fixes the $\mathbf{KV}$ cache size to a constant upper bound, thereby preventing excessive memory usage for long contexts. Next, it adopts a hybrid strategy in which the prefix region is computed rapidly in a single pass and stored in the cache, while the subsequent regions are processed using standard tree attention. This approach reduces the computational cost of tree attention in long token sequences. By combining these two techniques, LongSpec enhances both memory efficiency and decoding speed, enabling stable speculative decoding even for tasks that require extended contexts.}

\textbf{Support in Inference Engines.}
Several LLM inference frameworks already incorporate speculative decoding. Ollama~\cite{ollama}, PowerInfer~\cite{song2024powerinfer}, MAX~\cite{max} and llama.cpp~\cite{llamacpp} each expose this capability directly. vLLM~\cite{kwon2023efficient} performs offline speculative decoding, generating up to five tokens at a time and-including an n-gram-based suggestion module-splits the input string into n-grams to compute similarity scores. MLC LLM~\cite{mlcllm} supports speculative decoding with lightweight draft models, as well as Medusa~\cite{cai2024medusa} and EAGLE~\cite{li2024eagle}-family drafts, and TGI~\cite{tgi} also offers Medusa~\cite{cai2024medusa} and n-gram methods. SGLang~\cite{zheng2024sglang} implements speculative decoding with EAGLE~\cite{li2024eagle}, EAGLE-2~\cite{li2024eagle2}, and EAGLE-3~\cite{li2025eagle} and integrates smoothly with Radix Cache and chunked-prefill. TensorRT-LLM~\cite{tensorrtllm} accepts draft models such as EAGLE~\cite{li2024eagle}, EAGLE-2~\cite{li2024eagle2}, and Medusa~\cite{cai2024medusa} and additionally supports Recurrent Drafter (ReDrafter)~\cite{cheng2024recurrent}, which recursively predicts drafts, and look-ahead decoding, which performs parallel n-gram prediction and verification. Commercial engines, including Friendli Inference~\cite{friendli} and Fireworks AI~\cite{fireworks}, also leverage speculative decoding to accelerate inference.

\subsection{Structured Outputs} \label{sec:inference_optimization_structured_outputs}

In autoregressive LLMs, the generated tokens function as both the model's inputs and its outputs. During tokenization, however, meaningful units can be divided or Unicode characters fragmented. This limitation becomes especially problematic in applications that require problem solving or planning, such as reasoning tasks or AI agents, where structured outputs like JSON, function calls, or code blocks are essential.

\begin{figure}[tbp]
    \centering
    \resizebox{0.95\textwidth}{!}{%
    \begin{minipage}[t]{0.52\textwidth}
        \centering
        \includegraphics[width=\linewidth]{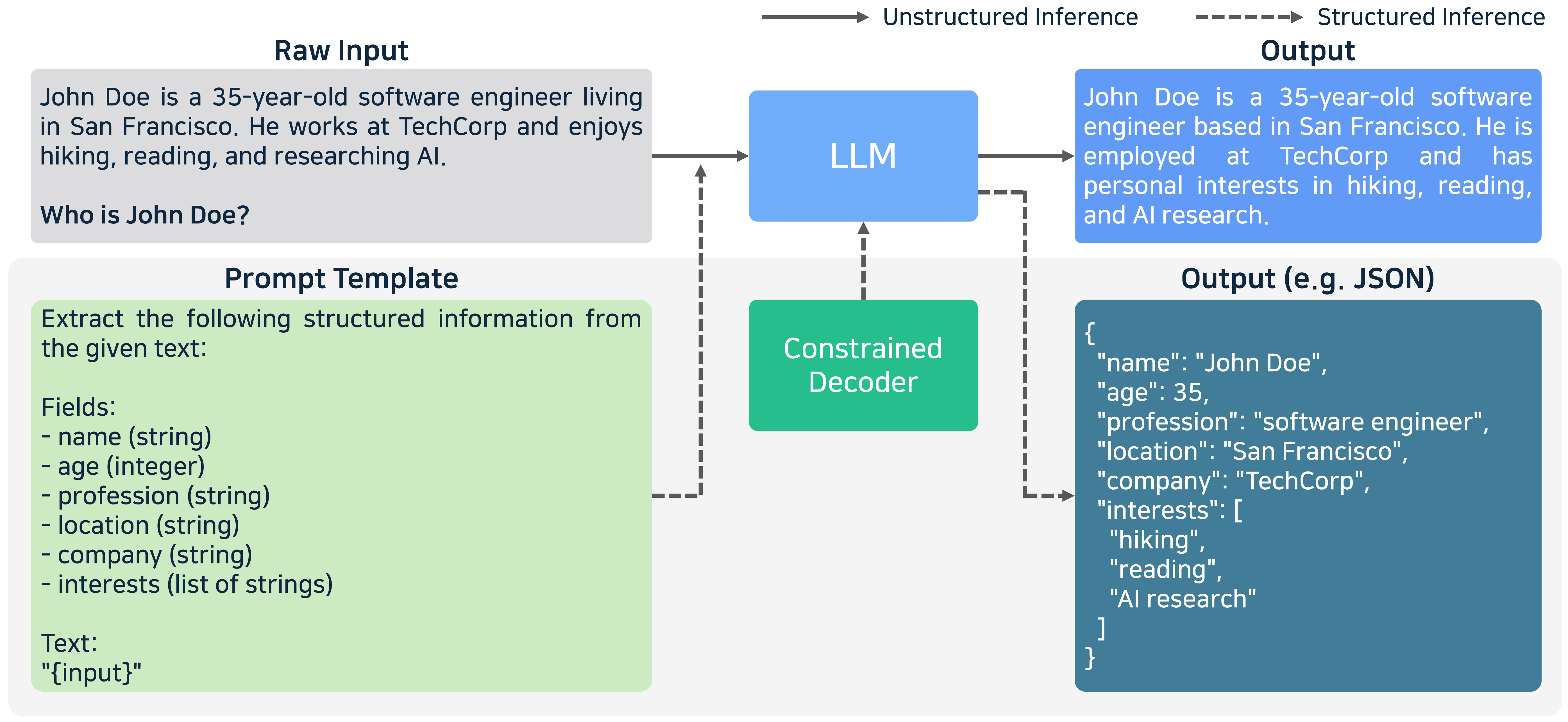}
        \captionof{figure}{Unstructured Outputs and Structured Outputs}
        \label{fig:structured_output}
    \end{minipage}
    \hspace{0.03\textwidth}
    \begin{minipage}[t]{0.42\textwidth}
        \centering
        \includegraphics[width=\linewidth]{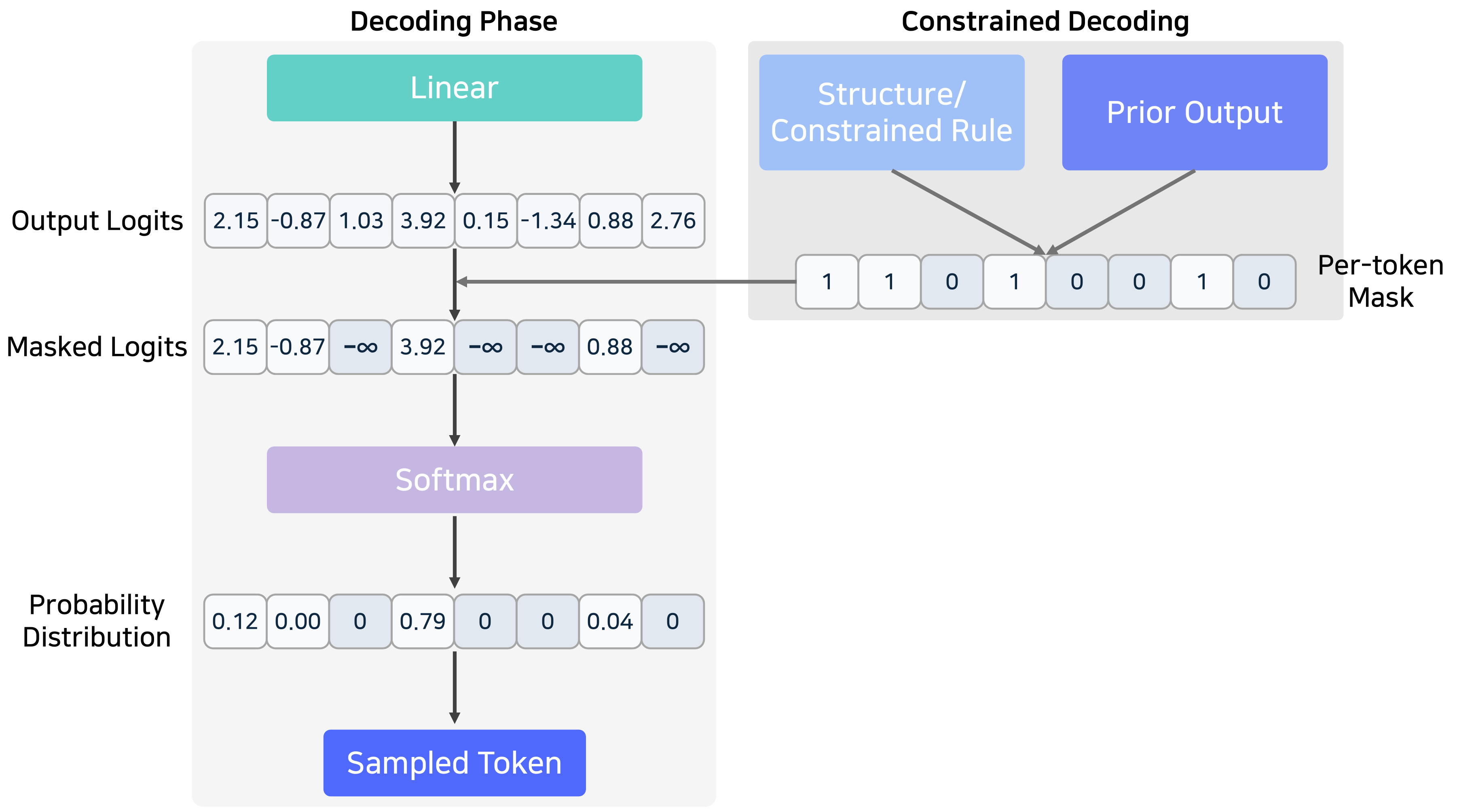}
        \captionof{figure}{Constrained Decoding in Decoding Phase}
        \label{fig:constrained_decoding}
    \end{minipage}
    }
\end{figure}

As shown in Fig.~\ref{fig:structured_output}, structured output~\cite{liu2024we, irugalbandara2024meaning} refers to generating text that follows a predefined format---JSON, XML, or another structured schema. Unlike typical free-form LLM output, structured generation ensures that the produced content conforms to constraints expected by downstream systems. For example, if a database entry or an API call requires JSON, structured output enables the LLM to return a valid JSON.

Modern LLMs have evolved beyond basic text generation to support tasks such as code creation, function invocation, and autonomous decision making. To enable these applications, inference engines must provide machine-readable structured outputs that integrate smoothly with other systems. Structured generation produces output that complies with explicit schemas---such as JSON or SQL---thus improving correctness and consistency, simplifying interpretation and integration, and reducing hallucinations by eliminating unnecessary or invalid information. Consequently, domain-specific output requirements are often satisfied without additional fine-tuning~\cite{wang2023grammar}.

Constrained decoding~\cite{wang2023grammar, willard2023efficient} is commonly used to generate structured outputs. As Fig.~\ref{fig:constrained_decoding} illustrates, the entire vocabulary is evaluated at each decoding phase and tokens that violate the output schema are masked out. After the logits are computed, a token-level mask invalidates any token that would break the structure; those logits are then set to zero before the softmax operation. This masking procedure directly influences both the performance and the speed of structured generation.

In structured generation with a finite-state machine (FSM)~\cite{willard2023efficient}, the LLM generates tokens sequentially while moving through the states defined by the machine. The FSM records the current state and assigns a probability of zero to any token that violates the required format, thereby filtering it out. Structures such as JSON schemas can be modeled as directed-graph FSMs, where each node represents a valid partial sequence and each edge represents an allowed next token. FSM decoding is fast, but it handles only simple patterns and cannot easily express recursive structures such as nested JSON.

Using a context-free grammar (CFG)~\cite{geng2023grammar, barke2024hysynth} supports more complex structured generation. A CFG defines language structure through production rules and can capture formats that an FSM cannot. Runtime guidance with a CFG, however, requires recursively checking these rules against the entire vocabulary and maintaining multiple parser stacks, which adds significant computational overhead.

\textbf{Library/Frameworks for Structured Outputs.}
The Outlines~\cite{outlines} library enables guided generation through a FSM based regular expression parser that can be started or stopped at any point in the decoding process. To efficiently identify valid tokens, it builds an index that lets each step run in amortized $\mathcal{O}(1)$ time. The library expands sequences with multinomial sampling until an EOS token appears, then applies a Boolean mask to produce an unnormalized conditional distribution; the same mask is reused in subsequent steps. In addition to FSM guidance, Outlines supports multiple-choice prompts and structured generation from CFGs written in Extended Backus-Naur Form (EBNF).

XGrammar~\cite{dong2024xgrammar} is a structured generation library designed for efficiency, flexibility, and portability. It supports CFGs for complex formats and includes system-level optimizations for high-speed execution. A C++ backend simplifies integration into diverse runtimes. To reduce recursion overhead, XGrammar converts each CFG into a byte-level pushdown automaton (PDA). The PDA separates context-independent tokens---validated by position alone---from context-dependent tokens that need entire stack inspection. Context-independent tokens are prevalidated and cached for reuse, while context-dependent tokens are dynamically checked by the PDA. XGrammar also maintains a persistent execution stack for rapid branching and rollback and expands context windows to decrease context-dependent tokens, achieving up to 100$\times$ lower per-token latency.

The LM Format Enforcer~\cite{lm-format-enforcer} guarantees output formatting through token-level filtering while preserving the expressive freedom of the model. It works independently of the base model and tokenizer, supports batch decoding and beam search, and enforces formats such as JSON Schema, JSON, or regular expressions. Internally, it merges a \texttt{\small {Character-Level Parser}}, which used to read character sets, with a \texttt{\small {Tokenizer Prefix Tree}} that stores every token the tokenizer can be generated. A token is accepted only if both structures allow it. After each token is generated, the parser and tree advance together, preparing the constraints for the next step. This approach maintains strict adherence to the target format, while allowing the model to manage details such as spacing and word order.

Low-level Guidance (llguidance)~\cite{llguidance} is a library for fast structured output generation using CFGs. It accepts a CFG, a tokenizer, and a token prefix, then computes the set of valid next tokens (a token mask) that can follow the prefix while still yielding strings valid under the grammar. The library is highly efficient, taking only about 50 \textmu s of CPU overhead per token for a 128 \(k\)-entry tokenizer. This speed comes from a CFG parser based on the Earley algorithm on top of regular expression derivatives~\cite{lexer} and a token prefix trie for mask computation. Supported grammar formats include the JSON Schema, regular expressions, CFGs derived from Lark~\cite{lark}, and the llguidance's own grammar syntax.

GGML Backus-Naur Form (GBNF)~\cite{gbnf} is a formal grammar format introduced in llama.cpp~\cite{llamacpp} to constrain the output. It merges traditional BNF notation with modern regex features, specifying grammars as production rules that link non-terminal and terminal symbols. GBNF is compatible with JSON Schema and can be used via CLI or API, allowing straightforward control of structured generation.

OpenAI Structured Outputs~\cite{openai_structure_outputs} enables structured generation through function calling and response-format specifications. Function calling permits a model to invoke external system functions safely, while response-format specification ensures that the output matches the desired JSON schema. Both features provide type safety, eliminating post-processing or retries due to format errors, and maintaining consistent formatting without extensive prompt engineering. Built-in safety policies allow the system to reject unsafe or inappropriate requests.

\textbf{Performance Evaluation of Structured Output.} 
Benchmarks such as The Beyond the Imitation Game Benchmark (BIG-Bench)~\cite{srivastava2022beyond} and Massive Multitask Language Understanding (MMLU)-Pro~\cite{wang2024mmlu} were used to evaluate multilingual and multitask performance of the LLM. As the quality of structured output has become critical for real-world LLM deployment, newer benchmarks target this capability more directly. JSONSchemaBench~\cite{geng2025generating} provides 10,000 real-world JSON schemas of varying complexity and constraints, enabling the evaluation of constrained-decoding techniques on metrics such as schema adherence, output efficiency, and generalization. Using the official JSON Schema test suite, JSONSchemaBench evaluates llguidance~\cite{llguidance}, GBNF~\cite{gbnf}, Outlines~\cite{outlines}, XGrammar~\cite{dong2024xgrammar}, OpenAI~\cite{openai_structure_outputs}, and Gemini~\cite{gemini}, providing detailed functional and accuracy analyses.

StructTest~\cite{chen2024structtest} is a rule-based evaluator that measures the ability of LLM to follow complex instructions while producing structured output. Emphasizing cost efficiency, ease of use, bias reduction, and robustness to data contamination, StructTest reports both accuracy and consistency and indirectly gauges instruction decomposition and reasoning capabilities. It has been applied to models such as GPT-3.5 and GPT-4~\cite{achiam2023gpt}, the Claude 3 family~\cite{anthropic2024claude}, and DeepSeek-v3~\cite{liu2024deepseek-v3} for tasks including summarization, code generation, HTML creation, and mathematical reasoning.

To further study structured generation, SoEval~\cite{liu2024llms} offers a structured output benchmark dataset covering 13 output types---such as JSON and XML---across more than 20 domains, including science, technology, literature, and healthcare.

\textbf{Support in Inference Engines.}
From the inference engine perspective, vLLM~\cite{kwon2023efficient} supports guided decoding with Outlines~\cite{outlines}, LM Format Enforcer~\cite{lm-format-enforcer} and XGrammar~\cite{dong2024xgrammar} in both online such as OpenAI Completions and Chat APIs~\cite{openai_structure_outputs} and offline modes. SGLang~\cite{zheng2024sglang} integrates Outlines~\cite{outlines}, XGrammar~\cite{dong2024xgrammar}, and llguidance~\cite{llguidance}, while LightLLM~\cite{lightllm} supports Outlines~\cite{outlines} and XGrammar~\cite{dong2024xgrammar}. MAX~\cite{max}, MLC LLM~\cite{mlcllm}, and TensorRT-LLM~\cite{tensorrtllm} adopt XGrammar~\cite{dong2024xgrammar}, whereas Ollama~\cite{ollama}, llama.cpp~\cite{llamacpp}, and PowerInfer~\cite{song2024powerinfer} rely on GBNF~\cite{gbnf} for format enforcement. LMDeploy~\cite{2023lmdeploy} provides structured output through its PyTorch~\cite{paszke2019pytorch} interface, and major commercial engines offer equivalent capabilities via proprietary solutions or OpenAI-compatible APIs~\cite{openai_structure_outputs}.


\section{\news{Empirical Evaluation}} \label{sec:experiments}

\news{In this section, we evaluated the performance of the inference engines introduced in Section~\ref{sec:detailed_review}. Table~\ref{tab:metrics} summarizes the four key metrics used in our experiments (TTFT, TBT, latency, and throughput) which are critical in real-world service environments and served as the basis for our comparison. The analysis covered 21 open-source inference engines and examined how the optimization techniques discussed in Section~\ref{sec:inference_optimization} influenced their practical performance.}

\news{All inference engines were installed and benchmarked under the environments described in Table~\ref{tab:env_specs}. To provide a comprehensive efficiency analysis, experiments were conducted separately on server-class hardware and edge devices.}


\begin{table}[tbp]
    \caption{\news{Server Specifications for Experimental Environment}}
    \label{tab:env_specs}
    \centering
    \resizebox{.80\textwidth}{!}{%
    \begin{tabular}{@{}>{\centering\arraybackslash}p{0.10\textwidth}
                        >{\centering\arraybackslash}p{0.14\textwidth}
                        >{\centering\arraybackslash}p{0.32\textwidth}
                        >{\centering\arraybackslash}p{0.32\textwidth}
                        >{\centering\arraybackslash}p{0.32\textwidth}@{}}
    \toprule
    \multicolumn{2}{c}{} & \multicolumn{2}{c}{\news{Server}} & \multirow{2.5}{*}{\news{Edge}} \\
    \cmidrule(lr){3-4}
    \multicolumn{2}{c}{\raisebox{1.8ex}[0pt]{\news{Category}}} & \news{Server 1 (high-spec)} & \news{Server 2 (mid-spec)} & \\
    \midrule
    \multicolumn{2}{c}{\news{CPU}}    
        & \news{Intel Xeon Platinum 8480+ $\times$ 2}   
        & \news{Intel Xeon Gold 6426Y $\times$ 2} 
        & \news{8-core Arm Cortex A78AE} \\
        
    \multicolumn{2}{c}{\news{Memory}} 
        & \news{2.0 TiB (DDR5)}                          
        & \news{1.5 TiB (DDR5)}                    
        & \news{32 GiB (Unified LPDDR5)} \\
        
    \multicolumn{2}{c}{\news{OS}}     
        & \news{Ubuntu 22.04 LTS}                     
        & \news{Ubuntu 22.04 LTS}                
        & \news{Ubuntu 22.04 LTS (JetPack 6.0)} \\
        
    \multirow{3}{*}{\news{GPU}} 
        & \news{Name}            
            & \news{NVIDIA H100 $\times$ 8}     
            & \news{NVIDIA RTX A6000 $\times$ 6}      
            & \news{NVIDIA Ampere c GPU} \\
            
        & \news{Memory}          
            & \news{640 GB (80 GB per GPU)}     
            & \news{288 GB (48 GB per GPU)}           
            & \news{32 GB (Unified LPDDR5)} \\
            
        & \news{Interconnection} 
            & \news{SXM}                        
            & \news{NVLink}                            
            & \news{-} \\
    \bottomrule
    \end{tabular}%
    }
\end{table}


\begin{table}[tbp]
    \caption{\news{Inference Engines Execution Environments}}
    \label{tab:engine_deploy}
    \centering
    \resizebox{.9\textwidth}{!}{%
    \begin{tabular}{@{}>{\centering\arraybackslash}p{0.27\textwidth} 
                            >{\centering\arraybackslash}p{0.21\textwidth} 
                            >{\centering\arraybackslash}p{0.23\textwidth} 
                            >{\centering\arraybackslash}p{0.20\textwidth} 
                            >{\centering\arraybackslash}p{0.18\textwidth} 
                            >{\centering\arraybackslash}p{0.13\textwidth} 
                            >{\centering\arraybackslash}p{0.25\textwidth}@{}}
    \toprule
    \multicolumn{2}{c}{\news{Inference Engine}} & 
    \multicolumn{3}{c}{\news{Installation}} & 
    \multicolumn{2}{c}{\news{API}} \\ 
    \cmidrule(lr){1-2} \cmidrule(lr){3-5} \cmidrule(lr){6-7}
    \news{Name} & \news{Version} & \news{Install Method} & \news{Environment} & \news{Install Ease} & \news{Support} & \news{Type} \\ 
    \midrule
        \news{Ollama}              & \news{0.11.10}           & \news{curl}                      & \news{venv}   & \news{\textcolor{Green}{Easy}}       & \news{\greencheck} & \news{OpenAI-compatible} \\
        \news{LLaMA.cpp}           & \news{b6423 (7057faf6$^\ast$)}  & \news{source build}              & \news{-}    & \news{\textcolor{Green}{Easy}}         & \news{\greencheck} & \news{OpenAI-compatible} \\
        \news{vLLM}                & \news{0.10.1.1}          & \news{pip/uv}                       & \news{venv}     & \news{\textcolor{Green}{Easy}}     & \news{\greencheck} & \news{OpenAI-compatible} \\
        \news{DeepSpeed-FastGen (DeepSpeed-MII)}   & \news{0.3.3}             & \news{pip}                       & \news{venv}    & \news{\textcolor{Green}{Easy}}      & \news{\greencheck} & \news{RESTful API (OpenAI-compatible)}  \\
        \news{unsloth}             & \news{2025.9.2}          & \news{pip}                       & \news{venv}     & \news{\textcolor{Green}{Easy}}     & \news{\redxmark}   & \news{\redxmark}   \\
        \news{MAX}                 & \news{25.4.0}            & \news{docker}                       & \news{venv}     & \news{\textcolor{Green}{Easy}}     & \news{\greencheck} & \news{OpenAI-compatible} \\
        \news{MLC LLM}             & \news{nightly-cu12}      & \news{pip}                       & \news{conda}    & \news{\textcolor{Orange}{Medium}}     & \news{\greencheck} & \news{RESTful API (OpenAI-compatible)} \\
        \news{llama2.c}            & \news{350e04f$^\ast$}    & \news{source build}              & \news{-}        & \news{\textcolor{Green}{Easy}}        & \news{\redxmark}   & \news{\redxmark}   \\
        \news{bitnet.cpp}          & \news{404980e$^\ast$}    & \news{source build}              & \news{conda}    & \news{\textcolor{Green}{Easy}}        & \news{\redxmark}   & \news{\redxmark}   \\
        \news{SGLang}              & \news{0.5.2rc2}          & \news{uv}                       & \news{venv}     & \news{\textcolor{Green}{Easy}}        & \news{\greencheck} & \news{OpenAI-compatible} \\
        \news{LitGPT}              & \news{0.5.10}            & \news{pip}                       & \news{venv}     & \news{\textcolor{Green}{Easy}}        & \news{\greencheck} & \news{Python API (OpenAI-compatible)}    \\
        \news{OpenLLM}             & \news{0.6.30}            & \news{pip}                       & \news{venv}     & \news{\textcolor{Green}{Easy}}        & \news{\greencheck} & \news{OpenAI-compatible} \\
        \news{TensorRT-LLM}        & \news{0.21.0}            & \news{docker}                  & \news{docker}   & \news{\textcolor{Orange}{Medium}}      & \news{\greencheck} & \news{OpenAI-compatible} \\
        \news{TGI}                 & \news{3.3.5}             & \news{docker}                  & \news{docker}    & \news{\textcolor{Orange}{Medium}}     & \news{\greencheck} & \news{OpenAI-compatible} \\
        \news{PowerInfer}          & \news{d3ebd7c$^\ast$}     & \news{source build}              & \news{venv}     & \news{\textcolor{Green}{Easy}}     & \news{\redxmark}   & \news{\redxmark}   \\
        \news{LMDeploy}            & \news{0.10.1}            & \news{pip}                       & \news{conda}    & \news{\textcolor{Green}{Easy}}     & \news{\greencheck} & \news{OpenAI-compatible} \\
        \news{LightLLM}            & \news{1.1.0}             & \news{prebuilt}                  & \news{docker/conda} & \news{\textcolor{Green}{Easy}}  & \news{\redxmark}   & \news{\redxmark}   \\
        \news{NanoFlow}            & \news{8a28a8c$^\ast$}   & \news{source build}   & \news{docker}  & \news{\textcolor{Red}{Hard}}      & \news{\redxmark}   & \news{\redxmark}   \\
        \news{DistServe}           & \news{82831f1$^\ast$}   & \news{source build}   & \news{conda}    & \news{\textcolor{Red}{Hard}}      & \news{\redxmark}   & \news{\redxmark}   \\
        \news{vAttention}          & \news{ef3fff2$^\ast$}   & \news{prebuilt}              & \news{docker}    & \news{\textcolor{Green}{Easy}}    & \news{\greencheck} & \news{OpenAI-compatible} \\
        \news{Sarathi-Serve}       & \news{786d144$^\ast$}   & \news{source build}              & \news{docker}   & \news{\textcolor{Green}{Easy}}     & \news{\redxmark}   & \news{\redxmark}   \\
    \bottomrule
    \end{tabular}
    }
    \flushleft
    \hspace{7mm} \tiny{\news{$^\ast$Commit number}}
\end{table}

\subsection{\news{Environment Setup}} \label{sec:exp_environment}

\news { We installed 21 open-source LLM inference engines on both server and edge hardware environments summarized in Table~\ref{tab:env_specs}, and verified successful installation by running example workloads. Most engines (e.g., Ollama~\cite{ollama}, vLLM~\cite{kwon2023efficient}) supported installation via pip~\cite{pip} or uv~\cite{uv}, while engines such as MAX~\cite{max}, TensorRT-LLM~\cite{tensorrtllm}, and TGI~\cite{tgi} provided official container images (Docker~\cite{docker}, Podman~\cite{podman}) which facilitated deployment. Lightweight C or C++-based engines such as LLaMA.cpp~\cite{llamacpp}, llama2.c~\cite{llama2c}, and bitnet.cpp~\cite{wang20241} were built directly using make~\cite{make} or ninja~\cite{ninja} to generate executable binaries. DeepSpeed-FastGen~\cite{holmes2024deepspeed} was distributed as an internal module of DeepSpeed-MII~\cite{deepspeedmii}, which required the installation of DeepSpeed-MII.}

\news { Some engines required additional configurations. MLC LLM~\cite{mlcllm} encountered build errors with third-party libraries, which were resolved by using a conda~\cite{conda} environment. TensorRT-LLM~\cite{tensorrtllm} and TGI~\cite{tgi} exhibited instability during source builds; therefore, we used their official container environments instead. DistServe~\cite{zhong2024distserve} supported only Hugging Face model weights in binary (.bin) format, which required modifying the loader code.
NanoFlow~\cite{zhu2024nanoflow} required the commercial Gurobi Optimizer~\cite{gurobi} for execution graph exploration, which meant that in environments without an academic license, the code had to be modified to use alternative libraries such as Google OR-Tools~\cite{or-tools}. After modification, both DistServe and NanoFlow successfully built and ran example workloads, but due to lack of correctness verification, they were excluded from the experimental evaluation.}

\news { In the edge environment, Ollama~\cite{ollama} and LLaMA.cpp~\cite{llamacpp} supported native installation and were directly tested. On NVIDIA Jetson devices, unofficial containers~\cite{jetson-containers} supported only limited inference engines and versions, so these were excluded from the performance comparison. The container-based execution of vLLM~\cite{kwon2023efficient} and MLC LLM~\cite{mlcllm} on edge environment resulted in inference errors and was therefore also excluded from the analysis.}

\news { Based on these observations, Table~\ref{tab:engine_deploy} categorizes installation difficulty into three levels: \texttt{Easy}, \texttt{Medium}, and \texttt{Hard}. \texttt{Easy} refers to cases requiring no additional setup beyond dependency installation and example execution; \texttt{Medium} includes cases requiring adjustments to build options or environment configuration; \texttt{Hard} applies when extra procedures such as source code modification or replacement of proprietary libraries were necessary.}

\news{\textbf{Documentation Quality vs. Real Installation Difficulty.} In Fig.~\ref{fig:inference_engine_spider_graph}, we previously assigned \texttt{Ease-of-Deploy} scores for each inference engine based on official documentation, guides, and repositories. However, empirical results revealed discrepancies for certain engines. Unsloth~\cite{unsloth}, llama2.c~\cite{llama2c}, bitnet.cpp~\cite{wang20241}, LitGPT~\cite{litgpt}, OpenLLM~\cite{openai_structure_outputs}, PowerInfer~\cite{song2024powerinfer}, vAttention~\cite{prabhu2025vattention}, and Sarathi-Serve~\cite{agrawal2023sarathi} received relatively low documentation-based scores, but were found to be easier to install in practice, with dependency setup and example execution proceeding smoothly. In contrast, NanoFlow~\cite{zhu2024nanoflow} and DistServe~\cite{zhong2024distserve} were as difficult to install as indicated by their documentation, requiring additional library configuration or code modification. Engines such as TensorRT-LLM~\cite{tensorrtllm}, TGI~\cite{tgi}, MLC LLM~\cite{mlcllm}, and vAttention~\cite{prabhu2025vattention} which had high documentation-based scores, provided official container or development environments (e.g., conda) that effectively offset the complexity of source builds. }

\subsection{\news{Evaluation of LLM Serving and Inference Performance}} \label{sec:exp_performance}

\subsubsection{\news{Evaluation Environments}} \label{sec:exp_evaluation_environment}

\news { In this study, we investigated the compatibility of each LLM inference engine with the OpenAI API, as summarized in Table~\ref{tab:engine_deploy}, to evaluate performance. Although benchmark scripts can be implemented individually for each inference engine, differences in supported libraries and dependencies between engines may lead to inconsistencies in environment configuration and code behavior.}

\news {To ensure a fair and reproducible performance evaluation, this study adopted an OpenAI API-compatible interface-based approach. This approach minimizes implementation-dependent variability while enabling measurement of end-to-end serving performance, which reflects actual inference efficiency in real service environments.}

\news {However, API-based evaluation inherently includes overhead from serving operations, such as server-side request handling, scheduling, and communication latency, in addition to the model's computational performance. Nevertheless, this is considered a practical evaluation approach, as it represents the end-to-end responsiveness perceived by users. Given that most LLM inference engines are provided as API-based services and standalone executables (e.g., CLI), measuring performance under serving mode aligns more closely with real-world application scenarios.}

\news {Therefore, the goal of this study is not only to assess computational efficiency within the model itself but also to evaluate the overall inference service pipeline, the end-to-end serving performance.}

\news {For performance measurement, we employed GuideLLM~\cite{guidellm}, a tool capable of simulating inference workloads that closely resemble real-world service conditions. GuideLLM enables analysis of throughput, latency, resource efficiency, and scalability during model deployment. It follows the OpenAI API request-response protocol and supports various workload conditions, including single, concurrent, and asynchronous requests.}

\news {Using GuideLLM, we evaluated 13 inference engines compatible with the OpenAI API (Ollama~\cite{ollama}, LLaMA.cpp~\cite{llamacpp}, vLLM~\cite{kwon2023efficient}, DeepSpeed-FastGen (DeepSpeed-MII~\cite{deepspeedmii})~\cite{holmes2024deepspeed}, MAX \cite{max}, MLC LLM~\cite{mlcllm}, SGLang~\cite{zheng2024sglang}, LitGPT~\cite{litgpt}, OpenLLM~\cite{openllm}, TensorRT-LLM~\cite{tensorrtllm}, TGI~\cite{tgi}, LMDeploy~\cite{2023lmdeploy}, and vAttention~\cite{prabhu2025vattention}) as listed in Table~\ref{tab:engine_deploy}. All engines were tested in the server environment (Table~\ref{tab:env_specs} - \texttt{Server}), while only Ollama~\cite{ollama} and LLaMA.cpp~\cite{llamacpp} were evaluated in the \texttt{Edge} environment. To ensure comparability, models with similar parameter scales and training datasets were selected from the officially supported weights for each engine. As mentioned earlier, the default options of each inference engine were used in this experiment, and the detailed configurations are summarized in Table~\ref{tab:llm_inference_engines_serving_configuration}.}

\begin{table}[tbp]
    \caption{\news {LLM Inference Engine Serving Default Configuration Comparison}}
    \label{tab:llm_inference_engines_serving_configuration}
    \centering
    \resizebox{.95\textwidth}{!}{
    \begin{tabular}{
        @{}>{\centering\arraybackslash}p{0.18\textwidth}
        >{\centering\arraybackslash}p{0.17\textwidth}
        >{\centering\arraybackslash}p{0.30\textwidth}
        >{\centering\arraybackslash}p{0.12\textwidth}
        >{\centering\arraybackslash}p{0.16\textwidth}
        >{\centering\arraybackslash}p{0.20\textwidth}
        >{\centering\arraybackslash}p{0.18\textwidth}@{}}
    \toprule
        Inference Engine & Context Length & Attention & Max Batch & GPU & API Server & Model Repository \\
    \midrule
        Ollama
        & 4,096
        & Disable (FlashAttention)
        & --
        & 1
        & Go Lang Server
        & Ollama \\

        LLaMA.cpp
        & 4,096
        & FlashAttention (Auto)
        & 2,048
        & Visible devices
        & HTTP server
        & HuggingFace \\

        vLLM
        & Model config
        & FlashAttention, Cascade Attention (FlashInfer)
        & --
        & 1
        & vLLM APIServer
        & HuggingFace \\

        DeepSpeed
        & Model config
        & --
        & --
        & 1
        & RESTful / Uvicorn
        & HuggingFace \\

        MAX
        & Model config
        & --
        & --
        & 1
        & --
        & MAX / HuggingFace \\

        MLC LLM
        & Model config
        & --
        & 128
        & 1
        & Uvicorn
        & MLC LLM / HuggingFace \\

        SGLang
        & Model config
        & Triton / FlashAttention / FlashInfer
        & --
        & 1
        & HTTP server
        & HuggingFace \\

        litgpt
        & --
        & --
        & --
        & 1
        & Uvicorn
        & HF \\

        openllm
        & 3,192
        & FlashAttention
        & 2,048
        & 1
        & --
        & Bento \\

        TRT
        & Model config
        & --
        & 2,048
        & 1
        & Uvicorn
        & HuggingFace \\

        TGI
        & 8,192
        & FlashInfer
        & 128
        & 1
        & --
        & HuggingFace \\

        LMDeploy
        & 8,192
        & --
        & --
        & 1
        & Uvicorn
        & HuggingFace \\

        vAttention
        & 8,192
        & FlashAttention
        & 128
        & 1
        & Uvicorn
        & HuggingFace \\
    \bottomrule
    \end{tabular}
    }
\end{table}

\news {Performance testing was conducted using GuideLLM by generating workloads with different combinations of concurrent request counts (1-64) and input/output token lengths (64-4096). Each test was run for 30 seconds, with the initial 10\% of the test period used for warmup and the final 5\% for cooldown. Data from this period were excluded from the analysis. In the edge environment, performance measurement data were collected for 240 seconds considering the limited computational capacity. The measured metrics, as described in Table~\ref{tab:metrics}, include TTFT, TBT, latency, and throughput, which are critical indicators of both LLM inference and service-level performance. The input prompts consisted of 1,000 synthetic samples generated by GuideLLM, and the results were averaged across all conditions for comparison. The corresponding GuideLLM result metrics for the LLM inference performance metrics defined in Table~\ref{tab:metrics} are summarized in Table~\ref{tab:guidellm_metrics_summary}.} 

\begin{table}[tbp]
    \caption{\news{LLM Inference Performance Metrics and GuideLLM Result Metrics}}
    \label{tab:guidellm_metrics_summary}
    \centering
    \resizebox{.95\textwidth}{!}{%
    \begin{tabular}{@{}>{\centering\arraybackslash}p{0.25\textwidth}
                    >{\centering\arraybackslash}p{0.25\textwidth}
                    >{\centering\arraybackslash}p{0.18\textwidth}
                    p{0.8\textwidth}@{}}
    \toprule
    \multicolumn{1}{c}{Category} &
    \multicolumn{1}{c}{GuideLLM Metric} &
    \multicolumn{1}{c}{Unit} &
    \multicolumn{1}{c}{Description} \\
    \midrule
    TTFT & TTFT & ms &
    The time it takes to generate the first token in the output. This reflects the model's initial response latency and the user's perceived responsiveness.
    \\
    TBT / ITL & ITL & ms &
    The average time interval between consecutive output tokens, excluding the first token. This indicates the smoothness of generation and the speed of decoding.
    \\
    End-to-End Latency & Request Latency & ms &
    End-to-end latency for each request is measured from submission to completion and is a key responsiveness metric for an inference system.
    \\
    \multirow[t]{3}{*}{Throughput} 
    & Requests Rate (Requests Per Second) & req/s &
    The number of requests successfully processed per second. This indicates the overall system service capacity and scheduling efficiency.
    \\
    & Total Tokens Per Second & Tot tok/s &
    The combined rate of prompts and output tokens processed per second. This measures the overall system efficiency, including input and output token processing.
    \\
    \bottomrule
    \end{tabular}%
    }
\end{table}

\news{To measure the performance metrics defined in Table~\ref{tab:guidellm_metrics_summary}, the following experimental procedures were applied.}

\begin{itemize}
  \item \news{First, to examine how TTFT varies with prompt length, the concurrency level was fixed at 16 and the output length at 1,024 tokens, while the prompt length was gradually increased to 64, 256, 512, 1,024, 2,048, and 4,096 tokens. In the edge device environment, TTFT was measured by varying the prompt length to 65, 128, and 256 tokens, while fixing the concurrency level at 2 and the output length at 512 tokens.}
  \item \news{Next, to observe how TBT changes as the output length increases, the prompt length was fixed at 1,024 tokens and the concurrency at 16, while the output length was varied across 512, 1,024, 2,048, and 4,096 tokens. In the edge environment, TBT was evaluated by fixing the prompt length at 128 tokens and the concurrency level at 2, while varying the output length to 256, 512, and 1,024 tokens.}
  \item \news{To evaluate request-handling capability in multi-user environments, both the prompt and output lengths were fixed at 1,024 tokens, and the concurrency level was increased to 1, 4, 8, 16, 32, and 64. For each setting, we collected the number of requests processed per second and the overall success rate. For performance evaluation on the edge device, the prompt length was fixed at 128 tokens and the output length at 512 tokens, while the concurrency level was increased to 1, 2, and 4 to measure request throughput and latency in multi-user scenarios.}
  \item \news{Under the same configurations, we additionally recorded the total number of tokens processed per second to measure token throughput and measured user-perceived latency at each concurrency level.}
\end{itemize}

\news {In the performance evaluation of this study, all inference engines were executed using their default configurations. Each engine differs in serving mechanisms and internal parameters, and although performance can be improved by adjusting these options, the range of supported configurations varies across engines, making it difficult to establish uniform conditions. To avoid potential bias from arbitrary tuning, the experiments focused on maintaining consistent evaluation conditions by preserving the default settings.}

\begin{table}[tbp]
    \centering
    \caption{\news{Inference Engine Model Support Matrix}}
    \label{tab:inference_engines_support_model}
    \resizebox{\textwidth}{!}{%
    \begin{tabular}{l*{26}{c}}
    \toprule
    \multirow{4.5}{*}{\centering Model} & \multicolumn{2}{c}{\multirow{2}{*}{Ollama}} & \multicolumn{2}{c}{\multirow{2}{*}{LLaMA.cpp}} & \multicolumn{2}{c}{\multirow{2}{*}{vLLM}} & \multicolumn{2}{c}{DeepSpeed-} & \multicolumn{2}{c}{\multirow{2}{*}{MAX}} & \multicolumn{2}{c}{\multirow{2}{*}{MLC LLM}} & \multicolumn{2}{c}{\multirow{2}{*}{SGLang}} & \multicolumn{2}{c}{\multirow{2}{*}{LitGPT}} & \multicolumn{2}{c}{\multirow{2}{*}{OpenLLM}} & \multicolumn{2}{c}{TensorRT-} & \multicolumn{2}{c}{\multirow{2}{*}{TGI}} & \multicolumn{2}{c}{\multirow{2}{*}{LMDeploy}} & \multicolumn{2}{c}{\multirow{2}{*}{vAttention}} \\
    & & & & & & & \multicolumn{2}{c}{FastGen} & & & & & & & & & & & \multicolumn{2}{c}{LLM} & & & & & & \\
    \cmidrule(lr){2-3} \cmidrule(lr){4-5} \cmidrule(lr){6-7} \cmidrule(lr){8-9} \cmidrule(lr){10-11} \cmidrule(lr){12-13} \cmidrule(lr){14-15} \cmidrule(lr){16-17} \cmidrule(lr){18-19} \cmidrule(lr){20-21} \cmidrule(lr){22-23} \cmidrule(lr){24-25} \cmidrule(lr){26-27}
     & \rotatebox{90}{A6000} & \rotatebox{90}{H100} & \rotatebox{90}{A6000} & \rotatebox{90}{H100} & \rotatebox{90}{A6000} & \rotatebox{90}{H100} & \rotatebox{90}{A6000} & \rotatebox{90}{H100} & \rotatebox{90}{A6000} & \rotatebox{90}{H100} & \rotatebox{90}{A6000} & \rotatebox{90}{H100} & \rotatebox{90}{A6000} & \rotatebox{90}{H100} & \rotatebox{90}{A6000} & \rotatebox{90}{H100} & \rotatebox{90}{A6000} & \rotatebox{90}{H100} & \rotatebox{90}{A6000} & \rotatebox{90}{H100} & \rotatebox{90}{A6000} & \rotatebox{90}{H100} & \rotatebox{90}{A6000} & \rotatebox{90}{H100} & \rotatebox{90}{A6000} & \rotatebox{90}{H100} \\
    \midrule
    Llama-2-7b-hf & \greencheck & \greencheck & \redxmark & \redxmark & \greencheck & \greencheck & \greencheck & \greencheck & \redxmark & \redxmark & \greencheck & \greencheck & \greencheck & \greencheck & \greencheck & \greencheck & \redxmark & \redxmark & \greencheck & \greencheck & \greencheck & \greencheck & \greencheck & \greencheck & \redxmark & \redxmark \\
    Llama-2-13b-hf & \greencheck & \greencheck & \redxmark & \redxmark & \greencheck & \greencheck & \greencheck & \redxmark & \redxmark & \redxmark & \greencheck & \greencheck & \greencheck & \greencheck & \greencheck & \greencheck & \redxmark & \redxmark & \greencheck & \greencheck & \greencheck & \greencheck & \greencheck & \greencheck & \redxmark & \redxmark \\
    Llama-2-70b-hf & \greencheck & \greencheck & \redxmark & \redxmark & \redxmark & \greencheck & \redxmark & \redxmark & \redxmark & \redxmark & \redxmark & \redxmark & \greencheck & \greencheck & \greencheck & \greencheck & \redxmark & \redxmark & \redxmark & \redxmark & \redxmark & \greencheck & \redxmark & \greencheck & \redxmark & \redxmark \\
    Meta-Llama-3-8B-Instruct & \redxmark & \redxmark & \redxmark & \redxmark & \greencheck & \redxmark & \redxmark & \redxmark & \redxmark & \redxmark & \redxmark & \redxmark & \greencheck & \redxmark & \redxmark & \redxmark & \redxmark & \redxmark & \greencheck & \greencheck & \greencheck & \greencheck & \greencheck & \greencheck & \greencheck & \redxmark \\
    Meta-Llama-3-70B & \greencheck & \greencheck & \redxmark & \redxmark & \greencheck & \greencheck & \redxmark & \redxmark & \redxmark & \greencheck & \greencheck & \greencheck & \greencheck & \greencheck & \redxmark & \greencheck & \redxmark & \redxmark & \redxmark & \redxmark & \redxmark & \greencheck & \redxmark & \greencheck & \redxmark & \redxmark \\
    Llama-3.1-8B & \greencheck & \greencheck & \greencheck & \greencheck & \greencheck & \greencheck & \greencheck & \redxmark & \redxmark & \greencheck & \greencheck & \greencheck & \greencheck & \greencheck & \greencheck & \greencheck & \greencheck & \greencheck & \greencheck & \greencheck & \redxmark & \greencheck & \greencheck & \greencheck & \redxmark & \redxmark \\
    Llama-4-Scout-17B-16E & \greencheck & \greencheck & \greencheck & \greencheck & \redxmark & \greencheck & \redxmark & \redxmark & \redxmark & \redxmark & \redxmark & \redxmark & \redxmark & \greencheck & \redxmark & \redxmark & \redxmark & \redxmark & \redxmark & \redxmark & \redxmark & \redxmark & \redxmark & \greencheck & \redxmark & \redxmark \\
    Mistral-7B-Instruct-v0.3 & \greencheck & \greencheck & \redxmark & \redxmark & \redxmark & \greencheck & \redxmark & \redxmark & \redxmark & \redxmark & \greencheck & \greencheck & \greencheck & \greencheck & \greencheck & \greencheck & \redxmark & \redxmark & \greencheck & \greencheck & \greencheck & \greencheck & \greencheck & \greencheck & \redxmark & \redxmark \\
    Mistral-Small-3.2-24B-Instruct-2506 & \greencheck & \greencheck & \greencheck & \greencheck & \redxmark & \greencheck & \redxmark & \redxmark & \redxmark & \greencheck & \redxmark & \redxmark & \redxmark & \greencheck & \redxmark & \redxmark & \redxmark & \redxmark & \greencheck & \greencheck & \redxmark & \redxmark & \redxmark & \redxmark & \redxmark & \redxmark \\
    Mixtral-8x7B-Instruct-v0.1 & \greencheck & \greencheck & \redxmark & \redxmark & \redxmark & \greencheck & \redxmark & \redxmark & \redxmark & \redxmark & \greencheck & \greencheck & \greencheck & \greencheck & \greencheck & \greencheck & \redxmark & \redxmark & \redxmark & \redxmark & \redxmark & \redxmark & \redxmark & \redxmark & \redxmark & \redxmark \\
    Qwen2.5-7B & \greencheck & \greencheck & \redxmark & \redxmark & \greencheck & \greencheck & \greencheck & \redxmark & \redxmark & \greencheck & \greencheck & \greencheck & \greencheck & \greencheck & \greencheck & \greencheck & \greencheck & \greencheck & \greencheck & \greencheck & \greencheck & \greencheck & \greencheck & \greencheck & \redxmark & \redxmark \\
    Qwen3-32B & \greencheck & \greencheck & \greencheck & \greencheck & \redxmark & \greencheck & \redxmark & \redxmark & \redxmark & \greencheck & \greencheck & \greencheck & \greencheck & \greencheck & \redxmark & \greencheck & \redxmark & \redxmark & \redxmark & \redxmark & \greencheck & \greencheck & \redxmark & \greencheck & \redxmark & \redxmark \\
    Phi-3-mini-4k-instruct & \greencheck & \greencheck & \greencheck & \greencheck & \greencheck & \greencheck & \redxmark & \redxmark & \redxmark & \redxmark & \redxmark & \redxmark & \redxmark & \redxmark & \greencheck & \greencheck & \redxmark & \redxmark & \redxmark & \greencheck & \greencheck & \greencheck & \greencheck & \greencheck & \redxmark & \redxmark \\
    falcon-40b & \greencheck & \greencheck & \redxmark & \redxmark & \redxmark & \greencheck & \redxmark & \redxmark & \redxmark & \redxmark & \redxmark & \redxmark & \redxmark & \redxmark & \greencheck & \greencheck & \redxmark & \redxmark & \redxmark & \redxmark & \redxmark & \redxmark & \redxmark & \redxmark & \redxmark & \redxmark \\
    Falcon3-1B-Instruct & \greencheck & \greencheck & \greencheck & \greencheck & \greencheck & \greencheck & \redxmark & \redxmark & \redxmark & \greencheck & \redxmark & \redxmark & \redxmark & \redxmark & \greencheck & \greencheck & \redxmark & \redxmark & \greencheck & \greencheck & \greencheck & \greencheck & \redxmark & \redxmark & \redxmark & \redxmark \\
    gpt-oss-20b & \greencheck & \greencheck & \greencheck & \greencheck & \redxmark & \greencheck & \redxmark & \redxmark & \redxmark & \redxmark & \redxmark & \redxmark & \greencheck & \greencheck & \redxmark & \redxmark & \redxmark & \redxmark & \redxmark & \greencheck & \redxmark & \redxmark & \greencheck & \greencheck & \redxmark & \redxmark \\
    DeepSeek-R1-Distill-Qwen-32B & \greencheck & \greencheck & \greencheck & \greencheck & \redxmark & \greencheck & \redxmark & \redxmark & \redxmark & \redxmark & \greencheck & \greencheck & \greencheck & \greencheck & \redxmark & \redxmark & \redxmark & \redxmark & \redxmark & \greencheck & \redxmark & \greencheck & \redxmark & \redxmark & \redxmark & \redxmark \\
    DeepSeek-V2 & \greencheck & \greencheck & \redxmark & \redxmark & \redxmark & \greencheck & \redxmark & \redxmark & \redxmark & \redxmark & \redxmark & \redxmark & \redxmark & \greencheck & \redxmark & \redxmark & \redxmark & \redxmark & \redxmark & \redxmark & \redxmark & \greencheck & \redxmark & \redxmark & \redxmark & \redxmark \\
    opt-6.7b & \redxmark & \redxmark & \redxmark & \redxmark & \greencheck & \greencheck & \greencheck & \redxmark & \redxmark & \redxmark & \redxmark & \redxmark & \greencheck & \greencheck & \redxmark & \redxmark & \redxmark & \redxmark & \redxmark & \redxmark & \redxmark & \greencheck & \redxmark & \redxmark & \redxmark & \redxmark \\
    \bottomrule
    \end{tabular}%
    }

\end{table}

\news{In this paper, as summarized in Table~\ref{tab:inference_engines_support_model}, we first verified which models each inference engine could successfully run on each GPU, and then conducted performance measurements. The experimental setup was standardized by applying a tensor parallelism factor of 4 for models with more than 20B parameters and 8 for models with more than 70B parameters.}

\news{As summarized in Table~\ref{tab:inference_engines_support_model}, we observed cases in which the same model and inference engine ran successfully on the NVIDIA A6000 but failed on the NVIDIA H100, and vice versa. These differences arise not only from architectural distinctions between the two GPUs but also from whether each inference engine provides the necessary runtime and kernel support required for the corresponding device. In general, the H100 offers greater memory capacity and computational performance than the A6000 and is widely adopted as a primary data-center GPU, resulting in broader support across many inference engines. However, some engines, such as vAttention~\cite{prabhu2025vattention}, do not yet provide H100-compatible kernels, making it impossible to execute certain models on the H100 in such cases.}

\subsubsection{\news{Evaluation Results on Servers with Quantized Models}} \label{sec:exp_results_server_quant}

\news {In this section, we evaluate the performance of various language models using server-grade inference engines. We first distinguish between engines that primarily support quantized models such as Ollama~\cite{ollama}, LLaMA.cpp~\cite{llamacpp}, and MLC LLM~\cite{mlcllm}, and engines that mainly operate on FP16/BF16 full-precision weights such as vLLM~\cite{kwon2023efficient}, SGLang~\cite{zheng2024sglang}, and TensorRT-LLM~\cite{tensorrtllm}, and compare their performance separately. 
Since the evaluated inference engines support 4-bit weight-only quantization, all experiments were conducted using this format for consistency.
To emulate real service workloads, most experiments were conducted with the concurrency level fixed at 16, while the prompt and output lengths were varied to measure the metrics defined in Table~\ref{tab:guidellm_metrics_summary} including TTFT, TBT, and token throughput.}

\news { Six models were selected, including Meta-Llama-3.1-8B, Llama-4-Scout-17B-16E, Qwen3-32B, Phi-3-mini-4k-Instruct, gpt-oss-20b, and DeepSeek-R1-Distill-Qwen-32B, and these models were executed on Ollama~\cite{ollama}, LLaMA.cpp~\cite{llamacpp}, and MLC LLM~\cite{mlcllm}.}

\news{\textbf{TTFT variation with respect to prompt length.} When measuring TTFT while increasing the prompt length from 64 to 4,096 tokens, all inference engines showed a general trend in which TTFT increased as the prompt length became longer. However, the magnitude of this increase differed depending on the model, the engine, and the underlying GPU architecture. For example, as shown in Fig.~\ref{fig:server_ttft_q}, Qwen3-32B and DeepSeek-R1-Distill-Qwen-32B exhibited a relatively linear TTFT growth across nearly all engine-hardware combinations, whereas other models showed more gradual increases or, in some cases, nearly constant TTFT across segments. When running Meta-Llama-3.1-8B with a prompt length of 1,024 tokens, LLaMA.cpp (A6000) recorded an average TTFT of approximately 5,454 ms, providing nearly twice the initial response speed of Ollama (A6000) at 10,706 ms. The difference persisted in the H100 environment, where LLaMA.cpp (H100) measured 11,980 ms, compared to 13,737 ms for Ollama (H100). Under the same conditions, MLC LLM (H100) achieved the lowest TTFT at 2,076 ms. For the Llama-4-Scout-17B-16E, the performance gap was even more pronounced. LLaMA.cpp (H100) recorded 658 ms, delivering a substantially faster response than Ollama (H100), which measured 14,349 ms. In certain large-scale models, the absolute TTFT values were unexpectedly low. This appears to be due to specific engines including optimized kernels for those models, or because some requests failed during execution, resulting in a smaller number of samples being included in the final statistics.}

\begin{figure*}[tbp]
    \centering
    \begin{subfigure}[t]{0.375\textwidth}
        \centering
        \input{result/server/quant/ttft/server_ttft_quant_llama3}   
        \vspace*{-4.2mm}
        \caption{Meta-Llama-3.1-8B}
        \label{fig:server_ttft_quant_llama3}
    \end{subfigure}%
    \hspace*{-3.0em}
    \begin{subfigure}[t]{0.375\textwidth}
        \centering
         \input{result/server/quant/ttft/server_ttft_quant_llama4}   
        \caption{Llama-4-Scout-17B-16E}
        \label{fig:server_ttft_quant_llama4}
    \end{subfigure}%
    \hspace*{-3.0em}
    \begin{subfigure}[t]{0.375\textwidth}
        \centering
        \input{result/server/quant/ttft/server_ttft_quant_qwen3}   
        \caption{Qwen3-32B}
        \label{fig:server_ttft_quant_qwen3}
    \end{subfigure}%

    \vspace{0.15cm}  
    
    \hspace*{-0.86em}
    \begin{subfigure}[t]{0.375\textwidth}
        \centering
        \input{result/server/quant/ttft/server_ttft_quant_phi3}   
        \caption{Phi-3-mini-4k-instruct}
        \label{fig:server_ttft_quant_phi}
    \end{subfigure}
    \hspace*{-2.74em}
    \begin{subfigure}[t]{0.375\textwidth}
        \centering
        \input{result/server/quant/ttft/server_ttft_quant_gpt}   
        \caption{gpt-oss-20b}
        \label{fig:server_ttft_quant_gpt}
    \end{subfigure}
    \hspace*{-3.3em}
    \begin{subfigure}[t]{0.375\textwidth}
        \centering
        \input{result/server/quant/ttft/server_ttft_quant_deepseek}   
        \caption{DeepSeek-R1-Distill-Qwen-32B}
        \label{fig:server_ttft_quant_deepseek}
    \end{subfigure}

    \caption{\news{TTFT variation with prompt length on server devices (4bit Quantized models, Concurrency=16, Output length=1024) }}
    \label{fig:server_ttft_q}

\end{figure*}

\news{\textbf{TBT variation with respect to output length.} When measuring TBT as the output length increased from 512 to 4,096 tokens, the absolute decoding speed varied across GPU-engine combinations, but the overall trend remained consistent. As shown in Fig.~\ref{fig:server_tbt_q}, when running Meta-Llama-3.1-8B with an output length of 1,024 tokens, Ollama and LLaMA.cpp achieved similar performance on the A6000, recording 11.24 ms and 10.12 ms per token, respectively. In contrast, on the H100, Ollama recorded 6.52 ms, and LLaMA.cpp recorded 4.54 ms, meaning both engines achieved more than double the decoding speed compared to the A6000. A similar trend was observed for the mid-sized model Phi-3-mini-4k-instruct, with LLaMA.cpp (H100) providing the lowest TBT among all tested combinations. Conversely, for larger models such as DeepSeek-R1-Distill-Qwen-32B, TBT remained relatively high even on the H100. For example, Ollama (H100) recorded 17.69 ms, and MLC LLM (H100) recorded 13.34 ms, both noticeably slower than other combinations. In addition, several inference failures occurred on the A6000 when running Qwen3-32B and DeepSeek-R1-Distill-Qwen-32B with output lengths exceeding 512 tokens. On the H100, Qwen3-32B consistently failed once the output length exceeded 1,024 tokens. These results indicate that for large models, memory and bandwidth requirements increase rapidly with output length, and the kernels and runtime configurations provided by some inference engines may be insufficient to handle such demands.}

\begin{figure*}[tbp]
    \centering
    \begin{subfigure}[t]{0.375\textwidth}
        \centering
        \input{result/server/quant/tbt/server_tbt_quant_llama3}   
        \vspace*{-4.2mm}
        \caption{Meta-Llama-3.1-8B}
        \label{fig:server_tbt_quant_llama3}
    \end{subfigure}%
    \hspace*{-3.0em}
    \begin{subfigure}[t]{0.375\textwidth}
        \centering
         \input{result/server/quant/tbt/server_tbt_quant_llama4}   
        \caption{Llama-4-Scout-17B-16E}
        \label{fig:server_tbt_quant_llama4}
    \end{subfigure}%
    \hspace*{-3.0em}
    \begin{subfigure}[t]{0.375\textwidth}
        \centering
        \input{result/server/quant/tbt/server_tbt_quant_qwen3}   
        \caption{Qwen3-32B}
        \label{fig:server_tbt_quant_qwen3}
    \end{subfigure}%

    \vspace{0.3cm}  
    
    \hspace*{-1.2em}
    \begin{subfigure}[t]{0.375\textwidth}
        \centering
        \input{result/server/quant/tbt/server_tbt_quant_phi3}   
        \caption{Phi-3-mini-4k-instruct}
        \label{fig:server_tbt_quant_phi}
    \end{subfigure}
    \hspace*{-2.4em}
    \begin{subfigure}[t]{0.375\textwidth}
        \centering
        \input{result/server/quant/tbt/server_tbt_quant_gpt}   
        \caption{gpt-oss-20b}
        \label{fig:server_tbt_quant_gpt}
    \end{subfigure}
    \hspace*{-3.4em}
    \begin{subfigure}[t]{0.375\textwidth}
        \centering
        \input{result/server/quant/tbt/server_tbt_quant_deepseek}   
        \caption{DeepSeek-R1-Distill-Qwen-32B}
        \label{fig:server_tbt_quant_deepseek}
    \end{subfigure}

    \caption{\news{TBT on server devices with varying output length (4bit Quantized models, Concurrency = 16, Prompt length = 1024) }}
    \label{fig:server_tbt_q}

\end{figure*}

\news{\textbf{Request throughput (Requests/s) with respect to concurrency.} As shown in Fig.~\ref{fig:server_rps_q}, the request throughput measured under varying concurrency levels differed across inference engines due to differences in batch scheduling strategies and internal pipeline designs. For Meta-Llama-3.1-8B with the concurrency level fixed at 16, Ollama (H100) achieved the highest throughput at approximately 0.43 requests/s, followed by Ollama (A6000) at 0.23 requests/s and LLaMA.cpp (H100) at 0.21 requests/s. The throughput improvement on the newer H100, which was roughly twice that of the A6000 for the same engine, indicates that faster GPU decoding speed directly translates into higher batch processing performance. For the Llama-4-Scout-17B-16E, the gap widened further. Under the same test conditions, Ollama (H100) reached 0.33 requests/s, whereas LLaMA.cpp (H100) achieved only 0.06 requests/s. This highlights the increasing importance of batch scheduling strategies and kernel optimizations as model size grows. For the smaller model Phi-3-mini-4k-instruct, the differences between engines were relatively small. Ollama (H100) achieved 0.33 requests/s, and LLaMA.cpp (H100) recorded 0.27 Requests/s, indicating similar levels of performance. This is likely because smaller models require fewer computations per token, making engine-level kernel differences less pronounced.}

\begin{figure*}[tbp]
    \centering
    \begin{subfigure}[t]{0.375\textwidth}
        \centering
        \input{result/server/quant/rps/server_req_quant_llama3}   
        \vspace*{-4.2mm}
        \caption{Meta-Llama-3.1-8B}
        \label{fig:server_rps_quant_llama3}
    \end{subfigure}%
    \hspace*{-3.0em}
    \begin{subfigure}[t]{0.375\textwidth}
        \centering
         \input{result/server/quant/rps/server_req_quant_llama4}   
        \caption{Llama-4-Scout-17B-16E}
        \label{fig:server_rps_quant_llama4}
    \end{subfigure}%
    \hspace*{-3.0em}
    \begin{subfigure}[t]{0.375\textwidth}
        \centering
        \input{result/server/quant/rps/server_req_quant_qwen3}   
        \caption{Qwen3-32B}
        \label{fig:server_rps_quant_qwen3}
    \end{subfigure}%

    \vspace{0.3cm}  
    
    \hspace*{-1.2em}
    \begin{subfigure}[t]{0.375\textwidth}
        \centering
        \input{result/server/quant/rps/server_req_quant_phi3}   
        \caption{Phi-3-mini-4k-instruct}
        \label{fig:server_rps_quant_phi}
    \end{subfigure}
    \hspace*{-2.4em}
    \begin{subfigure}[t]{0.375\textwidth}
        \centering
        \input{result/server/quant/rps/sever_req_quant_gpt}   
        \caption{gpt-oss-20b}
        \label{fig:server_rps_quant_gpt}
    \end{subfigure}
    \hspace*{-3.4em}
    \begin{subfigure}[t]{0.375\textwidth}
        \centering
        \input{result/server/quant/rps/server_req_quant_deepseek}   
        \caption{DeepSeek-R1-Distill-Qwen-32B}
        \label{fig:server_rps_quant_deepseek}
    \end{subfigure}

    \caption{ \news{Requests per second  on server devices with varying concurrency (4bit Quantized models, Prompt length = 1024, Output length = 1024) }}
    \label{fig:server_rps_q}

\end{figure*}

\news{\textbf{Token Processing Capability.} Total tokens/s represents the overall token processing throughput of the system and directly reflects hardware resource utilization and the efficiency of engine implementation. As shown in Fig.~\ref{fig:server_tts_q}, for the Meta-Llama-3.1-8B model with concurrency set to 16, Ollama (H100) achieved approximately 588 tokens/s, while LLaMA.cpp (H100) recorded 431 tokens/s. LLaMA.cpp (A6000) reached only 194 tokens/s, which is less than half of the throughput observed on the H100. For Llama-4-Scout-17B-16E, Ollama (H100) achieved 510 tokens/s, which is nearly four times higher than LLaMA.cpp (H100) at 132 tokens/s. In contrast, for the smaller model Phi-3-mini-4k-instruct, the two engines showed almost identical performance, reaching 589 tokens/s and 555 tokens/s, respectively. For the gpt-oss-20B model, LLaMA.cpp (H100) delivered 402 tokens/s, surpassing Ollama (H100), which achieved 271 tokens/s.}

\begin{figure*}[tbp]
    \centering
    \begin{subfigure}[t]{0.375\textwidth}
        \centering
        \input{result/server/quant/tts/server_tts_quant_llama3}   
        \vspace*{-4.2mm}
        \caption{Meta-Llama-3.1-8B}
        \label{fig:server_tts_quant_llama3}
    \end{subfigure}%
    \hspace*{-3.0em}
    \begin{subfigure}[t]{0.375\textwidth}
        \centering
         \input{result/server/quant/tts/server_tts_quant_llama4}   
        \caption{Llama-4-Scout-17B-16E}
        \label{fig:server_tts_quant_llama4}
    \end{subfigure}%
    \hspace*{-3.0em}
    \begin{subfigure}[t]{0.375\textwidth}
        \centering
        \input{result/server/quant/tts/server_tts_quant_qwen3}   
        \caption{Qwen3-32B}
        \label{fig:server_tts_quant_qwen3}
    \end{subfigure}%

    \vspace{0.3cm}  
    
    \hspace*{-1.2em}
    \begin{subfigure}[t]{0.375\textwidth}
        \centering
        \input{result/server/quant/tts/server_tts_quant_phi3}   
        \caption{Phi-3-mini-4k-instruct}
        \label{fig:server_tts_quant_phi}
    \end{subfigure}
    \hspace*{-2.4em}
    \begin{subfigure}[t]{0.375\textwidth}
        \centering
        \input{result/server/quant/tts/server_tts_quant_gpt}   
        \caption{gpt-oss-20b}
        \label{fig:server_tts_quant_gpt}
    \end{subfigure}
    \hspace*{-3.4em}
    \begin{subfigure}[t]{0.375\textwidth}
        \centering
        \input{result/server/quant/tts/server_tts_quant_deepseek}   
        \caption{DeepSeek-R1-Distill-Qwen-32B}
        \label{fig:server_tts_quant_deepseek}
    \end{subfigure}

    \caption{ \news{Total tokens per second on server devices with varying concurrency (4bit Quantized models, Prompt length = 1024, Output length = 1024) }}
    \label{fig:server_tts_q}

\end{figure*}

\news{\textbf{End-to-End Latency.} End-to-end latency represents the combined effect of TTFT, TBT, scheduling delays, and serving-layer delays, and therefore provides the most accurate indication of the response time perceived by end users. As shown in Fig.~\ref{fig:server_latency_q} the Meta-Llama-3.1-8B model measured at a concurrency level of 16, request latency showed similar values across engines and GPU configurations, generally falling within the range of 15-17 seconds. Specifically, Ollama (A6000) recorded 14.8 s, LLaMA.cpp (A6000) recorded 15.8 s, Ollama (H100) measured 15.9 s, and LLaMA.cpp (H100) recorded 16.6 s, with differences among engines being relatively small compared to those observed in TTFT and TBT. For Phi-3-mini-4k-instruct, Ollama (H100) measured 16.7s, while LLaMA.cpp (H100) measured 16.6 s, indicating nearly identical end-to-end latency. However, detailed metrics show that LLaMA.cpp achieved slightly faster token streaming due to its lower TBT. Overall, most model--engine--hardware combinations exhibited increasing request latency as concurrency increased. However, DeepSeek-R1-Distill-Qwen-32B showed intermittent fluctuations in latency. MLC LLM successfully executed the Qwen3-32B model only under specific configurations on both the A6000 and H100, while other configurations failed to run.}

\begin{figure*}[tbp]
    \centering
    \begin{subfigure}[t]{0.375\textwidth}
        \centering
        \input{result/server/quant/latency/server_latency_quant_llama3}   
        \vspace*{-4.2mm}
        \caption{Meta-Llama-3.1-8B}
        \label{fig:server_latency_quant_llama3}
    \end{subfigure}%
    \hspace*{-3.0em}
    \begin{subfigure}[t]{0.375\textwidth}
        \centering
         \input{result/server/quant/latency/server_latency_quant_llama4}   
        \caption{Llama-4-Scout-17B-16E}
        \label{fig:server_latency_quant_llama4}
    \end{subfigure}%
    \hspace*{-3.0em}
    \begin{subfigure}[t]{0.375\textwidth}
        \centering
        \input{result/server/quant/latency/server_latency_quant_qwen3}   
        \caption{Qwen3-32B}
        \label{fig:server_latency_quant_qwen3}
    \end{subfigure}%

    \vspace{0.3cm}  
    
    \hspace*{-1.2em}
    \begin{subfigure}[t]{0.375\textwidth}
        \centering
        \input{result/server/quant/latency/server_latency_quant_phi3}   
        \caption{Phi-3-mini-4k-instruct}
        \label{fig:server_latency_quant_phi}
    \end{subfigure}
    \hspace*{-2.4em}
    \begin{subfigure}[t]{0.375\textwidth}
        \centering
        \input{result/server/quant/latency/server_latency_quant_gpt}   
        \caption{gpt-oss-20b}
        \label{fig:server_latency_quant_gpt}
    \end{subfigure}
    \hspace*{-3.4em}
    \begin{subfigure}[t]{0.375\textwidth}
        \centering
        \input{result/server/quant/latency/server_latency_quant_deepseek}   
        \caption{DeepSeek-R1-Distill-Qwen-32B}
        \label{fig:server_latency_quant_deepseek}
    \end{subfigure}

    \caption{ \news{Request Latency on server devices with varying concurrency (4bit Quantized models, Prompt length = 1024, Output length = 1024) }}
    \label{fig:server_latency_q}

\end{figure*}

\news{\textbf{Capacity under increasing concurrency.} To evaluate stability as concurrency increased from 1 to 64, we measured success rates and partial-response rates. As shown in Fig.~\ref{fig:server_req_quant_request}, all engines exhibited a common pattern in which success rates dropped sharply and partial responses increased substantially as concurrency increased. This behavior became more pronounced for larger models. For example, with Meta-Llama-3.1-8B at concurrency 16, Ollama (H100) maintained a relatively stable success rate of 48.1\%, whereas LLaMA.cpp (A6000) dropped to 11.1\%. At concurrency 64, the success rate of Ollama (H100) further decreased to 16.2\%, and LLaMA.cpp (A6000) fell to approximately 3\%. MLC LLM failed to maintain high concurrency, reaching 0\% success starting from concurrency 4. For the Qwen3-32B model, at concurrency 16, Ollama (H100) and LLaMA.cpp (H100) reported success rates of 6.3\% and 6.0\%, respectively, and both fell below 1.5 at concurrency 64. DeepSeek-R1-Distill-Qwen-32B showed a similar pattern, with Ollama (H100) reaching only 11.8\% even at concurrency 16. These results indicate that when deploying large models in quantized form on a single server under high concurrency, a substantial portion of requests may fail or return incomplete responses regardless of the inference engine or hardware used.}

\begin{figure*}[t]
    \centering    
    \begin{subfigure}{\textwidth}
        \centering
        \input{result/server/quant/request/server_req_quant_llama3}  
        \caption{Meta-Llama-3.1-8B}
        \label{fig:server_req_quant_llama3}
    \end{subfigure}
    
    \begin{subfigure}{\textwidth}
        \centering
        \input{result/server/quant/request/server_req_quant_llama4}  
        \caption{Llama-4-Scout-17B-16E}
        \label{fig:server_req_quant_llama4}
    \end{subfigure}
    
    \begin{subfigure}{\textwidth}
        \centering
        \input{result/server/quant/request/server_req_quant_qwen3}  
        \caption{Qwen3-32B}
        \label{fig:server_req_quant_qwen3}
    \end{subfigure}
    
    \begin{subfigure}{\textwidth}
        \centering
        \input{result/server/quant/request/server_req_quant_phi}  
        \caption{Phi-3-mini-4k-instruct}
        \label{fig:server_req_quant_phi}
    \end{subfigure}
    
    \begin{subfigure}{\textwidth}
        \centering
        \input{result/server/quant/request/server_req_quant_gpt}  
        \caption{gpt-oss-20b}
        \label{fig:server_req_quant_gpt}
    \end{subfigure}
    
    \begin{subfigure}{\textwidth}
        \centering
        \input{result/server/quant/request/server_req_quant_deepseek}  
        \caption{DeepSeek-R1-Distill-Qwen-32B}
        \label{fig:server_req_quant_deepseek}
    \end{subfigure}
    \caption{ \news {Impact of concurrency scaling on inference capability (4bit Quantized models, Prompt length = 1024, Output length = 1024)}}
    \label{fig:server_req_quant_request}
\end{figure*}

\begin{findingbox}[]
   {\small \news{\textbf{Key Takeaways}}}
    \begin{itemize}[leftmargin=1.2em, label=--, itemsep=0.6em]

        \item \news{\textbf{Small Models (e.g., Phi-3, Meta-Llama-3.1-8B):}  
        Ollama and LLaMA.cpp both demonstrated performance levels suitable for real service deployment.  
        LLaMA.cpp (H100) achieved exceptionally low decoding latency (4.54 ms/token), enabling responsive streaming,  
        while Ollama (H100) provided higher throughput and more stable success rates.}

        \item \news{\textbf{Medium-Scale Models (e.g., Llama-4-Scout-17B-16E, gpt-oss-20B):}  
        Ollama (H100) maintained relatively stable throughput and reliability.  
        In contrast, LLaMA.cpp delivered fast decoding but suffered from significantly lower success rates under concurrency,  
        suggesting limited suitability for high-throughput multi-user scenarios.}

        \item \news{\textbf{Large Models (e.g., Qwen3-32B, DeepSeek-R1-Distill-Qwen-32B):}  
        Even with quantization, these models were unable to sustain a concurrency level of 16 or higher on a single server.  
        Success rates remained extremely low (1--10\%), indicating that practical deployment would require  
        distributed serving, reduced concurrency, or careful tuning of batching and scheduling parameters.}

        \item \news{\textbf{MLC LLM Stability Considerations:}  
        Although MLC LLM achieved very fast TTFT in specific configurations,  
        its overall low success rate makes it unsuitable for production workloads in its current state.  
        This emphasizes that backend stability, scheduler design, and memory management strategies  
        are essential factors that quantization alone cannot compensate for.}

    \end{itemize}
\end{findingbox}

\begin{table}[tbp]
    \caption{ \news {Hands-on Experience with Inference Engines for Quantization models)}}
    \label{tab:inference_engine_quant_hands-on}
    \centering
    \resizebox{.75\textwidth}{!}{
        \begin{tabular}{@{}p{0.2\textwidth} p{0.8\textwidth}@{}}
        \toprule
        Inference Engine & Feature Items \\
        \midrule
        Ollama &
        - Easy installation and execution \newline
        - Supports model hot-swap and multi-model serving during server operation \newline
        - Allocates only the memory required for inference
        \\
        \midrule
        LLaMA.cpp &
        - By default, utilizes multiple GPUs to optimize computation \newline
        - Stability degrades when prompt or output length becomes large \newline
        - Uses relatively low memory
        \\
        \midrule
        MLC LLM &
        - The server performance is slow because it wraps the RESTful API into an OpenAI compatible server \newline
        - Allocates only the necessary memory \newline
        - Lacks model-checking when receiving client requests, reducing reliability \newline
        - Offers strong synergy for users proficient with compiler-level optimizations
        \\
        \bottomrule
        \end{tabular}
    }
\end{table}

\subsubsection{\news{Evaluation Results on Servers}} \label{sec:exp_results_server}

\news {For server-side performance analysis, models were selected according to their parameter scale. Small-scale models included Falcon-3-1B-Instruct; medium-scale models included Llama-2-7B-hf, Meta-Llama-3-8B-Instruct, and Qwen-2.5-7B; medium-large models included gpt-oss-20B, DeepSeek-R1-Distill-Qwen-32B, and Meta-Llama-3-70B; large models included Llama-4-Scout-17B-16E; and the extra-large model category was represented by DeepSeek-V2. }

\news {The selection also included models with knowledge distillation (e.g., DeepSeek-R1-Distill-Qwen-32B) and MoE architectures (e.g., gpt-oss-20B and Llama-4-Scout-17B-16E) to evaluate how effectively each inference engine can accommodate diverse service requirements. In addition, because large- and larger-scale models often cannot be executed on a single GPU, this experiment also examines the degree of parallelization support and the effectiveness of optimization provided by each inference engine.}

\news{\textbf{TTFT variation with respect to prompt length.} In the H100-based environment, TensorRT-LLM, vLLM, LMDeploy, and TGI exhibited the shortest TTFT across most models. As shown in Fig.~\ref{fig:server_ttft}, when a 1,024-token prompt was provided to the Llama-2-7B-HF model, TensorRT-LLM achieved approximately 133 ms, followed by TGI at 186 ms, LMDeploy at 290 ms, and vLLM at 320 ms, making TensorRT-LLM the fastest in terms of initial response. Under the same configuration, A6000-based engines generally showed TTFT values 1.5$\times$ to 3$\times$ longer than their H100 counterparts. Notably, LitGPT (A6000) recorded delays of 18,000-19,000 ms even for prompt lengths between 64 and 512 tokens, indicating that additional optimization is required for real-time service scenarios. A similar trend was observed with the Qwen 2.5-7B model. On the A6000, vLLM and LMDeploy showed TTFT values of 898 ms and 1,053 ms, respectively, for a 1,024-token prompt, whereas on the H100 these values were reduced to 299 ms and 258 ms. For larger models such as Meta-Llama-3-70B and DeepSeek-R1-Distill-Qwen-32B, the absolute TTFT naturally increased, but TensorRT-LLM and vLLM still recorded the lowest TTFTs on the H100, followed by LMDeploy and TGI, maintaining the same ranking across configurations.}

\begin{figure*}[tbp]
    \centering
    
    \begin{subfigure}[]{0.375\textwidth}
        \centering
        \input{result/server/normal/ttft/server_ttft_llama2}   
        \vspace*{-4.0mm}
        \caption{Llama-2-7b-hf}
        \label{fig:server_ttft_llama2}
    \end{subfigure}%
    \hspace*{-3.2em}
    \begin{subfigure}[]{0.375\textwidth}
        \centering
        \vspace*{19mm}
        \input{result/server/normal/ttft/server_ttft_llama3_8b}   
        \caption{Meta-Llama-3-8B-Instruct}
        \label{fig:server_ttft_llama3_8b}
    \end{subfigure}%
    \hspace*{-3.2em}
    \begin{subfigure}[]{0.375\textwidth}
        \centering
        \vspace*{19mm}
        \input{result/server/normal/ttft/server_ttft_llama3_70b}   
        \caption{Meta-Llama-3-70B}
        \label{fig:server_ttft_llama3_70b}
    \end{subfigure}%

    \vspace{0.15cm}  
    
    \hspace*{-0.46em}
    \begin{subfigure}[t]{0.375\textwidth}
        \centering
        \input{result/server/normal/ttft/server_ttft_llama4}   
        \caption{Llama-4-Scout-17B-16E}
        \label{fig:server_ttft_llama4}
    \end{subfigure}
    \hspace*{-3.3em}
    \begin{subfigure}[t]{0.375\textwidth}
        \centering
        \input{result/server/normal/ttft/server_ttft_qwen}   
        \caption{Qwen2.5-7B}
        \label{fig:server_ttft_qwen}
    \end{subfigure}
    \hspace*{-3.35em}
    \begin{subfigure}[t]{0.375\textwidth}
        \centering
        \input{result/server/normal/ttft/server_ttft_falcon}   
        \caption{Falcon3-1B-Instruct}
        \label{fig:server_ttft_falcon}
    \end{subfigure}

    \vspace{0.15cm}  
    
    \hspace*{-0.95em}
    \begin{subfigure}[t]{0.375\textwidth}
        \centering
        \input{result/server/normal/ttft/server_ttft_gpt}   
        \caption{gpt-oss-20b}
        \label{fig:server_ttft_gpt}
    \end{subfigure}
    \hspace*{-2.65em}
    \begin{subfigure}[t]{0.375\textwidth}
        \centering
        \input{result/server/normal/ttft/server_ttft_deepseek_v2}   
        \caption{DeepSeek-V2}
        \label{fig:server_ttft_deepseek_v2}
    \end{subfigure}
    \hspace*{-3.35em}
    \begin{subfigure}[t]{0.375\textwidth}
        \centering
        \input{result/server/normal/ttft/server_ttft_deepseek_r1}   
        \caption{DeepSeek-R1-Distill-Qwen-32B}
        \label{fig:server_ttft_deepseek_r1}
    \end{subfigure}

    \caption{\news{TTFT variation with prompt length on server devices (Concurrency=16, Output length=1024) }}
    \label{fig:server_ttft}

\end{figure*}

\news{\textbf{TBT variation with respect to output length.} TBT directly reflects the level of kernel and KV-cache optimization in the decoding stage. For the Llama-2-7B-HF model, TBT was measured by fixing concurrency at 16 and the prompt length at 1,024 tokens while increasing the output length. The results are as shown in Fig.~\ref{fig:server_tbt}. In the H100 environment, TensorRT-LLM showed an average TBT of approximately 5.0 ms at an output length of 1,024 tokens, 2.65 ms at 2,048 tokens, and 1.23 ms at 4,096 tokens, indicating a decreasing per-token latency as output length increased. Under the same conditions, vLLM (H100) recorded 10.6 ms, 11.3 ms, and 12.6 ms at 512, 1,024, and 2,048 tokens, showing a gradual increase. TGI (H100) produced a relatively flat curve with values of 14.6 ms, 15.1 ms, 15.2 ms, and 14.5 ms at 512, 1,024, 2,048, and 4,096 tokens. LMDeploy (H100) measured 9.6 ms, 10.1 ms, and 11.4 ms at 512, 1,024, and 2,048 tokens, which is moderately lower than TGI but still roughly twice that of TensorRT-LLM. These results indicate that TensorRT-LLM applies the most aggressive kernel and cache optimizations on the H100, whereas vLLM, TGI, and LMDeploy provide more stable performance profiles. For medium-scale models such as Qwen 2.5-7B, the absolute TBT values were slightly lower, but the relative performance ranking among engines remained largely consistent.}

\begin{figure*}[tbp]
    \centering
    
    \begin{subfigure}[]{0.375\textwidth}
        \centering
        \input{result/server/normal/tbt/server_tbt_llama2}   
        \vspace*{-4.0mm}
        \caption{Llama-2-7b-hf}
        \label{fig:server_tbt_llama2}
    \end{subfigure}%
    \hspace*{-3.2em}
    \begin{subfigure}[]{0.375\textwidth}
        \centering
        \vspace*{19mm}
        \input{result/server/normal/tbt/server_tbt_llama3_8b}   
        \caption{Meta-Llama-3-8B-Instruct}
        \label{fig:server_tbt_llama3_8b}
    \end{subfigure}%
    \hspace*{-3.2em}
    \begin{subfigure}[]{0.375\textwidth}
        \centering
        \vspace*{19mm}
        \input{result/server/normal/tbt/server_tbt_llama3_70b}   
        \caption{Meta-Llama-3-70B}
        \label{fig:server_tbt_llama3_70b}
    \end{subfigure}%

    \vspace{0.15cm}  
    
    \hspace*{-0.7em}
    \begin{subfigure}[t]{0.375\textwidth}
        \centering
        \input{result/server/normal/tbt/server_tbt_llama4}   
        \caption{Llama-4-Scout-17B-16E}
        \label{fig:server_tbt_llama4}
    \end{subfigure}
    \hspace*{-2.7em}
    \begin{subfigure}[t]{0.375\textwidth}
        \centering
        \input{result/server/normal/tbt/server_tbt_qwen}   
        \caption{Qwen2.5-7B}
        \label{fig:server_tbt_qwen}
    \end{subfigure}
    \hspace*{-3.35em}
    \begin{subfigure}[t]{0.375\textwidth}
        \centering
        \input{result/server/normal/tbt/server_tbt_falcon}   
        \caption{Falcon3-1B-Instruct}
        \label{fig:server_tbt_falcon}
    \end{subfigure}

    \vspace{0.15cm}  
    
    \hspace*{-1.0em}
    \begin{subfigure}[t]{0.375\textwidth}
        \centering
        \input{result/server/normal/tbt/server_tbt_gpt}   
        \caption{gpt-oss-20b}
        \label{fig:server_tbt_gpt}
    \end{subfigure}
    \hspace*{-2.6em}
    \begin{subfigure}[t]{0.375\textwidth}
        \centering
        \input{result/server/normal/tbt/server_tbt_deepseek_v2}   
        \caption{DeepSeek-V2}
        \label{fig:server_tbt_deepseek_v2}
    \end{subfigure}
    \hspace*{-3.5em}
    \begin{subfigure}[t]{0.375\textwidth}
        \centering
        \input{result/server/normal/tbt/server_tbt_deepseek_r1}   
        \caption{DeepSeek-R1-Distill-Qwen-32B}
        \label{fig:server_tbt_deepseek_r1}
    \end{subfigure}

    \caption{\news{TBT on server devices with varying output length (Concurrency = 16, Prompt length = 1024) }}
    \label{fig:server_tbt}

\end{figure*}

\news{\textbf{Request throughput (Requests/s) with respect to concurrency.} Examining the throughput metrics Requests/s show that H100-based engines provide excellent scalability, particularly for small and medium-size models. As illustrated in Fig.~\ref{fig:server_rps}, for the Llama-2-7B-HF model, TensorRT-LLM (H100) achieved approximately 0.21, 1.06, 1.83, 2.74, 3.52, and 3.68 req/s at concurrency levels of 1, 4, 8, 16, 32, and 64, respectively. Under the same conditions, vLLM (H100) recorded 0.14, 0.50, 0.88, 1.34, 1.87, and 2.00 req/s, corresponding to about 50-70\% of TensorRT-LLM. LMDeploy (H100) achieved 1.50, 2.07, and 2.57 req/s at concurrency levels of 16, 32, and 64, slightly outperforming vLLM but still falling below TensorRT-LLM. TGI (H100) produced approximately 2.37 req/s at concurrency 64, placing it between vLLM and LMDeploy. In contrast, engines running on the A6000, such as DeepSpeed-FastGen, MAX, OpenLLM, and vAttention, showed substantially lower throughput, typically in the range of 0.03-0.10 req/s, and in some cases produced no measurable results. These observations indicate that, even on the same hardware, the amount of real-world service traffic that can be supported varies significantly depending on the inference engine being used.}

\begin{figure*}[tbp]
    \centering
    
    \begin{subfigure}[]{0.375\textwidth}
        \centering
        \input{result/server/normal/rps/server_rps_llama2}   
        \vspace*{-4.05mm}
        \caption{Llama-2-7b-hf}
        \label{fig:server_rps_llama2}
    \end{subfigure}%
    \hspace*{-3.2em}
    \begin{subfigure}[]{0.375\textwidth}
        \centering
        \vspace*{19mm}
        \input{result/server/normal/rps/server_rps_llama3_8b}   
        \caption{Meta-Llama-3-8B-Instruct}
        \label{fig:server_rps_llama3_8b}
    \end{subfigure}%
    \hspace*{-3.2em}
    \begin{subfigure}[]{0.375\textwidth}
        \centering
        \vspace*{19mm}
        \input{result/server/normal/rps/server_rps_llama3_70b}   
        \caption{Meta-Llama-3-70B}
        \label{fig:server_rps_llama3_70b}
    \end{subfigure}%

    \vspace{0.15cm}  
    
    \hspace*{-0.9em}
    \begin{subfigure}[t]{0.375\textwidth}
        \centering
        \input{result/server/normal/rps/server_rps_llama4}   
        \caption{Llama-4-Scout-17B-16E}
        \label{fig:server_rps_llama4}
    \end{subfigure}
    \hspace*{-2.65em}
    \begin{subfigure}[t]{0.375\textwidth}
        \centering
        \input{result/server/normal/rps/server_rps_qwen}   
        \caption{Qwen2.5-7B}
        \label{fig:server_rps_qwen}
    \end{subfigure}
    \hspace*{-3.4em}
    \begin{subfigure}[t]{0.375\textwidth}
        \centering
        \input{result/server/normal/rps/server_rps_falcon}   
        \caption{Falcon3-1B-Instruct}
        \label{fig:server_rps_falcon}
    \end{subfigure}

    \vspace{0.15cm}  
    
    \hspace*{-0.95em}
    \begin{subfigure}[t]{0.375\textwidth}
        \centering
        \input{result/server/normal/rps/server_rps_gpt}   
        \caption{gpt-oss-20b}
        \label{fig:server_rps_gpt}
    \end{subfigure}
    \hspace*{-2.65em}
    \begin{subfigure}[t]{0.375\textwidth}
        \centering
        \input{result/server/normal/rps/server_rps_deepseek_v2}   
        \caption{DeepSeek-V2}
        \label{fig:server_rps_deepseek_v2}
    \end{subfigure}
    \hspace*{-3.35em}
    \begin{subfigure}[t]{0.375\textwidth}
        \centering
        \input{result/server/normal/rps/server_rps_deepseek_r1}   
        \caption{DeepSeek-R1-Distill-Qwen-32B}
        \label{fig:server_rps_deepseek_r1}
    \end{subfigure}

    \caption{ \news{Requests per second  on server devices with varying concurrency (Prompt length = 1024, Output length = 1024) }}
    \label{fig:server_rps}

\end{figure*}

\news{\textbf{Token Processing Capability.} The total tokens per second metric provides a clearer indication of decoding efficiency across inference engines. As shown in Fig.~\ref{fig:server_tts}, when running the Llama-2-7B-HF model on the H100, TensorRT-LLM achieved approximately 7,535 tokens/s at concurrency 64 and 7,206 tokens/s at concurrency 32. Under the same conditions, vLLM reached 4,107 tokens/s, TGI achieved 3,058 tokens/s, and LMDeploy delivered 4,246 tokens/s. In other words, TensorRT-LLM provides roughly 1.8$\times$ the throughput of vLLM and 2.5$\times$ that of TGI. For the medium-scale Qwen 2.5-7B model, overall throughput was higher. At concurrency 64, LMDeploy achieved approximately 12,020 tokens/s, vLLM reached 10,928 tokens/s, TensorRT-LLM recorded 10,596 tokens/s, and TGI achieved 9,242 tokens/s. All four engines maintained between 7,000 and 12,000 tokens/s in the concurrency 32-64 range, suggesting that these engines may be suitable for high-concurrency production environments. In contrast, LitGPT and SGLang exhibited sharp drops in throughput beyond certain concurrency levels, with some configurations failing to produce any measurable output. For large models such as DeepSeek-R1-Distill-Qwen-32B, several engines were unable to achieve meaningful throughput, indicating the substantial challenges of serving large models efficiently at scale.}

\begin{figure*}[tbp]
    \centering
    
    \begin{subfigure}[]{0.375\textwidth}
        \centering
        \input{result/server/normal/tts/server_tts_llama2}   
        \vspace*{-4.05mm}
        \caption{Llama-2-7b-hf}
        \label{fig:server_tts_llama2}
    \end{subfigure}%
    \hspace*{-3.4em}
    \begin{subfigure}[]{0.375\textwidth}
        \centering
        \vspace*{19mm}
        \input{result/server/normal/tts/server_tts_llama3_8b}   
        \caption{Meta-Llama-3-8B-Instruct}
        \label{fig:server_tts_llama3_8b}
    \end{subfigure}%
    \hspace*{-3.2em}
    \begin{subfigure}[]{0.375\textwidth}
        \centering
        \vspace*{19mm}
        \input{result/server/normal/tts/server_tts_llama3_70b}   
        \caption{Meta-Llama-3-70B}
        \label{fig:server_tts_llama3_70b}
    \end{subfigure}%

    \vspace{0.15cm}  
    
    \hspace*{-1.1em}
    \begin{subfigure}[t]{0.375\textwidth}
        \centering
        \input{result/server/normal/tts/server_tts_llama4}   
        \caption{Llama-4-Scout-17B-16E}
        \label{fig:server_tts_llama4}
    \end{subfigure}
    \hspace*{-2.7em}
    \begin{subfigure}[t]{0.375\textwidth}
        \centering
        \input{result/server/normal/tts/server_tts_qwen}   
        \caption{Qwen2.5-7B}
        \label{fig:server_tts_qwen}
    \end{subfigure}
    \hspace*{-3.4em}
    \begin{subfigure}[t]{0.375\textwidth}
        \centering
        \input{result/server/normal/tts/server_tts_falcon}   
        \caption{Falcon3-1B-Instruct}
        \label{fig:server_tts_falcon}
    \end{subfigure}

    \vspace{0.15cm}  
    
    \hspace*{-1.0em}
    \begin{subfigure}[t]{0.375\textwidth}
        \centering
        \input{result/server/normal/tts/server_tts_gpt}   
        \caption{gpt-oss-20b}
        \label{fig:server_tts_gpt}
    \end{subfigure}
    \hspace*{-2.65em}
    \begin{subfigure}[t]{0.375\textwidth}
        \centering
        \input{result/server/normal/tts/server_tts_deepseek_v2}   
        \caption{DeepSeek-V2}
        \label{fig:server_tts_deepseek_v2}
    \end{subfigure}
    \hspace*{-3.35em}
    \begin{subfigure}[t]{0.375\textwidth}
        \centering
        \input{result/server/normal/tts/server_tts_deepseek_r1}   
        \caption{DeepSeek-R1-Distill-Qwen-32B}
        \label{fig:server_tts_deepseek_r1}
    \end{subfigure}

    \caption{ \news{Total tokens per second on server devices with varying concurrency (Prompt length = 1024, Output length = 1024) }}
    \label{fig:server_tts}

\end{figure*}

\news{\textbf{End-to-End Latency.} Analysis of request latency shows that, across most models, TensorRT-LLM, vLLM, LMDeploy, and TGI on the H100 maintain average latencies of approximately 10 seconds up to concurrency 32, and converge to around 10-12 seconds even at concurrency 64. For example, in Fig.~\ref{fig:server_latency}, with Qwen-2.5-7B, LMDeploy (H100) recorded 6.4 s, 6.7 s, 6.9 s, 7.4 s, 8.3 s, and 10.6 s at concurrency levels of 1, 4, 8, 16, 32, and 64, respectively. Under the same conditions, vLLM (H100) showed 6.6 s, 7.0 s, 7.4 s, 8.1 s, 9.0 s, and 12.0 s, while TensorRT-LLM (H100) achieved the lowest latencies at 5.6 s, 6.0 s, 6.2 s, 6.7 s, 7.6 s, and 10.1 s. A similar pattern was observed for Llama-2-7B-HF, where TensorRT-LLM and vLLM demonstrated both high Requests/s and Total tokens/s while keeping latency at comparatively low levels, highlighting their effectiveness as high-efficiency serving engines. In contrast, engines such as DeepSpeed-FastGen, MAX, OpenLLM, and vAttention exhibited low success rates or high partial-response ratios, making it difficult to assess service quality based solely on average latency.}

\begin{figure*}[tbp]
    \centering
    
    \begin{subfigure}[]{0.375\textwidth}
        \centering
        \hspace*{-0.3em}
        \input{result/server/normal/latency/server_latency_llama2}   
        \vspace*{-4.05mm}
        \caption{Llama-2-7b-hf}
        \label{fig:server_latency_llama2}
    \end{subfigure}%
    \hspace*{-3.4em}
    \begin{subfigure}[]{0.375\textwidth}
        \centering
        \vspace*{19mm}
        \input{result/server/normal/latency/server_latency_llama3_8b}   
        \caption{Meta-Llama-3-8B-Instruct}
        \label{fig:server_latency_llama3_8b}
    \end{subfigure}%
    \hspace*{-3.2em}
    \begin{subfigure}[]{0.375\textwidth}
        \centering
        \vspace*{19mm}
        \input{result/server/normal/latency/server_latency_llama3_70b}   
        \caption{Meta-Llama-3-70B}
        \label{fig:server_latency_llama3_70b}
    \end{subfigure}%

    \vspace{0.15cm}  
    
    \hspace*{-1.1em}
    \begin{subfigure}[t]{0.375\textwidth}
        \centering
        \input{result/server/normal/latency/server_latency_llama4}   
        \caption{Llama-4-Scout-17B-16E}
        \label{fig:server_latency_llama4}
    \end{subfigure}
    \hspace*{-2.7em}
    \begin{subfigure}[t]{0.375\textwidth}
        \centering
        \input{result/server/normal/latency/server_latency_qwen}   
        \caption{Qwen2.5-7B}
        \label{fig:server_latency_qwen}
    \end{subfigure}
    \hspace*{-3.4em}
    \begin{subfigure}[t]{0.375\textwidth}
        \centering
        \input{result/server/normal/latency/server_latency_falcon}   
        \caption{Falcon3-1B-Instruct}
        \label{fig:server_latency_falcon}
    \end{subfigure}

    \vspace{0.15cm}  
    
    \hspace*{-1.0em}
    \begin{subfigure}[t]{0.375\textwidth}
        \centering
        \input{result/server/normal/latency/server_latency_gpt}   
        \caption{gpt-oss-20b}
        \label{fig:server_latency_gpt}
    \end{subfigure}
    \hspace*{-2.65em}
    \begin{subfigure}[t]{0.375\textwidth}
        \centering
        \input{result/server/normal/latency/server_latency_deepseek_v2}   
        \caption{DeepSeek-V2}
        \label{fig:server_latency_deepseek_v2}
    \end{subfigure}
    \hspace*{-3.35em}
    \begin{subfigure}[t]{0.375\textwidth}
        \centering
        \input{result/server/normal/latency/server_latency_deepseek_r1}   
        \caption{DeepSeek-R1-Distill-Qwen-32B}
        \label{fig:server_latency_deepseek_r1}
    \end{subfigure}

    \caption{ \news{Total tokens per second on server devices with varying concurrency (Prompt length = 1024, Output length = 1024) }}
    \label{fig:server_latency}

\end{figure*}

\news{\textbf{Capacity under increasing concurrency.} In terms of overall serving capacity, the impact of increasing concurrency shows relatively stable behavior for vLLM, TGI, TensorRT-LLM, and LMDeploy when evaluated with the Llama-2-7B-hf model. As shown in Fig.~\ref{fig:server_req}, in the H100 environment, vLLM maintains success rates of 80\% at concurrency 1, 75\% at concurrency 4 and 8, and continues to record approximately 75\% at concurrency 16. When concurrency increases to 32 and 64, the success rates gradually drop to 66.7\% and 50\%, respectively, yet more than half of the requests are still completed successfully. TGI (H100) records success rates of 83.3\% and 87.5\% at concurrency 1 and 4, 82.5\% at 8, and then declines to 67.1\%, 56.5\%, and 25.\% at concurrency 16, 32, and 64. TensorRT-LLM (H100) achieves 85.7\%, 89.2\%, and 87.7\% at concurrency 1, 4, and 8, respectively, and maintains relatively high success rates of 78.5\%, 68.6\%, and 67.2\% at concurrency 16, 32, and 64. In contrast, some engines such as vAttention show near-zero success rates or a majority of partial responses for the same model, making them unsuitable for practical service deployment. For example, with Llama-2-7B-hf, DeepSpeed-FastGen (A6000) achieved only 5.9\% success at concurrency 16, with 94.1\% of requests resulting in incomplete responses. SGLang and LitGPT maintained success rates above 70\% at lower concurrencies (1, 4, 8) on the A6000, but produced almost no valid responses when concurrency increased beyond 16. These patterns were consistently observed across other models, including Meta-Llama-3-8B-Instruct, Qwen-2.5-7B, Falcon-3-1B-Instruct, GPT-OSS-20B, and DeepSeek-V2, indicating that each inference engine has a different upper bound for stable concurrency.}

\begin{figure*}[tbp]
    \centering    
    
    \begin{subfigure}{\textwidth}
        \centering
        \input{result/server/normal/request/server_request_llama2_1}\par

        \vspace{-0.2cm} 
        
        \input{result/server/normal/request/server_request_llama2_2}
        \caption{Llama-2-7b-hf}
        \label{fig:server_request_llama2_7b_hf}
    \end{subfigure}
    
    \begin{subfigure}{\textwidth}
        \centering
        \input{result/server/normal/request/server_request_llama3_8b_1}

        \vspace{-0.2cm} 
        
        \input{result/server/normal/request/server_request_llama3_8b_2}
        \caption{Meta-Llama-3-8B-Instruct}
        \label{fig:server_request_llama3_8b_instruct}
    \end{subfigure}
    
    \begin{subfigure}{\textwidth}
        \centering
        \input{result/server/normal/request/server_request_llama3_70b_1}

        \vspace{-0.2cm} 
    
        \input{result/server/normal/request/server_request_llama3_70b_2}
        \caption{Meta-Llama-3-70B}
        \label{fig:server_request_llama3_70b}
    \end{subfigure}
    
    \caption{ \news {Impact of concurrency scaling on inference reliability on servers (Prompt length = 1024, Output length = 1024)}}
    \label{fig:server_req}
\end{figure*}

\begin{figure*}[tbp]
    \ContinuedFloat 
    \centering    
    
    \begin{subfigure}{\textwidth}
        \centering
        \input{result/server/normal/request/server_request_llama4_1}

         \vspace{-0.2cm} 

        \input{result/server/normal/request/server_request_llama4_2}
        \caption{Llama-4-Scout-17B-16E}
        \label{fig:server_request_llama4}
    \end{subfigure}
    
    \begin{subfigure}{\textwidth}
        \centering
        \input{result/server/normal/request/server_request_qwen_1}

         \vspace{-0.2cm} 

        \input{result/server/normal/request/server_request_qwen_2}
        \caption{Qwen2.5-7B}
        \label{fig:server_request_qwen2_5_7b}
    \end{subfigure}
    
    \begin{subfigure}{\textwidth}
        \centering
        \input{result/server/normal/request/server_request_falcon_1}

         \vspace{-0.2cm}

        \input{result/server/normal/request/server_request_falcon_2}
        \caption{Falcon3-1B-Instruct}
        \label{fig:server_request_falcon3_1b_instruct}
    \end{subfigure}

    \caption{ \news { (continued) Impact of concurrency scaling on inference reliability on servers (Prompt length = 1024, Output length = 1024)}}
    \label{fig:server_req}
\end{figure*}

\begin{figure*}[tbp]
    \ContinuedFloat 
    \centering  
    
    \begin{subfigure}{\textwidth}
        \centering
        \input{result/server/normal/request/server_request_gpt_1}

         \vspace{-0.2cm}

        \input{result/server/normal/request/server_request_gpt_2}
        \caption{gpt-oss-20b}
        \label{fig:server_request_gpt_oss_20b}
    \end{subfigure}
    
    \begin{subfigure}{\textwidth}
        \centering
        \input{result/server/normal/request/server_request_deepseek_v2_1}

         \vspace{-0.2cm}

        \input{result/server/normal/request/server_request_deepseek_v2_2}
        \caption{DeepSeek-V2}
        \label{fig:server_request_deepseek_v2}
    \end{subfigure}
    
    \begin{subfigure}{\textwidth}
        \centering
        \input{result/server/normal/request/server_request_deepseek_r1_1}

         \vspace{-0.2cm}

        \input{result/server/normal/request/server_request_deepseek_r1_2}
        \caption{DeepSeek-R1-Distill-Qwen-32B}
        \label{fig:server_request_deepseek_r1_qwen32b}
    \end{subfigure}
    
    \caption{ \news {(continued) Impact of concurrency scaling on inference reliability on servers (Prompt length = 1024, Output length = 1024)}}
\end{figure*}


\begin{findingbox}[]
   {\small \news{\textbf{Key Takeaways}}}
    \begin{itemize}[leftmargin=1.2em, label=--, itemsep=0.6em]

        \item \news{\textbf{Top Server-Side Performance:}  
        On H100 servers, TensorRT-LLM, vLLM, LMDeploy, and TGI consistently delivered the strongest performance among full-precision inference engines.}

        \item \news{\textbf{TensorRT-LLM Shows Peak Performance with Minor Stability Trade-Offs:}  
        TensorRT-LLM achieved the highest performance across nearly all metrics—TTFT, TBT, Requests/s, Total tokens/s, and request latency.  
        However, it occasionally showed higher failure or partial-response rates for certain models or at high concurrency.}

        \item \news{\textbf{vLLM, LMDeploy, and TGI: Balanced and Stable Alternatives:}  
        Although these engines did not reach TensorRT-LLM’s peak performance, they exhibited smoother degradation and more stable behavior under load, making them strong general-purpose choices.}

        \item \news{\textbf{SGLang, LitGPT, and DeepSpeed-FastGen offer competitive performance but reduced stability:}  
        These engines achieved competitive performance under specific hardware–model–concurrency combinations.  
        Their stability, however, decreased noticeably as concurrency levels or model size increased, leading to reduced throughput and lower success rates.}

        \item \news{\textbf{Consistency Across Diverse Models:}  
        The observed performance–stability patterns were consistent across a wide range of architectures, including  
        Llama-2-7B-hf, Meta-Llama-3-8B/70B, Llama-4-Scout-17B, Qwen-2.5-7B, Falcon-3-1B, GPT-OSS-20B, DeepSeek-V2, and DeepSeek-R1-Distill-Qwen-32B.}

        \item \news{\textbf{Implication for Engine Selection:}  
        Choosing an inference engine for server-side deployment involves balancing peak speed against operational stability.  
        In practice, TensorRT-LLM offers unmatched performance, while vLLM, LMDeploy, and TGI provide more stable and reliable behavior under sustained workloads.}

    \end{itemize}
\end{findingbox}

\begin{table}[tbp]
    \caption{\news {Hands-on Experience with Inference Engines}}
    \label{tab:inference_engine_hands-on}
    \centering
    \resizebox{.75\textwidth}{!}{
        \begin{tabular}{@{}p{0.22\textwidth} p{0.85\textwidth}@{}}
        \toprule
        Inference Engine & Feature Items \\
        \midrule

        Ollama &
        - Easy installation and execution \newline
        - Supports model hot-swap and multi-model serving during server operation \newline
        - Allocates only the memory required for inference
        \\
        \midrule

        LLaMA.cpp &
        - Utilizes multiple GPUs by default for optimized computation \newline
        - Stability degrades when prompt or output length becomes large \newline
        - Uses relatively low memory
        \\
        \midrule

        MLC LLM &
        - Server performance is slow due to RESTful API wrapping into an OpenAI-compatible server \newline
        - Allocates only necessary memory \newline
        - Lacks model-checking when receiving client requests \newline
        - Strong synergy for users proficient in compiler-level optimizations
        \\
        \midrule

        vLLM &
        - Supported context length is strict and tied to the model's maximum window \newline
        - Allocates GPU memory aggressively (1B models may reach \~90 percent of total GPU memory)
        \\
        \midrule

        DeepSpeed-FastGen &
        - Wraps a RESTful interface as an OpenAI API, resulting in slow serving performance \newline
        - Unstable with large models or large input/output sequences \newline
        - Limited support for long-context inference \newline
        - Aggressive memory allocation (7B models may reach \~90 percent of GPU memory)
        \\
        \midrule

        MAX &
        - Supports only specific accelerators (H100, MI300, CPUs) \newline
        - A6000 is not supported \newline
        - Multi-GPU execution unavailable for many models
        \\
        \midrule

        SGLang &
        - Requires an additional Python server implementation \newline
        - Moderate memory usage (\~40 percent of GPU memory for 7B models)
        \\
        \midrule

        LitGPT &
        - Not directly compatible with GuideLLM's OpenAI API interface \newline
        - Requires a proxy server or code modifications \newline
        - Moderate GPU memory usage (\~40 percent for 7B models)
        \\
        \midrule

        TensorRT-LLM &
        - Compatible with OpenAI API (GuideLLM request format requires modification) \newline
        - Supports only chat-oriented models in current evaluation \newline
        - Utilizes full available GPU memory
        \\
        \midrule

        TGI &
        - Provides optimizations such as prefix caching and FlashInfer \newline
        - Consumes substantial GPU memory (\~90 percent for 8B models)
        \\
        \midrule

        LMDeploy &
        - Based on the Turbomind backend \newline
        - Tensor parallelism can be unstable \newline
        - High GPU memory usage (\~90 percent for 8B models)
        \\
        \midrule

        vAttention &
        - Uses Sarathi as backend \newline
        - Serving via Ray Workers \newline
        - Moderate GPU memory usage (\~50 percent for 8B models) \newline
        - Tensor parallelism is unstable \newline
        - Does not support H100 due to incompatible kernel versions
        \\
        \bottomrule
        \end{tabular}
    }
\end{table}

\subsubsection{\news{Evaluation Results on Edge}} \label{sec:exp_results_edge}

\news{Edge-device performance evaluation was conducted only for Ollama and LLaMA.cpp, the two engines that can run natively on the NVIDIA Jetson Orin AGX 32GB board. Due to hardware constraints, the evaluation was limited to quantized models, and OpenAI API-style requests were generated using GuideLLM, consistent with the server-side experiments. However, considering the limited computational resources and throughput of the edge device, the experimental configuration was adjusted by reducing concurrency levels, prompt lengths, and output lengths.}

\news{\textbf{TTFT variation with respect to prompt length.} On the Jetson Orin AGX 32GB board, all models exhibited multi-second to multi-tens-of-seconds delays before returning the first token, with the magnitude of latency varying significantly across engine-model combinations. As shown in Fig.~\ref{fig:jetson_ttft}, for the Llama-3.1 8B model under concurrency 2 with prompt lengths of 64, 128, and 256 tokens, Ollama recorded TTFT values of approximately 4.97 s, 7.18 s, and 5.96 s, respectively. Under the same conditions, LLaMA.cpp measured 18.3 s, 18.4 s, and 18.6 s, indicating that Ollama provided initial response times roughly 2.5-3.5$\times$ faster for this model. In contrast, the trend reversed for the Qwen3 0.6B model. For prompt lengths of 64, 128, and 256 tokens, Ollama recorded TTFT values of 6.59 s, 6.33 s, and 6.06 s, while LLaMA.cpp achieved 4.41 s, 4.44 s, and 4.51 s, respectively, offering about 20-30\% lower latency. For medium-to-large models such as Qwen3 8B and Qwen3 14B, TTFT increased sharply for both engines. On Qwen3 8B, Ollama reached 20.6-20.8 s and LLaMA.cpp 19.3-19.7 s. On Qwen3 14B, Ollama recorded 35.3-40.3 s, while LLaMA.cpp measured 33.5-34.5 s. Although LLaMA.cpp consistently provided 1-6 s lower TTFT in this range, the absolute delay exceeding 30 seconds indicates substantial initial response latency for practical use. Clear differences in initial response characteristics between the two engines were also observed for models such as the Gemma family, DeepSeek, and gpt-oss-20B. For Gemma3-1B, both Ollama and LLaMA.cpp recorded TTFT values in the 7.1-8.2 s range, showing no meaningful gap between the two engines. However, for Gemma3-4B, Ollama achieved 10.2 s (prompt 64), 14.6 s (128), and 14.7 s (256), whereas LLaMA.cpp recorded 14.2 s, 14.4 s, and 14.5 s, indicating that Ollama provided initial response times approximately 20-30\% faster. The opposite trend appeared for DeepSeek-R1-Distill-Qwen-1.5B. While Ollama produced TTFT values around 13.5-14.0 s, LLaMA.cpp achieved 6.7-6.8 s, more than twice as fast. A similar pattern was observed for gpt-oss-20B, where Ollama recorded 26.8-32.8 s, whereas LLaMA.cpp measured 14.2-14.7 s, delivering roughly half the TTFT. In summary, on the Jetson Orin platform, Ollama provides significantly faster TTFT for certain models such as Llama-3.1-8B and Gemma3-4B, whereas LLaMA.cpp consistently outperforms Ollama for other model families including Qwen3-0.6B, DeepSeek-1.5B, and gpt-oss-20B. This indicates that the two engines differ in their optimization pathways, and their performance advantages vary depending on model architecture characteristics.}

\begin{figure*}[tbp]
    \centering
    \begin{subfigure}[t]{0.3\textwidth}
        \centering
        \input{result/edge/ttft/edge_ttft_llama3}   
        \vspace*{-4.2mm}
        \caption{Meta-Llama-3.1-8B}
        \label{fig:edge_ttft_llama3}
    \end{subfigure}%
    \hspace*{-3.1em}
    \begin{subfigure}[t]{0.3\textwidth}
        \centering
         \input{result/edge/ttft/edge_ttft_qwen3_0.6b}   
        \caption{Qwen3-0.6B}
        \label{fig:edge_ttft_qwen3_0.6b}
    \end{subfigure}%
    \hspace*{-3.1em}
    \begin{subfigure}[t]{0.3\textwidth}
        \centering
        \input{result/edge/ttft/edge_ttft_qwen3_8b}   
        \caption{Qwen3-8B}
        \label{fig:edge_ttft_qwen3_8b}
    \end{subfigure}%
    \hspace*{-3.05em}
    \begin{subfigure}[t]{0.3\textwidth}
        \centering
        \input{result/edge/ttft/edge_ttft_qwen3_14b}   
        \caption{Qwen3-14B}
        \label{fig:edge_ttft_qwen3_14b}
    \end{subfigure}%

    \vspace{0.15cm}  
    
    \hspace*{-1.15em}
    \begin{subfigure}[t]{0.3\textwidth}
        \centering
        \input{result/edge/ttft/edge_ttft_gemma3_1b}   
        \caption{gemma-3-1b-it}
        \label{fig:edge_ttft_gemma3_1b}
    \end{subfigure}
    \hspace*{-2.55em}
    \begin{subfigure}[t]{0.3\textwidth}
        \centering
        \input{result/edge/ttft/edge_ttft_gemma3_4b}   
        \caption{gemma-3-4b-it}
        \label{fig:edge_ttft_gemma3_4b}
    \end{subfigure}
    \hspace*{-3.35em}
    \begin{subfigure}[t]{0.3\textwidth}
        \centering
        \input{result/edge/ttft/edge_ttft_deepseek}   
        \caption{\centering DeepSeek-R1-\\Distill-Qwen-1.5B}
        \label{fig:edge_ttft_deepseek}
    \end{subfigure}
    \hspace*{-3.3em}
    \begin{subfigure}[t]{0.3\textwidth}
        \centering
        \input{result/edge/ttft/edge_ttft_gpt}   
        \caption{gpt-oss-20b }
        \label{fig:edge_ttft_gpt}
    \end{subfigure}

    \caption{ \news{TTFT variation with prompt length on edge devices (Concurrency=2, Output length=512) }}
    \label{fig:jetson_ttft}

\end{figure*}

\news{\textbf{TBT variation with respect to output length.} In the Jetson Orin environment, the measured TBT latency generally ranged from several milliseconds to several tens of milliseconds per token. For example, as shown in Fig.~\ref{fig:jetson_tbt}, for Llama-3.1-8B, even when the output length increased to 256, 512, and 1,024 tokens, Ollama maintained TBT values of 38.4 ms, 38.5 ms, and 38.4 ms, respectively, while LLaMA.cpp recorded 35.2 ms, 35.2 ms, and 35.3 ms, showing an almost flat profile. For Qwen3-8B, Ollama remained around 40 ms, and LLaMA.cpp around 37 ms, resulting in a difference of only about 3 ms between the engines. For larger models such as Qwen3-14B, per-token latency increased to around 60 ms, but the performance gap between engines remained within 10 percent. Small models exhibited a different pattern. With Qwen3-0.6B, LLaMA.cpp remained nearly constant at 8.4-8.6 ms across all output lengths, but Ollama decreased from 10.0 ms (256 tokens) to 6.0 ms (512 tokens) and 2.5 ms (1,024 tokens), showing lower latency as output length increased. For Gemma-3 1B and 4B, LLaMA.cpp recorded 35-36 ms and 27-28 ms, respectively, which were 2-3 ms faster than Ollama. Conversely, for DeepSeek-1.5B and gpt-oss-20B, Ollama's TBT dropped to as low as 1 ms in some segments, whereas LLaMA.cpp maintained approximately 13 ms and 27 ms, respectively. These results indicate that per-token latency varies depending on the model family and scale, and no single engine dominates across all cases. However, since overall response time is more heavily influenced by TTFT, differences in TBT may have a relatively limited impact on perceived user performance.}

\begin{figure*}[tbp]
    \centering
    \begin{subfigure}[t]{0.3\textwidth}
        \centering
        \input{result/edge/tbt/edge_tbt_llama3}   
        \vspace*{-4.2mm}
        \caption{Meta-Llama-3.1-8B}
        \label{fig:edge_tbt_llama3}
    \end{subfigure}%
    \hspace*{-3.1em}
    \begin{subfigure}[t]{0.3\textwidth}
        \centering
         \input{result/edge/tbt/edge_tbt_qwen3_0.6b}   
        \caption{Qwen3-0.6B}
        \label{fig:edge_tbt_qwen3_0.6b}
    \end{subfigure}%
    \hspace*{-3.1em}
    \begin{subfigure}[t]{0.3\textwidth}
        \centering
        \input{result/edge/tbt/edge_tbt_qwen3_8b}   
        \caption{Qwen3-8B}
        \label{fig:edge_tbt_qwen3_8b}
    \end{subfigure}%
    \hspace*{-3.05em}
    \begin{subfigure}[t]{0.3\textwidth}
        \centering
        \input{result/edge/tbt/edge_tbt_qwen3_14b}   
        \caption{Qwen3-14B}
        \label{fig:edge_tbt_qwen3_14b}
    \end{subfigure}%

    \vspace{0.15cm}  
    
    \hspace*{-0.95em}
    \begin{subfigure}[t]{0.3\textwidth}
        \centering
        \input{result/edge/tbt/edge_tbt_gemma3_1b}   
        \caption{gemma-3-1b-it}
        \label{fig:edge_tbt_gemma3_1b}
    \end{subfigure}
    \hspace*{-2.9em}
    \begin{subfigure}[t]{0.3\textwidth}
        \centering
        \input{result/edge/tbt/edge_tbt_gemma3_4b}   
        \caption{gemma-3-4b-it}
        \label{fig:edge_tbt_gemma3_4b}
    \end{subfigure}
    \hspace*{-3.3em}
    \begin{subfigure}[t]{0.3\textwidth}
        \centering
        \input{result/edge/tbt/edge_tbt_deepseek}   
        \caption{\centering DeepSeek-R1-\\Distill-Qwen-1.5B}
        \label{fig:edge_tbt_deepseek}
    \end{subfigure}
    \hspace*{-3.4em}
    \begin{subfigure}[t]{0.3\textwidth}
        \centering
        \input{result/edge/tbt/edge_tbt_gpt}   
        \caption{gpt-oss-20b }
        \label{fig:edge_tbt_gpt}
    \end{subfigure}

    \caption{ \news{TBT on edge devices with varying output length (Concurrency = 2, Prompt length = 128) }}
    \label{fig:jetson_tbt}

\end{figure*}

\news{\textbf{Request throughput (Requests/s) with respect to concurrency.} The request throughput on edge devices is one to two orders of magnitude lower than in server environments. Due to the limited GPU compute capability and memory bandwidth of the Jetson Orin platform, increasing concurrency does not meaningfully improve throughput; for some models, throughput remained nearly unchanged even as concurrency was raised. For example, as shown in Fig.~\ref{fig:jetson_rps}, the throughput of Ollama on Llama-3 8B remained almost flat at 0.14, 0.14, and 0.15 req/s for concurrency levels 1, 2, and 4, respectively, indicating that 8B models cannot benefit from batch parallelism on Jetson-class hardware. Under the same settings, LLaMA.cpp achieved 0.05, 0.05, and 0.04 req/s, approximately one-third of Ollama's throughput. For smaller models such as Qwen3 0.6B, both engines achieved around 0.20 req/s, though at concurrency 4, Ollama showed a 5-10 percent lower throughput than LLaMA.cpp. This reflects the architectural characteristics of Jetson devices, where front-end and model execution threads become less of a bottleneck as model size decreases. For mid-size models (Qwen3 8B and 14B), throughput dropped sharply for both engines. In particular, Qwen3 14B reached only 0.07 req/s at concurrency 4, making real-time service effectively infeasible. Gemma-3 1B and 4B also remained within 0.15-0.25 req/s, even when concurrency increased. For the large-scale gpt-oss-20B model, Ollama achieved 0.06 req/s and LLaMA.cpp 0.08 req/s at concurrency 4, maintaining a 20-30 percent performance gap.}

\begin{figure*}[t]
    \centering
    \begin{subfigure}[t]{0.3\textwidth}
        \centering
        \input{result/edge/rps/edge_rps_llama3}   
        \vspace*{-4.2mm}
        \caption{Meta-Llama-3.1-8B}
        \label{fig:edge_rps_llama3}
    \end{subfigure}%
    \hspace*{-3.1em}
    \begin{subfigure}[t]{0.3\textwidth}
        \centering
         \input{result/edge/rps/edge_rps_qwen3_0.6b}   
        \caption{Qwen3-0.6B}
        \label{fig:edge_rps_qwen3_0.6b}
    \end{subfigure}%
    \hspace*{-3.1em}
    \begin{subfigure}[t]{0.3\textwidth}
        \centering
        \input{result/edge/rps/edge_rps_qwen3_8b}   
        \caption{Qwen3-8B}
        \label{fig:edge_rps_qwen3_8b}
    \end{subfigure}%
    \hspace*{-3.05em}
    \begin{subfigure}[t]{0.3\textwidth}
        \centering
        \input{result/edge/rps/edge_rps_qwen3_8b}   
        \caption{Qwen3-14B}
        \label{fig:edge_rps_qwen3_14b}
    \end{subfigure}%

    \vspace{0.15cm}  
    
    \hspace*{-0.95em}
    \begin{subfigure}[t]{0.3\textwidth}
        \centering
        \input{result/edge/rps/edge_rps_gemma3_1b}   
        \caption{gemma-3-1b-it}
        \label{fig:edge_rps_gemma3_1b}
    \end{subfigure}
    \hspace*{-2.9em}
    \begin{subfigure}[t]{0.3\textwidth}
        \centering
        \input{result/edge/rps/edge_rps_gemma3_4b}   
        \caption{gemma-3-4b-it}
        \label{fig:edge_rps_gemma3_4b}
    \end{subfigure}
    \hspace*{-3.3em}
    \begin{subfigure}[t]{0.3\textwidth}
        \centering
        \input{result/edge/rps/edge_rps_deepseek}   
        \caption{\centering DeepSeek-R1-\\Distill-Qwen-1.5B}
        \label{fig:edge_rps_deepseek}
    \end{subfigure}
    \hspace*{-3.4em}
    \begin{subfigure}[t]{0.3\textwidth}
        \centering
        \input{result/edge/rps/edge_rps_gpt}   
        \caption{gpt-oss-20b}
        \label{fig:edge_rps_gpt}
    \end{subfigure}

    \caption{ \news{Requests per second on edge devices with varying concurrency (Prompt length = 128, Output length = 512) }}
    \label{fig:jetson_rps}

\end{figure*}

\news{\textbf{Token Processing Capability.} On the Jetson Orin platform, the significantly lower memory bandwidth and compute capability compared to server-grade hardware lead to a pronounced decline or stagnation in throughput as concurrency increases. This effect is particularly evident in Ollama, which shows stronger sensitivity to higher concurrency levels. As shown in Fig.~\ref{fig:jetson_tps}, for Llama-3 8B, Ollama maintained an almost flat throughput of 42.9, 43.3, and 42.2 tokens/s at concurrency 1, 2, and 4, respectively. In contrast, LLaMA.cpp dropped from 35.0 to 32.0 and 25.6 tokens/s, showing notable degradation as concurrency increased. For this model, Ollama holds a clear advantage in token throughput. The trend reverses for the smaller Qwen3 0.6B model. LLaMA.cpp sustained 145.6, 143.3, and 137.2 tokens/s across concurrency levels 1, 2, and 4, demonstrating high stability. Meanwhile, Ollama decreased sharply from 117.6 to 63.3 and 31.8 tokens/s. At concurrency 4, the throughput gap widens beyond a factor of four, suggesting that Jetson's memory and cache hierarchy align more favorably with LLaMA.cpp for smaller models. For larger models such as Qwen3 8B and 14B, throughput dropped significantly for both engines. In Qwen3 14B, Ollama recorded 18.2, 12.0, and 7.3 tokens/s, whereas LLaMA.cpp achieved 19.1, 15.2, and 9.6 tokens/s, consistently maintaining a 20-30 percent advantage. A similar trend appeared in Gemma-3 1B, where LLaMA.cpp delivered 89.4, 87.0, and 81.0 tokens/s, while Ollama fluctuated between 71.6, 76.9, and 71.8 tokens/s. The same pattern persists in DeepSeek R1-1.5B and gpt-oss-20B. For DeepSeek-1.5B, LLaMA.cpp maintained 95.4, 92.5, and 86.6 tokens/s, whereas Ollama dropped sharply to 73.7, 39.3, and 19.7 tokens/s. In gpt-oss-20B, LLaMA.cpp achieved 45.1, 41.9, and 35.5 tokens/s, compared to Ollama's 37.2, 19.3, and 9.8 tokens/s, resulting in more than a threefold gap at concurrency 4.}

\begin{figure*}[tbp]
    \centering
    \begin{subfigure}[t]{0.3\textwidth}
        \centering
        \input{result/edge/tps/edge_tps_llama3}   
        \vspace*{-4.2mm}
        \caption{Meta-Llama-3.1-8B}
        \label{fig:edge_tps_llama3}
    \end{subfigure}%
    \hspace*{-3.1em}
    \begin{subfigure}[t]{0.3\textwidth}
        \centering
         \input{result/edge/tps/edge_tps_qwen3_0.6b}   
        \caption{Qwen3-0.6B}
        \label{fig:edge_tps_qwen3_0.6b}
    \end{subfigure}%
    \hspace*{-3.1em}
    \begin{subfigure}[t]{0.3\textwidth}
        \centering
        \input{result/edge/tps/edge_tps_qwen3_8b}   
        \caption{Qwen3-8B}
        \label{fig:edge_tps_qwen3_8b}
    \end{subfigure}%
    \hspace*{-3.05em}
    \begin{subfigure}[t]{0.3\textwidth}
        \centering
        \input{result/edge/tps/edge_tps_qwen3_8b}   
        \caption{Qwen3-14B}
        \label{fig:edge_rps_qwen3_14b}
    \end{subfigure}%

    \vspace{0.15cm}  
    
    \hspace*{-0.95em}
    \begin{subfigure}[t]{0.3\textwidth}
        \centering
        \input{result/edge/tps/edge_tps_gemma3_1b}   
        \caption{gemma-3-1b-it}
        \label{fig:edge_tps_gemma3_1b}
    \end{subfigure}
    \hspace*{-2.9em}
    \begin{subfigure}[t]{0.3\textwidth}
        \centering
        \input{result/edge/tps/edge_tps_gemma3_4b}   
        \caption{gemma-3-4b-it}
        \label{fig:edge_tps_gemma3_4b}
    \end{subfigure}
    \hspace*{-3.3em}
    \begin{subfigure}[t]{0.3\textwidth}
        \centering
        \input{result/edge/tps/edge_tps_deepseek}   
        \caption{\centering DeepSeek-R1-\\Distill-Qwen-1.5B}
        \label{fig:edge_tps_deepseek}
    \end{subfigure}
    \hspace*{-3.4em}
    \begin{subfigure}[t]{0.3\textwidth}
        \centering
        \input{result/edge/tps/edge_tps_gpt}   
        \caption{gpt-oss-20b}
        \label{fig:edge_tps_gpt}
    \end{subfigure}

    \caption{\news{ Total tokens per second on edge devices with varying concurrency (Prompt length = 128, Output length = 512) }}
    \label{fig:jetson_tps}

\end{figure*}

\news{\textbf{End-to-End Latency.} Request latency trends closely follow the TTFT and TBT characteristics described earlier. In Fig.~\ref{fig:jetson_latency} ,for Llama-3.1-8B, the average response latency of Ollama was 7.2 s, 13.7 s, and 24.8 s at concurrency levels 1, 2, and 4, respectively, whereas LLaMA.cpp recorded 18.3 s, 36.4 s, and 72.7 s, showing that latency nearly doubled whenever concurrency doubled. Smaller models such as Qwen3-0.6B exhibited notably shorter delays, with LLaMA.cpp measuring 4.4 s, 8.7 s, and 17.5 s under the same conditions. In contrast, mid- to large-scale models such as Qwen3-8B and Qwen3-14B experienced sharp latency increases, reaching 76-83 s and 134-142 s, respectively, at concurrency 4. For gpt-oss-20B, Ollama reported 18.9 s, 35.3 s, and 69.1 s, while LLaMA.cpp recorded 14.2 s, 28.4 s, and 56.7 s. These results indicate that serving large models under high concurrency on Jetson Orin is effectively infeasible. In practical edge deployments, concurrency must be limited to 1-2, and model size realistically restricted to the 1B-4B range to maintain acceptable latency. }

\begin{figure*}[tbp]
    \centering
    \begin{subfigure}[]{0.3\textwidth}
        \centering
        \vspace*{-11.35mm}
        \input{result/edge/latency/edge_latency_llama3}   
        \vspace*{-6.5mm}
        \caption{ Meta-Llama-3.1-8B}
        \label{fig:edge_latency_llama3}
    \end{subfigure}%
    \hspace*{-3.15em}
    \begin{subfigure}[t]{0.3\textwidth}
        \centering
         \input{result/edge/latency/edge_latency_qwen3_0.6b}   
        \caption{ Qwen3-0.6B}
        \label{fig:edge_latency_qwen3_0.6b}
    \end{subfigure}%
    \hspace*{-3.2em}
    \begin{subfigure}[t]{0.3\textwidth}
        \centering
        \input{result/edge/latency/edge_latency_qwen3_8b}   
        \caption{Qwen3-8B}
        \label{fig:edge_latency_qwen3_8b}
    \end{subfigure}%
    \hspace*{-3.15em}
    \begin{subfigure}[t]{0.3\textwidth}
        \centering
        \input{result/edge/latency/edge_latency_qwen3_8b}   
        \caption{Qwen3-14B}
        \label{fig:edge_latency_qwen3_14b}
    \end{subfigure}%

    \vspace{0.15cm}  
    
    \hspace*{-0.95em}
    \begin{subfigure}[t]{0.3\textwidth}
        \centering
        \vspace*{-19.35mm}
        \input{result/edge/latency/edge_latency_gemma3_1b}   
        \vspace*{-2.1mm}
        \caption{\centering  gemma-3-1b-it}
        \label{fig:edge_latency_gemma3_1b}
    \end{subfigure}
    \hspace*{-3.0em}
    \begin{subfigure}[t]{0.3\textwidth}
        \centering
        \input{result/edge/latency/edge_latency_gemma3_4b}   
        \caption{\centering  gemma-3-4b-it}
        \label{fig:edge_latency_gemma3_4b}
    \end{subfigure}
    \hspace*{-3.45em}
    \begin{subfigure}[t]{0.3\textwidth}
        \centering
        \input{result/edge/latency/edge_latency_deepseek}   
        \caption{\centering DeepSeek-R1-\\Distill-Qwen-1.5B}
        \label{fig:edge_latency_deepseek}
    \end{subfigure}
    \hspace*{-3.45em}
    \begin{subfigure}[t]{0.3\textwidth}
        \centering
        \input{result/edge/latency/edge_latency_gpt}   
        \caption{gpt-oss-20b}
        \label{fig:edge_latency_gpt}
    \end{subfigure}

    \caption{\news{ Request Latency on edge devices with varying concurrency (Prompt length = 128, Output length = 512) }}
    \label{fig:jetson_latency}

\end{figure*}

\news{\textbf{Capacity under increasing concurrency.} The request latency measured on the Jetson Orin closely reflects the previously observed TTFT and TBT characteristics. As shown in Fig.~\ref{fig:edge_req_quant_request}, for Llama-3 8B, Ollama recorded average latencies of 7.2 s, 13.7 s, and 24.8 s at concurrency levels 1, 2, and 4, respectively. In contrast, LLaMA.cpp showed 18.3 s, 36.4 s, and 72.7 s, with latency nearly doubling each time concurrency doubled. For the small-scale Qwen3 0.6B, both engines generally exhibited short delays. LLaMA.cpp measured 4.4 s, 8.7 s, and 17.5 s at concurrency 1, 2, and 4, with Ollama showing similar values. However, as model size increases, latency grows sharply. At concurrency 4, Qwen3 8B and 14B reached average latencies of 76-83 s and 134-142 s, respectively. For gpt-oss-20B, Ollama recorded 18.9 s, 35.3 s, and 69.1 s, while LLaMA.cpp measured 14.2 s, 28.4 s, and 56.7 s across concurrency levels 1, 2, and 4-indicating that a single request can take from one to nearly two minutes to complete at concurrency 4. Overall, serving large models under high concurrency on the Jetson Orin is practically infeasible. The realistic operational range is limited to concurrency levels 1-2 with 1B-4B scale models, which offer manageable latency and consistent responsiveness.}

\begin{figure*}[tbp]
    \centering    
    \begin{subfigure}[t]{0.29\textwidth}
        \centering
        \input{result/edge/request/edge_request_llama3}  
        \vspace*{-4.2mm}
        \caption{Meta-Llama-3.1-8B}
        \label{fig:edge_req_quant_llama3}
    \end{subfigure}%
    \hspace*{-2.7em}
    \begin{subfigure}[t]{0.29\textwidth}
        \centering
        \input{result/edge/request/edge_request_qwen3_0.6b}  
        \caption{Qwen3-0.6B}
        \label{fig:edge_req_quant_qwen3_0.6b}
    \end{subfigure}%
    \hspace*{-2.7em}
    \begin{subfigure}[t]{0.29\textwidth}
        \centering
        \input{result/edge/request/edge_request_qwen3_8b}  
        \caption{Qwen3-8B}
        \label{fig:edge_req_quant_qwen3_8b}
    \end{subfigure}
    \hspace*{-2.7em}
    \begin{subfigure}[t]{0.29\textwidth}
        \centering
        \input{result/edge/request/edge_request_qwen3_14b}  
        \caption{Qwen3-14B}
        \label{fig:edge_req_quant_qwen3_14b}
    \end{subfigure}
    
    \vspace{0.15cm}  

    \hspace*{-1.0em}
    \begin{subfigure}[t]{0.29\textwidth}
        \centering
        \input{result/edge/request/edge_request_gemma3_1b}  
        \caption{gemma-3-1b-it }
        \label{fig:edge_req_quant_gemma3_1b}
    \end{subfigure}
    \hspace*{-2.4em}
    \begin{subfigure}[t]{0.29\textwidth}
        \centering
        \input{result/edge/request/edge_request_gemma3_4b}  
        \caption{gemma-3-4b-it}
        \label{fig:edge_req_quant_gemma3_4b}
    \end{subfigure}
    \hspace*{-2.9em}
    \begin{subfigure}[t]{0.29\textwidth}
        \centering
        \input{result/edge/request/edge_request_deepseek}  
        \caption{\centering DeepSeek-R1-\\Distill-Qwen-1.5B}
        \label{fig:edge_req_quant_deepseek}
    \end{subfigure}
    \hspace*{-2.8em}
    \begin{subfigure}[t]{0.29\textwidth}
        \centering
        \input{result/edge/request/edge_request_gpt}  
        \caption{gpt-oss-20b }
        \label{fig:edge_req_quant_gpt}
    \end{subfigure}
    
    \caption{\news{Impact of concurrency scaling on inference reliability on edge device (4bit Quantized models, Prompt length = 128, Output length = 512)}}
    \label{fig:edge_req_quant_request}
\end{figure*}

\begin{findingbox}[]
   {\small \textbf{\news{Key Takeaways}}}
    \begin{itemize}[leftmargin=1.2em, label=--, itemsep=0.6em]

        \item \news{\textbf{Ollama excels for 4--8B models on edge devices:}  
        On Jetson Orin AGX 32GB, Ollama consistently outperformed LLaMA.cpp for models such as Llama-3.1-8B and Gemma-3 4B,  
        achieving lower TTFT, higher request/token throughput, and lower latency—making it well suited for interactive assistant-style workloads.}

        \item \news{\textbf{LLaMA.cpp performs better for sub-2B models:}  
        For smaller models including Qwen-3 0.6B, Gemma-3 1B, DeepSeek R1 1.5B, and gpt-oss-20B,  
        LLaMA.cpp delivered lower TTFT and higher token throughput.  
        Its latency also scaled more smoothly up to concurrency 4, favoring batch-oriented, resource-efficient edge deployments.}

        \item \news{\textbf{Large models become impractical on Jetson hardware:}  
        Once model size exceeds 7B, both engines exhibit TTFTs in the tens of seconds,  
        while end-to-end latency exceeds one minute at concurrency 4.  
        Success rates also decline to the 60--80\% range, indicating severe performance constraints.}

        \item \news{\textbf{Implication for edge deployment:}  
        Effective edge operation requires small-model-centric design and low concurrency.  
        Large-scale models remain impractical in constrained edge environments due to latency, throughput, and stability limitations.}

    \end{itemize}
\end{findingbox}



\section{Future Directions and Open Challenges} \label{sec:future_direction}

LLM inference engines continue to support new optimization methods and models, but to flexibly respond to the rapidly changing LLM ecosystem and diverse service requirements, the following additional considerations are necessary.

\subsection{\news{Extended Context Windows and Memory Management}} \label{sec:future_direction_context_window}

There is a growing trend in LLMs toward handling extremely long context windows, ranging from tens of thousands to millions of tokens. For example, ChatGPT o1~\cite{chatgpt} supports up to 128K tokens, Google Gemini 2.0 Flash~\cite{gemini} supports 1M tokens, and the recently introduced Llama 4 Scout~\cite{llama4} claims to handle up to 10M tokens.

This expansion leads to a dramatic increase in the KV cache size, posing significant challenges for memory management. To address this, techniques such as hierarchical cache management~\cite{zhao2024alisa}, partial offloading to CPU memory~\cite{lee2024infinigen}, and memory-efficient attention mechanisms~\cite{kwon2023efficient} have been proposed. However, these methods are not yet sufficient to fully cope with the increasing context length, and further research is required.

\news{
To optimally handle long-context scenarios, multiple inference optimization techniques must be applied in combination. For example, PagedAttention~\cite{kwon2023efficient} manages the $\mathbf{KV}$ cache to reduce internal fragmentation, while chunked prefill~\cite{ye2024chunkattention} divides long prompts into chunks to mitigate TTFT without sacrificing throughput. In addition, speculative decoding~\cite{li2024eagle,cai2024medusa} accelerates generation by verifying and accepting tokens proposed in advance by a draft model, thereby avoiding rollbacks.
}
\news{
Efforts to compress the context itself are also actively explored. Selective-Context~\cite{li2023compressing} reduces input length by removing redundancy or irrelevant information within the prompt, but this approach risks information loss and semantic distortion. To overcome these limitations, LLMLingua~\cite{jiang2023llmlingua} employs a coarse-to-fine compression scheme. It incorporates a budget controller to adjust the target compression ratio, applies a token-level bidirectional compression algorithm to preserve interdependencies across compressed segments, and aligns semantic fidelity through instruction-tuning-based distribution matching. Experiments have demonstrated up to a 20$\times$ compression ratio without performance degradation.}

\news{
In real-world services, multi-turn dialogues and streaming generation essentially require handling inputs of unbounded length~\cite{gao2024cost}. However, tokens that exceed the context window not only degrade model quality but also cause the $\mathbf{KV}$ cache to grow explosively. Traditional sliding-window methods~\cite{beltagy2020longformer} retain only the most recent \texttt{L} tokens while recomputing the $\mathbf{KV}$ of earlier tokens, but this incurs substantial overhead. StreamingLLM~\cite{xiao2023efficient} addresses this by fixing the $\mathbf{KV}$ of the initial tokens, maintaining only the most recent \texttt{L} tokens in a rolling cache, and discarding the rest. Combined with relative position encoding methods such as RoPE~\cite{su2024roformer} or ALiBi~\cite{press2021train}, this design enables infinite-length inputs without additional training and achieves up to a 22.2$\times$ speedup compared to conventional sliding-window approaches. Nevertheless, since the context window size is not actually extended during training, its performance gains are limited for tasks requiring long-range dependencies, such as document summarization or long-term memory-based queries.}

\news{
In the case of vLLM~\cite{kwon2023efficient}, when the input sequence length exceeds the maximum position embedding length supported by the model, the prompt is automatically split into multiple chunks. Each chunk is converted into an embedding vector using the model's original pooling strategy (either CLS token~\cite{kim2021self} or mean pooling).
In the final step, chunk embeddings are aggregated into a single representation by computing their weighted average according to the number of tokens per chunk. This approach is simple to implement and computationally efficient, but its reliance on averaging limits the ability to capture cross-chunk contextual interactions.}


\awesomebox[violet]{2pt}{\faRocket}{violet}{
\textbf{Keywords}
\news{Long-context processing, Memory-efficient attention, KV cache optimization, 
Context compression, Retrieval-based attention}

\textbf{Solutions}

\news{PagedAttention~\cite{kwon2023efficient} and chunked prefill~\cite{ye2024chunkattention} optimize KV cache usage and TTFT, while speculative decoding~\cite{li2024eagle,cai2024medusa} speeds up generation. LLMLingua~\cite{jiang2023llmlingua} compresses context effectively, and RetrievalAttention~\cite{liu2024retrievalattention} or MoBA~\cite{lu2025moba} adaptively balance sparsity and accuracy for long-context processing.}
}


\subsection{\news{Complex Logical Reasoning}} \label{sec:future_direction_reasoning}

Modern LLMs are evolving beyond simple response generation toward performing complex logical reasoning. This includes guiding users step by step through problem solving processes, autonomously generating CoT reasoning~\cite{wei2022chain}, and interacting with external tools to complete tasks.

\news{CoT prompting has become a key technique for improving accuracy and interpretability in complex mathematical, logic, and code problems, generalizing structured reasoning such as problem decomposition, proof generation, and verification to language agents~\cite{zhang2025igniting}. In such scenarios, a large number of tokens may be consumed during intermediate reasoning steps, and multi-turn dialogues are often required to refine answers.}

\news{CoT induces long reasoning chains, which significantly increase the number of tokens in the decode phase, thereby causing quasi-linear growth in FLOPs and memory access. Since each token requires access to the KV cache, cache capacity and bandwidth bottlenecks become more severe.}

\news{To mitigate this, methods have been proposed to keep Keys in low-rank representations while offloading Values and dynamically reconstructing sparse $\mathbf{KV}$ caches as needed~\cite{sun2024shadowkv}. Additionally, long CoT requests can cause head-of-line blocking in the queue, worsening the TTFT for shorter interactive requests. This interference can be reduced by separating prefill and decode phases and batching them across heterogeneous devices~\cite{patel2024splitwise}.}

\news{An important thing to note is that overly verbose CoT outputs can lead to bloated responses rather than higher quality~\cite{nayab2024concise}. Therefore, it is necessary to optimize prompts using metrics such as correct-conciseness, and to design reward functions that encourage models to suppress unnecessary tokens autonomously.}

Along with it, managing session continuity and context preservation becomes critical. In response, inference engines are being developed with support for streaming output~\cite{xiao2023efficient} and multi-turn dialogue optimization~\cite{agarwal2024symphony, gao2024cost}. The ability to stably manage long token sequences and complex reasoning flows is becoming a key competency for LLM inference systems.

\awesomebox[violet]{2pt}{\faRocket}{violet}{
\textbf{Keywords}
\news{Chain-of-Thought (CoT), Logical Reasoning, KV Cache Optimization, 
Efficient Decoding, Reward Optimization}

\textbf{Solutions}

\news{Low-rank and sparse KV caching~\cite{sun2024shadowkv} and phase-splitting~\cite{patel2024splitwise} reduce reasoning latency, while conciseness-aware reward modeling~\cite{nayab2024concise} improves CoT efficiency. Combined with streaming~\cite{xiao2023efficient} and session management~\cite{agarwal2024symphony,gao2024cost}, these enable stable long reasoning flows.}
}

\subsection{\news{Inference Engine Selection Based on Application Needs}} \label{sec:future_direction_engine}

The selection and design of LLM inference engines should be based on a balance between application requirements and system constraints. For applications like translation services or conversational agents where real-time interaction is critical, latency optimization is the top priority. On the other hand, server-side applications that must handle high-volume traffic will prioritize throughput maximization.

Looking ahead, it will be increasingly important to develop inference engines that support both hardware acceleration for multimodal data and general-purpose compatibility with diverse model architectures. Persistent challenges include optimizing memory for extended context windows, designing architectures that can flexibly handle complex reasoning, and developing strategies that strike the right balance between latency and throughput.

\news{Meeting the accuracy and latency requirements demanded by applications and services cannot be achieved solely by enhancing the structure of inference engines, the model architecture itself must also be optimized. Compared to traditional CNNs or DNNs, LLMs exhibit a lower computational density relative to their parameter scale~\cite{moar2024characterizing}. As a result, inefficiencies in computation and memory arise that cannot be fully resolved by quantization or pruning alone. Thus, low-rank decomposition has emerged as a key complementary technique~\cite{kim2025efficient}.}

\news{Low-rank decomposition~\cite{kaushal2023lord,jaiswal2024galore,yuan2023asvd} reconstructs the weight matrices or tensors into lower-dimensional components, thereby reducing both memory usage and computational cost. A representative approach applies singular value decomposition (SVD)~\cite{wang2024svd,yuan2023asvd} to retain only dominant singular values as an approximation, while another approach constrains rank during training via non-convex regularization~\cite{qin2024nonconvex}. In addition, tensor decomposition methods such as Tensor Train (TT)~\cite{qin2025computational}, Tensor Ring (TR)~\cite{he2023scalable}, and Tucker~\cite{koike2025latentllm} represent high-order tensors as products of smaller cores. Within Transformer-based LLMs, applying these decompositions to attention weights and FFN weights can substantially reduce both computation and memory demands.}

\news{The low-rank decomposition can be applied in two primary stages. The first is during pre-training, where the layers are parameterized in a low-rank form so that the model maintains a low-rank structure throughout training~\cite{qin2025computational,li2025lost,makni2025a}. The second is post-training decomposition applied to a pre-trained model, where SVD or tensor decomposition is used to adjust the rank of each layer and balance the trade-off between latency and accuracy. For example, CALDERA~\cite{saha2024compressing} progressively applies decomposition across multiple stages, yielding more stable results than one-shot compression~\cite{kaushal2023lord}.}

\news{The impact of low-rank decomposition on model performance depends on the model architecture, data, and chosen rank value. Studies on the accuracy-efficiency trade-off show that applying Tucker decomposition to the Llama 2 family reduces model size by approximately 9\% while maintaining precision losses within 4.5 to 10\% points, thereby lowering both latency and energy consumption~\cite{moar2024characterizing}.}

\news{To effectively apply low-rank decomposition in real-world inference services, the decomposition scheme and kernel implementation must be co-designed in accordance with the memory hierarchy and computational units of the target hardware. First, when determining the dimension along which matrices or tensors are decomposed, the rank value must be selected with consideration of micro-level resource constraints such as GPU/NPU warp size, memory bank organization, and shared memory capacity. Second, although the decomposed low-dimensional matrix multiplications are computationally lighter, the increased number of calls may lead to excessive kernel launch overhead and frequent accesses to global memory. To mitigate this, multiple multiplications should be fused within a single kernel or repacked into tensor-core-friendly block structures arranged within contiguous memory ranges. Finally, the execution scheduler should analyze the computation graph before and after decomposition and reorder operations so that parts with low data reuse are retained in shared memory or registers, thereby alleviating memory bandwidth bottlenecks. In summary, optimal low-rank decomposition yields maximum benefit only when the decomposition algorithm, kernel fusion strategy, and scheduling policy are integrated with hardware characteristics.}

\news{Thus, a low-rank decomposition at the model level complements inference engine optimizations. For instance, if inference engines supporting post-training quantization, such as Unsloth~\cite{unsloth}, additionally provide low-rank decomposition modules, LLMs could be executed more efficiently even on personal devices or edge hardware.}


\awesomebox[violet]{2pt}{\faRocket}{violet}{
\textbf{Keywords}
\news{Inference Engine Optimization, Application-driven Design, 
Low-rank Decomposition, Latency-Throughput Trade-off, 
Hardware-aware Kernel Co-design}

\textbf{Solutions}

\news{Low-rank decompositions such as SVD, TT, and Tucker~\cite{wang2024svd,qin2025computational,he2023scalable,koike2025latentllm} reduce model cost with minimal accuracy loss. Progressive compression~\cite{saha2024compressing} and kernel fusion tuned to hardware improve throughput, while integration with quantization engines like Unsloth~\cite{unsloth} enhances edge efficiency.}
}


\subsection{\news{Increasing Importance of LLM Alignment}} \label{sec:future_direction_engine}

\news{As LLMs spread across various domains and services, LLM alignment~\cite{ji2023ai} has become as important as the accuracy of the model. Alignment guides the outputs toward responses that users prefer and that meet policy or style rules. Fine-tuning usually improves prediction accuracy on tasks with ground truth, while alignment pursues broader goals such as usefulness, policy compliance, and domain tone. Thus, fine-tuning helps answer oriented tasks, and alignment covers those tasks while adding higher-level constraints.}

\news{Several alignment techniques have been widely studied. First, after Supervised Fine-Tuning (SFT)~\cite{shengyu2023instruction}, Reinforcement Learning with Human Feedback (RLHF)~\cite{zheng2023secrets, dai2023safe} trains a reward model~\cite{wang2024secrets} using human-preference pairwise data and then optimizes the policy using Proximal Policy Optimization (PPO)~\cite{xu2024dpo}. Second, RL from AI Feedback (RLAIF)~\cite{lee2023rlaif}/Constitutional AI~\cite{weyssow2024codeultrafeedback, li2024llms} that replaces human feedback with an LLM-based judge to generate preference data and perform alignment accordingly. Third, Direct Preference Optimization (DPO)~\cite{xu2024dpo}, which directly optimizes the policy from preference pairs without a separate reward model or PPO, has recently gained significant attention.}

\news{Aligning LLMs, which often have billions or even trillions of parameters, requires a downstream phase after supervised or instruction fine-tuning. This alignment phase uses preference-based methods such as RLHF, RLAIF/Constitutional AI with AI feedback, or DPO to embed user or organizational policies. Because the process requires large amounts of hardware, several large-scale alignment frameworks have appeared.}

\news{Verl~\cite{sheng2025hybridflow} combines PPO-based RLHF, DPO, and AI feedback in a single pipeline after fine-tuning and supports parallel training at scale. LlamaRL~\cite{wu2025llamarl} is a light alignment framework tailored to LLaMA models that allows researchers to run quick tests on small GPU configurations. LlamaRL is less suitable for very large distributions, memory or communication optimization, and flexible pipelines. Other tools, including Transformer Reinforcement Learning (TRL)~\cite{trl}, OpenRLHF~\cite{hu2024openrlhf}, and DeepSpeed-Chat~\cite{yao2023deepspeed}, also support alignment training.}

\news{LLM Alignment remains crucial at inference time. An aligned model better matches user intent and policy, reducing post-processing and retries. It also produces outputs with more stable formats and lengths, easing batch processing and scheduling. Even so, alignment does not change the parameter count, so inference engines still need techniques such as quantization, caching, and careful batch scheduling to meet real-time service goals.}


\awesomebox[violet]{2pt}{\faRocket}{violet}{
\textbf{Keywords}
\news{LLM Alignment, RLHF, RLAIF, DPO, Preference Optimization, 
Human/AI Feedback, Policy Compliance}

\textbf{Solutions}

\news{RLHF~\cite{zheng2023secrets,dai2023safe}, RLAIF~\cite{lee2023rlaif}, and DPO~\cite{xu2024dpo} align models with user preferences and policies. Frameworks like Verl~\cite{sheng2025hybridflow}, LlamaRL~\cite{wu2025llamarl}, and DeepSpeed-Chat~\cite{yao2023deepspeed} scale alignment efficiently, ensuring consistent, compliant outputs at inference time.}
}

\subsection{Hardware-Aware Fusion and Mixed-Precision Kernels for Efficiency} \label{sec:future_direction_hardware}

In traditional AI workloads that primarily relied on convolution operations, optimization was often achieved through simple operator fusions, such as ReLU. However, in the era of generative AI based on Transformers and diffusion models, more sophisticated fusion strategies tailored to specific hardware architectures are required. A representative example is FlashAttention-3~\cite{shah2024flashattention}, which is highly optimized for NVIDIA's H100 hardware. These complex fusion techniques involve advanced tiling strategies, carefully designed by expert engineers to align with the hardware's shared memory and cache size constraints.

\news{Low-precision microscaling data types, such as MXFP4, can accelerate GEMM operations while simultaneously reducing both training and inference costs. The NVIDIA Blackwell architecture natively supports multiple low-precision formats, including FP4, MXFP4~\cite{rouhani2023microscaling}, and NVFP4~\cite{chmiel2025fp4}. Since the training process of LLMs also relies heavily on large-scale matrix multiplications, a recent study has proposed leveraging MXFP4 to perform training without accuracy degradation~\cite{tseng2025training}. This study computes low-bias, low-variance gradient estimates, enabling more accurate parameter updates, and applies the Random Hadamard transform to mitigate the impact of rare outliers on the overall training process. As a result, when pretraining a 6.7B-parameter GPT model, they achieved an accuracy virtually identical to that obtained using BF16 mixed-precision.}

\news{OpenAI's open-source model gpt-oss~\cite{agarwal2025gpt} is an autoregressive MoE Transformer based on the GPT-2~\cite{radford2019language} and GPT-3~\cite{brown2020language} architectures, supporting both CoT reasoning and structured output generation. It employs RoPE for positional encoding and can process up to 128K tokens of context. This model reduces memory requirements substantially by quantizing MoE weights into the MXFP4 format through post-training quantization, storing them at approximately 4.25 bits per parameter.}

\news{With the growing adoption of MoE and the continued expansion of LLM scales, the use of microscaling data types is expected to become increasingly important. Currently, vLLM~\cite{kwon2023efficient} provides limited support for MXFP4, and effectively deploying such ultra-low-precision formats in LLM services requires kernel-level optimizations within inference engines that are deeply aligned with hardware architectures.}

Moreover, as shown in Table~\ref{tab:frameworks_data_type}, generative AI models demand support for a wide range of data type precisions to reduce model size while preserving accuracy. Therefore, it is essential to develop operator kernels that can flexibly and efficiently handle mixed-precision computation.

\awesomebox[violet]{2pt}{\faRocket}{violet}{
\textbf{Keywords}
\news{Hardware-aware Fusion, Mixed-Precision Kernels, Microscaling Data Types, 
FP4/MXFP4/NVFP4, FlashAttention, Low-Precision Optimization}

\textbf{Solutions}
%
\news{FlashAttention-3~\cite{shah2024flashattention} maximizes GPU memory efficiency, while MXFP4 and NVFP4~\cite{rouhani2023microscaling,chmiel2025fp4} enable faster, smaller models. MXFP4 training~\cite{tseng2025training} maintains BF16-level accuracy, and MoE quantization~\cite{agarwal2025gpt} further reduces cost. Hardware-aligned kernels in engines like vLLM~\cite{kwon2023efficient} are key for FP4 deployment.}
}

\subsection{\news{Support for On-Device Inference}} \label{sec:future_direction_on_device}

Most LLM services perform inference using large-scale resources in cloud or data center environments, delivering the results to users. Although this approach enables fast computation, it is network-dependent and requires transmitting user data to servers, raising privacy concerns. As a result, demand for on-device (or on-premise) LLM inference on edge and mobile devices has been increasing.

Traditionally, LLM models were too large to run on a single device, but the emergence of small language models (SLMs) such as Llama 3.2~\cite{grattafiori2024llama}, Gemma~\cite{team2024gemma}, Phi-3~\cite{abdin2024phi}, and Pythia~\cite{biderman2023pythia} has enabled LLM execution on embedded systems, mobile devices, and IoT systems, as well as single-GPU environments.

Since edge environments (e.g., embedded and mobile systems) have lower hardware specifications than servers, both model compression and hardware-specific parallelization and memory optimizations are essential. For example, mobile inference optimizations include tolerance-aware compression, I/O recomputation pipeline loading, and chunk life cycle management~\cite{yin2024llm}. Research has also explored collaborative inference, where multiple edge devices share computational workloads~\cite{zhang2024edgeshard}. For single GPU environments, 4-bit quantization and memory offloading techniques that distribute weights, activations, and $\mathbf{KV}$ caches across CPU, disk, and other memory resources~\cite{sheng2023flexgen} are being investigated. These advances help reduce server power consumption and enable personalized models and inference in network-limited environments. However, edge devices vary significantly by manufacturer and environment, making generalized optimization difficult. Additionally, developing compiler and hardware-dependent transformation tools incurs additional development costs.

\news{
In on-device inference environments, achieving efficiency requires not only optimization techniques and supporting inference engines but also lightweight yet high-performing models. Several models such as DistilBERT~\cite{sanh2019distilbert}, the DeepSeek-R1-Distill family~\cite{guo2025deepseek}, and DistilQwen-2.5~\cite{wang2025distilqwen2} have been released, which use KD~\cite{yang2024survey} to reduce the number of parameters while maintaining accuracy close to that of the original LLM.}

\news{KD is a representative model compression technique in which a large teacher model transfers its trained knowledge to a smaller student model. Through this process, smaller models can be trained to approximate the performance of larger models, thus achieving strong accuracy with a relatively compact size. This approach enables straightforward training of domain-specific models suited for on-device environments and helps narrow the performance gap between proprietary large-scale models and open-source models~\cite{xu2024survey}.}

\news{There are various ways to extract knowledge from teacher models: using teacher output labels directly as supervision (\texttt{Labeling}), generating additional input-output pairs from a few examples (\texttt{Expansion}), transferring probability distributions or intermediate representations (\texttt{Feature}), synthesizing data from external meta-information (\texttt{Data curation}), having the teacher provide feedback on student outputs (\texttt{Feedback}), or adopting self-knowledge methods where the student filters and refines its own outputs (\texttt{Self-Knowledge})~\cite{xu2024survey}. To transfer such knowledge into student models, fine-tuning can be employed with supervised or semi-supervised learning~\cite{meng2025collaborative,agarwal2024policy}, divergence-based loss for distribution alignment~\cite{gu2023minillm,yang2025feature}, or reinforcement learning-based policy optimization~\cite{luo2023wizardmath, liang2024yes} can be employed.}

\news{
When internal weights, logits, or attention values of the teacher model are accessible, white-box distillation~\cite{shum2024first,liu2024multi} can be applied to achieve fine-grained distribution matching and structural preservation. In this case, it is easier to bring student performance close to that of the teacher, though transfer efficiency may degrade when there is a large-scale gap between the teacher and student. In contrast, in black-box environments, such as commercial APIs where internal states are inaccessible, distillation is based solely on the final output~\cite{yang2024distillseq,wang2024don}. Sequence-level imitation learning and self-training methods fall into this category, but the absence of logits or intermediate representations often limits the performance gain.}

\news{KD can be applied either during the fine-tuning stage of a pre-trained model or throughout the entire pre-training process. For inference engines to support KD, they must be able to handle loss computation and backpropagation within the training loop. Alternatively, lightweight distillation techniques, such as zero-shot or few-shot prompt-based Self-Instruct distillation or prompt-supervised generation, rely only on the teacher's output. These methods enable the creation of student models without requiring additional computation, even in engines that do not support training functionalities.}

Among existing LLM inference engines, llama.cpp~\cite{llamacpp}, MLC LLM~\cite{mlcllm}, and TensorRT-LLM~\cite{tensorrtllm} offer partial support for edge environments. llama.cpp~\cite{llamacpp}, implemented in C/C++, is highly portable across different platforms. MLC LLM~\cite{mlcllm} uses the TVM~\cite{chen2018tvm} compiler to support GPU, mobile, and web environments, although its hardware compatibility is limited. TensorRT-LLM~\cite{tensorrtllm} supports only specific edge devices, such as NVIDIA Jetson series.

\awesomebox[violet]{2pt}{\faRocket}{violet}{
\textbf{Keywords} 
\news{On-device LLM Inference, Edge/Mobild device optimization, Model Compression}

\textbf{Solutions}
\news{Small language models (SLMs) (Llama 3.2~\cite{grattafiori2024llama}, Gemma ~\cite{team2024gemma}, Phi-3 ~\cite{abdin2024phi}), tolerance-aware compression, I/O recomputation, and chunk life cycle management~\cite{yin2024llm}, collaborative inference~\cite{zhang2024edgeshard}, knowledge distillation~\cite{yang2024survey}}
}

\subsection{Support Diverse Hardware for Inference Optimization} \label{sec:future_direction_optimization}

LLM inference, which was traditionally centered around NVIDIA GPUs, is now expanding to heterogeneous hardware with the emergence of TPU~\cite{jouppi2023tpu}, Neural Processing Units (NPUs), and various LLM accelerators. In addition to widely used AWS Inferentia~\cite{inferentia} and Google TPU~\cite{jouppi2023tpu}, new accelerators such as AMD Instinct MI300X~\cite{smith2024amd}, Furiosa AI RNGD (Tensor Contraction Processor)~\cite{kim2024tcp}, and Cerebras CS-2 (WSE-2)~\cite{lie2024inside} are being developed. Furthermore, next-generation memory technologies such as Processing-in-Memory (PIM)~\cite{ortega2024pim, kim2024breakthrough} are also under development.

To support heterogeneous hardware, engines must incorporate pipeline execution, batch optimization, and load balancing. However, differences in performance, synchronization, and communication overhead across hardware types can pose challenges. Research has explored LLM inference optimizations in heterogeneous GPU clusters using Phase-Aware Partitioning and Adaptive Quantization~\cite{zhao2024llm} and hardware-aware allocation of prefill and decode processes for optimized inference~\cite{patel2024splitwise}. Furthermore, techniques such as sub-batch interleaving~\cite{heo2024neupims} have been proposed to optimize inference across systems with multiple NPUs and PIM devices.

\news{To optimize LLM inference on accelerators beyond GPUs, studies~\cite{li2024large, guo2025survey, kachris2025survey} have also analyzed the characteristics of CPUs, GPUs, FPGAs, ASICs, and PIM or NDP platforms. Especially in~\cite{li2024large}, by synthesizing previous work, the reasearch summarized hardware-specific optimization strategies for both the prefill and decode phases and compared throughput (tokens per second) relative to power consumption. In particular, they examined in detail how optimizations such as quantization, sparsity, and fast decoding (e.g., speculative decoding) exhibit different performance behaviors across hardware depending on batch size. Based on the findings, the research paper argued that addressing recent inference trends such as longer input sequences and expanded prefill phase requires hardware-software-algorithm co-design to simultaneously achieve real-time responsiveness, high-throughput, and resource efficiency.}

\news{Moreover, most accelerators are accompanied by dedicated compilers and frameworks to generate hardware-friendly code. For example, Google TPU~\cite{jouppi2023tpu} leverages the XLA~\cite{sabne2020xla} compiler and JAX~\cite{frostig2019compiling}, while Groq LPU~\cite{abts2020think} provides a dedicated software stack consisting of GroqWare/GroqFlow~\cite{groqflow}, and the LPU Inference Engine. Although attempts have been made to support multiple accelerators beyond NVIDIA GPUs, such as AMD MI300X~\cite{smith2024amd}, Google TPU~\cite{jouppi2023tpu}, and Huawei Ascend~\cite{liao2021ascend} by vLLM~\cite{kwon2023efficient}, many inference engines still provide official support only for a limited set of hardware. Integrating new accelerators requires coordination across engine-level optimizations, runtimes, compilers, and libraries, which often entails significant delays before official repository integration. Consequently, it has become increasingly common for hardware vendors to adapt existing engines and provide wrappers or bindings tailored to their accelerators. Realizing broad hardware support thus necessitates these preparatory steps.}

\awesomebox[violet]{2pt}{\faRocket}{violet}{
\textbf{Keywords} 
\news{Heterogeneous Hardware, LLM Accelerator, Hardware-Aware Optimization, Compiler and Runtime Integration}

\textbf{Solutions}
\news{Phase-Aware Partitioning and Adaptive Quantization~\cite{zhao2024llm}, hardware-aware allocation of prefill and decode processes~\cite{patel2024splitwise}, sub-batch interleaving for multi-NPU and PIM inference~\cite{heo2024neupims}, hardware-specific optimization across CPUs, GPUs, FPGAs, ASICs, and PIM/NDP platforms~\cite{li2024large, guo2025survey, kachris2025survey}}
}

\subsection{Requirements for Multimodal LLMs} \label{sec:future_direction_multimodal}

Most current LLM inference engines are optimized for text-based models. However, relying solely on text has limitations when processing information. To achieve human-level intelligence, it is essential to support multiple data types such as images, audio, and video. In response to this, multimodal models like Qwen2-VL~\cite{wang2024qwen2} and LLaVA-1.5~\cite{liu2024improved} have been developed. To support such models effectively, inference engines must be designed to handle multimodal data preprocessing and multi-stream parallel execution efficiently.

In this context, existing model compression techniques, such as quantization must be adapted to preserve modality-specific information while still reducing model size. Software-level methods like hybrid parallelization are not sufficient on their own. Therefore, new hardware-accelerated kernels and decoding methods, such as speculative decoding tailored for multimodal inputs, need to be considered.

A good example of adapting model architecture to hardware for multimodal tasks is Multimodal Rotary Position Embedding (M-RoPE), introduced in Qwen2-VL\cite{wang2024qwen2} and LLaVA-1.5~\cite{liu2024improved}. M-RoPE extends the traditional positional embedding used in Transformer models to more effectively capture the positional relationships in various multimodal inputs.

\awesomebox[violet]{2pt}{\faRocket}{violet}{
\textbf{Keywords} 
\news{Multimodal LLM Inference, Multimodal Model Compression, Hardware-Accelerated Decoding}

\textbf{Solutions}
\news{Multimodal models (Qwen2-VL~\cite{wang2024qwen2} and LLaVA-1.5~\cite{liu2024improved}), multimodal data preprocessing and multi-stream parallel execution support in inference engines, modality-preserving quantization for model compression, hardware-accelerated kernels and speculative decoding for multimodal inputs, Multimodal Rotary Position Embedding (M-RoPE)~\cite{wang2024qwen2, liu2024improved}}
}


\subsection{Alternative Architectures Beyond Transformers} \label{sec:future_direction_architecture}

While Transformer-based models still dominate LLM inference, new architectures tailored for multimodal LLM are emerging. Models like RetNet~\cite{sun2023retentive} and RWKV~\cite{peng2023rwkv} propose alternatives to the standard Transformer design. Another notable direction is Mamba~\cite{gu2023mamba}, a sequence modeling architecture developed to overcome the limitations of Transformers. Mamba uses a Selective State Space Model (SSM) to process long sequences more efficiently, achieving linear time complexity without relying on the standard attention mechanism.

Jamba~\cite{lieber2024jamba} is a hybrid model that combines the Mamba~\cite{gu2023mamba} and Transformer architectures, aiming to take advantage of both. It also integrates a MoE strategy that increases model capacity while keeping the number of active parameters manageable during inference.

\news{IBM Granite 4.0~\cite{granite4.0} is a hybrid architecture that combines Mamba~\cite{gu2023mamba} and Transformer~\cite{vaswani2017attention} models, achieving over 70\% reduction in memory usage while maintaining performance comparable to conventional Transformer-based models. It alleviates the quadratic computational growth of traditional Transformers with respect to sequence length by leveraging Mamba's linear scaling, thereby enabling efficient processing of long contexts without compromising in-context learning capability. Because of its Mamba-based structure, positional embeddings are also unnecessary. Granite 4.0 is distributed under the Apache 2.0 open-source license and has obtained ISO 42001 certification. The model family includes Granite-4.0-H-Small with 32B parameters, Granite-4.0-H-Tiny, a 7B MoE hybrid model, Granite-4.0-H-Micro with a 3B hybrid configuration, and Granite-4.0-Micro, a pure 3B Transformer variant. It can operate across diverse hardware platforms, including AMD Instinct MI300X~\cite{smith2024amd} and Qualcomm Hexagon NPU~\cite{mahurin2023qualocmm}, supporting inference in both server and edge environments. Furthermore, Granite 4.0 is compatible with inference engines such as Ollama~\cite{ollama}, vLLM~\cite{kwon2023efficient}, and llama.cpp~\cite{llamacpp}.}

These trends highlight the growing need for general-purpose inference engines that can support diverse architectures. Future inference systems must not only be optimized for internal operations of Transformer models but also be scalable and flexible enough to support new and evolving model structures.

\awesomebox[violet]{2pt}{\faRocket}{violet}{
\textbf{Keywords} 
\news{Transformer Alternatives, Hybrid Models, MoE}

\textbf{Solutions}
\news{Selective State Space Models (SSM) (RetNet~\cite{sun2023retentive}, RWKV~\cite{peng2023rwkv}, Mamba~\cite{gu2023mamba}), hybrid architectures (Jamba~\cite{lieber2024jamba}, IBM Granite 4.0~\cite{granite4.0})}
}


\subsection{Security Support in Inference} \label{sec:future_direction_security}

During LLM inference, vulnerabilities such as prompt injection attacks, jailbreak attacks, and data leaks have emerged~\cite{yao2024survey}. Prompt injection attacks occur when an attacker manipulates inputs to override the model's system prompts or objectives. In environments handling sensitive data, such as finance and healthcare, personal data exposure risks are significant. Additionally, if malicious attacks generate abnormal or harmful data, it can severely impact user experience and system stability.

To mitigate these risks, robustness techniques, such as adversarial training~\cite{liu2020adversarial} can be applied during the model training phase. During inference, tools like OpenAI Moderation~\cite{markov2023holistic}, instruction manipulation prevention, and input sanitization methods~\cite{ning2024cheatagent, moraffah2024adversarial} can be used to block harmful or malicious inputs.

From a service security perspective, role-based access control (RBAC) and multi-factor authentication (MFA) can be implemented to prevent unauthorized access. In addtion, access tokens can be set to expire after a certain period to enhance security policies.

Currently, most LLM inference engines focus primarily on performance and do not include dedicated security features. However, they aim to reduce risks through methods such as data filtering and strengthened ethical policies.

\awesomebox[violet]{2pt}{\faRocket}{violet}{
\textbf{Keywords} 
\news{Prompt Injection, Jailbreak Attack, Data Leakage, Model Robustness, Security in LLM Inference}

\textbf{Solutions}
\news{Adversarial Training~\cite{liu2020adversarial}, OpenAI Moderation~\cite{markov2023holistic}, Instruction Manipulation Prevention, Input Sanitization~\cite{ning2024cheatagent, moraffah2024adversarial}, Role-Based Access Control (RBAC), Multi-Factor Authentication (MFA), Access Token Expiration, Data Filtering, Ethical Policy Enforcement}
}

\subsection{\news{Support for Cloud Orchestration and Multi-node Serving Platforms}} \label{sec:future_direction_orchestration}

Cloud orchestration and serving strategies are critical for LLM inference services. When deploying large-scale inference services in the cloud, orchestration platforms such as Kubernetes~\cite{burns2016borg} enable autoscaling, hardware resource monitoring (Prometheus~\cite{turnbull2018monitoring}, Grafana~\cite{chakraborty2021grafana}), and failover recovery. To facilitate this, inference engines should provide containerized environments, multi-node deployment, and load balancing tools that allow easy configuration based on service requirements and SLOs.

Most LLM inference engines offer built-in serving functionalities, but large-scale serving systems require additional workload distribution, scheduling, and autoscaling optimizations. vLLM~\cite{kwon2023efficient}, TensorRT-LLM~\cite{tensorrtllm}, DistServe~\cite{zhong2024distserve}, and Sarathi-Serve~\cite{zhong2024distserve} utilize Ray~\cite{moritz2018ray} to support distributed runtime and serving. Additionally, TensorRT-LLM~\cite{tensorrtllm} integrates with NVIDIA Triton Inference Server~\cite{tritoninferenceserver} and NVIDIA Dynamo~\cite{dynamo} for model deployment and execution, while TGI~\cite{tgi} enables model deployment via Hugging Face Spaces~\cite{huggingfacespaces}.

\news{ With the growing adoption of MoE models~\cite{cai2024survey} and multi-agent~\cite{li2024survey} inference environments, serving platforms are rapidly expanding beyond single devices or nodes to multi-device and multi-node architectures. MegaScale-Infer~\cite{zhu2025megascale}, proposed for efficient large-scale MoE model serving, addresses the problem that as model size increases, the number of experts grows and sparsity intensifies, resulting in fewer tokens being assigned to each expert within a batch and consequently lower GPU utilization of the FFN modules. This framework disaggregates the attention and FFN modules and employs a ping-pong pipeline parallelism strategy that overlaps the two computations to hide communication latency. As a result, it achieves up to 1.9$\times$ higher decoding throughput, 4.2$\times$ higher communication throughput compared to NCCL, and 68\% lower latency on an 8-node cluster of NVIDIA A100 GPUs connected via NVLink, as well as on a heterogeneous cluster consisting of NVIDIA H20 and L40S GPUs. }

\news{ In complex multi-agent systems, $\mathbf{KV}$ cache sharing between models often becomes a major bottleneck. To alleviate this, KVCOMM~\cite{ye2025kvcomm} defines offsets that allow reuse of overlapping $\mathbf{KV}$ cache segments across different prefix regions without retraining, thereby reducing prefill latency. Another research, Cache-to-Cache~\cite{fu2025cache}, projections, fuses $\mathbf{KV}$ caches between source and target models, significantly reducing inter-model data transfer overhead. }

\news{ In multi-node environments scaled across tens to thousands of GPUs, low-latency inter-node communication is essential for efficiently handling user requests. To address this requirement, the collective communication framework NCCLX~\cite{si2025collective} has been proposed as an enhancement to NCCL. NCCLX introduces a host-based customizable transport layer (CTran) that enables zero-copy, SM-free transmission, and fault-tolerant All-Reduce operations. As a result, NCCLX achieves up to 2.7$\times$ higher throughput compared to NCCL's copy-based communication and reduces latency by up to 1.57$\times$ in tensor parallel workloads, which are critical for inference. Furthermore, in multi-node inference settings, NCCLX introduces the AllToAllvDynamic operation as a replacement for traditional AlltoAll computation, yielding up to 43\% performance improvement. }

\news{ As LLM inference environments continue to evolve toward multi-node and heterogeneous device architectures, existing inference engines must incorporate capabilities such as distributed expert placement, cache-sharing optimization, and scalable communication layers to meet the demands of real-world service deployment.}


\awesomebox[violet]{2pt}{\faRocket}{violet}{
\textbf{Keywords} 
\news{Cloud Orchestration, Multi-node/-agent LLM Serving, Autoscaling, Load Balancing, Distributed Inference, Containerization}

\textbf{Solutions}
\news{Kubernetes~\cite{burns2016borg}, Prometheus~\cite{turnbull2018monitoring}, Grafana~\cite{chakraborty2021grafana}, Ray~\cite{moritz2018ray}, NVIDIA Triton Inference Server~\cite{tritoninferenceserver}, NVIDIA Dynamo~\cite{dynamo}, Hugging Face Spaces~\cite{huggingfacespaces}, KVCOMM~\cite{ye2025kvcomm}, Cache-to-Cache~\cite{fu2025cache}, NCCLX~\cite{si2025collective}}
}


\section{Conclusion} \label{sec:conclusion}
This paper systematically analyzed the optimization methods and hardware adaptation strategies of LLM inference engines. First, we identified the memory and computation bottlenecks of decoder-only transformers and summarized mitigation methods such as batching, parallelism, caching, and compression. Second, we classified 25 open source and commercial inference engines along two axes, single-node versus multi-node, and homogeneous versus heterogeneous device support, and compared their architectural goals and supported hardware. In particular, we analyzed the inference engine with a focus on ease-of-use, ease-of-deployment, general-purpose support, scalability, throughput-aware, and latency-aware. Our analysis showed that selecting an inference engine required balancing multiple factors, including latency-throughput trade-offs, hardware diversity, inference engine-level optimization support, and SLO. Additionally, we outlined future directions that included multi-agent, multimodal inference support, alternative transformer architectures, longer context windows, improved logical reasoning, application-specific design, stronger security, on-device execution, heterogeneous acceleration, and cloud orchestration. In general, this study provided a practical foundation for designing and operating next-generation inference infrastructure.


\begin{acks}
This work was supported by Institute of Information \& communications Technology Planning \& Evaluation (IITP) grant funded by the Korea government(MSIT) (No. RS-2024-00402898, Simulation-based High-speed/High-Accuracy Data Center Workload/System Analysis Platform) and  Institute of Information \& communications Technology Planning \& Evaluation (IITP) grant funded by the Korea government(MSIT) (No.RS-2023-00277060, Development of open edge AI SoC hardware and software platform).
\end{acks}

\bibliographystyle{ACM-Reference-Format}
\bibliography{ref}










\end{document}